\pdfoutput=1
\def\TTwoArxivBuild{}
\documentclass{article} 
\usepackage{iclr2027_conference,times}

\usepackage{amsmath}
\usepackage{amssymb}
\usepackage{amsthm}

\usepackage{iftex}
\ifXeTeX
  \usepackage{fontspec}
\fi
\usepackage{microtype}

\usepackage{graphicx}
\usepackage{float}
\usepackage{wrapfig}
\usepackage{needspace}
\usepackage{subcaption}
\usepackage{adjustbox}

\usepackage{booktabs}
\usepackage{multirow}
\usepackage{makecell}
\usepackage{array}
\usepackage{tabularx}
\newcolumntype{Y}{>{\raggedright\arraybackslash}X}

\usepackage{xcolor}
\usepackage{colortbl}
\definecolor{t2wRowGray}{HTML}{F1F2F4}
\definecolor{t2wlink}{RGB}{48,54,199}
\definecolor{t2wAppMethod}{RGB}{31,78,140}
\definecolor{t2wAppExp}{RGB}{17,110,88}
\definecolor{t2wAppRes}{RGB}{146,88,24}
\usepackage[
  colorlinks=true,
  linkcolor=t2wlink,
  citecolor=t2wlink,
  urlcolor=t2wlink,
  pdfborder={0 0 0}
]{hyperref}
\usepackage{url}
\usepackage[capitalize,noabbrev]{cleveref}
\crefname{equation}{Eq.}{Eqs.}
\Crefname{equation}{Eq.}{Eqs.}

\usepackage{algorithm}
\usepackage{algorithmicx}
\usepackage{algpseudocode}

\usepackage[most]{tcolorbox}
\usepackage{listings}
\lstdefinestyle{t2wprompt}{
  basicstyle=\ttfamily\small,
  columns=fullflexible,
  keepspaces=true,
  breaklines=true,
  breakatwhitespace=true,
  frame=none,
  backgroundcolor=\color{white},
  xleftmargin=0pt,
  xrightmargin=0pt,
  aboveskip=2pt,
  belowskip=2pt,
  showstringspaces=false
}
\newtcolorbox{t2wpromptbox}[1]{
  enhanced,
  breakable,
  enhanced jigsaw,
  width=\textwidth,
  colback=white,
  colframe=black,
  coltitle=white,
  colbacktitle=black,
  fonttitle=\bfseries,
  title={#1},
  boxrule=0.8pt,
  arc=6pt,
  top=6pt,
  bottom=6pt,
  left=8pt,
  right=8pt
}

\usepackage{enumitem}
\IfFileExists{soul.sty}{\usepackage{soul}}{}
\usepackage{placeins}

\makeatletter
\newcommand{\TTwoAppendixEntries}{}
\newcommand{\TTwoAppendixMethodologyEntries}{}
\newcommand{\TTwoAppendixExperimentEntries}{}
\newcommand{\TTwoAppendixResourceEntries}{}
\newcommand{\TTwoRegisterAppendixSection}[3]{}
\newcommand{\TTwoRegisterAppendixLandmark}[2]{}
\newcommand{\TTwoRegisterAppendixEntry}[5]{%
  \g@addto@macro\TTwoAppendixEntries{%
    \TTwoAppendixEntry{#1}{#2}{#3}{#4}{#5}}}
\newcommand{\TTwoAppendixStoreMethodology}[5]{%
  \g@addto@macro\TTwoAppendixMethodologyEntries{%
    \TTwoAppendixEntry{#1}{#2}{#3}{#4}{#5}}}
\newcommand{\TTwoAppendixStoreExperiments}[5]{%
  \g@addto@macro\TTwoAppendixExperimentEntries{%
    \TTwoAppendixEntry{#1}{#2}{#3}{#4}{#5}}}
\newcommand{\TTwoAppendixStoreResources}[5]{%
  \g@addto@macro\TTwoAppendixResourceEntries{%
    \TTwoAppendixEntry{#1}{#2}{#3}{#4}{#5}}}
\newcommand{\TTwoRegisterGroupedAppendixEntry}[6]{%
  \csname TTwoAppendixStore#1\endcsname{#2}{#3}{#4}{#5}{#6}}
\newcommand{\TTwoWriteAppendixEntry}[6]{%
  \protected@write\@auxout{}{%
    \string\TTwoRegisterGroupedAppendixEntry{#1}{#2}{#3}{#4}{#5}{#6}}}

\newcommand{\appsection}[3]{%
  \xdef\TTwoCurrentAppendixGroup{\csname TTwoAppendixGroupFor#1\endcsname}%
  \section{#3}\label{#2}%
  \TTwoWriteAppendixEntry{\TTwoCurrentAppendixGroup}{section}{\@currentHref}{\thesection}{#3}{\thepage}}
\let\TTwoOriginalSubsection\subsection
\NewDocumentCommand{\TTwoCapturedSubsection}{s o m}{%
  \IfBooleanTF{#1}{%
    \TTwoOriginalSubsection*{#3}%
  }{%
    \IfNoValueTF{#2}{%
      \TTwoOriginalSubsection{#3}%
    }{%
      \TTwoOriginalSubsection[#2]{#3}%
    }%
    \TTwoWriteAppendixEntry{\TTwoCurrentAppendixGroup}{subsection}{\@currentHref}{\thesubsection}{#3}{\thepage}%
  }}
\newcommand{\TTwoBeginAppendixNavigation}{%
  \let\subsection\TTwoCapturedSubsection}

\makeatother

\usepackage{amsmath,amsfonts,bm}

\def\eqref#1{equation~\ref{#1}}

\def\1{\bm{1}}

\DeclareMathAlphabet{\mathsfit}{\encodingdefault}{\sfdefault}{m}{sl}
\SetMathAlphabet{\mathsfit}{bold}{\encodingdefault}{\sfdefault}{bx}{n}

\graphicspath{{figures/generated/}{figures/method/}{figures/}{./}}

\title{Trace2World: Reconstructing App Worlds from Real Interaction Traces for Scalable GUI Agent Training}
\makeatletter\let\TTwoTitleText\@title\def\@title{\raggedright\TTwoTitleText}\makeatother
\newif\ifTTwoArxiv
\ifdefined\TTwoArxivBuild\TTwoArxivtrue\fi
\ifTTwoArxiv
  \iclrfinalcopy 

\newcommand{\TTwoArxivHeader}{Preprint}

\newcommand{\TTwoArxivAuthors}{%
Zezhou Wang$^{1,2\ast}$, Hanzhang Zhou$^{2\ddagger}$, Panrong Tong$^{2}$, Quyu Kong$^{2}$, Pengxiang Li$^{2}$,\\
\bfseries Steven Hoi$^{2}$, Zhuzhong Qian$^{1\dagger}$, Yue Wang$^{2\dagger\ddagger}$\\[2pt]
\small $^{1}$State Key Laboratory of Novel Software Technology, Nanjing University, Nanjing 210023, China\\
\small $^{2}$Alibaba Token Hub, Alibaba Group%
}

\newcommand{\TTwoArxivFooterReserve}{8pt}
\newcommand{\TTwoArxivFootnotes}{%
\footnotetext[1]{Work was done during internship at Alibaba Token Hub, Alibaba Group.}%
\footnotetext[2]{Corresponding authors: \texttt{qzz@nju.edu.cn}, \texttt{yue.w@alibaba-inc.com}.\quad$^{\ddagger}$Project leads.}%
}

\newcommand{\TTwoArxivProjectPage}{}

\newcommand{\TTwoAbstractLink}{ Project page: \url{https://tongyi-mai.github.io/trace2app}.}
\newcommand{\TTwoAbstractLinkLines}{1}

\newcommand{\TTwoAcknowledgments}{\section*{Acknowledgments}
This work was supported by the Programme of Introducing Talents of Discipline to Universities (the
``111 Center'', No.~B26023). We thank the operators who recorded the real-app interaction traces on
which this work is built.}

\hypersetup{
  pdftitle={Trace2World: Reconstructing App Worlds from Real Interaction Traces for Scalable GUI Agent Training},
  pdfauthor={Zezhou Wang, Hanzhang Zhou, Panrong Tong, Quyu Kong, Pengxiang Li, Steven Hoi, Zhuzhong Qian, Yue Wang},
}

  \author{\TTwoArxivAuthors}
  \makeatletter
  \newsavebox\TTwoAnonBox\newsavebox\TTwoAuthorBox\newsavebox\TTwoFootBox
  \newlength\TTwoAuthorExcess\newlength\TTwoFootHeight
  \newlength\TTwoTitleLift
  \newsavebox\TTwoProjectBox\newlength\TTwoProjectLift
  \newlength\TTwoProjectSkip
  \newcommand{\TTwoMeasureProject}{%
    \setbox\TTwoProjectBox\vbox{\hsize\textwidth\normalfont\normalsize\vskip\TTwoProjectSkip
      \centerline{\TTwoArxivProjectPage}}%
    \setlength\TTwoProjectLift{\dimexpr\ht\TTwoProjectBox+\dp\TTwoProjectBox\relax}}
  \newlength\TTwoFooterLift
  \long\def\TTwoMeasureNote#1{\vbox{\reset@font\footnotesize\hsize\columnwidth\@parboxrestore
    \@makefntext{\rule\z@\footnotesep\ignorespaces#1\@finalstrut\strutbox}}}
  \newcommand{\TTwoArxivFirstPage}{%
    \sbox\TTwoAnonBox{\begin{tabular}[t]{l}\bf\rule{0pt}{24pt}Anonymous authors\\Paper under double-blind review\end{tabular}}%
    \sbox\TTwoAuthorBox{\begin{tabular}[t]{l}\bf\rule{0pt}{24pt}\TTwoArxivAuthors\end{tabular}}%
    \setlength\TTwoAuthorExcess{\dimexpr\dp\TTwoAuthorBox-\dp\TTwoAnonBox\relax}%
    \setbox\TTwoFootBox\vbox{\offinterlineskip\renewcommand\thefootnote{\fnsymbol{footnote}}%
      \let\@footnotetext\TTwoMeasureNote\let\H@@footnotetext\TTwoMeasureNote\TTwoArxivFootnotes}%
    \setlength\TTwoFootHeight{\dimexpr\skip\footins+\ht\TTwoFootBox+\dp\TTwoFootBox\relax}%
    \begingroup\renewcommand\thefootnote{\fnsymbol{footnote}}\TTwoArxivFootnotes\endgroup
    \vspace*{-\dimexpr\TTwoAuthorExcess+\TTwoFooterLift-\TTwoTitleLift\relax}%
    \par{\dimen@\prevdepth\nointerlineskip\box\TTwoProjectBox\prevdepth\dimen@}%
    \enlargethispage{\dimexpr\TTwoFootHeight-\TTwoFooterLift-\TTwoArxivFooterReserve\relax}}
  \let\TTwoIclrTitle\@title
  \newlength\TTwoLinkLift
  \begingroup\normalsize\global\TTwoLinkLift=\TTwoAbstractLinkLines\baselineskip\endgroup
  \def\@title{\vskip-\dimexpr\TTwoTitleLift+\TTwoProjectLift+\TTwoLinkLift\relax\vskip-\parskip\TTwoIclrTitle}
  \let\TTwoIclrMaketitle\maketitle
  \renewcommand{\maketitle}{\TTwoMeasureProject\TTwoIclrMaketitle\lhead{\TTwoArxivHeader}\TTwoArxivFirstPage}
  \makeatother
\else
  \author{Anonymous authors} 
  \newcommand{\TTwoAcknowledgments}{}
  \newcommand{\TTwoAbstractLink}{} 
\fi

\newcommand{\appcover}[2]{%
\begin{minipage}[t]{0.235\linewidth}\centering
{\small\bfseries #1}\par\smallskip
\setkeys{Gin}{width=\linewidth,height=0.26\textheight,keepaspectratio}#2%
\end{minipage}}
\newcommand{\TraceSupport}{\Omega_R}
\newcommand{\Implemented}{\Omega_I}


\begin{document}
\raggedbottom

\maketitle

\begin{abstract}
Scaling GUI reinforcement learning requires environments that preserve real-app behavior. Yet coupling
behavior inference with coding can turn partial traces into runnable mocks with incorrect state
transitions. We introduce \textbf{Trace2World}, which separates global behavior reconstruction from
implementation. It reconciles traces into a shared, source-linked app model, then compiles an
executable environment. Independent replay of source workflows localizes mismatches for bounded
implementation repair. Validated worlds then support learner-adaptive task and environment
augmentation within source-supported behavior. From 1,570 traces, Trace2World constructs 50
environments. On 200 workflows across ten apps, 91.1\% are reproduced end to end, versus 59.8\%
for the strongest matched task-local baseline. Source-grounded validation retains 94.0\% of
simulated training gains on real devices, versus 54.7\% under self-validation. Learner guidance
raises benchmark success from 56.9\% to 61.1\%. Training in Trace2World further improves the
model by 20.2 points on our benchmark, 5.4--16.2 points across four external GUI benchmarks, and
16.3 points on 100 real-device tasks. Reliable reconstruction separates what behavior source
evidence supports from whether the implementation preserves it.\TTwoAbstractLink
\end{abstract}

\noindent\begin{minipage}{\linewidth}
\centering
\includegraphics[width=0.90\linewidth]{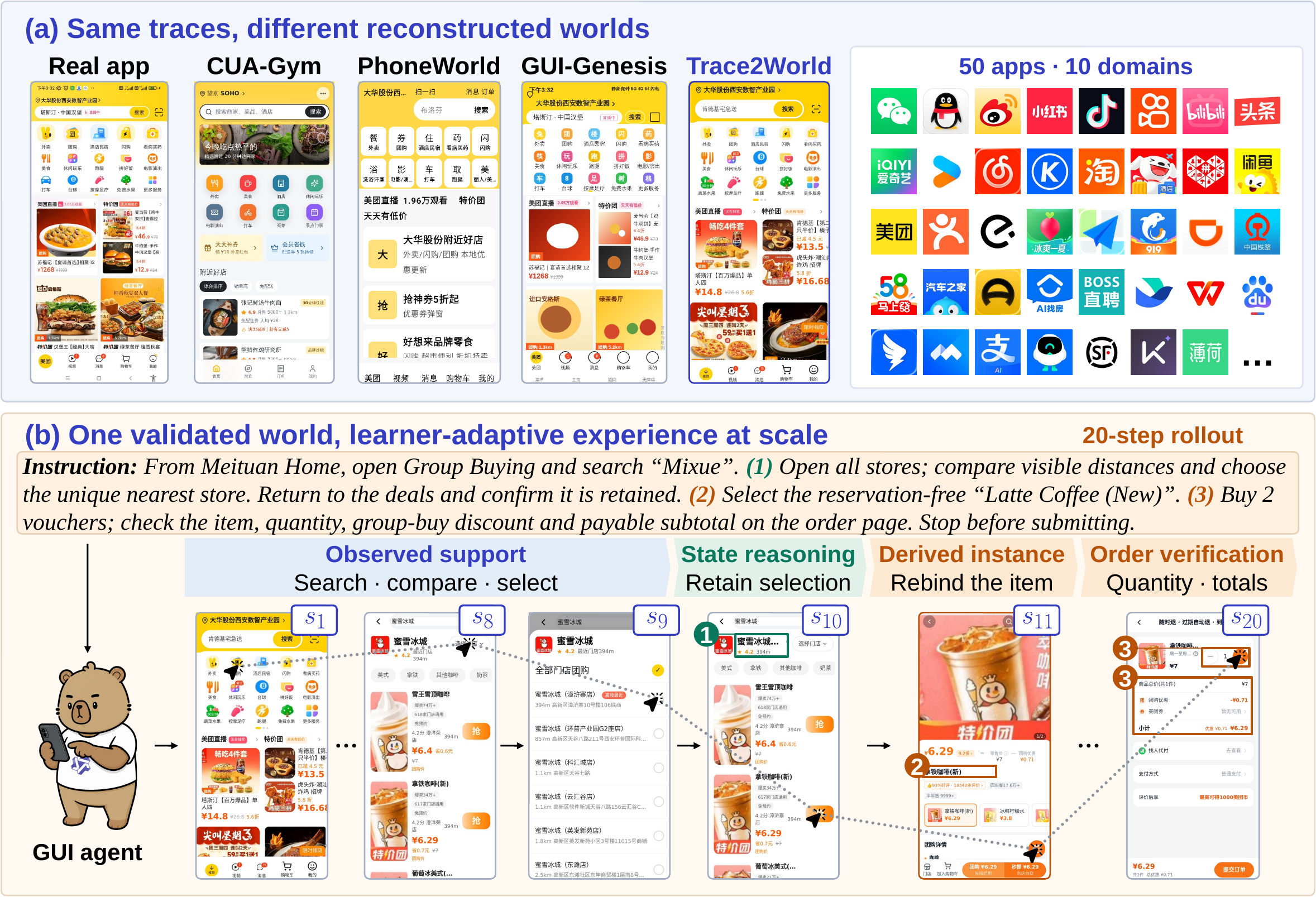}
\captionsetup{type=figure,font=small,skip=2pt}
\caption{\textbf{Faithful app worlds, learner-adaptive experience at scale.}
(a) Left: Baseline environments constructed from the same source traces. Right: 50 reconstructed apps across ten domains.
(b) Learner-adaptive tasks recombine source-supported behavior with state reasoning, derived
instances, and order verification. A \emph{derived instance} reuses observed page structure and
purchase actions with new item data; here, a latte replaces the source trace's snow-top coffee
without changing app semantics.}
\label{fig:portfolio-overview}
\end{minipage}

\newpage

\section{Introduction}
\label{sec:introduction}
\suppressfloats[t]

Online reinforcement learning improves GUI agents through repeated interaction
\citep{bai2024digirl,qi2024webrl,qin2025uitars,xu2025mobilerl}. Real apps are hard to reset, and
exploratory actions can change remote state or commit transactions
\citep{rawles2024androidworld,xie2024osworld,kong2026mobileworld,yan2025stepgui}. Simulators offer
controlled starts, yet training value depends on transfer: visual \citep{rusu2017sim2real} or dynamics
\citep{peng2018sim2real} gaps can break learned policies. For pixel-based GUI agents, these gaps lie in
what the policy sees and what its actions do, so we use \emph{fidelity} to mean preserving the source
app's task-relevant observations and action semantics, including state, data, intermediate effects,
and outcomes, not only its look or final value.

\begin{wrapfigure}{R}{0.38\textwidth}
\vspace{-10pt}
\centering
\includegraphics[width=\linewidth]{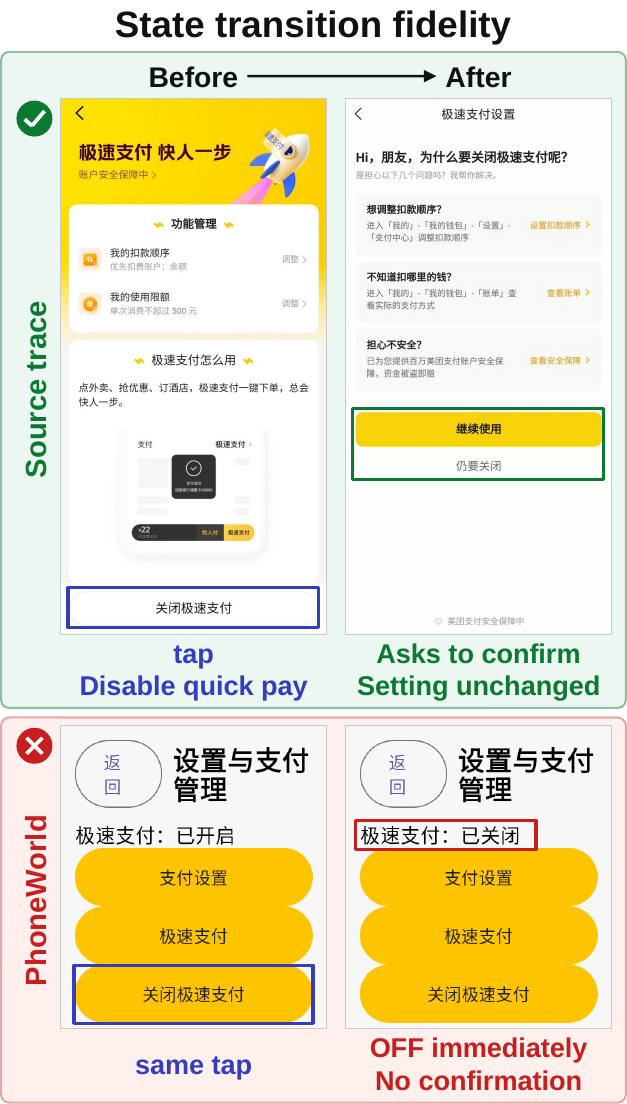}
\captionsetup{justification=raggedright,singlelinecheck=false,skip=3pt}
\caption{\textbf{Same goal, different first transition.} The source's first \emph{Disable quick
pay} tap opens a confirmation; our matched PhoneWorld-style reconstruction saves the disabled
value on that tap.}
\label{fig:choice-boundary}
\vspace{-8pt}
\end{wrapfigure}

Obtaining faithful environments at scale, however, remains difficult. MobileGym authors app behavior by
hand \citep{wu2026mobilegym}. Coding agents reduce this manual work, and every existing route hands
them the semantics as well: CUA-Gym codes mocks from topic specifications \citep{wang2026cuagym},
GUI-GENESIS codes a task-local site from one trace \citep{cao2026guigenesis}, and PhoneWorld, in
concurrent work, codes reusable Android pages from frequency-ranked screen types, consecutive-screen
transitions, and page specifications written from two or three screenshots, then checks
representative flows with human QA \citep{liu2026phoneworld}. None asks which shared identities,
initial states, data, guards, or conflicting effects the traces support: the coding agent settles
them while writing the mock, and builder tests or a human comparison of representative flows then
close fidelity. Reconstruction reduced to implementation fails on the app's own workflows even at
the level of structure: on the same recordings, our matched PhoneWorld-style build preserves source
page surfaces on only 39.3\% of workflows, 1.8 pp above a topic-specification builder
(Table~\ref{tab:modeling-main}). Automating code has not settled which app behavior the code may
commit to, and two failures follow.

\textbf{Partial paths do not define the app.} Each trace exposes only one path from one initial
state \citep{rawles2023aitw,deng2023mind2web}: Home screens showing Xi'an and Beijing may be two
pages or one page in two city states, while one control can have state-dependent effects
(Section~\ref{sec:anchored-construction}). A page-transition graph records what was observed, not
the state condition that explains it, so coding before reconciliation hardens one interpretation
into the app. Across our matched builds of this family, routes reach 50.4--94.8\% of endpoints yet
preserve only 16.7--59.8\% of source workflows (Table~\ref{tab:modeling-main};
Figure~\ref{fig:portfolio-overview}a and Appendix~\ref{app:matched-baseline-trajectories}).

\textbf{Passing tests do not establish fidelity.} An implementation and the tests derived
from the same interpretation can agree while both diverge from the source \citep{barr2015oracle}.
In Meituan, tapping \textit{Disable quick pay} opens a confirmation without changing the saved
setting; our PhoneWorld-style reconstruction disables it immediately yet passes a final-value check
(Figure~\ref{fig:choice-boundary}), rewarding a policy for a transition the real app does not
perform. Thus the terminal value alone cannot show that the workflow matches the source.

\begin{figure}[t]
\centering
\includegraphics[width=0.9\linewidth]{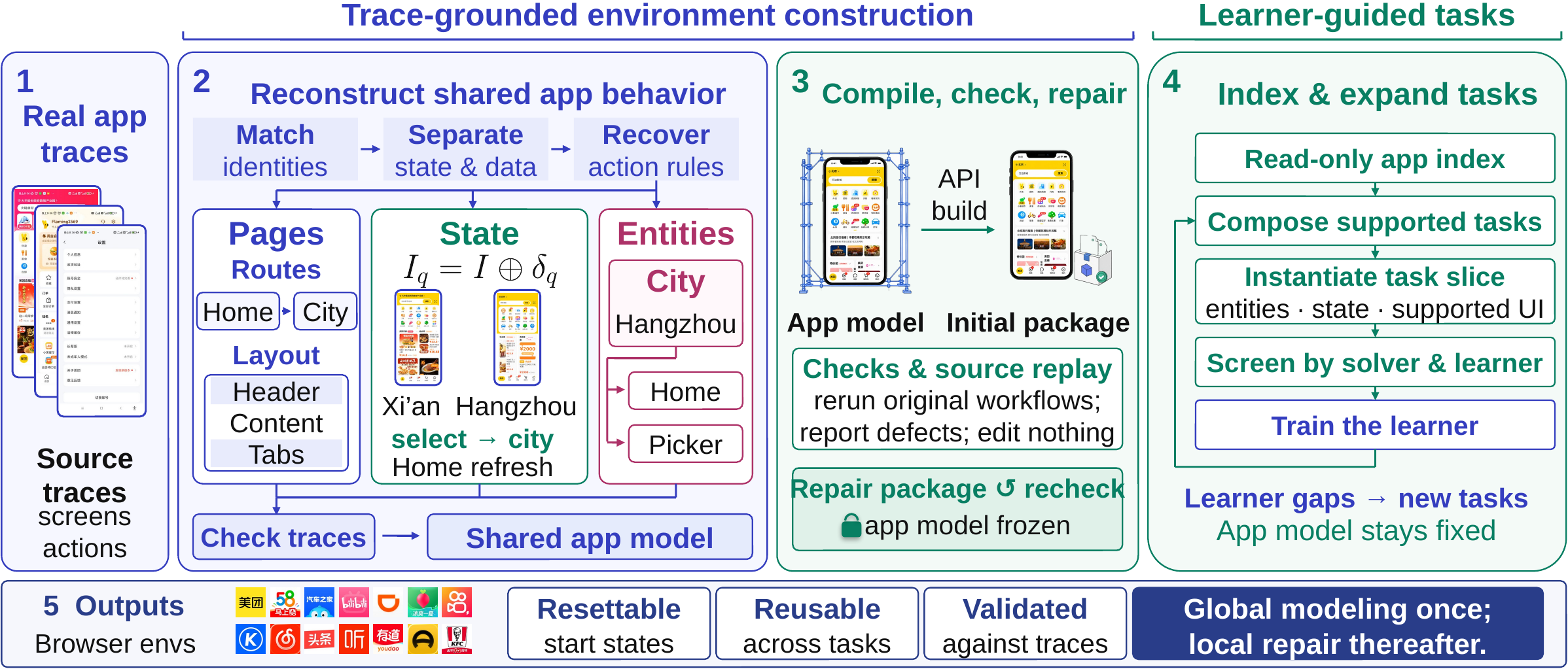}
\caption{\textbf{Semantic commitment before implementation, source acceptance afterward.}
Partial traces are reconciled into one source-linked model of shared app behavior; unresolved
behavior stays non-executable. Bounded API calls compile the model into a package; code checks and
replay of the original workflows only report defects, and repair edits the package while the model
stays frozen. The validated world is indexed read-only and expanded with learner-guided tasks.}
\label{fig:method-overview}
\end{figure}

\begin{samepage}
Both failures arise when an ambiguous page identity, state value, or action effect becomes a coding
choice before the traces are reconciled or an independent check can contest it. The remaining
challenge is therefore not a better coding loop but a question no existing route poses:
\begin{center}
\emph{Which app semantics do partial traces actually support, and what evidence may accept an
implementation of them?}
\end{center}
\end{samepage}

Trace2World answers with two boundaries, \textbf{semantic commitment before implementation
and source acceptance afterward}: it resolves open-ended app semantics before construction, then
reduces later failures to local repair over a concrete package. It first reconciles repeated
observations into shared page and entity identities, then separates mutable state, reusable data,
and task-specific initial values. When observed state explains different effects, the model records a
conditional rule; otherwise the conflict remains unresolved and outside executable scope. Because
every supported relation keeps its source links, additional traces widen supported behavior without
admitting unsupported shortcuts (Figure~\ref{fig:method-overview}).

Bounded API calls then compile the fixed model into an initial browser environment per app. Code
checks and independent replay of each original workflow compare information, intermediate effects,
and terminal state with the source and only report defects; a bounded repair loop edits the package
under the frozen model, and a workflow that cannot pass without changing the model is excluded.
Finally, learner feedback guides task composition from a read-only index of the validated world: it
changes what is practiced or instantiates supported behavior, but never redefines what the app
means. Our contributions follow this sequence:

\begin{enumerate}[leftmargin=*,nosep]
\item \textbf{Global behavior reconstruction as the semantic contract.} Source-linked semantics are
fixed from evidence before any code exists. From 1,570 partial traces, we build 50 shared worlds;
on 200 workflows across ten apps, 91.1\% pass all six source checks versus 59.8\% for the strongest
existing route, our matched task-local GUI-GENESIS-style build.
\item \textbf{Source-grounded detection with bounded local repair.} Independent replay of every
original workflow checks six source obligations, apart from builder tests and RL rewards, and shows
that every existing route reaches endpoints it cannot preserve. Replacing replay with
self-validation lowers the share passing all six from 91.1\% to 79.1\% and real-device gain
retention from 94.0\% to 54.7\%.
\item \textbf{Validated worlds that index and expand cheaply.} Forty validated environments supply
1,280 seed and 768 learner-guided tasks within fixed source support. Learner guidance raises
benchmark success from 56.9\% to 61.1\%; our final policy improves on four external GUI benchmarks
and by 16.3 points on real devices.
\end{enumerate}

\section{Related Work}
\label{sec:related-work}

\paragraph{Environment construction and semantic fidelity.}
Constructed GUI environments differ less in how they are coded than in where their semantics come
from and what evidence establishes correctness (Table~\ref{tab:env-landscape}). AndroidWorld
executes the real application, while MobileGym, WebArena, and AppWorld author environment behavior
by hand \citep{rawles2024androidworld,wu2026mobilegym,zhou2023webarena,trivedi2024appworld}. CUA-Gym
synthesizes topic-based mocks together with their verifiers \citep{wang2026cuagym}, GUI-GENESIS
builds one task-local site and reward per trace \citep{cao2026guigenesis}, and PhoneWorld, in
concurrent work, codes reusable Android mocks from recovered screen types and transitions and closes
representative flows by human side-by-side QA \citep{liu2026phoneworld}. These coding-agent routes commit
semantics inside implementation. Trace2World changes the semantic contract: every
trace of an app is reconciled into one source-linked model before any code exists, unresolved
conflicts stay outside executable scope, and independent replay of every source workflow decides
acceptance.

\paragraph{Experience supply and evolving tasks.}
A separate line of work adapts training experience once an environment exists. Online GUI RL and
task synthesis draw new experience from interaction
\citep{bai2024digirl,qi2024webrl,yang2025zerogui,xue2026evocua,xu2025mobilerl,shi2025mobileguirl}.
Qwen-UI-Agent synthesizes tasks, environments, and executable verifiers and iterates on failure analysis
\citep{zhou2026qwenuiagent}; Agent-World uses capability gaps to evolve tool-environment curricula
\citep{dong2026agentworld}. Learnability-driven curricula adapt task selection to
learner competence \citep{sukhbaatar2018selfplay,wang2019poet,dennis2020paired,jiang2021plr}. The
pipeline connects the two lines through a validated app world: semantics are reconstructed and accepted once
per app, while learner feedback changes only which supported behavior is practiced.

\section{Trace2World}
\label{sec:method}

\subsection{From Traces to One Shared App}
\label{sec:problem-setting}

A \emph{source trace} records one successful human workflow on a real app: its instruction,
observations, actions, terminal criterion, and collection metadata. It exposes only one path from one
initial state. Trace2World reconciles traces from the same app into \emph{one shared app model}:
common pages and entities, mutable state, reusable data, task-specific initial values, and
state-dependent action effects, each linked to supporting source steps. Ambiguous or unsupported
behavior remains non-executable. This model is the semantic commitment boundary between source
evidence and browser code.

To separate reconstruction, implementation, and task authoring, Trace2World maintains three objects.
The \emph{app model} $S$ is reconstructed once from source traces and then frozen; the
\emph{executable package} $E_r$ is the browser world and the only object repair edits; and the
\emph{task-facing index} $I_r$ is a read-only projection of a validated package for task authoring.
Source replay checks $E_r$ against the recordings, while learner feedback changes which supported
experience is generated. This separation yields one resettable environment per app with task-specific
initial state rather than task-specific semantics; definitions and the algorithm appear in
Appendix~\ref{app:induction-details}.

\subsection{Reconciling App Semantics Across Traces}
\label{sec:anchored-construction}
\label{sec:scaffold-induction}

\begin{wrapfigure}[20]{R}{0.40\textwidth}
\vspace{-8pt}
\centering
\includegraphics[width=\linewidth]{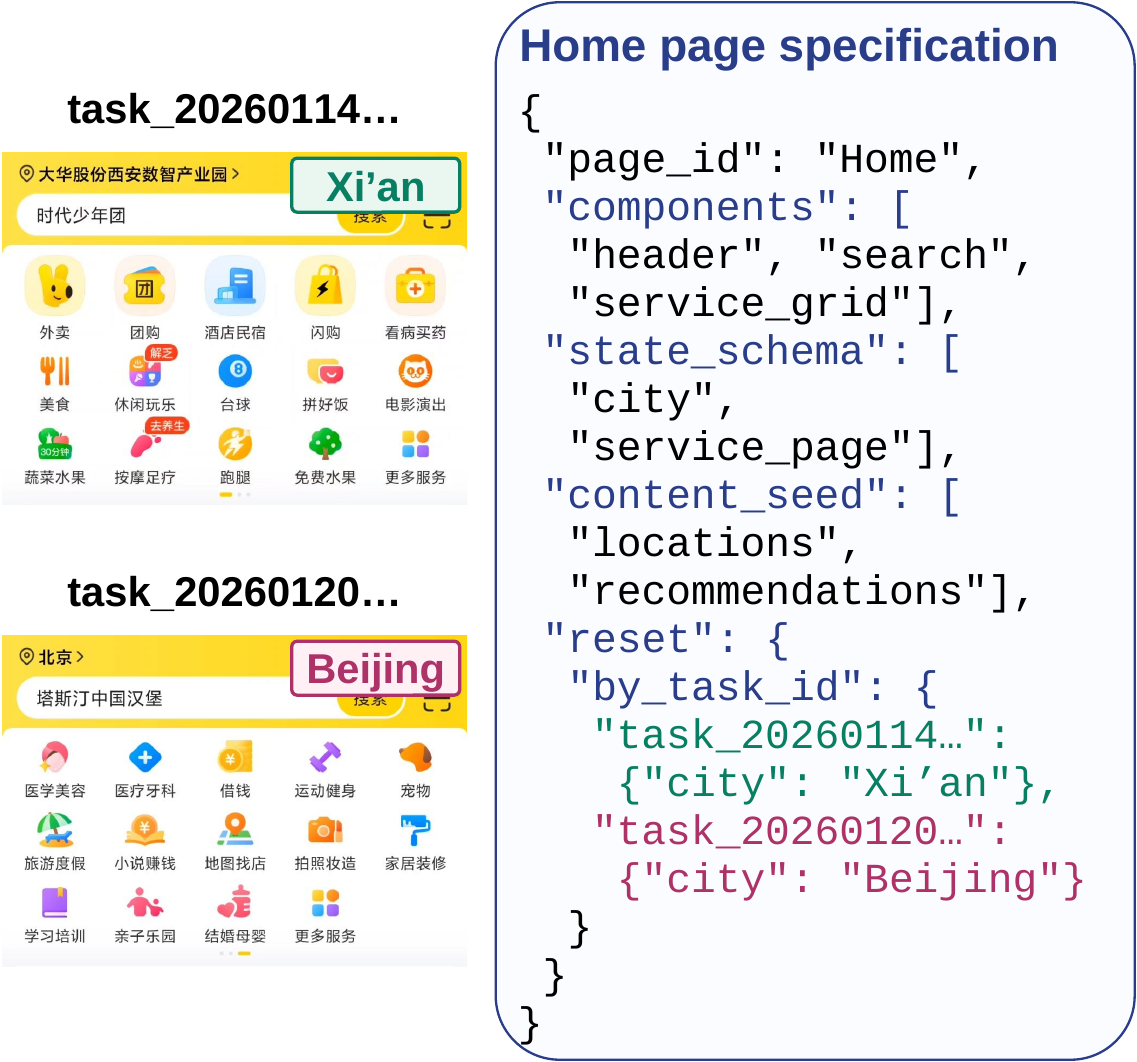}
\caption{\textbf{Two traces, one Home page.} Xi'an and Beijing are supported initial values of the
same city state field. The task selects a reset value; subsequent behavior uses the shared page and
state-dependent action rules.}
\label{fig:home-contract}
\vspace{-8pt}
\end{wrapfigure}

Two observations may look different because they show different pages, different values of the same
state, or one entity rendered on different surfaces. The pipeline first matches repeated pages, action hosts,
and entities across traces using their containment, routes, lifecycle, and observed transitions. It
then separates stable interface structure from mutable state and reusable data. For example, the
Xi'an and Beijing entries in Figure~\ref{fig:home-contract} share one Home page; city is state set at
task reset, so changing the task does not duplicate the page or alter its action rules.

Actions that appear identical can still have different effects. The model links each candidate effect to
the source step that shows its precondition and result. If observed state explains a difference, the
app model records that condition. If no observed condition explains conflicting effects, it keeps the
conflict unresolved instead of choosing one interpretation. Concretely, each supported interaction
has a page and control, an action and guard, state or data reads and writes, a destination, and a
visible postcondition. The resulting action rule is reused across workflows.
A final support check requires complete page and entity mappings, compatible effects, their data
dependencies, and coverage of the cited evidence
(Appendices~\ref{app:trace-preprocessing} and~\ref{app:induction-algorithm}).

An app's traces may exceed one model context, so reconciliation uses bounded shards and reductions
that preserve source links, unresolved conflicts, and uncovered evidence. Across all 50 apps, the largest
model-visible call is 33.2K tokens under a 40K-token budget, with 99.1\% audited semantic coverage
(Appendices~\ref{app:bounded-execution} and~\ref{app:scaling-detail}).

\subsection{Building and Validating the Environment}
\label{sec:package-closure}

Bounded API calls compile the supported part of the app model into page instructions and then into
an initial standalone, DOM-backed browser package; no open-ended coding agent builds the world. A
central state kernel owns business state and applies guarded updates; page code chooses only
rendering, routes, and storage. Unsupported controls stay visible but inert, and unsupported clicks
or failed guards leave business state unchanged. Code checks confirm that required pages, resources,
hosts, bindings, and resets are present; they establish that the package runs, not that it matches
the source app (Appendix~\ref{app:package-substrate}).

Source validation replays each recorded workflow from reset in the package. A read-only checker
compares the original trace with the resulting information, intermediate effects, and terminal
state without seeing the app model, generated verifier, or repair log; contract probes also check
reset, repeated actions, guards, and supported alternatives. Detection and editing are separate
roles: neither check edits anything, and each failure becomes a defect packet naming the step and
the expected and observed state.

Compilation gives repair a concrete executable hypothesis. Reconstruction jointly infers identity,
state, data, and action dynamics from partial evidence; repair localizes mismatches within the
package. A bounded agent receives the package, defect packet, and source evidence, edits only the
package, and reruns both checks. It may correct state updates, control handlers, or other
implementation errors exposed by the source, but cannot change the frozen app model or add
unsupported behavior. Freezing reduces joint revision of semantics and implementation to local
repair of one artifact under a fixed contract.

Repair remains bounded by reconstructed support. A workflow that cannot pass without changing the
app model is excluded after three repair waves; later model revisions begin a separate construction
revision. The evaluator for RL reward and held-out success is separate from source replay
(Appendices~\ref{app:closure-algorithm}--\ref{app:closure-status}), so validation can reject a workflow
accepted by the package's own success check (Figure~\ref{fig:choice-boundary}). After trace collection,
acceptance requires no human QA; human fidelity labels serve only the independent evaluation audit
(Appendix~\ref{app:human-routes}).

\subsection{Learner-Guided Task Generation}
\label{sec:policy-expansion}

\begin{figure}[!t]
\centering
\includegraphics[width=0.9\linewidth]{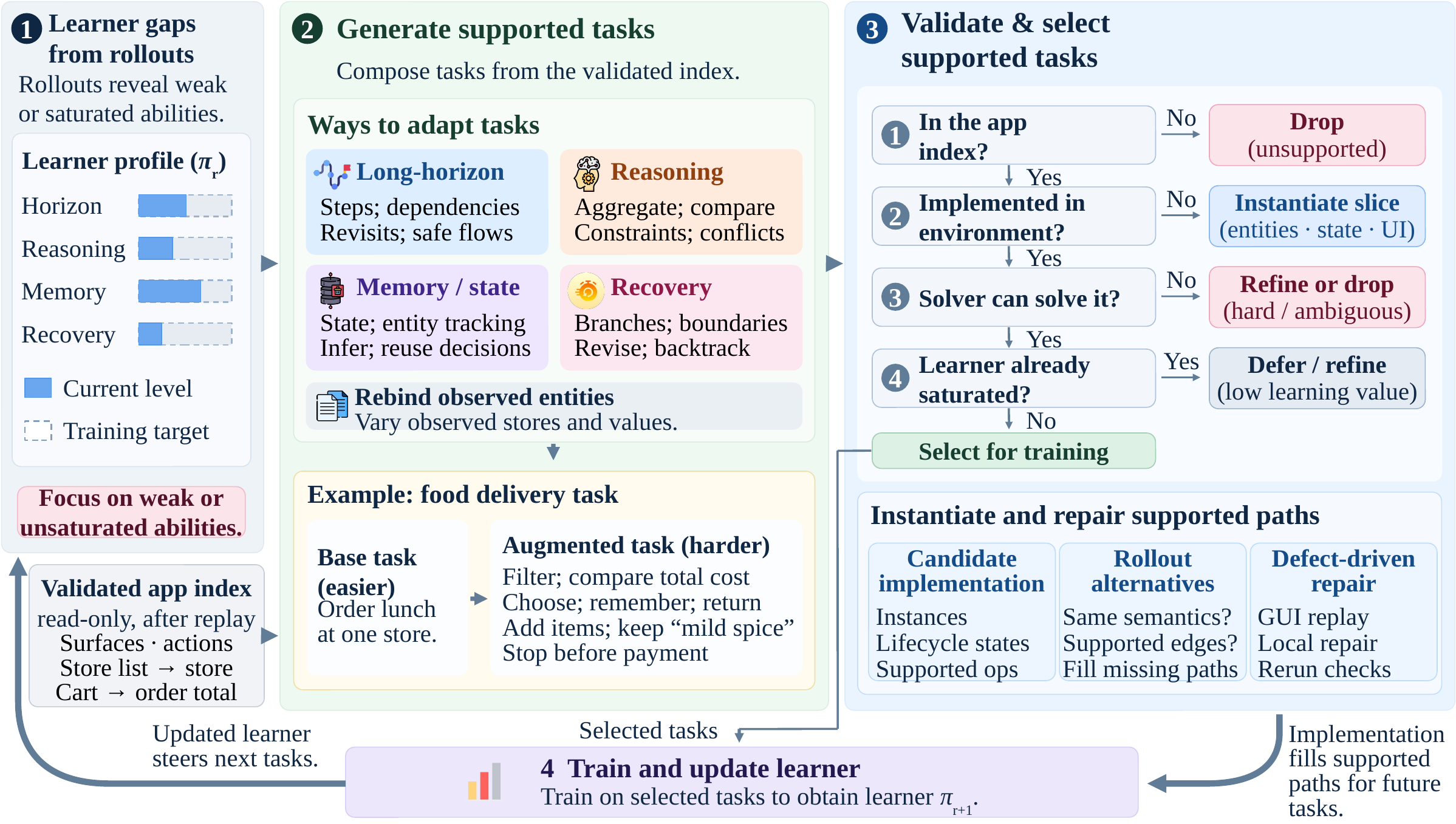}
\caption{\textbf{Learner-guided tasks from a validated app index.} Task generation combines
supported surfaces, operations, and dependencies from the validated package's read-only index into
a food-delivery task adapted to the current learner. Learner gaps guide sampling; source, GUI,
solver, and learner-signal checks screen candidates. Missing entity, state, or UI slices are
instantiated in the package; the app model is unchanged. Other icons by Freepik from Flaticon.}
\label{fig:composition}
\end{figure}

After source replay, Trace2World extracts a read-only index of the validated package's surfaces,
entities, observable fields, supported operations, and already-modeled dependencies, retaining
source and implementation tags. The indexer reconstructs no behavior. Task generation queries the
index to compose tasks, rebind observed entities or initial states, and select combinations relevant
to the current learner (Figure~\ref{fig:composition}). Candidates requiring unsupported app-level
transitions, guards, or effects are rejected.

A supported task may still need content or implementation absent from the package. Trace2World
instantiates schema-backed entities, values, and state or materializes already-supported operations
as a task slice. GUI execution and affected seed-task regressions check the expanded package before
the index is refreshed. Learner feedback may expose these gaps but cannot change the app model
(Appendices~\ref{app:expansion-details} and~\ref{app:alternative-path-results}). Augmentation thus
amortizes one semantic reconstruction and source acceptance across tasks.

A fixed external solver filters unsolvable candidates; four current-learner attempts estimate reward
SD, with values above 0.1 marking useful variation. Saturated candidates may be refined twice within
source support. Training on 1,280 seeds yields $\pi_1$; two 384-task guided batches then yield
$\pi_2$ and $\pi_3$ (the final policy). Task membership, environment revision, and evaluator freeze before each
stage (Appendix~\ref{app:policy-optimization}).

Evaluation-only cross-app tasks join two source-validated app-local indexes: a GUI-visible output
in $A$ conditions work in $B$, and a visible result in $B$ changes the final action after returning to
$A$, adding a frozen task-level dependency but no app-local behavior or cross-app source recording.
All 2,048 optimization tasks remain single-app (Appendix~\ref{app:cross-app-runner}).

\section{Experiments}
\label{sec:experiments}
\label{sec:exp-setup}

The pipeline constructs one shared environment per app from 1,570 real-app traces across 50 apps. Forty apps
supply 1,280 seed and 768 learner-guided tasks for policy training; the other ten contribute
construction traces but no optimization tasks and form the unseen-app split. We test three questions:
(i) do reconstruction and independent source
validation preserve recorded workflows on a fixed 200-workflow, ten-app set; (ii) do validated worlds
support useful adaptive training on Trace2World-Bench and four external GUI benchmarks; and
(iii) do simulation gains transfer to real devices on 100 paired tasks across 20 policy-training apps?
Qwen3.5-9B \citep{qwen2026qwen35} is trained with DAPO \citep{yu2025dapo} across three frozen stages;
Qwen3.7-Plus is used only for solvability screening. Full split, horizon, model-role, and training
contracts appear in Appendices~\ref{app:model-roles}, \ref{app:portfolio}, and~\ref{app:training-config}.

\Needspace{6\baselineskip}
\subsection{End-to-End Reconstruction Fidelity}
\label{sec:reconstruction-fidelity}
\label{sec:modeling-fidelity}
\label{sec:artifact-fidelity}
\label{sec:scaling-results}

\paragraph{Setup and metrics.}
All conditions use the same 200 source workflows from the ten apps held out from policy training, one
per domain. We operationalize fidelity as \emph{workflow fidelity}, the share of source workflows that
pass all six checks: reachability, surface, state, effect, terminal outcome, and data. Equivalent
routes and page partitions pass when they preserve the required decisions and effects. A blinded
GPT-5.6 Sol reviewer compares each anonymous execution with its source recording and labels every
check P/F; missing evidence is F, and rates are macro-averaged over apps (Appendix~\ref{app:gold-matching}).

We reimplement Direct Agent, CUA-Gym, PhoneWorld-style, and GUI-GENESIS-style on the same substrate,
tools, runtime, and recordings, with matched model capability per builder role and each route's
construction rules (Appendix~\ref{app:baselines}). Builder rows compare complete construction
contracts; only the controls isolate one factor: no-trace variants withhold the recordings, and
self-validation changes only the validation source (Table~\ref{tab:contract-comparison}).

\begin{table}[!t]
\centering
\caption{\textbf{End-to-end reconstruction fidelity (\%).} Ten-app macro means over the same 200 source workflows. All six (workflow fidelity) requires every check; Reach checks only the endpoint.}
\label{tab:modeling-main}
\footnotesize
\setlength{\tabcolsep}{3pt}
\begin{tabular*}{\linewidth}{@{\extracolsep{\fill}}lccccccc@{}}
\toprule
\multirow{2}{*}[-0.5\dimexpr\aboverulesep+\belowrulesep+\cmidrulewidth\relax]{Method} & \multicolumn{1}{c}{Overall} & \multicolumn{2}{c}{Structure} & \multicolumn{3}{c}{State} & \multirow{2}{*}[-0.5\dimexpr\aboverulesep+\belowrulesep+\cmidrulewidth\relax]{Data} \\
\cmidrule(lr){2-2}\cmidrule(lr){3-4}\cmidrule(lr){5-7}
 & All six & Reach & Surface & State & Effect & Terminal &  \\
\midrule
Direct w/o traces & 5.2 & 28.4 & 14.8 & 22.7 & 15.0 & 15.9 & 12.8 \\
Direct + traces & 16.7 & 53.6 & 35.3 & 41.2 & 34.6 & 33.7 & 31.7 \\
CUA-Gym w/o traces & 8.2 & 34.4 & 20.0 & 29.0 & 22.6 & 21.5 & 19.6 \\
CUA-Gym + traces & 19.9 & 50.4 & 37.5 & 38.0 & 35.0 & 35.3 & 35.0 \\
PhoneWorld-style & 22.4 & 57.0 & 39.3 & 46.8 & 41.5 & 39.8 & 38.5 \\
GUI-GENESIS-style & 59.8 & 94.8 & 83.9 & 76.6 & 77.0 & 79.4 & 78.4 \\
\textbf{Trace2World} & 91.1 & 98.0 & 93.7 & 93.3 & 93.8 & 93.7 & 94.1 \\
\rowcolor{t2wRowGray}\multicolumn{8}{@{}l}{\emph{Control (not our method)}} \\
Self-validation & 79.1 & 98.5 & 90.4 & 87.0 & 85.0 & 88.0 & 89.1 \\
\bottomrule
\end{tabular*}
\end{table}

\textbf{Reachability alone overstates reconstruction quality.} With the same recordings, Direct Agent,
CUA-Gym, and PhoneWorld-style reach 50.4--57.0\% of endpoints, yet only 16.7--22.4\% of their
workflows pass all six checks (Table~\ref{tab:modeling-main}). The task-local GUI-GENESIS-style build
reaches 94.8\% but passes 59.8\%, with the State that workflows share at 76.6\%. Trace2World reaches
98.0\% and passes 91.1\% (per-app contrasts in Appendix~\ref{app:coverage-correctness}).

\textbf{The gap reflects where app behavior is committed.} When semantic inference and implementation
are coupled, an uncertain page identity, state value, or action effect becomes a coding choice before
conflicting source evidence is reconciled. A builder can therefore recover a reachable surface while
encoding the wrong state or effect behind it. Trace2World instead reconstructs source-linked behavior
before compilation and leaves unresolved conflicts outside executable support.

\textbf{The ordering is not explained by a stronger constructor or reviewer bias.} GPT-5.2 performs
Trace2World's reconciliation and initial compilation, and GPT-5.5 only its verification and repair,
whereas compared builders receive GPT-5.5 in every role they define. Blind human relabeling agrees
with the automated reviewer on 90.2\% of the same 1,200 packets and preserves the ordering
(Appendices~\ref{app:model-roles} and~\ref{app:human-routes}).

\textbf{More source evidence expands supported behavior without collapsing fidelity.} Growing the
trace set of the same ten apps from 100 to 290 traces expands supported routes from 270 to 512 and
visual obligations from 553 to 1,076, while workflow fidelity, scored against each trace budget's own
supported scope, moves only from 92.6\% to 91.1\% (Figure~\ref{fig:artifact-trace-scaling} in
Appendix~\ref{app:artifact-scaling}). Across all 1,570 source workflows, 74.6\% pass source replay on
the first attempt and all close within the $K=3$ automated repair cap without human QA
(Appendix~\ref{app:automation-yield}). A seed build takes an estimated 3.6 hours and 1,160 model calls
per app, comparable to PhoneWorld-style and below GUI-GENESIS-style (Appendix~\ref{app:construction-cost}).

\FloatBarrier

\subsection{Source-Grounded Validation}
\label{sec:source-validation}

\paragraph{Setup and metrics.}
The self-validation control changes only validation evidence under the same construction stages
and inputs (Table~\ref{tab:modeling-main}). We compare stage-1 policies on the same 100 paired
sim/device tasks, with matched goals, restorable starts, terminal criteria, and action budgets; the
50 entity/state rebindings and 50 held-out workflows never enter RL, and results average training
seeds (Appendix~\ref{app:real-device-protocol}).

\textbf{Self-consistency does not establish source fidelity.} Replacing source-trace validation with
self-validation lowers workflow fidelity from 91.1\% to 79.1\% while Reach stays near 98\%. Independent
source rechecks reject 22/160 (13.8\%) self-accepted executions versus 1/176 (0.6\%) source-accepted
ones (Appendix~\ref{app:generated-closure-control}). A separate 400-execution audit isolates the
mechanism: giving the same model reviewer the source trace raises defect coverage from 43.2\% to
68.4\% and precision from 73.6\% to 88.2\% while flagging fewer clean executions
(Appendix~\ref{app:review-repair-results}).

\textbf{Source-trace validation decides how much environment gain transfers.} On the shared 100-task
panel, source validation retains 94.0\% of its 16.7-pp simulated gain on real devices (15.7 pp),
versus 54.7\% of 17.7 pp under self-validation (9.7 pp), a 6.0-pp paired difference (seed SD 1.0).
With training matched and simulated gains nearly equal, the rise in simulation-only passes from
10.0\% to 18.3\% locates the transfer loss in behavior the simulator accepts but the device rejects.

\subsection{Learner-Guided Task Generation}
\label{sec:task-supply}
\label{sec:expansion-results}

\paragraph{Setup and metrics.}
Four branches fork from $\pi_1$ for the same 20 RL iterations (32--52), environment revisions, sampler,
and evaluator: learner-guided generation uses feedback; static generation removes it from the same generator;
entity rebinding substitutes seed-template entities; and seed continuation redraws seed tasks, each with two
384-task training-only batches. Static and learner-guided tasks also share $H=45$; seed-based tasks use $H=30$.
Four attempts per candidate under its class budget give current-policy difficulty (mean pass) and
learning signal (reward SD above 0.1); the held-out benchmark reports SR and PR.

\textbf{Learner feedback makes tasks harder for the learner, not easier.} Cross-surface composition,
lifecycle revisit, and conditional recovery become more common than in the 1,280 seeds, and the
dependency-heavy share rises from 46\% to 71\% across the two guided batches. The current learner's
pass rate among validated candidates falls from 41\% to 29\% while the fixed solver's moves only from
71\% to 66\%, and 194 of 200 audited accepted candidates are independently solvable by humans; the
new tasks recombine supported capabilities rather than add semantics (Appendices~\ref{app:task-dimensions}
and~\ref{app:generation-results}).

\textbf{New tasks expand implementation, not semantics.} The support gate rejects 77 and 90 of the
576 candidates proposed per guided round (Table~\ref{tab:stage-funnel}), so selected tasks bind only
indexed, supported items. Of 611 terminal-reaching attempt cases on paths absent from source traces, 486 are
source-equivalent, 402 already executing and 84 materialized from supported edges; the remaining
125 ambiguous, invalid, or unresolved cases stay outside support (Appendix~\ref{app:alternative-path-results}).

\begin{figure}[!t]
\centering
\includegraphics[width=0.94\linewidth]{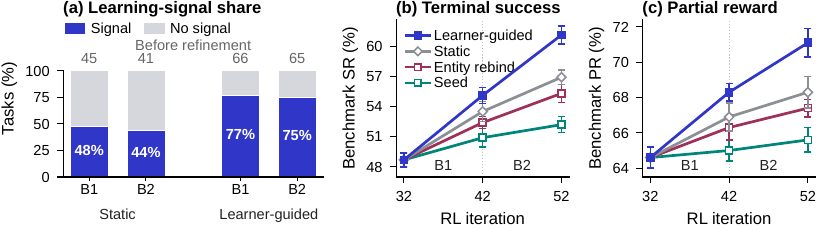}
\caption{\textbf{Learner-guided task supply.} (a) Learning-signal share in selected tasks
(four-attempt reward SD $>0.1$); gray labels above bars show the corresponding first-valid share
before bounded refinement. Only learner-guided generation refines candidates; B1/B2 use the current
branch learner. (b,c) Trace2World-Bench SR/PR for
iteration/rollout-matched continuations; only the two augmentation branches also share $H=45$.
Error bars show three-seed SD.}
\label{fig:task-supply-adaptation}
\end{figure}

\textbf{Learner-conditioned generation supplies tasks with learning signal.}
Across the two batches, 76.8/75.3\% of guided tasks carry learning signal versus 47.9/44.0\% for the
static control. The gap precedes bounded refinement, which is sparse: only 14.1/14.5\% of validated
lineages enter it, and about 60\% of those become selectable within two rounds. Feedback thus changes
which supported tasks are proposed, not just which candidates are rescued
(Figure~\ref{fig:task-supply-adaptation}a, Appendix~\ref{app:pool-results}).

\textbf{Learner-guided batches outperform matched continuations.}
At iteration 52, SR is 61.1\% guided, 56.9\% static, 55.3\% entity rebinding, and 52.2\% seed
continuation. The two $H=45$-matched augmentation branches differ by 4.2 pp, isolating learner
feedback (Figure~\ref{fig:task-supply-adaptation}b,c); external MobileGym keeps the ordering
($25.2\pm1.2$ versus $22.2\pm1.3$ SR). Budget does not substitute for feedback: training on all
426/427 validated candidates for 54 iterations reaches 57.8\% SR, and seed continuation matched to
the guided run's 1,351 retained groups needs 79 iterations to reach 54.3\%
(Appendix~\ref{app:task-supply-performance}).

\subsection{Downstream Policy Performance and Real-App Transfer}
\label{sec:downstream-transfer}
\label{sec:downstream-performance}
\label{sec:real-application-transfer}

\begin{table}[!t]
\centering
\caption{\textbf{Downstream policy performance (\%).} Trace2World-Bench and MobileGym (MG) report SR/PR; AndroidWorld (AW), MobileWorld (MW), and AndroidDaily (AD) report SR (MW: GUI-only SR@50). Reference agents give capability context; the fixed Qwen3.5-9B block isolates training gain. Bold marks the best in each block (Appendices~\ref{app:external-performance},~\ref{app:benchmark-tables}).}
\label{tab:downstream-main}
\footnotesize
\setlength{\tabcolsep}{0.8pt}
\begin{tabular*}{\linewidth}{@{\extracolsep{\fill}}lccccccccc@{}}
\toprule
\multirow{2}{*}[-0.5\dimexpr\aboverulesep+\belowrulesep+\cmidrulewidth\relax]{Model} & \multicolumn{5}{c}{Trace2World-Bench (SR / PR)} & \multirow{2}{*}{\makecell{MG\\SR / PR}} & \multirow{2}{*}{\makecell{AW\\SR}} & \multirow{2}{*}{\makecell{MW\\SR@50}} & \multirow{2}{*}{\makecell{AD\\SR}} \\
\cmidrule(lr){2-6}
& \makecell{Entity\\repl.} & \makecell{Held-out\\workflow} & Cross-app & \makecell{Unseen\\apps} & Overall & & & & \\
\midrule
\rowcolor{t2wRowGray}\multicolumn{10}{@{}l}{\emph{Proprietary models}} \\
Gemini 3.1 Pro & 75.1 / 82.2 & 69.0 / 76.2 & 58.0 / 67.0 & 65.7 / 71.9 & 68.9 / 76.1 & 31.3 / 44.6 & -- & -- & -- \\
GPT-5.5 & 74.5 / 82.5 & 68.3 / 76.8 & 57.3 / 66.3 & 66.1 / 73.8 & 68.4 / 76.7 & -- & -- & 62.4 & -- \\
Qwen3.7-Plus & \textbf{76.3} / \textbf{83.4} & \textbf{70.6} / \textbf{78.4} & \textbf{59.0} / \textbf{67.8} & \textbf{68.8} / \textbf{76.4} & \textbf{70.6} / \textbf{78.3} & 32.8 / \textbf{46.0} & \textbf{81.0} & 62.3 & \textbf{79.8} \\
\rowcolor{t2wRowGray}\multicolumn{10}{@{}l}{\emph{Open-source GUI agents}} \\
UI-TARS-1.5-7B & 54.4 / 66.0 & 45.2 / 58.5 & 30.3 / 44.2 & 39.6 / 50.8 & 45.1 / 57.7 & 13.8 / 26.3 & 30.0 & -- & -- \\
MAI-UI-8B & 66.3 / 76.9 & 58.1 / 69.8 & 45.0 / 57.5 & 51.9 / 62.7 & 57.8 / 69.2 & 21.5 / -- & 70.7 & 27.5 & -- \\
UI-Venus-2-9B & 64.2 / 75.0 & 56.8 / 67.5 & 44.7 / 56.0 & 52.8 / 61.4 & 56.8 / 67.2 & \textbf{52.7} / -- & 80.2 & \textbf{65.8} & -- \\
\rowcolor{t2wRowGray}\multicolumn{10}{@{}l}{\emph{Controlled environment training (Qwen3.5-9B)}} \\
Base policy & 49.2 / 63.5 & 40.2 / 55.1 & 26.0 / 41.3 & 39.2 / 52.6 & 40.9 / 55.4 & 9.0 / 19.7 & 57.8 & 18.5 & 47.2 \\
Stage-1 policy & 57.0 / 72.6 & 48.6 / 65.1 & 35.3 / 52.0 & 44.3 / 58.9 & 48.7 / 64.6 & 15.8 / 28.4 & 60.1 & 21.9 & 51.9 \\
\textbf{Ours} & \textbf{68.4 / 78.9} & \textbf{62.4 / 71.8} & \textbf{50.7 / 60.5} & \textbf{52.8 / 63.7} & \textbf{61.1 / 71.1} & \textbf{25.2 / 40.7} & \textbf{63.2} & \textbf{27.4} & \textbf{58.3} \\
\bottomrule
\end{tabular*}
\end{table}

\paragraph{Setup and metrics.}
Trace2World-Bench has 1,044 tasks: 868 on training apps (256 entity replacements, 512 held-out
workflows, 100 evaluation-only two-app returns) and 176 on unseen apps; none enters training, task
generation, or checkpoint selection. Action caps are $H=30/45/60$ for source-derived, augmented, and
two-app tasks. We report terminal SR and final-state PR over three attempts per task and three
training seeds; external suites are evaluation-only (Appendices~\ref{app:portfolio}
and~\ref{app:held-out-protocol}).

\begin{wraptable}[13]{R}{0.50\textwidth}
\centering
\caption{\textbf{Real-device transfer on 100 paired tasks.} Gains over base (pp; three-seed mean $\pm$ SD); real/sim is gain retention, and Sim-only pass is simulation-only success.}
\label{tab:transfer-main}
\small
\setlength{\tabcolsep}{0.7pt}
\begin{tabular*}{\linewidth}{@{\extracolsep{\fill}}lccc@{}}
\toprule
& \multicolumn{2}{c}{Matched stage-1 source} & Stage 3 \\
\cmidrule(lr){2-3}\cmidrule(l){4-4}
Metric & \makecell{Self-\\validation} & \makecell{Source-trace\\validation} & Ours \\
\midrule
Real gain & 9.7$\pm$2.1 & 15.7$\pm$1.5 & 16.3$\pm$2.1 \\
Sim.\ gain & 17.7$\pm$1.5 & 16.7$\pm$2.1 & 18.7$\pm$1.5 \\
Real/sim gain & 54.7\% & 94.0\% & 87.5\% \\
Sim-only pass & 18.3\% & 10.0\% & 10.7\% \\
\bottomrule
\end{tabular*}
\end{wraptable}
\textbf{Training transfers beyond reconstructed environments.}
Our final policy gains 20.2 pp SR and 15.7 pp PR over the fixed-backbone base on Trace2World-Bench
(Table~\ref{tab:downstream-main}), and the same 52-iteration run improves SR on all four external
suites: MobileGym by 16.2 pp, AndroidWorld by 5.4 pp, MobileWorld by 8.9 pp, and AndroidDaily by
11.1 pp (Appendix~\ref{app:external-performance}). On the 176 tasks from apps excluded from policy
training it gains 13.6 pp, against 21.6 pp on training-app tasks of different difficulty
(Appendix~\ref{app:application-holdout}).

\textbf{Gains also reach compositions that no training task contains.} All 2,048 optimization tasks
are single-app, yet the largest slice gain is on the 100 evaluation-only two-app tasks: 24.7 pp SR
(26.0 to 50.7), 15.4 pp of it from the two learner-guided stages. The share of failures carrying a
handoff tag is unchanged (32.4 versus 33.1\%), so competence learned in one validated world carries
across the return dependency rather than only handoff behavior
(Table~\ref{tab:downstream-main}, Appendix~\ref{app:horizon-error-analysis}).

\textbf{Training halves hard loops; remaining failures are not budget-limited.} Across 3,132
Trace2World-Bench attempts, hard loops fall from 16.1\% under the base policy to 8.5\% under our final
policy and false completions from 12.7\% to 10.2\%. Of the 613 final-policy failures reaching their
horizon, 511 make no new checkpoint in their last ten actions and a 15-action extension rescues only
18: the step budget is not the binding constraint (Appendix~\ref{app:horizon-error-analysis}).

\textbf{The final policy keeps most of its simulated gain on real devices.} It retains 87.5\%
of its simulated gain and lifts sim/device agreement from 81.3 to 85.0\%
(Table~\ref{tab:transfer-main}, Appendix~\ref{app:sim2real-detail}).

\Needspace{8\baselineskip}
\section{Conclusion}
\label{sec:conclusion}

Trace2World separates two problems that agentic reconstruction conflates: global reasoning
reconstructs app behavior once from partial traces, leaving unresolved effects non-executable,
while later source mismatches become local repairs over one world that independent replay accepts.
This contract reaches 91.1\% workflow fidelity versus 59.8\% for the strongest existing route and
keeps 94.0\% of simulated gains on real devices versus 54.7\% under self-validation.
Reusing validated worlds amortizes reconstruction and source validation across learner-guided tasks.
The bottleneck
for scalable GUI RL is no longer producing runnable mocks but reconstructing app worlds whose state
and effects are fixed by evidence, checked on the source workflows, and trusted without human QA.

\section*{Ethics Statement}

We use source traces collected by consenting operators from accounts they control, with consent covering
research use and public release. We release the source recordings and derived trace records used for
construction together with the generated environments and evaluation artifacts
(Appendix~\ref{app:resource-contract}). Before public release, every recording is reviewed for credentials
and session material, personal identifiers, private communications, precise locations, payment or order
information, and third-party personal data; affected regions are redacted when this does not alter the
research-relevant interaction evidence, and a trace is withheld and listed when safe release is not possible.

The released material provides no production credentials, privileged backend access, or automation
authority over the original services; environments are isolated browser reimplementations with synthetic
content that use application names and examples only to identify the systems studied, without implying
affiliation or endorsement. Real-device evaluation begins
from restorable states and excludes irreversible actions
(Appendix~\ref{app:real-device-protocol}). Recordings, included visual resources, and environments
are additionally reviewed for applicable service terms, redistribution rights, and trade-dress concerns; an
artifact that fails review is withheld and reported. The artifacts are intended for controlled GUI-agent
research; use against live services requires account-holder authorization and compliance with applicable
rules.

\needspace{9\baselineskip}
\section*{AI Use Statement}
Generative models are components of Trace2World, supporting trace interpretation and semantic induction,
environment implementation and repair, task generation and refinement, and rollout-based screening and
evaluation (Section~\ref{sec:exp-setup}). In preparing this work, generative AI tools also assisted code
development, figure preparation, manuscript drafting and revision, and review of methodological and
experimental descriptions. Generated environments and tasks are assessed under the shared evaluator
boundaries in Appendix~\ref{app:evaluation} and the study-specific protocols in
Appendix~\ref{app:extended-results}; human assessment covers the sampled evaluation packets in
Appendix~\ref{app:human-routes}. The authors retain responsibility for the study design, the
verification and interpretation of results, and the final manuscript and artifacts.

\TTwoAcknowledgments 

\needspace{7\baselineskip}
\bibliography{iclr2026_conference}
\bibliographystyle{iclr2027_conference}

\appendix
\makeatletter
\@addtoreset{table}{section}
\@addtoreset{figure}{section}
\makeatother
\renewcommand{\thetable}{\thesection\arabic{table}}
\renewcommand{\thefigure}{\thesection\arabic{figure}}
\clearpage
\phantomsection
\thispagestyle{plain}
\begin{center}
  {\Large\bfseries Appendix}\\[0.45em]
  {\large\bfseries Table of Contents}
\end{center}
\vspace{0.7em}
\label{sec:appendix-contents}

\begingroup
\footnotesize
\setlength{\parindent}{0pt}
\setlength{\tabcolsep}{2pt}
\renewcommand{\arraystretch}{0.95}
\newcommand{\TTwoAppendixEntry}[5]{\csname TTwoAppendix#1Row\endcsname{#2}{#3}{#4}{#5}}
\newcommand{\TTwoAppendixsectionRow}[4]{%
  \addlinespace[0.18em]
  \hyperlink{#1}{\textcolor{black}{\bfseries #2}} &
  \hyperlink{#1}{\textcolor{black}{\bfseries #3}} &
  \hyperlink{#1}{\textcolor{black}{\bfseries #4}}\\}
\newcommand{\TTwoAppendixsubsectionRow}[4]{%
  \hyperlink{#1}{\textcolor{black}{\hspace*{0.75em}#2}} &
  \hyperlink{#1}{\textcolor{black}{#3}} &
  \hyperlink{#1}{\textcolor{black}{#4}}\\}
\newcommand{\TTwoRenderAppendixGroup}[3]{%
  \begin{tcolorbox}[
    enhanced,
    width=\textwidth,
    colback=white,
    colframe=#3,
    boxrule=1pt,
    arc=2.5pt,
    outer arc=2.5pt,
    boxsep=0pt,
    left=5pt,
    right=5pt,
    top=7pt,
    bottom=4pt,
    before skip=0.25em,
    after skip=0.7em,
    title={\MakeUppercase{#1}},
    fonttitle=\scriptsize\sffamily\bfseries,
    coltitle=black,
    attach boxed title to top left={xshift=0.115\linewidth,yshift=-1.2mm},
    boxed title style={
      colback=white,
      colframe=white,
      boxrule=0pt,
      boxsep=0pt,
      left=2pt,
      right=2pt,
      top=0pt,
      bottom=0pt,
    },
    ]
    \begin{tabularx}{\linewidth}{@{}>{\raggedright\arraybackslash}p{0.11\linewidth}
      >{\raggedright\arraybackslash}X
      >{\raggedleft\arraybackslash}p{0.06\linewidth}@{}}
    #2
    \end{tabularx}
  \end{tcolorbox}}
\TTwoRenderAppendixGroup{Methodology}{\TTwoAppendixMethodologyEntries}{t2wAppMethod}
\TTwoRenderAppendixGroup{Experiments}{\TTwoAppendixExperimentEntries}{t2wAppExp}
\TTwoRenderAppendixGroup{Cases and Resources}{\TTwoAppendixResourceEntries}{t2wAppRes}
\endgroup
\vfill
\clearpage
\TTwoBeginAppendixNavigation
\appsection{environment}{app:reference-evidence}{Semantic Reconciliation Across Traces}

\label{app:reference}
\label{app:induction}
\label{app:induction-details}

This appendix formalizes the reconciliation summarized in Section~\ref{sec:anchored-construction}, from
source-trace preprocessing through bounded-context scheduling. It opens by placing that contract among
existing environment builders.

\paragraph{Comparison scope.}
Table~\ref{tab:env-landscape} expands the main-text positioning along five axes: where app behavior
comes from, what is represented before coding, how cross-trace effects are handled, how behavior is
implemented, and which evidence is used for acceptance.
It preserves each system's native target rather than treating the absence of an external-app fidelity
objective as a failed check.
Appendix~\ref{app:baselines} defines the builders reimplemented under one experimental contract.

\begin{table}[H]
\centering
\caption{\textbf{Semantics and acceptance across GUI environments.} Columns group systems by the
source of app behavior; rows distinguish input, representation, cross-trace semantic rules,
implementation, and native acceptance evidence. Task checks in authored or synthesized worlds do not
certify fidelity to an external app.
PhoneWorld uses agentic construction with representative human QA; Trace2World fixes source-linked semantics
before coding and replays each recorded workflow. CUA-Gym targets desktop and web software; the other
systems target mobile applications. Counts use native units;
paired denotes same-task simulation/device evaluation and unpaired a separate real benchmark
(Appendix~\ref{app:transfer-results}). AndroidWorld and MobileGym memory figures are reported by
MobileGym \citep{wu2026mobilegym}; Trace2World's is marginal PSS at 16 concurrent episodes across
the 50-app portfolio (Appendix~\ref{app:runtime-cost}). These are contextual cross-host figures;
dashes mean no comparable figure here.}
\label{tab:env-landscape}
\footnotesize
\setlength{\tabcolsep}{1.8pt}
\renewcommand{\arraystretch}{1.00}
\rowcolors{3}{t2wRowGray}{white}
\begin{tabularx}{\linewidth}{@{}>{\raggedright\arraybackslash\bfseries}p{0.17\linewidth}*{6}{>{\raggedright\arraybackslash}X}@{}}
\toprule
& \multicolumn{2}{c}{Real / designed} & \multicolumn{2}{c}{Topic / task-local} & \multicolumn{2}{c}{Trace-grounded} \\
\cmidrule(lr){2-3}\cmidrule(lr){4-5}\cmidrule(lr){6-7}
Axis & AndroidWorld & MobileGym & CUA-Gym & GUI-GENESIS & PhoneWorld & \textbf{Trace2World} \\
\midrule
Construction input & real Android apps and task definitions & authored simulator and tasks & topic request & one traced task & real trajectories and screenshots & all app traces \\
Behavior representation & real app itself & authored state and actions & topic-based mock & task-local site & screen graph, page PRDs, mutable state & \textbf{shared IDs, state/data, guards and effects} \\
Cross-trace state/effect rule & native app & authored & topic specification & task-local & no explicit source-linked rule & \textbf{guarded or unresolved} \\
Implementation path & native apps & hand-authored simulator & Plan--Dev--Web agents & code agent & PRDs $\to$ coding agent & \textbf{model $\to$ API compile $\to$ agentic repair} \\
Acceptance evidence & real-app task oracle & authored task checks & generated verifier and filter & code-native reward and golden path & smoke tests and human comparison & \textbf{replay of every source workflow} \\
\addlinespace[1pt]
Memory per episode & $\sim$4.5\,GB & $\sim$400\,MB & -- & -- & -- & 218\,MiB \\
\addlinespace[1pt]
How training tasks are produced & fixed authored set & parameterized templates & co-generated with environment & one task per environment & generated from content, state schema, and PRDs & \textbf{learner-guided within fixed support} \\
\addlinespace[1pt]
Sim-to-Real evidence & n/a; emulator is target & paired 59-task signal subset & held-out benchmark transfer & held-out real-app tasks (unpaired) & real-benchmark gains (unpaired) & paired 100-task balanced panel \\
Scale (native units) & 20 apps / 116 templates & 28 apps / 416 templates & 110 environments / 32,112 tuples & one site per task & 34 apps / 16 domains & 50 apps / 1,280 seed $+$ 768 learner-guided tasks \\
\bottomrule
\end{tabularx}
\rowcolors{1}{white}{white}
\end{table}

\subsection{Source-Trace Preprocessing}
\label{app:trace-preprocessing}

Reconciliation consumes recorded traces, not raw recordings. Preprocessing selects the traces that
enter construction, gives every recorded step a typed disposition and a semantic description, and
extracts the visual resources those steps contain. It produces only trace- and step-linked local evidence:
canonical page identity, State fields, guards, and verifier predicates are decided later by
reconciliation and are never authored here.

\begin{figure}[H]
\centering
\includegraphics[width=0.52\linewidth]{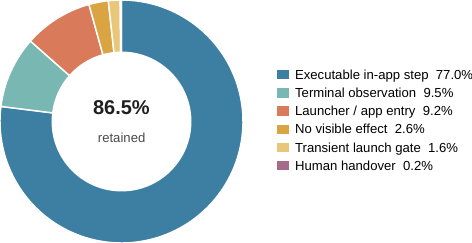}
\caption{\textbf{Disposition of 16,472 recorded frames across the 1,570-trace corpus.}
Executable steps (77.0\%) and terminal observations (9.5\%) are retained as induction evidence;
launcher/app-entry, transient-gate, no-effect, and human-handover frames are removed before semantic
calls (13.5\% combined).}
\label{fig:trace-disposition}
\end{figure}

Preprocessing covers the fixed 50-application construction corpus: 19 multi-service applications
contribute 950 traces and 31 single-service applications contribute 620, for 1,570 traces in total.
Figure~\ref{fig:trace-disposition} visualizes step dispositions. Table~\ref{tab:trace-selection-projection}
then jointly reports trace selection and the visual-evidence volume it produces. Resource extraction is
summarized in text.

\begin{figure}[H]
\centering
\includegraphics[width=\linewidth]{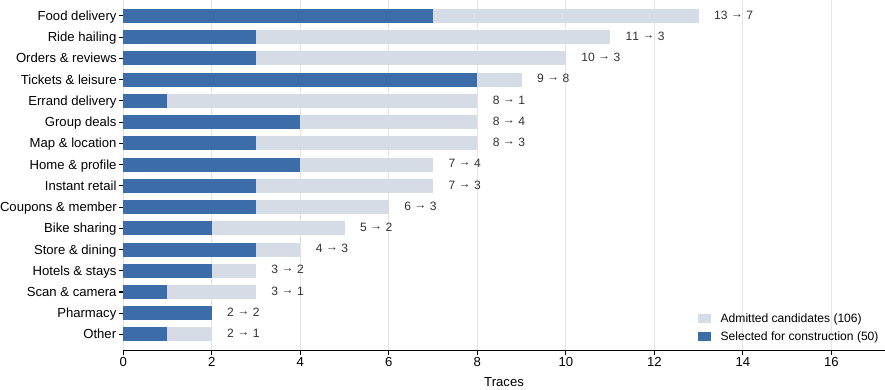}
\caption{\textbf{Selection halves volume while keeping workflow coverage (Meituan).}
Light bars count admitted candidate recordings and blue overlays count the traces selected for
construction; right-side counts read candidate $\rightarrow$ selected. All 16 occupied buckets retain
at least one selected trace. Buckets are a post-hoc keyword grouping of the recorded instruction text,
not a registered product taxonomy.}
\label{fig:meituan-trace-coverage}
\end{figure}

\paragraph{Trace selection and diversity control.}
Operators record complete workflows on one real application from a frozen app-level task catalog that
varies goals, entry routes, entity bindings, launch states, and supported compositions. A recording is
admitted only when visible terminal evidence or an independent receipt establishes its outcome at the
recorded application revision; a recorder flag or a generated description is insufficient, and
agent-generated exploration is excluded. Admission certifies the outcome, not the necessity of each
intermediate action, so retries and corrections stay in the evidence. Admitted recordings exceed the
per-application budget, and selection keeps diversity rather than volume: candidates must contain at
least four in-app interaction steps once launch, terminal, and handover rows are discounted;
task instructions are embedded with the bge-large-zh-v1.5 encoder \citep{xiao2023cpack} and clustered,
and traces are drawn to cover clusters while avoiding duplicate action-signature prefixes. Across the
corpus, this process retains the fixed
1,570-trace construction set from 6,541 admitted recordings, a 24.0\% retention rate
(Table~\ref{tab:trace-selection-projection}). Figure~\ref{fig:meituan-trace-coverage} instantiates the
diversity side of that trade on Meituan: halving the pool from 106 admitted candidates to 50 selected
traces leaves every occupied workflow bucket populated, and the selected traces carry 50 distinct
action signatures with no repeated three-action prefix.

\paragraph{Step disposition and semantic description.}
Each recorded step stores a device event, one screenshot, and a marked screenshot in which pointer
geometry, typed text, and drag paths are drawn onto the observation. One bounded API pass per trace
(Appendix~\ref{app:preprocessing-prompt}) reads the raw action payload together with the marked
screenshot and returns, for every step, a
verdict in \{\textsc{keep}, \textsc{drop}, \textsc{observe}\}, a semantic description naming the
visible target rather than a coordinate, and a reason; at trace level it returns a normalized
instruction and a terminal contract. A deterministic pass then reconciles those verdicts with the raw
event: launcher and app-entry rows become \textsc{drop}, a recorded completion becomes
\textsc{observe} so its observation is retained as terminal evidence while its action stays
non-executable, and privacy or login handovers are excluded from behavior. Every raw row keeps exactly
one typed disposition with its reason; nothing is deleted, so a later stage can still audit why a step
left the executable set. These descriptions remain step-local hypotheses subject to
cross-trace reconciliation and cannot override the raw event.

\paragraph{Corpus-wide step statistics.}
One recorded step carries one screenshot, so the disposition count is also the screenshot budget.
The full construction corpus contains 16,472 recorded frames
(Figure~\ref{fig:trace-disposition}). Executable in-app steps account
for 12,680 frames (77.0\%), while 1,570 terminal observations (9.5\%) remain as evidence without
an executable terminal action. The four pre-semantic removal categories total 2,222 frames
(13.5\%). Their composition shows that launch boundaries dominate the reduction: launcher or app-entry
frames contribute 9.2\%, compared with 1.6\% transient gates, 2.6\% no-effect interactions, and
0.2\% human handovers.

\paragraph{Visual-evidence selection.}
Visual-evidence selection (Appendix~\ref{app:evidence-projection}) chooses a page-level visual basis
from retained observations after page obligations have been projected; it does not change step
dispositions or decide app semantics. Across the corpus, this basis contains 9,918 selected screenshots
(Table~\ref{tab:trace-selection-projection}), 39.8\% fewer than the recorded frames and 30.4\% fewer
than the retained disposition inputs.

\begin{table}[H]
\centering
\caption{\textbf{Source-trace and visual-evidence selection across the full construction corpus.}
Candidate recordings form the admissible pool, selected traces constitute the fixed 1,570-trace
construction set, and selected screenshots form the page-level visual basis. Retention is selected
traces divided by candidate recordings within each collection-budget stratum.}
\label{tab:trace-selection-projection}
\small
\setlength{\tabcolsep}{3pt}
\begin{tabular*}{\linewidth}{@{\extracolsep{\fill}}lcccccc@{}}
\toprule
& & \multicolumn{3}{c}{Trace selection} & \multicolumn{2}{c}{Visual evidence} \\
\cmidrule(lr){3-5}\cmidrule(lr){6-7}
Stratum & Apps & \makecell{Candidate\\recordings} & \makecell{Selected\\traces} & Retention & \makecell{Selected\\screenshots} & \makecell{Screens /\\trace} \\
\midrule
Multi-service & 19 & 2,005 & 950 & 47.4\% & 5,206 & 5.48 \\
Single-service & 31 & 4,536 & 620 & 13.7\% & 4,712 & 7.60 \\
\midrule
Full corpus & 50 & 6,541 & 1,570 & 24.0\% & 9,918 & 6.32 \\
\bottomrule
\end{tabular*}
\end{table}

\paragraph{Visual-resource extraction.}
Generated pages reuse the icons, product images, banners, and background art displayed in the selected
evidence. The MinerU API \citep{wang2024mineru} extracts and classifies these resources as icon, image,
banner, or background. Each record is deduplicated by content digest, bound to its source step and
observation, and labeled per page so the builder can address resources by role. MinerU output is local
evidence: it supplies what a surface displayed,
never which canonical page owns it or what an action does. A resource whose owning relation fails
source support is retained in $\Sigma$ and not rendered.
\FloatBarrier

\subsection{Behavior Support and Implementation Scope}
\label{app:trace-support}

$\TraceSupport$ is the set of Structure, State, and Data relations supported by the frozen traces,
not the complete commercial application. Unsupported or irreducibly ambiguous behavior stays outside
that set. A rendered region without an app-specification binding cannot trigger an action, route, or
transaction. At round $r$, the implemented behavior scope
$\Implemented^{(r)}\subseteq\TraceSupport$ records the supported behavior currently available for
training. The support set determines what may be implemented; the implemented scope determines what
is available. A held-out task may compose familiar
supported primitives without becoming construction evidence.
Within an episode, only components bound to the current task within that scope can accept interactions.
A visible component outside that boundary returns \texttt{NOT\_INTERACTIVE} when acted on and cannot
change business state; it is not an implied cross-task affordance.

\subsection{Semantic Reconciliation and Support Validation}
\label{app:induction-algorithm}
\label{app:candidate-restoration}
\label{app:factorization}

Algorithm~\ref{alg:scaffold-induction} orders the four semantic operations; Appendix~\ref{app:bounded-execution}
specifies the shared context-bounded schedule used to execute its reconciliation and factorization phases.
Appendix~\ref{app:reconciliation-prompt} condenses their semantic decision rule.
\textsc{Restore} pools transition candidates from pre/post observations, marked geometry, action payloads,
and lifecycle context, retaining links to the originating trace and step alongside alternative host/effect hypotheses, including no
effect. \textsc{Reconcile} distinguishes structural owners, runtime variants, and presentations of the
same entity; \textsc{Factorize} expresses the reconciled candidates as Structure--State--Data relations;
\textsc{Supported} retains only relations that satisfy the source-support contract below.

\begin{algorithm}[h]
\caption{App-specification reconciliation}
\label{alg:scaffold-induction}
\algtext*{EndFor}
\algtext*{EndIf}
\begin{algorithmic}[1]
\Require real app traces $\mathcal R$, phase budgets $\{B_\phi\}$
\Ensure reconciled app specification $S$ of supported Structure--State--Data relations with evidence map $\Sigma$
\State $E\gets\Call{Restore}{\mathcal R}$
\State $O\gets\Call{HierarchicalPhase}{\textsc{Reconcile},E,B_{\mathrm{owner}}}$
\State $C\gets\Call{HierarchicalPhase}{\textsc{Factorize},O,B_{\mathrm{semantic}}}$
\State $S\gets\varnothing$
\ForAll{$c\in C$}
  \If{\Call{Supported}{$c$}}
    \State $S\gets S\cup\{c\}$
  \Else
    \State \Call{RecordFailure}{$c$}
  \EndIf
\EndFor
\State \Return $S$
\end{algorithmic}
\end{algorithm}

Each candidate transition has an execution key and a proposed effect:
\begin{equation}
\begin{aligned}
k&=(\mathrm{page},\mathrm{host},\mathrm{action},\mathrm{guard}),\\
k&\mapsto(\mathrm{reads},\mathrm{writes},\mathrm{destination},
\mathrm{postcondition},\mathrm{projection}).
\end{aligned}
\label{eq:execution-key}
\end{equation}
For each execution key, the supported consequence is compatible
across all supporting observations.
Stable shell and route continuity, entry/exit conditions, lifecycle, and entity lineage determine
canonical identity. Conflicting consequences remain unresolved until evidence supplies a discriminator;
neither filenames nor majority agreement can settle semantic ownership. Table~\ref{tab:ssd-factorization}
defines the residual factorization.

\begin{table}[H]
\centering
\caption{\textbf{Structure--State--Data factorization.} Rows distinguish stable interaction topology,
mutable task/action-conditioned fields, and reusable domain entities after owner reconciliation. The final
column states the supported app-specification relation, preventing task-local variation from becoming duplicated pages
or data.}
\label{tab:ssd-factorization}
\small
\setlength{\tabcolsep}{4pt}
\renewcommand{\arraystretch}{1.08}
\begin{tabularx}{\linewidth}{@{}p{0.13\linewidth}p{0.39\linewidth}X@{}}
\toprule
Layer & Evidence rule & App-specification commitment \\
\midrule
Structure & Interaction topology that remains stable across tasks and runtime variation. &
Canonical pages/surfaces, routes, containment, visible hosts, overlays, and bindings. \\
State & Launch-, guard-, or action-conditioned mutable variation owned by a canonical surface or entity. &
Initialization deltas, fields, guards, reads/writes, reducers, projections, terminal conditions, and reset. \\
Data & Identity and attributes reused across pages or tasks, including stable resource bindings. &
Canonical entities, attributes, relations, and resources rather than task-local aliases. \\
\bottomrule
\end{tabularx}
\end{table}

\paragraph{Factorization.}
\textsc{Factorize} operates over reconciled observations of complete traces, so every step of a path
is interpreted with its canonical page and entity identities. It proposes typed State fields with their
owning surface or entity, guards, and one consequence per execution key. Deterministic validation
requires every field to name a canonical owner, every guard to reference an observed field and its
evidence steps, and every consequence to cite the source transitions that support it. The phase also
records, per canonical page and trace, field values at first entry and values written by later transitions;
initialization (Appendix~\ref{app:task-init}) uses this separation to distinguish launch values from
runtime writes. Relations merge only on canonical identities, and incompatible supported consequences
remain unresolved rather than being selected by frequency.

\paragraph{Support-validation contract.}
\label{app:admission-gate}
\textsc{Supported} requires qualified source support, total local-to-canonical mappings,
compatible effects or an explicit non-executable contradiction, compiler-complete fields and
dependencies, and coverage of required evidence. \textsc{RecordFailure} retains an unresolved or
rejected relation and its failed gate; unsupported or incomplete relations are not compiled.
For example, a menu and cart may expose one item under different aliases:
item identity is Data, quantity is State, and the two hosts remain separate Structure surfaces.

\subsection{Task-Conditioned Initialization and Reset}
\label{app:task-init}

The supported State schema and transition relations fix the dynamics of one shared environment;
initialization fixes where each task starts in it. A browser environment keeps one canonical state $x$
over the fields supported as State; pages are views of $x$, never stores. An episode of task $q$ evolves as
\begin{align}
x_0^{(q)}&=\operatorname{Reset}\bigl(I_{\mathrm{env}},\delta_q^{\mathrm{launch}}\bigr),\qquad
o_t=\operatorname{View}_{p_t}(x_t),\notag\\
x_{t+1}&=\begin{cases}T(x_t,a_t)&\text{if the guard of }(p_t,h_t,a_t)\text{ holds at }x_t,\\
x_t&\text{otherwise,}\end{cases}
\label{eq:episode-state}
\end{align}
where $I_{\mathrm{env}}$ assigns every mutable field its default, $T$ applies the supported consequence
for the execution key (Eq.~\ref{eq:execution-key}), and $\operatorname{View}_p$ renders the fields that
surface $p$ projects. $\operatorname{View}_p$ acts on state at run time; the compiler's
$\operatorname{Project}_p$ (Appendix~\ref{app:compilation}) acts on the app specification.
The equation abstracts the GUI as follows. The current page $p_t$, open overlay, focus, scroll offset,
and pending selection are view variables kept beside $x_t$: a legal action may change them without
mutating $x_t$ (a route, tab, or overlay transition), and the consequence $T$ may set the destination
page. A coordinate action is resolved to a host $h_t$ by hit-testing the rendered view at the moment of
the action; a coordinate that hits no host is an unbound attempt and fails closed. Typing targets the
focused host. The runtime waits up to 5\,s for a declared navigation target and 50\,ms after pointer
or gesture actions; screenshot capture waits for fonts and an animation frame. Later asynchronous
updates can appear in a subsequent observation, and the public \texttt{wait} action can advance them
(Appendix~\ref{app:agent-interface}).

\paragraph{Dynamic interaction mask.}
\label{app:interaction-mask}
The compiler binds route and effect authority only to admitted hosts and removes navigation authority when
a referenced target page was not materialized. For a coordinate $(u,v)$, an actionability preflight obtains
the ordered DOM stack from \texttt{elementsFromPoint}, identifies the active interaction layer (the page or the topmost
open modal, drawer, or sheet), and examines only visible, enabled candidates. A deeper authorized element
cannot be selected through an opaque or pointer-intercepting element; a topmost descendant or activation
label may resolve to its owning host. The candidate must then match the permitted event type. If none
passes, the harness returns \texttt{NOT\_INTERACTIVE} without issuing the browser pointer event. If one
passes, only that host receives the event, and the transition runtime checks its guard before route,
reducer, or transaction dispatch. Recomputing this predicate after layout, scroll, and overlay changes
keeps unsupported chrome visually present while preventing it from acquiring behavior through browser
defaults or hidden hotspots.

\begin{figure}[!t]
\centering
\includegraphics[width=0.98\linewidth]{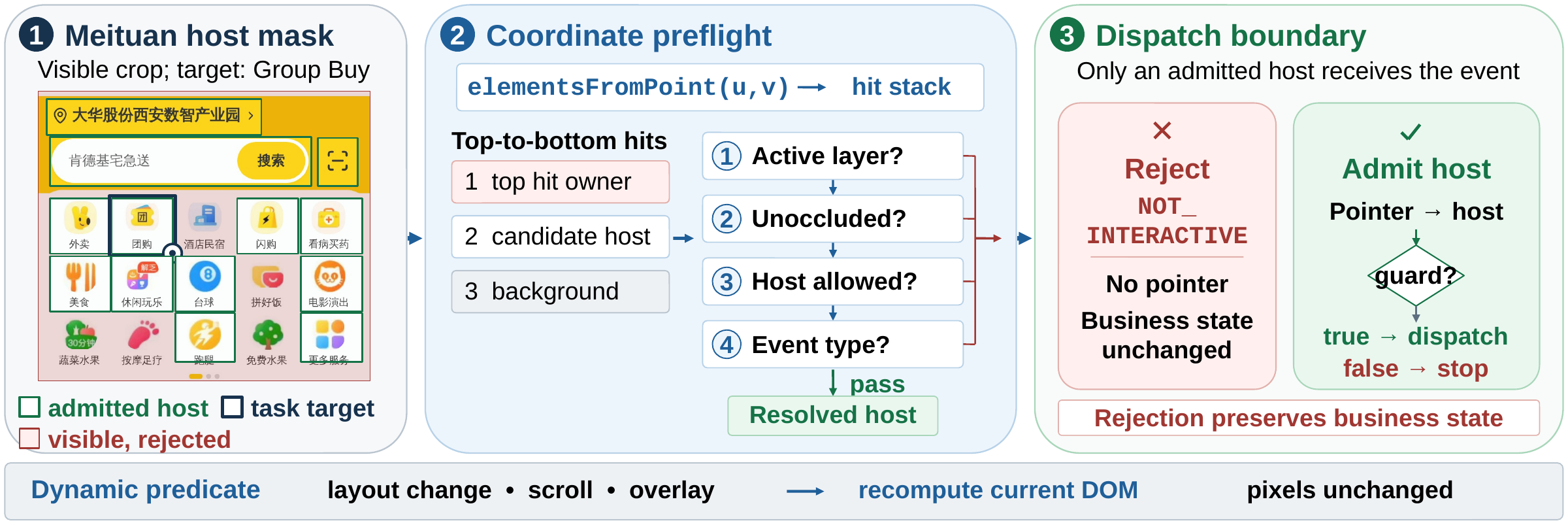}
\caption{\textbf{Dynamic interaction authority on the generated Meituan Home.}
Green marks every admitted host in the visible crop; blue singles out Group Buy as the example task
target, and red marks unbound regions for illustration rather than a runtime overlay. The current DOM
stack admits only visible, unobscured hosts on the active layer; other coordinates return
\texttt{NOT\_INTERACTIVE} before event dispatch.}
\label{fig:interaction-mask}
\end{figure}
\FloatBarrier

\paragraph{Launch values.}
Let $v_i(f)$ be the value of State field $f$ when trace $i$ first enters canonical page $p$ without
having written $f$; an earlier write makes the value runtime state and removes the trace from the launch
set of $(p,f)$. Traces with $v_i(f)\ne v_j(f)$ raise a launch conflict on $(p,f)$, resolved by the first
applicable rule:
\begin{enumerate}[leftmargin=*,nosep]
\item \textbf{Launch field.} If every differing value is visible in its trace's launch observation or is
a premise of the recorded task (account or persona, region, target entity, or pre-existing data such as
a saved listing, an unrated order, or a contact), $f$ is a launch field: $I_{\mathrm{env}}(f)$ is the
neutral value when some trace shows one (empty cart, default filter, no selection) and otherwise the
value most traces share, and each trace whose value differs stores it as
$\delta_{q_i}^{\mathrm{launch}}(f)=v_i(f)$. Page $p$ keeps one identity.
\item \textbf{Conflict to guard.} If $v_i(f)=g(c_i)$ for a canonical field $c$ and a supported map $g$
(price by region, button label by lifecycle, detail fields by selected entity), $f$ is projected from
$c$, the affected execution keys gain a guard on $c$, and the conflict moves to $(p,c)$, which the same
rules resolve.
\item \textbf{Unresolved.} Otherwise the conflict is recorded on $(p,f)$: neither value is compiled, no
value is keyed by task identifier, and tasks that depend on $f$ stay outside $\Implemented$ until
evidence arrives.
\end{enumerate}
A direct launch observation precedes an inferred guard. The default in Rule~1 is not a vote: it is
itself an observed value, it affects only tasks that leave $f$ undeclared, and every trace's task
carries its own value, so each source task starts exactly as its recording did. Conflicts between
action consequences under one execution key follow Rule~2 or remain non-executable
(Appendix~\ref{app:factorization}).

\paragraph{Launch support for generated tasks.}
A generated task may declare $\delta_q^{\mathrm{launch}}(f)=v$ only if some trace observed $v$ as a
launch value of $f$ or a supported operation sequence from $I_{\mathrm{env}}$ reaches $v$; the second
case uses the effect values of the operations that write $f$ and the records the read model exposes for
the entity type of $f$. Declared fields must already belong to State: a delta cannot add pages, routes,
entities, or attributes.

\paragraph{Runtime state ownership.}
Each episode runs in a fresh browser context: the state kernel creates a session for the task, applies
$\delta_q^{\mathrm{launch}}$ over $I_{\mathrm{env}}$, seals the initial snapshot, and renders the entry
surface; Reset restores the fixed reset snapshot and clears runtime state, events, transactions, and reward.
The session identifier is the task's storage namespace and appears nowhere in $T$,
$\operatorname{View}_p$, or the reward; only guarded transitions mutate $x$ during an episode,
guard-false attempts fail closed, and page-local caches never define business state. Instrumentation reads
the task identifier only to map checkpoints. The zero-action and reset probes of
Table~\ref{tab:probe-matrix} test this contract directly: a launch must render the declared fields
without progress or reward, and repeated Reset must reproduce $x_0^{(q)}$.

\subsection{Context-Bounded Reconciliation}
\label{app:bounded-execution}

\textsc{Reconcile} and \textsc{Factorize} use the same hierarchical execution schedule. Sharding changes
how evidence is assigned to API calls, while the semantic mappings, support gates, and contradiction
handling of Appendix~\ref{app:factorization} remain invariant.
For phase input $N_\phi$, per-call budget $B_\phi$, $K_\phi=\lceil N_\phi/B_\phi\rceil$ shards,
reducer fan-in $b_\phi$, and maximum partial size $S_\phi$, balanced execution gives
\begin{equation}
C^{\mathrm{peak}}_\phi=O\!\left(\max\{B_\phi,b_\phi S_\phi\}\right),\qquad
N^{\mathrm{calls}}_\phi=O(K_\phi),\qquad
D_\phi=O(\log_{b_\phi}K_\phi).
\label{eq:scaling-complexity}
\end{equation}
In Algorithm~\ref{alg:hierarchical-phase}, \textsc{Shard} forms source-complete groups within budget $B$;
the typed API producer $F$ emits bounded partials. \textsc{Reduce} constructs the next level with fan-in
$b$, preserving source-trace links, local-to-global mappings, contradictions, and uncovered obligations.
\textsc{Validate} returns the sole remaining partial only if coverage, total mappings, source support,
and contradiction preservation hold against input $E$. Any helper failure aborts induction rather than
yielding an empty app specification. With fixed budgets, peak per-call context is constant in trace-pool size;
aggregate calls, tokens, and cost can still grow linearly.

\begin{algorithm}[h]
\caption{Context-bounded hierarchical phase}
\label{alg:hierarchical-phase}
\algtext*{EndWhile}
\algtext*{EndFunction}
\begin{algorithmic}[1]
\Function{HierarchicalPhase}{$F,E,B$}
  \State $Q\gets\Call{Shard}{E,B}$
  \State $Q\gets\{F(q;B):q\in Q\}$
  \While{$|Q|>1$}
    \State $Q\gets\Call{Reduce}{F,Q,b}$
  \EndWhile
  \State \Return \Call{Validate}{$Q,E$}
\EndFunction
\end{algorithmic}
\end{algorithm}

\appsection{package}{app:agentic-closure}{Environment Construction and Validation}

\subsection{Environment Architecture and Export}
\label{app:package-substrate}
\label{app:compilation}

The compiler projects each page specification
$\mathcal C_p=\operatorname{Project}_p(S,\Implemented^{(r)})$ from validated canonical semantics:
page ownership, hosts, guards, reads/writes, destinations, projections, reset, and verifier-visible
consequences. Bounded API compilation combines this specification with its visual evidence set to emit
a DOM-backed HTML/CSS/JavaScript application in Chromium \citep{si2024design2code}. Each canonical route
becomes a static page served under one origin with the packaged state kernel, and every page uses the
same state client. Table~\ref{tab:package-architecture} makes the runtime authority boundaries explicit.

\begin{table}[!t]
\centering
\caption{\textbf{Runtime architecture of an exported application package.} All layers are served under
one origin. ``Owns'' denotes authoritative responsibility; ``Boundary'' states what that layer cannot
decide.}
\label{tab:package-architecture}
\footnotesize
\setlength{\tabcolsep}{3pt}
\renewcommand{\arraystretch}{1.12}
\begin{tabularx}{\linewidth}{@{}>{\raggedright\arraybackslash}p{0.14\linewidth}
  >{\raggedright\arraybackslash}p{0.21\linewidth}Y
  >{\raggedright\arraybackslash}p{0.24\linewidth}@{}}
\toprule
Layer & Packaged form & Owns & Boundary \\
\midrule
Presentation & Static HTML/CSS/JavaScript; one page per canonical route & Renders
$\operatorname{View}_p(x)$ and maps pointer or keyboard input to registered reads and actions &
Disposable UI state only; no business state or task-specific product branches \\
State interface & One shared JavaScript client and typed same-origin API & Stable entity and affordance
identity; registered reads, queries, and guarded transactions & No direct database or browser-storage
writes; a failed guard cannot mutate state \\
State kernel & Packaged state module or same-origin service; immutable world plus one writable overlay per episode &
Canonical entities, relations, fields, atomic transitions, revisions, diffs, and receipts & The world is
immutable; overlays are session-isolated \\
Episode control & Private lifecycle and initial-state template & Creates, initializes, seals, resets,
finalizes, and deletes an episode & Unreachable from pages; reset rebuilds the overlay and clears its
ledger \\
Evaluation & Task/evaluator manifests and a read-only state view & Computes checkpoints and reward from
finalized snapshots and diffs; receipts supplement order or count & Cannot mutate state or branch product
behavior on task identity \\
\bottomrule
\end{tabularx}
\end{table}
\FloatBarrier

Packages realize the kernel either inside the page runtime over packaged world data or as a separate
same-origin service over a SQLite world; both expose the same client and boundaries, and the service form
additionally separates the kernel from page code by a process boundary. The standalone package also carries
task definitions, evaluators, resources, and manifests. The app specification, visual evidence, source traces,
and evidence map $\Sigma$ stay outside the execution path as audit sidecars.

\paragraph{Visual-evidence selection.}
\label{app:evidence-projection}
For each materialized page $p$, the page-environment projection and source-linked visual signatures
define obligations $U_p$ for its shell, action hosts, selected states, overlays, and appearance
variants. Each retained source screenshot covers only the obligations visible in that observation;
render role and lifecycle distinguish the canonical initial layout from later states. The selector
solves a minimum set cover over these signatures, retains a canonical-layout observation when the
initial view depends on one, and records the complete candidate universe and uncovered obligations.
When coverage cannot be established, it keeps a conservative visual basis but marks the selection
uncertified; a missing image never becomes evidence for a semantic relation
(Algorithm~\ref{alg:visual-evidence-selection}).

\begin{algorithm}[h]
\caption{Page-level visual-evidence selection}
\label{alg:visual-evidence-selection}
\algtext*{EndIf}
\begin{algorithmic}[1]
\Require retained screenshots $F_p$, page projection $C_p$, source-linked visual signatures $V_p$
\Ensure selected visual basis $B_p$ and coverage record $L_p$
\State $U_p\gets\Call{NormalizeObligations}{C_p,V_p}$
\ForAll{$f\in F_p$}
  \State $r_f\gets\Call{RenderRole}{f}$; $A_f\gets\Call{VisibleCoverage}{f,U_p,r_f}$
\EndFor
\State $W_p\gets U_p\setminus\bigcup_{f\in F_p}A_f$
\If{$U_p\neq\varnothing$ and $W_p=\varnothing$}
  \State $B_p\gets\Call{MinimumCover}{\{A_f:f\in F_p\},U_p}$
\Else
  \State $B_p\gets\Call{ConservativeBasis}{F_p}$; mark coverage uncertified
\EndIf
\If{initial layout requires visual authority and $B_p$ lacks its canonical observation}
  \State add the eligible canonical-layout observation to $B_p$, or record its absence
\EndIf
\State $L_p\gets\Call{RecordCoverage}{F_p,B_p,U_p,\{A_f\},W_p,\text{source steps}}$
\State \Return $(B_p,L_p)$
\end{algorithmic}
\end{algorithm}

Export checks canonical-consumer coverage, identifier and field agreement, build/resource integrity,
independent host/reset entry, and a smoke execution. Compilation consumes page specifications,
but the exported runtime reads only packaged pages, the registered state interface, and episode state;
it does not read the reconciled app specification, source traces, or repair history.
Algorithm~\ref{alg:agentic-closure} then establishes behavioral validity on the exported revision.

\subsection{Source-Trace Validation and Repair}
\label{app:closure-algorithm}
\label{app:contrastive-review}
\label{app:execution-review}
\label{app:repair-invalidation}

\textsc{BuildObligations} derives a per-trace checklist from source-trace evidence. \textsc{Execute} starts
each affected trace at native reset in a fresh browser context, records visible GUI operations, and
stops at the first blocker; it cannot access the review oracle, inject state, or repair. \textsc{Review}
is independent and operates on a frozen evidence packet: it reads the complete source trace before the
execution and compares required information, decisions, effects, lifecycle, and terminal state, allowing
valid alternative paths and unequal sequence lengths. It may emit \textsc{Pass} or the first material
mismatch with its owner, but it cannot modify the source trace, checklist, environment, or execution, drive
actions, extend a stopped execution, or perform repairs.

\begin{algorithm}[h]
\caption{Environment validation and repair}
\label{alg:agentic-closure}
\algtext*{EndFor}
\begin{algorithmic}[1]
\Require browser environment $E_0$, frozen page specifications $\mathcal C$, source traces $R$, repair budget $K$
\Ensure validated environment or dropped slice
\State $O\gets\Call{BuildObligations}{R}$; $E\gets E_0$
\For{$k=0,1,\ldots,K$}
  \State $X\gets\Call{Execute}{E,O}$
  \State $z\gets\Call{Review}{R,X,O}$
  \State \textbf{if} $z=\textsc{Pass}$ \textbf{then} $z\gets\Call{Probe}{E,\mathcal C,O}$
  \State \textbf{if} $z=\textsc{Pass}$ \textbf{then} \Return \Call{Verified}{$E$}
  \State \textbf{if} $k=K$ \textbf{or not} \Call{Repairable}{$z\mid\mathcal C$} \textbf{then break}
  \State $(E,\Delta)\gets\Call{Repair}{E,z\mid\mathcal C}$
  \State \Call{Invalidate}{$\Delta$}
\EndFor
\State \Return \Call{DropSlice}{$E,z$}
\end{algorithmic}
\end{algorithm}

\textsc{Probe} applies the verification contract below and Table~\ref{tab:probe-matrix}, returning
\textsc{Pass} or an attributed failure. \textsc{Review} and \textsc{Probe} only report: a failure is a
defect packet naming the step, the expected and observed state, and the affected obligations.
\textsc{Repairable} holds when the packet can be closed by editing the package under fixed
$\mathcal C$: rendering, route wiring, storage, bindings, projection, a missing implementation of a
supported relation, or a supported relation whose realized state update or control behavior differs
from the source. A packet whose closure would require behavior outside $\mathcal C$, that is, a
new page identity, state or data relation, guard, or effect that no source evidence supports, is not
repairable, and $\mathcal C$ is never edited inside the loop. \textsc{Repair} receives the
existing package, the packet, and the source evidence and edits only the package; correcting a
wrongly realized effect changes the package, not the model. \textsc{Invalidate} removes execution,
review, probe, and
checkpoint evidence intersecting edited dependencies $\Delta$; affected traces restart at step zero,
while identical consumed slices retain their evidence. Shared runtime/reset edits reach all dependent
tasks. The default $K=3$ allows an initial check, three repairs, and a final check. \textsc{Verified}
returns revision-matched evidence with the environment. \textsc{DropSlice} records the packet and its
owner and freezes the environment without the slice that failed validation; a dropped slice can
re-enter only through a new reconciliation over its source evidence and a fresh compilation, which is
a construction revision rather than a repair. Appendix~\ref{app:source-review-prompt} gives the
condensed review prompt; full prompts and routing records are in the artifact.

\subsection{State-Based Verification}
\label{app:supervision-probes}
\label{app:closure-status}

Verification maps supported obligations to business-state checkpoints, canonical producers and
reducers, visible projections, and evaluator consumers. Partial progress and terminal success remain
separate: checkpoints represent completed subgoals, not actions or hidden reasoning. A valid
alternative path passes when it realizes the same contracted state. Legal actions must realize their
contracted projection; illegal attempts preserve business state and dispatch no route or transaction:
\begin{align}
a\in\mathcal A_P^+(x)&\Rightarrow
\operatorname{Proj}_{\mathrm{biz}}(\operatorname{Exec}_P(x,a))=
\operatorname{Proj}_{\mathrm{biz}}(\mathcal C_P(x,a)),\label{eq:action-contract}\\
a\in\mathcal A_P^-(x)&\Rightarrow
\operatorname{Proj}_{\mathrm{biz}}(\operatorname{Exec}_P(x,a))=
\operatorname{Proj}_{\mathrm{biz}}(x),\quad\operatorname{Dispatch}(x,a)=\varnothing.
\label{eq:action-contract-illegal}
\end{align}
Here $\mathcal A_P^+$ and $\mathcal A_P^-$ are legal and guard-false, wrong-type, or unbound attempts;
$\mathcal C_P$ is the prescribed transition. $\operatorname{Proj}_{\mathrm{biz}}$ is compiled from the
app specification, not hand-specified: it keeps the State fields and Data records supported by $\TraceSupport$ and
drops the route, overlay, focus, scroll, and selection variables of the view. Ephemeral UI changes do not count as business effects.
The same digest must pass natural GUI execution, independent sequence review, and every applicable
probe in Table~\ref{tab:probe-matrix} before validation passes.

\begin{table}[h]
\centering
\caption{\textbf{Mandatory probe matrix.} Every probe runs on the same environment digest; the required verdict
is judged on canonical business state, never on a screenshot.}
\label{tab:probe-matrix}
\small
\setlength{\tabcolsep}{4pt}
\renewcommand{\arraystretch}{1.06}
\begin{tabularx}{\linewidth}{@{}p{0.26\linewidth}p{0.20\linewidth}X@{}}
\toprule
Probe & Required verdict & False pass it exposes \\
\midrule
Zero-action initial state & no progress, no reward & reward granted at launch; leaked prior state \\
Each legal partial stage & progress only & terminal credit for route-only arrival \\
Wrong entity or value & fail & terminal check that ignores the bound object \\
Premature terminal & fail & success declared before the required mutation \\
Repeated action & judged on final state & double credit from a historical maximum \\
Correct terminal & pass & missing verifier-visible consequence \\
Fresh reset, back-to-back episodes & state, events, transactions, reward cleared & cross-episode leakage \\
Legitimate alternative path & pass iff same contracted state & trace-path imitation as the reward \\
\bottomrule
\end{tabularx}
\end{table}

\FloatBarrier
\subsection{Cross-Application Workflows}
\label{app:train-test-contract}
\label{app:cross-app-runner}

Source recordings are application-local. Cross-app benchmark authoring composes exactly two validated
app-local task-facing indexes; it may add a task-level dependency, but no new local field, route, entity
relation, state transition, or effect. A typed, GUI-visible output in $A$ supplies an input or guard
in $B$, and a visible result in $B$ must affect a later action after returning to $A$.
Independent $A$-then-$B$ completions and hidden-state handoffs are rejected. The runner preserves each
app's state while backgrounded. The joint terminal contract checks both local outcomes, both handoff
values, and the final decision in $A$:
$G=g_A^{\mathrm{pre}}\land h_{A\to B}\land g_B\land h_{B\to A}\land g_A^{\mathrm{final}}$.
Local behavior is validated against its source traces; the
composition is validated against this frozen joint contract, since no source recording spans the
complete cross-app episode. Benchmark authoring creates the composition graph before evaluation;
after freezing, environment construction and repair, learner-guided task generation, training, and
checkpoint selection cannot inspect the graph or joint oracle
(Appendix~\ref{app:held-out-protocol}).

\appsection{supply}{app:expansion}{Learner-Guided Task Supply under Fixed Semantics}

\label{app:expansion-details}

\subsection{Task Candidate Authoring}
\label{app:expansion-loop}

\paragraph{Task-facing index.}
\label{app:task-index}
After a package passes source replay, the indexer extracts a read-only index
$I_r=\Pi(E_r,\text{replay records},S_{\mathrm{anchors}})$ from the source-validated package $E_r$;
the frozen app model contributes only stable identifiers, provenance, and support anchors, so the
index never reads a stale specification once repair has moved the package closer to the source.
The index lists surface, host, and control identifiers, entity schemas, observable fields, supported
operation identifiers, observable outcomes, implementation status, and source links; guards,
effects, and destinations appear as references to already-established operations rather than as a
second copy of transition semantics. A dependency index records producer, consumer, value type,
entity scope, freshness, and the supported discovery path that makes an instruction-implicit value
observable. Reconstructing dynamics is the modeling problem; the indexer does not reconstruct app
behavior but extracts addressable structure from the validated world and attaches references to
established operations and dependencies, so it has no semantic authority. Every emitted item must
reference a supported relation and its evidence, an item that cannot be resolved is dropped rather
than inferred, and no transition, guard, or effect is introduced. Because extraction is cheap, the
index is refreshed after each instantiated task slice rather than re-derived by modeling
(Appendix~\ref{app:structure-extraction-prompt}).

\paragraph{Candidate binding.}
Task authoring reads this index and performs no structure extraction of its own. A candidate binds
its user goal and terminal value, launch state, business subtasks and decisions, cross-state
dependencies, evidence revisits, acceptable terminal set, reset boundary, and evaluator-observable
outcome to indexed items and their sources. A missing support binding rejects the candidate; a
binding whose item is supported but not yet implemented routes to implementation rather than to
reconciliation. Appendices~\ref{app:structure-extraction-prompt} and~\ref{app:task-authoring-prompt}
give the condensed indexing and task-authoring instructions; Algorithm~\ref{alg:task-screening-repair}
owns screening and selection.

\paragraph{Cross-application benchmark authoring.}
The learner-guided branch produces only single-app training tasks. A separate pre-evaluation branch
joins two source-validated app-local indexes through typed, GUI-visible handoffs $A\rightarrow B\rightarrow A$.
Its local requirements satisfy $\operatorname{Req}_A(c)\subseteq\TraceSupport^A$ and
$\operatorname{Req}_B(c)\subseteq\TraceSupport^B$.
Pairs are selected before policy outcomes by typed handoff compatibility in both directions;
within a pair, tasks must differ in dependency structure rather than only entity bindings.
The producer's surface, field, type, freshness, and entity scope must match the consumer's input or
guard; the second handoff must change the final action in the preserved state of $A$. The branch
freezes a joint terminal contract and rejects independent task concatenations. It adds a task-level
dependency over supported local semantics, not an application transition inferred from a cross-app
source trace (Appendix~\ref{app:cross-app-runner}). The resulting 100 tasks enter only
Trace2World-Bench, before any policy evaluation.

Difficulty follows the four capabilities in Figure~\ref{fig:composition}: long-horizon, reasoning,
memory/state, and recovery/backtracking. Schema-backed instance variation changes entities and values
without claiming a new workflow. Deliberately long tasks require at least seven business subtasks,
four non-trivial decisions, three cross-state dependencies, and two evidence revisits, plus two
required difficulty phenomena with at least one information- or state-bearing. Navigation and repeated
clicks do not satisfy this floor. Full prompts and fine-grained annotations are in the artifact.

\subsection{Task Implementation and Screening}
\label{app:expansion-gate}
\label{app:repair-budget}
\label{app:rollout-attribution}
\label{app:screening-formal}

\textsc{Implement} instantiates a candidate's task slice in the package: schema-backed entity,
value, and state instances, hosts and state-dependent transitions of already-supported operations,
projections, reset/session isolation, and evaluator-visible consequences. It adds no app-level
transition, guard, or effect beyond the frozen model. It checks fresh GUI execution and affected
seed-task regression, then refreshes the task-facing index. For \textsc{EnvDefect}, it reproduces
the failure on the current revision and calls Algorithm~\ref{alg:agentic-closure} with the remaining
repair budget. Its persistent per-slice ledger checks the three-wave cap before each repair and
retains counts across requeues and refinements. It returns a validated environment, edited
dependencies $\Delta$, and a status; failed slices are excluded. \textsc{RequeueAffected} invalidates
intersecting queued evidence and moves affected selections back to $Q$, preserving their repair counts.

\textsc{PassAtFour} returns four valid attempts each from the preceding learner $\pi_L$ and a fixed
stronger solver $\pi_S$ (instantiated in Appendix~\ref{app:model-roles}), or early failure records for diagnosis. Invalid infrastructure or environment
attempts are retained but do not count. \textsc{Diagnose} first attributes non-learner faults. A complete,
fault-free screen is held if the solver never succeeds, marked \textsc{Refinable} if learner mean reward
exceeds $0.8$, and otherwise \textsc{Selectable}. Screening attempts never enter training, reward,
or formal evaluation credit.

\begin{algorithm}[H]
\caption{Learner-guided task selection}
\label{alg:task-screening-repair}
\algtext*{EndWhile}
\begin{algorithmic}[1]
\Require validated environment $E$, supported candidates $\mathcal C$, learner $\pi_L$, solver $\pi_S$
\Ensure next task batch $A$
\State $A\gets\varnothing$; $Q\gets\{(c,\textsc{New}):c\in\mathcal C\}$
\While{$Q\ne\varnothing$}
  \State $(c,z)\gets\Call{Pop}{Q}$
  \State $(E,\Delta,z)\gets\Call{Implement}{E,c,z}$
  \State $(A,Q)\gets\Call{RequeueAffected}{A,Q,\Delta}$
  \State \textbf{if} $z\ne\textsc{Ready}$ \textbf{then} \Call{Record}{$c,z$}; \textbf{continue}
  \State $(A_L,A_S)\gets\Call{PassAtFour}{E,c,\pi_L,\pi_S}$
  \State $z\gets\Call{Diagnose}{c,A_L,A_S}$
  \State \textbf{if} $z=\textsc{Selectable}$ \textbf{then} $A\gets A\cup\{c\}$; \textbf{continue}
  \State \textbf{if} $z=\textsc{EnvDefect}$ \textbf{then} $Q\gets Q\cup\{(c,z)\}$; \textbf{continue}
  \State $Q\gets Q\cup\Call{RefineOrHold}{c,z}$
\EndWhile
\State \Return \Call{Freeze}{$E,A$}
\end{algorithmic}
\end{algorithm}

\textsc{RefineOrHold} returns $\{(c',\textsc{New})\}$ when an actionable task issue or saturation allows
a concrete supported refinement; otherwise it records a deferral/hold and returns $\varnothing$.
Refinement is one-directional and bounded: a lineage is refined at most twice, saturated lineages are
made more demanding within fixed support, and hard or solver-unsolved lineages are held, never
simplified (Appendix~\ref{app:pool-results}).
Refinements repeat implementation and screening. \textsc{Record} logs failed validation and its owner;
exhausted slices are dropped. \textsc{Freeze} binds selected tasks and current
evidence to the final environment revision. A discovered path is implemented only when its edges are
supported and its task semantics are equivalent. App-specification support stays fixed; the artifact retains
routing decisions and immutable attempts used for independent failure reproduction.

\subsection{Augmentation Batches and Policy Training}
\label{app:fixed-policy}
\label{app:policy-optimization}

A \emph{training stage} is one pass over a frozen task batch; an \emph{RL iteration} is one DAPO
optimizer step within that stage. The three stages use seed tasks, augmented batch 1, and augmented
batch 2. For stage $r$, $\mathcal B_1=\mathcal N$, $\mathcal B_2=\mathcal A_1$, and
$\mathcal B_3=\mathcal A_2$, with $\pi_r=\operatorname{DAPO}(\pi_{r-1};\mathcal B_r)$.
After seed training, generation reads $\pi_1$'s task outcomes; after the first augmented batch,
it reads $\pi_2$'s task outcomes, saturation, solver failures, and attributed defects. Each task
is screened with the same preceding learner. Batch manifests freeze task signatures, membership,
launch specifications, preceding checkpoint, screening attempts, and environment revisions before optimization.
Appendix~\ref{app:learner-gap-prompt} specifies how prior practice outcomes condition the next
authoring pass.
Between stages, supported implementation changes must be reverified before the next freeze.

Each batch is shuffled once and consumed without replacement; the learner continues its preceding
weights and never replays earlier batches. The optimized reward is the environment task evaluator's
checkpoint credit. Task $t$ declares $k_t$ checkpoints $c_{t1},\dots,c_{tk_t}$, each a predicate over
the final business state $x_T$ of an attempt, with $c_{tk_t}$ the task's terminal condition; checkpoints
are unordered and equally weighted, so an alternative path that reaches the same final state earns the
same credit, and a state that is undone before the episode ends earns none. With
$s=\mathbb 1[c_{tk_t}(x_T)]$ the terminal verdict,
\begin{equation}
q=\begin{cases}1 & s=1,\\[2pt] \frac{1}{k_t}\sum_{i=1}^{k_t}\mathbb 1[c_{ti}(x_T)] & s=0.\end{cases}
\label{eq:task-reward}
\end{equation}
A task group contains $G=8$ sampled trajectories, and it is retained only if the population SD of its
eight rewards exceeds 0.1: all-success and all-fail groups are dropped before optimization, as is a
group whose rewards differ by one checkpoint in only one sample of a task with four or more checkpoints.
DAPO \citep{yu2025dapo} uses token-level, group-mean-centered and group-std-normalized advantages,
asymmetric PPO clipping \citep{schulman2017ppo,shao2024deepseekmath}, and dynamic sampling over the
retained groups. No format reward, KL, entropy bonus, or overlong-reward shaping is added. The terminal
checkpoint is fixed by the schedule, and held-out scores never guide generation, stopping, or selection.
Batch sizes, RL-iteration counts, and the remaining run settings are fixed in
Appendix~\ref{app:training-config}.

\appsection{settings}{app:settings}{System Configuration and Evaluation Protocols}
\label{app:evaluation}
\label{app:evaluation-details}

This section records what the release contains and the shared settings and evaluation boundaries needed
to reproduce the experiments: the resource contract, the agent interface and model roles, the
application/task/evaluation partitions, the policy-optimization schedule, and the evaluator roles and
evaluation-record binding.
Appendix~\ref{app:extended-results} co-locates each study's setup, baselines or controls, metrics, and
supplementary results. Within each matched comparison, the settings below are held fixed unless the
local protocol names the changed factor.

\subsection{Trace2World Resource Contract}
\label{app:resource-contract}

Trace2World is released as an environment-construction resource rather than as a benchmark alone. The
release binds the source evidence, reconstructed environments, training-task supply, evaluation sets, and
fidelity records used in this study (Table~\ref{tab:resource-contract}). Trace2World-Bench is one frozen
evaluation view over the resulting environment portfolio; it is not the construction target. Two terms
separate evidence from its use. A \emph{source trace} is the semantic unit $(q,o_0,a_0,\ldots,o_T)$ of
Section~\ref{sec:problem-setting}; a \emph{source recording} is the released evidence behind it: the
screenshot sequence, raw device actions, marked observations, instruction, terminal contract, and
collection metadata. Source recordings and derived trace records are released together and preserve the
complete research-relevant chronology, including retries and corrections, subject only to the privacy,
security, and redistribution screening in the Ethics Statement. Each trace binds a \texttt{source\_digest},
a \texttt{release\_digest}, a release-review status, and a redaction manifest that declares every redacted
region with its reason and its semantic effect; a trace whose safe release would alter its interaction
evidence is withheld and listed rather than trimmed.

\begin{table}[H]
\centering
\caption{\textbf{Trace2World resource contract.} Scale, role, and released artifact of each component. Source traces are construction evidence and the external fidelity authority; the environment portfolio is the reusable resource; Trace2World-Bench is one evaluation view over it.}
\label{tab:resource-contract}
\small
\setlength{\tabcolsep}{4pt}
\begin{tabularx}{\linewidth}{@{}>{\raggedright\arraybackslash}p{2.35cm}>{\raggedright\arraybackslash}p{2.55cm}>{\raggedright\arraybackslash}p{3.5cm}>{\raggedright\arraybackslash}X@{}}
\toprule
Component & Scale & Role & Released artifact \\
\midrule
Source traces & 1,570 / 50 apps & construction evidence; external fidelity authority & source recordings, action records, instructions, terminal contracts, collection metadata \\
Construction pipeline & one shared pipeline & trace $\rightarrow$ reconciled specification $\rightarrow$ environment & preprocessing, reconciliation, compilation, validation, and repair code \\
Environment portfolio & 50 apps / 10 domains & resettable GUI RL substrate & standalone packages with state and evaluator manifests \\
Training-task supply & 1,280 seed + 768 learner-guided & policy optimization & task, launch, support, and evaluator manifests \\
Trace2World-Bench & 1,044 tasks & frozen policy evaluation & task definitions, split identities, evaluators \\
Fidelity study & 200 workflows / 10 apps & reconstruction correctness & source/execution packets, reviewer labels, human adjudication \\
Real-device panel & 100 tasks / 20 apps & paired sim-to-real evidence & frozen task manifests, paired outcomes \\
\bottomrule
\end{tabularx}
\end{table}

\subsection{Agent Interface and Model Roles}
\label{app:agent-interface}
\label{app:model-roles}

The learner observes the instruction, action history, and the three most recent screenshots,
and emits one call from Table~\ref{tab:action-space}. Coordinates use $[0,999]^2$, projected to the
current screenshot and browser CSS frame. Trace records, evidence maps, checkpoints, hidden state,
repair diagnoses, and oracle predicates are not observations. An action on a component outside the
current task binding or implemented behavior scope, or any other invalid or non-interactive action,
consumes a step, changes no business state, and returns only \texttt{NOT\_INTERACTIVE}, revealing
neither the correct target nor the next action. A missing or inconsistent legal host is an
environment blocker, not a learner error. Step budgets are given in Appendix~\ref{app:training-config};
\texttt{open\_app} is restricted to the episode manifest.

\begin{table}[H]
\centering
\caption{\textbf{Action space.} The same public interface serves training and evaluation rollouts.}
\label{tab:action-space}
\small
\setlength{\tabcolsep}{5pt}
\begin{tabularx}{\linewidth}{@{}l>{\raggedright\arraybackslash}p{4.2cm}X@{}}
\toprule
Action & Arguments & Definition \\
\midrule
\texttt{click} / \texttt{double\_click} / \texttt{long\_press} & coordinate & tap, double-tap, or 500\,ms press at the point \\
\texttt{type} & text & enter text into the focused editable control \\
\texttt{swipe} / \texttt{scroll} & direction; optional coordinate & finger-motion direction / content-motion direction \\
\texttt{drag} & start, end coordinates & drag from start to end \\
\texttt{open\_app} & app name & open the named application (episode manifest only) \\
\texttt{system\_button} & \{home, back, enter\} & press the system button \\
\texttt{wait} & seconds ($\le 30$) & wait \\
\texttt{answer} & text & submit a textual answer and end the trajectory \\
\texttt{terminate} & \{success, fail\} & end the trajectory with the declared status \\
\bottomrule
\end{tabularx}
\end{table}

\begin{table}[H]
\centering
\caption{\textbf{Policy-facing rollout protocol.} A step is one public action call, including rejected
calls and terminal submissions; it cannot batch several GUI actions. The task construction class fixes
the cap before policy outcomes.}
\label{tab:rollout-protocol}
\footnotesize
\setlength{\tabcolsep}{4pt}
\rowcolors{3}{t2wRowGray}{white}
\begin{tabularx}{\linewidth}{@{}>{\raggedright\arraybackslash}p{0.23\linewidth}X@{}}
\toprule
Field & Episode contract \\
\midrule
Policy observation & Instruction, action history, three most recent screenshots, and feedback for the preceding rejected action; no trace, hidden state, or evaluator access. \\
Action unit & One call from Table~\ref{tab:action-space} per policy decision. Invalid actions, \texttt{wait}, \texttt{open\_app}, \texttt{answer}, and \texttt{terminate} each use one decision; a nonterminal call is followed by a new observation. \\
Geometry & The model uses $[0,999]^2$ coordinates, mapped through the current screenshot to the browser CSS viewport; each package declares its viewport and device scale. \\
Settling & Declared navigation targets wait up to 5\,s; pointer and gesture actions add 50\,ms. Before a screenshot, the host waits for fonts and an animation frame. \texttt{wait} is a separate, budgeted action. \\
Stop and score & A terminal call or the class cap ends the attempt. In both cases the task evaluator scores the final business state and required receipts; a declared \texttt{success} does not override its verdict. \\
Horizon & $H=30$ source-derived single-app; $45$ augmented single-app; $60$ two-app return composition. Single-app caps match across screening, RL, and evaluation; two-app tasks enter only the benchmark. \\
Reset and return & A fresh browser context starts from the task's fixed launch state; reset clears episode state, events, transactions, and reward. Cross-app switching preserves each app's state while backgrounded. \\
\bottomrule
\end{tabularx}
\rowcolors{1}{white}{white}
\end{table}

The 50-app runtime profile fixes a $430\times884$ CSS-pixel viewport at device scale factor two,
producing $860\times1768$ raw PNG observations. Model-side image processing is
configured independently of this environment geometry. Training and evaluation decoding, token limits,
and attempt counts appear in Table~\ref{tab:dapo-config} and Appendix~\ref{app:held-out-protocol}.

\paragraph{Model roles.}
Construction uses two distinct model roles. GPT-5.2 serves every bounded API call in the construction
pipeline, including source-trace preprocessing (Appendix~\ref{app:trace-preprocessing}), semantic
reconciliation, and initial package compilation; GPT-5.5 drives package-level verification and
brownfield repair in the agentic Codex flow. The bounded API stage therefore runs on the weaker of
the two models, while every compared builder receives GPT-5.5 in each role it defines, so the
fidelity margin does not follow from model capability.
The initial environment,
agentic fixloop, and task-augmentation stages remain separate. Qwen3.5-9B is the learner;
Qwen3.7-Plus is the fixed solvability solver; GPT-5.6 Sol is the frozen offline fidelity reviewer
(Appendix~\ref{app:gold-matching}) and, in isolated contexts, the PhoneWorld-style QA reviewer.
It never supplies training rewards or held-out SR/PR; those use the environment's programmatic task evaluator.
The policy-optimization schedule and hyperparameters are specified in Appendix~\ref{app:training-config}.

\subsection{Application, Task, and Evaluation Splits}
\label{app:portfolio}

The experimental design separates environment construction from policy optimization.
All 50 applications and all 1,570 traces are used to construct the 50 environments.
Policy optimization uses tasks from 40 applications only. The other
ten applications, selected one per domain before task generation, form the \emph{unseen-application}
test split: they supply construction traces but no seed or learner-guided task, candidate slot, screening
rollout, or RL update. Table~\ref{tab:roster} lists all 50 applications by domain and makes the
40/10 policy split explicit.

\begin{table}[H]
\centering
\caption{\textbf{Application roster by domain and policy split.} All 50 applications provide
construction traces. The 40 policy-training applications supply optimization tasks; the ten held-out
evaluation applications, one per domain, supply no seed, candidate, screening, or RL-update task.}
\label{tab:roster}
\footnotesize
\setlength{\tabcolsep}{3pt}
\renewcommand{\arraystretch}{1.08}
\begin{tabularx}{\linewidth}{@{}>{\raggedright\arraybackslash}p{0.22\linewidth}>{\raggedright\arraybackslash}X>{\raggedright\arraybackslash}p{0.16\linewidth}@{}}
\toprule
Domain & Policy-training apps (40) & Held-out eval app (10) \\
\midrule
Social \& community & WeChat; QQ; Weibo; Xiaohongshu & Zhihu \\
News \& short video & Kuaishou; Bilibili; Toutiao; Tencent News & Douyin \\
Long video \& audio & iQIYI; Youku; NetEase Cloud Music; Kugou Music; Ximalaya & QQ Music \\
Commerce \& resale & Taobao; JD; Xianyu; Dewu & Pinduoduo \\
Local services \& food & Meituan; Dianping; Dingdong Maicai; KFC; Hema & Ele.me \\
Mobility, travel \& tickets & Amap; Ctrip; Didi; Railway 12306; Damai & Taopiaopiao \\
Auto, housing \& jobs & 58Tongcheng; Autohome; Dongchedi; BOSS Zhipin & Beike \\
Work, learning \& search & Feishu; WPS Office; Baidu; DingTalk; Youdao Dictionary; Baidu Netdisk & Tencent Meeting \\
Payment \& logistics & Alipay; SF Express & Cainiao \\
Health \& fitness & Boohee Health & Keep \\
\bottomrule
\end{tabularx}
\end{table}

\paragraph{Construction-trace allocation.}
The 1,570 trace count is a pre-collection evidence quota, not a train/test partition. An application
with at least three independent service lines receives 50 traces; the other applications receive 20.
Thus 19 multi-service applications contribute 950 traces and 31 single-service applications
contribute 620. The multi-service set is WeChat, QQ, Douyin, Kuaishou, Taobao, JD, Pinduoduo,
Meituan, Amap, Ctrip, 58Tongcheng, Autohome, Dongchedi, Beike, Xiaohongshu, Bilibili, Dianping,
Baidu, and Alipay. A service line is fixed from the product surface before collection and has its own
catalog or service plus a transactional record and lifecycle. Domain is descriptive and is used to
stratify held-out and audit sets; it does not determine the collection quota.

\begin{table}[H]
\centering
\caption{\textbf{Evaluation sets, partitions, and denominators.} This table is the sole size owner for
the Trace2World-Bench result columns in Table~\ref{tab:agym-family-models}; SR there averages three attempts
per task. Entity replacements sample training-side workflow skeletons and do not reduce the complete
20\% held-out-workflow split. Fidelity and device transfer retain separate populations.}
\label{tab:evaluation-sets}
\small
\setlength{\tabcolsep}{4pt}
\begin{tabularx}{\linewidth}{@{}XXr@{}}
\toprule
Evaluation set & Application coverage & Size / partition \\
\midrule
Fidelity (construction-exposed) & 10-app reconstruction set & 200 workflows (16--26/app) \\
Trace2World-Bench & all 50 apps & 1,044 tasks \\
\quad Policy-training-app subtotal & 40 policy-training apps / 10 frozen pairs & 868 tasks \\
\qquad Entity replacement & training-side workflow skeletons & 256 tasks \\
\qquad Held-out workflow & 20\% signature split (320 seed + 192 aug.) & 512 tasks \\
\qquad Cross-app & 10 policy-training-app pairs & 100 tasks (10/pair) \\
\quad Tasks from unseen apps & 10 apps excluded from policy training & 176 tasks \\
Device transfer & 20 policy-training apps & 100 tasks (5/app) \\
\bottomrule
\end{tabularx}
\end{table}

\paragraph{How evaluation tasks are separated.}
The fidelity evaluation set samples supported workflows from ten applications without replacement and scores the
same 200 construction-exposed workflows with the same six binary diagnostic questions for every
construction condition at equal application weight. The ten-app roster coincides with the policy-held-out
roster, but the estimator uses only these paired workflow judgments and equal application weighting; it is
not a held-out policy test. The primary policy uses 2,048 optimization tasks: 1,280 seed tasks followed by
two 384-task learner-guided batches, all on the 40 policy-training applications. Entity-rebinding variants
enter only the matched training control: two additional 384-task batches change entities while retaining
seed workflow structure, and replace rather than accompany the learner-guided batches in that branch.

Before any outcome is observed, the 40 policy-training apps define 2,560 single-app workflow signatures.
The split is stratified by source family: 1,600 representative seed-family signatures yield 1,280 train / 320 test,
and 960 augmentation-family signatures yield 768 train / 192 test. A seed or augmentation task enters the
512-task test slice whenever its complete goal, launch state, terminal predicate, primitive order, recovery
structure, and composition signature is absent from training. Entity replacement then samples 160 seed-family
and 96 augmentation-family skeletons from the 2,048-task training side while freezing new task IDs and canonical entity
lineages that appear in no RL batch; it therefore tests entity shift without consuming the workflow test split.

The four Trace2World-Bench slices in Table~\ref{tab:evaluation-sets} are disjoint as evaluation task
identities and exhaust the 1,044-task denominator. Tasks from unseen apps are
single-application only. Cross-application tasks use policy-training applications and form a
100-task Trace2World-Bench slice with ten disjoint two-app pairs and ten structurally distinct,
return-dependent tasks per pair
(Appendix~\ref{app:application-holdout}). No benchmark identity enters an optimization batch or guides
learner-guided generation, screening, or checkpoint selection. The pair roster, orientation, typed
handoffs, and joint terminal contracts are frozen before policy outcomes.

The fixed device-transfer population is the 100-task, 20-app panel in
Table~\ref{tab:evaluation-sets}: 50 unseen entity/state rebindings and 50 workflows drawn from the
512-task held-out-workflow slice. The latter are an overlapping real-device subset of Trace2World-Bench rather
than an independent benchmark. Appendix~\ref{app:real-device-protocol} owns the within-app allocation,
counterbalancing, eligibility, pairing, and estimands. The benchmark's cross-app slice is simulation-only and
adds no real-device executions.
Batch freezing follows Appendix~\ref{app:policy-optimization}; the artifact manifest records the
roster with class, service-line count, domain, trace identities, collection quotas, and split identities.

\subsection{Policy Optimization and Rollout Configuration}
\label{app:training-config}

The three frozen batches contain 1,280/384/384 single-app tasks on the 40 training apps. With at most
40 task groups per RL iteration,
\begin{equation}
K_r=\left\lceil|\mathcal B_r|/40\right\rceil,\qquad(K_1,K_2,K_3)=(32,10,10).
\label{eq:iteration-budget}
\end{equation}
The final iterations therefore contain 40/24/24 tasks. Table~\ref{tab:dapo-config} fixes the group
size, retention and no-refill rules, checkpoints, and remaining settings for every matched condition.

\begin{table}[H]
\centering
\caption{\textbf{Policy-optimization and rollout configuration.} All matched training conditions
use these settings; changing any field defines a new condition. The public-action horizon follows
the frozen construction class in training, screening, Trace2World-Bench, and real-device episodes.}
\label{tab:dapo-config}
\small
\setlength{\tabcolsep}{4pt}
\begin{tabular}{@{}lll@{}}
\toprule
Field & Value & Note \\
\midrule
Initial policy & Qwen3.5-9B & no SFT warm-up \\
Objective & DAPO & clip-higher, dynamic sampling, token-level loss \\
Reward & checkpoint credit $q$ & Eq.~\ref{eq:task-reward}; no format or length terms \\
$\epsilon_{\mathrm{low}}$ / $\epsilon_{\mathrm{high}}$ & 0.2 / 0.28 & DAPO defaults \\
KL / entropy coefficient & 0 / 0 & no reference model \\
Tasks / RL iteration & at most 40 & shuffled once per stage; no replacement \\
Retained groups / RL iteration $N$ & at most 32 & reward SD $\le0.1$ dropped; no refill \\
Task-group size $G$ & 8 trajectories & sampled for each task group \\
Task-supply stages & 3 & seed, augmented batch 1, augmented batch 2 \\
RL iterations / stage & 32 / 10 / 10 & $K_r=\lceil|\mathcal B_r|/40\rceil$ \\
Final-iteration tasks & 40 / 24 / 24 & no padding or repeated draws \\
Optimizer / learning rate & AdamW / $1\times10^{-6}$ & constant, no warmup \\
Mini-batch & 64 trajectories & up to 4 gradient steps per RL iteration \\
Training / evaluation temperature & 1.0 / 0.7 & Trace2World-Bench SR/PR; external-suite protocols \\
Evaluation checkpoints & 0 / 32 / 42 / 52 & cumulative RL iterations; base and stage ends \\
Screenshots in context & 3 & most recent \\
Episode step budget $H$ & 30 / 45 / 60 & source-derived / augmented / cross-app \\
Max context / response length & 32K / 4K & tokens \\
Seeds & 3 (0, 1, 2) & all training and policy-evaluation runs; mean $\pm$ std \\
\bottomrule
\end{tabular}
\end{table}

\paragraph{Construction-class horizons.}
An episode step is one public GUI action, including \texttt{open\_app} and switches. The class is fixed
from the task manifest before any model outcome: 1,280 source-derived training tasks use $H=30$ and
768 augmented training tasks use $H=45$; no cross-app task enters RL. Trace2World-Bench contains 656
source-derived single-app tasks (160 seed-family entity replacements, 320 seed-family workflow
holdouts, and 176 unseen-app tasks), 288 augmented single-app tasks (96 entity replacements and 192
workflow holdouts), and 100 two-app compositions. The paired device panel has 80 source-derived and
20 augmentation-family tasks, each with the same cap in simulation and on device. The 30/45/60
assignment follows provenance and structural composition rather than model success.

\begin{table}[H]
\centering
\caption{\textbf{Legal-path calibration of the frozen Trace2World-Bench horizons.} Path lengths count
public GUI actions from reset to terminal state. Every cross-app task is completed by a human through
the agent's action interface before admission, with a 45-action human cap; the agent receives 60.
The two single-app classes use frozen legal paths and GUI checks.}
\label{tab:horizon-calibration}
\small
\setlength{\tabcolsep}{4pt}
\begin{tabular}{@{}lrrrrrr@{}}
\toprule
Task class & Tasks & Certified & Median & P90 & Max & $H$ \\
\midrule
Source-derived single-app & 656 & 656 & 9 & 17 & 24 & 30 \\
Augmented single-app & 288 & 288 & 18 & 29 & 36 & 45 \\
Two-app return composition & 100 & 100 & 27 & 39 & 44 & 60 \\
\bottomrule
\end{tabular}
\end{table}

\FloatBarrier

\subsection{Evaluator Roles and Non-Interference}
\label{app:behavioral-oracle}

We distinguish three checks. Construction-time validation verifies trace obligations on builder
executions. The environment task evaluator programmatically evaluates business-state checkpoints
to supply the RL reward and held-out SR/PR. The independent fidelity reviewer compares raw source-trace
evidence with
generated executions and supplies binary reconstruction labels; it never scores training or held-out
rollouts.

The three outputs are not interchangeable: construction-time validation can trigger environment repair,
task-evaluator credit can train or score a policy, and fidelity-review labels can support only the
reconstruction study. Final evaluation and human annotations never return to construction, task
filtering, or checkpoint selection.

\subsection{Evaluation Record Binding}
\label{app:evaluation-records}

Every evaluation event binds its input identities, artifact or policy revision, native start or reset
state, execution packet digest, retry history, evaluator revision, seed, and reducer.
A score is valid only for its declared packet and estimand; it cannot be reused as evidence for a
different construction, policy, or transfer population.

\FloatBarrier

\appsection{results}{app:extended-results}{Extended Results and Analysis}

This appendix reports the setup, metrics, baselines, controls, and supporting results for
Sections~\ref{sec:reconstruction-fidelity}--\ref{sec:downstream-transfer}: reconstruction fidelity and
validation first, then construction scaling, policy gains and task supply, and real-device transfer.
Appendix~\ref{app:evaluation} defines the shared settings and evaluator boundaries; complete qualitative cases
are in Appendix~\ref{app:qualitative-examples}.

\subsection{Environment Fidelity and Reviewer Validation}
\label{app:fidelity-results}

\subsubsection{Fidelity protocol and evaluation packets}
\label{app:gold-matching}

\paragraph{Evaluation unit and labels.}
The frozen GPT-5.6 Sol reviewer compares each anonymous execution with its task-aligned source trace and
cannot see constructor identity, generated checks, or repair history. Six P/F diagnostics cover
Structure (Reach and Surface), State (initialization/lifecycle, action Effect, and Terminal outcome),
and Data (entities, relations, and values). Missing or contradicted required evidence is F; equivalent
routes and page partitions are allowed when they preserve the required decision and effect
\citep{zhou2023webarena,chai2026a3}. The all-six workflow verdict is P only when every diagnostic passes. Every diagnostic
uses the same tasks with no N/A label, and rates first average within app and then equally across apps.

\paragraph{Execution and scope.}
Evaluation starts from the native reset or declared task preset and permits at most $\max(80,3L)$ GUI
actions for a source trace of length $L$. Up to two recollections address stale or incomplete capture; the
first complete packet is scored, while an absent valid packet remains incomplete. The implemented behavior scope
fixes which visible components may act for that episode, so a correct
\texttt{NOT\_INTERACTIVE} outside the task boundary is not a capability failure. State invariance and
the absence of unsupported effects serve as negative obligations when no positive mutation is required.

\paragraph{Validation and records.}
Builder-side validation is reapplied to the independently collected evaluation packet before the
review label is exposed. This recheck is separate from construction-time validation; the false-pass rate is
the app-macro fraction of recheck-passing tasks whose all-six verdict is F. Each final record binds the
artifact and source trace digests, native start, packet, six labels, cited observations, and decisive
discrepancy. The full instruction and record schema remain in the artifact.

\subsubsection{Compared builders and controls}
\label{app:baselines}
\label{app:contract-comparison}

Two kinds of comparison appear in the fidelity study. Comparisons among builders evaluate complete
end-to-end construction contracts and support no stage-wise causal attribution; only the
interface-preserving controls below change one declared factor against Trace2World, with all
other inputs, model settings, evaluator evidence, and the execution-time limit shared unless stated below.
Oracle-assisted runs are excluded from primary comparisons, and a defect in a shared environment
invalidates all affected conditions and requires matched reruns.

\paragraph{Environment builders and controls.}
We use \emph{builder} for an end-to-end environment-construction method and \emph{condition} for one
scored configuration; the no-trace and self-validation conditions are controls, not additional builders.
All conditions share the app-level request, model revision, tools, runtime substrate, native recording
pool, and a 6-hour per-app wall-clock limit covering construction, review, and repair. Model requests
are metered but not capped. At the limit, the current artifact is frozen; unfinished builds remain failures.
Within the Direct Agent and CUA-Gym pairs, w/o traces withholds only recordings and derived summaries; the
construction query and research access stay fixed. These pairs isolate evidence access. Comparisons
among builders test complete end-to-end alternatives (Table~\ref{tab:contract-comparison}).

\begin{table}[H]
\centering
\caption{\textbf{Environment builders and controls under the shared experimental contract.} Input,
behavior source, and validation are distinct. Trace-access variants and the validation-evidence control each change one factor.}
\label{tab:contract-comparison}
\small
\setlength{\tabcolsep}{3pt}
\begin{tabularx}{\linewidth}{@{}p{0.22\linewidth}p{0.21\linewidth}p{0.23\linewidth}X@{}}
\toprule
Condition & Input & Behavior source & Validation \\
\midrule
Direct Agent $\pm$ traces & app request; recordings in +traces & implicit in generated code & self-execution and debugging \\
CUA-Gym $\pm$ traces & topic/research; recordings in +traces & Plan-agent specification & Dev--Web loop against the plan \\
PhoneWorld-style (reimpl.) & recorded screens and transitions & screen types and page specifications & two independent screen-QA rounds with fixes \\
GUI-GENESIS-style (reimpl.) & one traced task per site & task-local site and co-written reward & static checks and own golden path \\
Trace2World, self-validation & same traces and construction stages & shared app specification & tests generated from the same interpretation decide validation \\
Trace2World & same real-app traces & reconciled Structure, State, and Data & independent source-trace review and state probes \\
\bottomrule
\end{tabularx}
\end{table}

Direct Agent is an iterative coding process, not a single call. CUA-Gym follows its Plan--Dev--Web
process \citep{wang2026cuagym}. GUI-GENESIS-style creates standalone task-local sites with up to five
retries \citep{cao2026guigenesis}; costs are aggregated per app and unbuilt sites remain failures.
PhoneWorld-style follows 25--30 frequency-ranked screen types, consecutive-screen transitions,
and page specifications based on two or three screenshots \citep{liu2026phoneworld}. Its two
side-by-side QA rounds replace the original human audit with an isolated model reviewer;
only construction evidence is visible, QA time counts toward the limit, and the final evaluator
never sees QA transcripts. At our September 2026 artifact check, the linked public PhoneWorld repository
provides app source, tasks, and an evaluation runner, but not the construction agent's orchestration,
prompts, and repair workflow.
This row therefore tests a matched reimplementation of the described construction route, rather than
an execution of the authors' original agentic pipeline.

\paragraph{Validation-evidence control.}
The self-validation control keeps stages, schemas, models, evidence, and the execution-time limit fixed,
replacing only independent source-trace validation with the environment's generated verifiers; the frozen artifact is
then scored by the independent evaluator on the same 200-workflow sample
(Appendix~\ref{app:generated-closure-control}).

\subsubsection{Workflow fidelity across environment builders}

\begin{table}[p]
\centering
\caption{\textbf{Diagnostic fidelity and reviewer agreement (\%).} Method rows use the primary 200-workflow, ten-app evaluation set and report workflow fidelity plus the six P/F diagnostics assigned by the frozen GPT-5.6 Sol reviewer; missing or contradicted evidence is F, and the All six column requires every diagnostic. The final row reports exact automated--adjudicated-human label agreement on the same 1,200 method--workflow packets. It validates the labels and is not a second performance condition.}
\label{tab:fidelity-full}
\footnotesize
\setlength{\tabcolsep}{3pt}
\begin{tabular*}{\linewidth}{@{\extracolsep{\fill}}lccccccc@{}}
\toprule
\multirow{2}{*}[-0.5\dimexpr\aboverulesep+\belowrulesep+\cmidrulewidth\relax]{Method / audit statistic} & \multicolumn{1}{c}{Overall} & \multicolumn{2}{c}{Structure} & \multicolumn{3}{c}{State} & \multirow{2}{*}[-0.5\dimexpr\aboverulesep+\belowrulesep+\cmidrulewidth\relax]{Data} \\
\cmidrule(lr){2-2}\cmidrule(lr){3-4}\cmidrule(lr){5-7}
 & All six & Reach & Surface & State & Effect & Terminal &  \\
\midrule
Direct Agent w/o traces & 5.2 & 28.4 & 14.8 & 22.7 & 15.0 & 15.9 & 12.8 \\
Direct Agent + traces & 16.7 & 53.6 & 35.3 & 41.2 & 34.6 & 33.7 & 31.7 \\
CUA-Gym w/o traces & 8.2 & 34.4 & 20.0 & 29.0 & 22.6 & 21.5 & 19.6 \\
CUA-Gym + traces & 19.9 & 50.4 & 37.5 & 38.0 & 35.0 & 35.3 & 35.0 \\
PhoneWorld-style (reimpl.) & 22.4 & 57.0 & 39.3 & 46.8 & 41.5 & 39.8 & 38.5 \\
GUI-GENESIS-style (reimpl.) & 59.8 & 94.8 & 83.9 & 76.6 & 77.0 & 79.4 & 78.4 \\
\textbf{Trace2World} & 91.1 & 98.0 & 93.7 & 93.3 & 93.8 & 93.7 & 94.1 \\
\rowcolor{t2wRowGray}\multicolumn{8}{@{}l}{\emph{Controls (not our method)}} \\
Self-validation & 79.1 & 98.5 & 90.4 & 87.0 & 85.0 & 88.0 & 89.1 \\
\multicolumn{8}{@{}l}{\emph{Reviewer agreement (same packets)}} \\
Exact automated--human agreement & 90.2 & 97.4 & 96.2 & 96.3 & 95.8 & 96.7 & 96.0 \\
\bottomrule
\end{tabular*}
\end{table}

\begin{figure}[p]
\centering
\includegraphics[width=\linewidth]{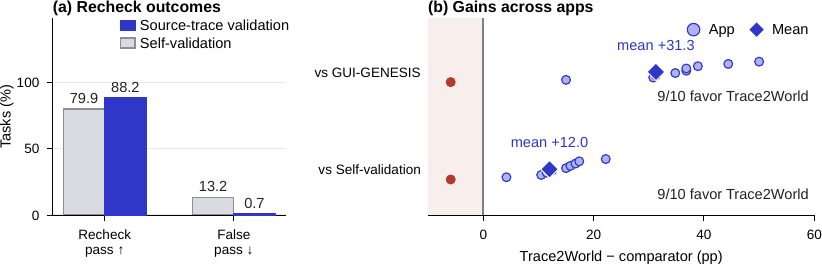}
\caption{\textbf{Validation-evidence control and per-application robustness.}
Both panels use the primary 200-task, ten-app evaluation set. (a) Grouped bars compare recheck pass and
false-pass rates for source-trace validation and the matched self-validation control; higher and lower are better,
respectively. (b) Each circle is one app's paired workflow-fidelity difference between Trace2World and the
comparator; diamonds are macro-app means, positive values favor our method, and the shaded region
marks reversals. No resampled interval is estimated.}
\label{fig:fidelity-reviewer-summary}
\end{figure}

\paragraph{Where each contract loses correctness.}
\label{app:diagnostic-profile}
Table~\ref{tab:fidelity-full} separates reaching an endpoint from the obligations that follow it. Reach is
the highest diagnostic for every condition, so the informative quantity is the distance from a condition's
Reach to its own weakest diagnostic: 21.9 pp for Direct Agent + traces, 15.4 pp for CUA-Gym + traces,
18.5 pp for PhoneWorld-style, 18.2 pp for GUI-GENESIS-style, and 4.3 pp for Trace2World. Which diagnostic
sits at that minimum also differs. The three builders that target one shared application without
reconciling observations across traces bottom out on Data or Effect, that is, on the values and
transitions behind a surface that already looks right. For these three, every diagnostic except Reach
fails more often than it passes, and given the same recordings PhoneWorld-style's screen taxonomy and
transition graph add only 1.8 pp in Surface and 2.5 pp in workflow fidelity over CUA-Gym + traces.
GUI-GENESIS-style is the only builder whose
weakest diagnostic is State, the obligation a task-local site is never required to carry from one
workflow to the next. The self-validation control reopens the distance to 13.5 pp while holding Reach at
98.5\%, so the compression is a property of the validation evidence rather than of Trace2World's stages
alone. Withholding recordings moves structure more than state: from w/o traces to + traces, Direct Agent
gains 22.9 pp on average over Reach and Surface against 18.6 pp over State, Effect, and Terminal, and
CUA-Gym gains 16.8 pp against 11.7 pp, while workflow fidelity stays below 20\% in both. Recordings therefore
buy reachable structure faster than they buy semantics. These are outcome profiles over complete
construction contracts: they locate which obligations fail, not which construction stage caused it.

\paragraph{Why the profiles take this shape.}
One interpretation is that the difficult commitment is semantic rather than implementation-level. When
semantic inference and implementation are coupled, an uncertain page identity, state value, or action
effect must become a concrete coding choice before it can remain an explicit conflict, and later checks
mainly test that implementation of the chosen reading. Trace2World moves the commitment upstream:
model-proposed relations are factorized into explicit Structure, State, and Data, unresolved conflicts
stay outside the compiled support, bounded API compilation emits the initial package from the
resulting page specifications, and agentic repair edits that package without redefining their
semantics. This is consistent with the high reach but low workflow fidelity above
and with Appendix~\ref{app:matched-baseline-trajectories}, where a task-local builder reproduces one
workflow exactly while task-local correctness does not determine state shared across workflows. The
ordering is also not explained by assigning Trace2World the stronger constructor: its reconciliation and
initial compilation run on GPT-5.2, whereas every compared builder receives GPT-5.5 in the construction roles it defines
(Appendix~\ref{app:model-roles}). These remain mechanism-consistent interpretations of complete
construction contracts, not a stage-wise causal ablation.

A second factor may amplify the difficulty for native applications. A website ships its production
client to any crawler, and open pretraining corpora are correspondingly dominated by web content at
trillion-token scale \citep{soldaini2024dolma}. A real mobile application ships a binary: the largest
Android research collections distribute APK artifacts \citep{allix2016androzoo}, while source-linked
collections necessarily cover the open-source subset \citep{liu2020androzooopen}, which excludes the
large consumer applications studied here. Rendered interaction is therefore the broadly available
evidence of a real application's production semantics, which is what source traces provide. We treat
this as an amplifier rather than an explanation of Trace2World's advantage: cross-trace ambiguity is a
property of partial evidence rather than of the constructor, and a builder-written verifier remains
self-referential however strong that constructor becomes.

Figure~\ref{fig:fidelity-reviewer-summary}a isolates the matched validation-evidence control.
The source-trace recheck passes 176 evaluation executions: 175 P and one F; its 24
rejections comprise 19 F and five P. The self-validation recheck passes 160:
138 P and 22 F. These joint labels measure recheck precision. They do not establish why
an independently passing execution was rejected or where a failing condition diverged; those diagnoses
require the paired execution packets. Construction-time validation covers all 1,570 seed workflows
(Table~\ref{tab:automation-yield}); evaluation rechecks use the separate 200-task population.

\paragraph{Per-application direction of the contrasts.}
\label{app:coverage-correctness}
On the 200-task sample, Trace2World has 180 passes and 20 confirmed failures; every verdict is P or F,
so workflow fidelity is a complete count rather than a bound. The ten-app set is the complete evaluation
population, so paired contrasts are reported per application rather than as resampled intervals: with
ten applications as the unit of replication, Figure~\ref{fig:fidelity-reviewer-summary}b shows all ten
paired app differences and marks the macro-app mean; positive values favor Trace2World. Trace2World
exceeds the trace-conditioned and no-trace baselines in all ten apps. Against GUI-GENESIS-style (a
31.3-pp macro-app lead) and self-validation (12.0 pp), it is higher in nine apps and lower in one
(Taopiaopiao, $-5.9$ pp in both contrasts).
The only inferential statement is design-based. Under the sharp null that the construction contract
changes no application's workflow fidelity, the sign of each paired difference is exchangeable; a sign-flip
permutation test enumerates all $2^{10}=1{,}024$ sign assignments and counts those whose mean is at
least as extreme as the observed mean, which gives exact $p=4/1{,}024=0.004$ for the GUI-GENESIS-style contrast and $6/1{,}024=0.006$ for
self-validation, while the all-positive contrasts sit at the floor of $2/1{,}024=0.002$. The test assumes
no population of applications beyond the ten evaluated, and no interval is estimated.

\paragraph{A final value can conceal a wrong transition.}
The proportion of all task packets whose value-only check passes while their Effect
judgment is a confirmed failure is 14.3\% for PhoneWorld-style,
10.9\% for CUA-Gym + traces, 15.6\% for Direct + traces,
13.2\% for GUI-GENESIS-style, and 4.5\% for Trace2World.
Figure~\ref{fig:choice-boundary} illustrates the missing-confirmation form of this error.

\label{app:generated-closure-control}

\subsubsection{Human validation and automated-reviewer agreement}
\label{app:reviewer-validation}
\label{app:human-routes}
\label{app:evaluator-agreement}

The human audit relabels the same 200 workflows from the ten-app reconstruction roster under each of
the six trace-conditioned construction conditions, yielding 1,200 anonymous method--workflow packets.
Two annotators inspect the raw source trace and execution with the same six P/F questions; a third
adjudicates disagreements. No additional applications or performance population are introduced;
Table~\ref{tab:human-workflow} reports the adjudicated labels per condition only to locate the automated
reviewer's bias.

\begin{table}[H]
\centering
\caption{\textbf{Adjudicated human workflow-fidelity labels on the same 1,200 packets (\%).} Automated and human-adjudicated workflow fidelity are ten-app macro means over the 200 workflows of each condition; $\Delta$ is human minus automated in pp. Joint counts are human/automated labels pooled over the 200 workflows. The human column qualifies the automated reviewer's labels; it is not a second performance condition.}
\label{tab:human-workflow}
\footnotesize
\setlength{\tabcolsep}{4pt}
\begin{tabular}{@{}lccrrrrr@{}}
\toprule
Method & Automated & Human & $\Delta$ & P/P & P/F & F/P & F/F \\
\midrule
Direct Agent + traces & 16.7 & 19.4 & 2.7 & 26 & 12 & 7 & 155 \\
CUA-Gym + traces & 19.9 & 22.1 & 2.2 & 31 & 12 & 7 & 150 \\
PhoneWorld-style (reimpl.) & 22.4 & 23.8 & 1.4 & 37 & 10 & 7 & 146 \\
GUI-GENESIS-style (reimpl.) & 59.8 & 63.6 & 3.8 & 108 & 18 & 10 & 64 \\
\textbf{Trace2World} & 91.1 & 93.5 & 2.4 & 178 & 9 & 4 & 9 \\
\rowcolor{t2wRowGray}\multicolumn{8}{@{}l}{\emph{Control (not our method)}} \\
Self-validation & 79.1 & 80.7 & 1.6 & 149 & 12 & 9 & 30 \\
\bottomrule
\end{tabular}
\end{table}

The final row of Table~\ref{tab:fidelity-full} reports exact automated--human
agreement for workflow fidelity and each diagnostic on the same audit packets. Workflow-label macro-F1 is
90.2\%, with 92.3/87.9\% pass
precision/recall; the lower recall is consistent with the automated reviewer's conservative bias.
Under adjudicated human labels every condition scores 1.4--3.8 pp higher than under
the automated reviewer, so the bias is shared across routes rather than favoring one of them; the ordering of
Table~\ref{tab:modeling-main} is unchanged, and our method's margins over GUI-GENESIS-style and self-validation are
29.9 and 12.8 pp
under human labels (Table~\ref{tab:human-workflow}).
The two original annotators agree on
87.8\% of all-six labels before adjudication, and 160 packets
require adjudication in at least one dimension. These agreement statistics qualify the automated scores;
they do not create a second application or performance population \citep{lu2025agentrewardbench}.

\subsection{Construction Scaling, Validation, and Cost}
\label{app:construction-results}

These quantities describe construction inputs, context, and cost; they are separate from policy
improvement under additional training tasks.

\subsubsection{Context and call scaling with longer trace prefixes}
\label{app:scaling-detail}

\begin{figure}[H]
\centering
\includegraphics[width=\linewidth]{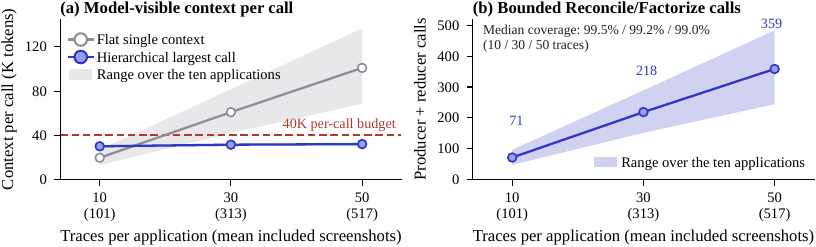}
\caption{\textbf{Context-bounded induction across ten applications and nested 10/30/50-trace prefixes.}
The horizontal axis is the trace prefix; the value in parentheses is the mean number of in-app screenshots
included at that prefix, which is the unit the context and call estimates are computed from. Lines are
ten-application means and shaded bands span those applications. (a) Flat single-context input (gray) exceeds
the 40K budget at every 30- and 50-trace point, whereas the largest hierarchical call (blue) remains within
28.7--33.2K tokens. (b) Bounded \textsc{Reconcile}/\textsc{Factorize} producer-plus-reducer calls; in-panel
values are application means. Median audited semantic coverage at 10/30/50 traces is shown in the panel.}
\label{fig:scaling-main}
\end{figure}

Across all 50 environments, peak model-visible context has a median/P95/maximum of
28.9/31.7/33.2K tokens, no call exceeds the 40K budget, semantic coverage is 99.1\%, and the
conservative fallback rate is 1.8\%. The ten-application replay resolves this by trace prefix while still
estimating from included screenshots (101, 313, and 517 on average).
All flat inputs fit at 10 traces (12.7--26.8K tokens), but every 30- and 50-trace input exceeds the 40K
budget (42.0--136.8K), whereas the largest hierarchical call stays within 28.7--33.2K; producer/reducer
calls grow from 47--96 at 10 traces to 244--484 at 50. The near-linear growth in traces is compositional
rather than assumed: about 0.70 calls per included screenshot, and 10.1--10.4 screenshots per trace at
every prefix. Screenshots per trace range 7.0--13.3 across applications, which the band width carries.
These call counts cover \textsc{Reconcile} and \textsc{Factorize}, not later implementation, verification, or
repair. Every bounded
result must still pass the source-coverage, mapping, support, and contradiction audits, so partitioning
changes work scheduling without changing which behavior the evidence supports. Coverage is evaluated against the
source-linked obligations of the current prefix; later traces contribute to a separate full-evidence
coverage denominator. Flat execution is a comparison only where its input fits, not an assumed
reference beyond the context limit.

\subsubsection{Reconstruction quality and evidence compression with more source traces}
\label{app:artifact-scaling}

\begin{figure}[H]
\centering
\includegraphics[width=\linewidth]{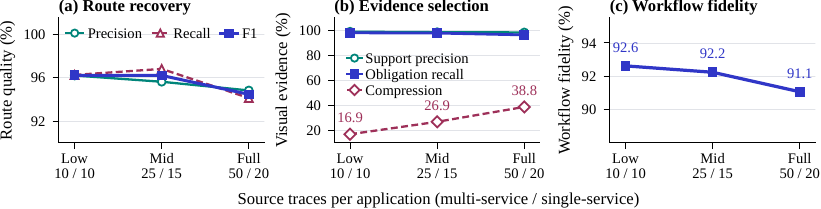}
\caption{\textbf{Shared-semantics reconstruction under expanding trace support.} Low/Mid/Full use
10/25/50 traces per multi-service app and 10/15/20 per single-service app; panels report route recovery,
evidence selection, and workflow fidelity against each budget's own induced target.}
\label{fig:artifact-trace-scaling}
\end{figure}

Figure~\ref{fig:artifact-trace-scaling} uses the ten applications excluded from policy training and the
per-application workflow counts of Table~\ref{tab:modeling-main}. Nested Low/Mid/Full subsets use
10/25/50 traces per multi-service app and 10/15/20 per single-service app (100, 180, and 290 traces in
total), and each subset is scored against its own induced routes, visual obligations, and workflows.
For each subset, route precision and recall divide matched routes by predicted and subset-supported
routes. For the visual evidence set of Appendix~\ref{app:evidence-projection}, support precision is the
supported share of selected screenshots, obligation recall is the covered share of source-supported
visual obligations, and compression is one minus selected over candidate screenshots. All displayed
rates are macro-averaged over applications.

\begin{figure}[H]
\centering
\includegraphics[width=\linewidth]{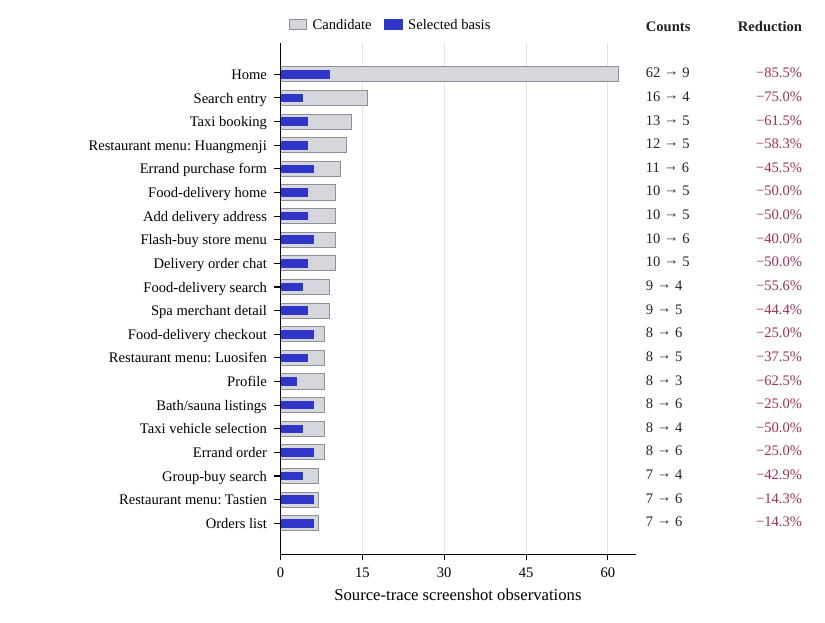}
\caption{\textbf{Top-20 pagewise screenshot compression in Meituan.} Pages from the 50-trace
source-trace set are ranked by candidate observations; ties use the first source-trace occurrence and then
filename. Light bars count candidate observations and blue overlays count the selected visual evidence set;
right-side counts use candidate $\rightarrow$ selected order. These pages retain 105 of 241 observations
and account for 136 of the 166 reductions across all 159 pages (81.9\%).}
\label{fig:meituan-top20-page-compression}
\end{figure}

From Low to Full, supported routes grow from 270 to 512 and visual obligations from 553 to 1,076. Route
F1 moves from 96.2\% to 94.5\%, visual-support precision and obligation recall from 99.0/98.2\% to
98.4/96.5\%, and workflow fidelity from 92.6\% to 91.1\%, while evidence compression rises from 16.9\%
to 38.8\%. Figure~\ref{fig:meituan-top20-page-compression} below localizes the visual-selection
reduction on the twenty Meituan pages with the most candidate observations.

\subsubsection{Automated validation and repair-wave accounting}
\label{app:automation-yield}

Trace-backed seed workflows and generated candidates follow related but distinct funnels.
Every seed workflow is a source obligation and must validate within the $K=3$ repair cap. Generated
candidates must first pass support, then validation, and finally learner--solver filtering.
Table~\ref{tab:automation-yield} keeps those denominators separate and reports newly closed cases at
each repair wave.

\begin{table}[H]
\centering
\caption{\textbf{Validation and admission funnels.} Read each population from left to right.
Percentages name their denominator; $+$ counts are cases newly closed at that repair wave rather than
cumulative totals. Fresh-GUI validation is capped at $K=3$.}
\label{tab:automation-yield}
\footnotesize
\setlength{\tabcolsep}{3pt}
\renewcommand{\arraystretch}{1.12}
\begin{tabularx}{\linewidth}{@{}>{\raggedright\arraybackslash}p{0.17\linewidth}>{\raggedright\arraybackslash}p{0.16\linewidth}>{\raggedright\arraybackslash}X>{\raggedright\arraybackslash}p{0.21\linewidth}@{}}
\toprule
Population & Evidence support & Fresh-GUI validation and repair & Training use \\
\midrule
\textbf{Seed workflows}\newline 1,570 across 50 apps &
1,570 pass\newline (100\% of seeds) &
1,171 at wave 0 (74.6\%); $+261/+103/+35$ at waves 1/2/3;
1,570 validated, 0 unresolved &
1,280 seeds from the 40 training apps \\
\addlinespace[2pt]
\textbf{Generated candidates}\newline 1,152 proposals across 40 apps &
985 pass\newline (85.5\% of proposals) &
673 at wave 0 (68.3\% of supported); $+113/+46/+21$ at waves 1/2/3;
853 validated (86.6\%), 132 not validated by $K=3$ &
768 admitted\newline (90.0\% of validated) \\
\bottomrule
\end{tabularx}
\end{table}

The 1,280 seed tasks are the workflows belonging to the 40 policy-training applications, so that
last step is an application partition rather than a rejection gate. Of the 132 support-passing
candidates not validated by $K=3$, 60 have an incomplete dependency slice, 43 lack an
evaluator-observable outcome, and 29 retain an unreproduced environment defect. Of the 85 validated
candidates not admitted, 49 are saturated for the learner, 24 remain unsolved by the solver, and 12
are deferred.

The self-validation control also validates all 1,570 builder executions: 1,294 (82.4\%)
at wave 0, then 173, 72, and 31 after one, two, and three repair waves. Its mean 0.26 waves per
workflow is lower than Trace2World's 0.36. These are builder-execution events; the independent
execution rechecks in Figure~\ref{fig:fidelity-reviewer-summary}a use separate packets. We do not
convert repair waves into model calls because the preserved logs do not expose a comparable
stage-level call boundary for this control.

\paragraph{Repair-round interpretation.} A repair round is one edit made from failure evidence followed
by fresh re-verification, whatever the number of requests inside it; a shared fix that closes several
tasks is counted once. Table~\ref{tab:automation-yield} gives the complete wave distribution. All seed
workflows close by $K=3$ (mean 0.36 rounds), whereas 13.4\% of support-passing augmented candidates
reach the cap and are dropped. A round is an observable repair outcome, not a fixed number of model turns.

\subsubsection{Construction cost}
\label{app:construction-cost}

The matched construction comparison uses the same ten-app, 200-workflow roster as Table~\ref{tab:modeling-main}
under the same construction models, reasoning setting, tools, trace bundle, and eight-worker schedule,
with a 6-hour per-app wall-clock limit. Model calls are metered but not capped. Following CUTV, one call
is one completed model-execution step (one model response); multiple tool actions emitted by that response
still count once. Trace2World applies this unit to two disjoint parts: the bounded API pipeline that produces
the initial Package, and an agentic workflow that performs package-level initial verification followed by
the verify--repair--reverify fixloop.
Elapsed time includes each app's verification and repair but excludes independent scoring and human
pauses.

\begin{table}[H]
\centering
\caption{\textbf{Mean construction budget and Trace2World time/call decomposition.} (a) Per-application totals
on the matched ten-app roster; Trace2World reports each total on the first line and (API pipeline + agentic
workflow) below. (b) Share of the 602 mean API-pipeline calls by phase; wedge area and legend percentages
use the pipeline-only denominator. Generated-task construction is accounted separately.}
\label{tab:construction-budget}
\footnotesize
\begin{minipage}[t]{0.47\linewidth}
\vspace{0pt}
\centering
\textbf{(a) Mean construction cost}\par\smallskip
\setlength{\tabcolsep}{2pt}
\begin{tabularx}{\linewidth}{@{}>{\raggedright\arraybackslash}Xrr@{}}
\toprule
Builder & Hours & Calls/app \\
\midrule
Direct w/o traces & 0.92 & 248 \\
Direct + traces & 1.27 & 319 \\
CUA-Gym w/o traces & 1.82 & 344 \\
CUA-Gym + traces & 2.14 & 392 \\
PhoneWorld-style & 5.02 & 1,118 \\
GUI-GENESIS-style & 5.90 & 2,415 \\
Trace2World seed & \makecell[r]{3.60\\(0.50 + 3.10)} & \makecell[r]{1,160\\(602 + 558)} \\
\bottomrule
\end{tabularx}
\end{minipage}\hfill
\begin{minipage}[t]{0.51\linewidth}
\vspace{0pt}
\centering
\textbf{(b) API-pipeline phase share}\par\smallskip
\includegraphics[width=\linewidth]{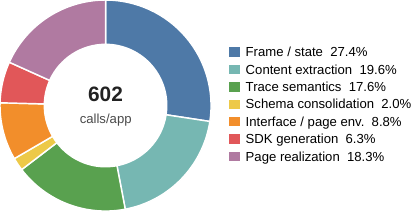}
\end{minipage}
\end{table}

At the reported 1,160-call mean, 602 calls (51.9\%) belong to the API pipeline and 558 (48.1\%) to the
agentic workflow. The 3.60-hour elapsed-time estimate separates a 0.50-hour highly concurrent pipeline
from a 3.10-hour agentic workflow, estimated as 558 completed model-execution steps at 20 seconds per
step. The agentic workload covers initial verification of all 20 matched workflows plus projected repair
and re-verification work on the shared Package. Generated-task construction remains outside this matched
seed comparison; the DAPO runs use the RL-iteration schedules of Appendix~\ref{app:training-config}.

\subsubsection{Runtime cost across the 50-app portfolio}
\label{app:runtime-cost}

We profile all 50 applications on one eight-core Xeon host with 61.4\,GiB of RAM. All apps share one
serving process and read-only packages. Table~\ref{tab:runtime-cost} reports per-application package
size (50 apps), process launch to a healthy API (ten trials), first task initialization after display
acquisition (one task per app), warm task reinitialization after teardown (three per app), and
alternating click/scroll \texttt{/step} probes (ten per app). The warm timer starts \emph{after}
teardown; the step probe excludes screenshot acquisition and model inference. P95 is an empirical
nearest-rank value.

\begin{table}[H]
\centering
\caption{\textbf{Runtime cost of the full 50-app portfolio.} Each app contributes to package and
task-timing summaries; host starts use ten trials. Units and sample counts are explicit for each row.}
\label{tab:runtime-cost}
\footnotesize
\setlength{\tabcolsep}{4pt}
\rowcolors{3}{t2wRowGray}{white}
\begin{tabularx}{\linewidth}{@{}>{\raggedright\arraybackslash}Xrrr@{}}
\toprule
Metric & $n$ & Median & P95 \\
\midrule
Package on disk, raw / app & 50 & 18.9\,MiB & 46.3\,MiB \\
Package on disk, tar.gz / app & 50 & 8.2\,MiB & 20.8\,MiB \\
Writable state / episode & 50 & 0\,KiB & 412\,KiB \\
Host process to healthy API & 10 & 3.06\,s & 3.42\,s \\
First task initialization after display acquisition & 50 & 258\,ms & 555\,ms \\
Warm task reinitialization, init only & 150 & 229\,ms & 512\,ms \\
Click/scroll \texttt{/step} round trip & 500 & 94\,ms & 365\,ms \\
\bottomrule
\end{tabularx}
\rowcolors{1}{white}{white}
\end{table}

The idle host occupies 420\,MiB PSS. At 16 concurrent episodes, each additional episode accounts
for 164\,MiB of system \texttt{MemAvailable} loss or 218\,MiB of marginal PSS. A 24-episode,
six-call-per-episode click/scroll probe reaches 82.6 calls/s with 174/915\,ms median/P95 round-trip
latency. These are API probes, not complete agent rollouts. For context, MobileGym reports about
400\,MB per browser instance, while its comparison puts AndroidWorld at about 4.5\,GB, AndroidLab at
about 6\,GB, and MobileWorld at at least 4.5\,GB \citep{wu2026mobilegym}. The latter values use the
source paper's respective measurement or estimate and are not a matched hardware comparison.

\subsubsection{Source-trace evidence for defect review and reproduce-before-repair}
\label{app:review-repair-results}

\paragraph{Source evidence improves defect coverage and review precision together.}
We freeze 400 executions from a separately selected 20-app construction-review set: ten defective and ten clean task/revision
pairs per app. Two independent annotators establish defects and blocking severity, with adjudication,
before the model reviewers run. Each execution receives three isolated reviews with the same model,
output schema, and token cap: execution only; source trace plus execution with an open-ended prompt;
and the same source-trace/execution evidence with the contrastive rubric. The last pair isolates the
rubric; the first pair isolates source-trace access. No reviewer sees another reviewer's findings. The balanced
sample estimates sensitivity and clean-case rejection separately, not their deployment prevalence.

\begin{table}[H]
\centering
\caption{\textbf{Construction-time review ablation.} All rows use the same model, output schema, and token cap; each lower row adds the named evidence or instruction. The 200 defective executions contain 599 adjudicated defects. Defects found is defect-level coverage; reported defects correct is precision. The last two columns use separate sets of 200 defective and 200 clean executions.}
\label{tab:review-coverage}
\footnotesize
\setlength{\tabcolsep}{4pt}
\begin{tabular*}{\linewidth}{@{\extracolsep{\fill}}lcccc@{}}
\toprule
Review condition & \makecell{Defects found\\(\%) $\uparrow$} & \makecell{Reported defects\\correct (\%) $\uparrow$} & \makecell{Defective execs.\\missed / 200 $\downarrow$} & \makecell{Clean execs.\\flagged / 200 $\downarrow$} \\
\midrule
Execution only & 43.2\% & 73.6\% & 34 & 42 \\
$+$ source trace & 68.4\% & 88.2\% & 8 & 28 \\
\textbf{$+$ contrastive rubric} & \textbf{86.0\%} & \textbf{93.0\%} & \textbf{1} & \textbf{20} \\
\bottomrule
\end{tabular*}
\end{table}

Across the 20 applications, defect-coverage ranges are
29.0--58.6\%,
51.6--79.3\%, and
77.4--96.6\% for the three rows.
The ordering is stable across applications: the rubric reviewer covers more defects than the plain
reviewer in 19 of 20 apps (tied in 1), and the plain reviewer more than the execution-only reviewer in
18 of 20 (tied in 1, lower in 1).


\paragraph{Regressions caught by re-execution from step zero.} Over the 50 applications, 1,884 environment
edits were made during repair. Re-executing every dependent trace from step zero after each edit
found 233 (12.4\%) that broke a previously passing dependent packet; reusing the stale suffix of the
earlier execution would have certified 214 of them. Fresh execution incurs the reported 1.9$\times$ repair-loop execution overhead.

\paragraph{Reproduce-before-repair.}
We sample 400 initially environment-attributed failures, 20 per audit app, with one task/revision per
sample. Each receives three fresh executions with the same reset and account state. Independent
adjudicators inspect all attempts and state evidence and assign every case an owner: 253 are persistent environment defects, 32
intermittent environment defects, and 115 model/task errors. Thus one non-reproduction
does not justify reclassification. Of the 285 confirmed environment defects, 274 close within $K=3$;
the remaining 11 retain their status. These counts describe the sampled
failure population, not the incidence of defects in all rollouts.


\subsection{Policy Gains from Learner-Guided Task Supply}
\label{app:policy-results}
\label{app:task-supply-results}

\subsubsection{Evaluation metrics, benchmark protocols, and matched conditions}
\label{app:held-out-protocol}

The 1,044 Trace2World-Bench identities enter no training batch. Table~\ref{tab:evaluation-sets}
fixes its four disjoint slices and signature split. Each task has three complete evaluation attempts
under the construction-class budget of Table~\ref{tab:dapo-config}: seed-family and unseen-app tasks
use 30 steps, augmentation-family tasks use 45, and evaluation-only cross-app tasks use 60.
The same family cap applies to single-app training and evaluation, and both metrics use all 1,044 tasks.
Infrastructure, provider, and environment failures are resolved and retried, with their counts
reported separately; they neither count as agent failures nor remove a task from the denominator.
No validation split is used; Appendix~\ref{app:policy-optimization} fixes the terminal checkpoint.
The 100 cross-application tasks follow the persistent runner and held-out joint-oracle
boundary of Appendix~\ref{app:train-test-contract} and contribute once each to the benchmark denominator.

\paragraph{Partial reward and terminal success.}
For task $t$ and valid attempt $j$, let $q_{tj}\in[0,1]$ be the programmatic task evaluator's checkpoint credit
of Eq.~\ref{eq:task-reward} on the attempt's final business state, and let $s_{tj}\in\{0,1\}$ be its terminal verdict.
For the full held-out task set $T$ and its three completed attempts per task, we report
\begin{equation}
\mathrm{PR}=\frac{100}{3|T|}\sum_{t\in T}\sum_{j=1}^{3}q_{tj},
\qquad
\mathrm{SR}=\frac{100}{3|T|}\sum_{t\in T}\sum_{j=1}^{3}s_{tj}.
\label{eq:agym-pr-sr}
\end{equation}
Both metrics average the same three attempts within each task and weight all tasks equally. PR retains
partial completion; SR credits each attempt only for complete terminal success. The three training
seeds form the outer mean and SD, separately from a task's three evaluation attempts.

\label{app:external-performance}
\paragraph{Absolute capability and controlled training gain.}
The reference GUI agents in Table~\ref{tab:downstream-main} provide absolute capability context;
their backbones, GUI pretraining, post-training objectives, and interaction scaffolds can differ
\citep{qin2025uitars,zhou2025maiui,zhou2026qwenuiagent}. UI-Venus-2-9B, for example, also starts
from Qwen3.5-9B but adds multimodal trajectory mid-training, domain-specific step-level offline RL,
and multi-teacher on-policy distillation over mobile, web, desktop, grounding, and CAPTCHA tasks
\citep{venus2026uivenus2}. Our final policy starts from Qwen3.5-9B without SFT warm-up and completes 52 DAPO
iterations on 2,048 single-app training tasks (Table~\ref{tab:dapo-config}); the external suites are
reserved for evaluation. Comparing it with the base holds initialization and evaluation
fixed for this training intervention. UI-Venus-2 supplies absolute capability context under a different
post-training program, rather than a matched training control.
Agent frameworks that wrap a general model are not compared because each suite's protocol fixes the agent template.

\paragraph{Protocols.} Each external column is produced under its benchmark's own protocol: task set,
agent template, observation, step budget, trial count, decoding temperature, and success criterion.
Our policies enter every suite through the suite's screenshot-only observation and its released action
adapter, with no Trace2World environment, evaluator, or step budget carried over.
\emph{MobileGym-Bench} resets before each task and grants 15/30/45/60 steps by level (+15 for
AnswerSheet tasks); general models use the \texttt{generic\_v2} template and specialized GUI agents
their native adapters; SR averages four parameter-resampled trials for open-weight policies and one
run for proprietary policies, and partial reward (PR) is the suite evaluator's partial credit over those
trials. \emph{AndroidWorld} runs the standard 116-task suite on the reference emulator with its
programmatic checkers, a single run per task, and the suite's per-task step budget, which the
maintainers set at about twice the human completion length. \emph{MobileWorld} evaluates its GUI-only
track of 117 tasks with a 50-round limit per task, single run, and backend-state verification, which
is what SR@50 denotes; its 44 user-interaction and 40 MCP tasks are outside this evaluation.
\emph{AndroidDaily} runs the 235 end-to-end tasks on live applications with its LLM-based completion
judge and reports single-run SR; the released protocol fixes no step limit beyond the judge's
termination, so none is imposed. Reference models are run by us only on Trace2World-Bench; their external-suite cells are the
published values at each suite's reported setting and are marked $\dagger$ in the Appendix owner
tables. These
suite-specific metrics are separate from Trace2World-Bench's mean of three complete attempts per task
under the budget of Table~\ref{tab:dapo-config}.

Our three policies are means over three training seeds; the reference rows are single published
checkpoints and carry no seed spread, so Table~\ref{tab:downstream-main} prints means only. Across
seeds our policies vary by 0.5--1.5 points on the suite headline scores and by 1.0--3.9 points on the
slices, widest on AndroidDaily's 24-task high-ambiguity slice and narrowest on its 164-task
low-ambiguity slice.

Each complete result table in Appendix~\ref{app:benchmark-tables} states its native slice
denominators. These released partitions are protocol-specific and distinct from Trace2World's four
overlapping task capabilities; no slice is pooled across suites.

\paragraph{Matched task-supply conditions.}
\label{app:task-supply-baselines}

Seed continuation draws 384 seed tasks without replacement within each later stage; entity rebinding uses
two new 384-task training batches that retain sampled seed workflow templates while changing their bound
entities. These 768 variants are training-only identities and do not enter Trace2World-Bench. Static augmentation uses two new 384-task
batches generated with the learner-guided condition's prompts, difficulty principles, support checks,
environment validation, and fixed-solver screen, but authored before $\pi_2$ exists, without learner
task outcomes, saturation estimates, or learner-guided refinement, and selected by random draw from the solver-solvable
candidates. All three fork from $\pi_1$, follow the same 20-iteration continuation to RL iteration 52,
and are evaluated at the same iteration checkpoints;
initialization at the fork, sampler, environment revisions, and evaluator are matched. Two
longer-schedule controls select all validated candidates without screening, or continue the seed tasks
until their retained groups match the learner-guided run; these population-matched and signal-matched
comparisons are reported with the task-generation results below.

\subsubsection{Learner feedback targets harder tasks while preserving solvability}
\label{app:generation-results}

A useful learner-guided supply must fill each batch with tasks that expose current-policy weaknesses
while a fixed stronger solver can still complete them. Each round realizes 576 of at most 640 candidate
slots ($24\times10+16\times25$); Table~\ref{tab:stage-funnel} separates the resulting funnel from the
screening and composition shift.

\begin{table}[H]
\centering
\caption{\textbf{Feedback rounds preserve training yield while shifting learner-relative difficulty and task composition.}
(a) Candidate funnel for R1/R2, which follow stages 1/2. (b) Mean pass over four fresh attempts per
validated task and dependency-heavy share among selected; lower learner pass means harder tasks and fixed-solver pass measures solvability.
Dependency-heavy comprises cross-surface, lifecycle-revisit, and recovery tasks.}
\label{tab:stage-funnel}
\scriptsize
\noindent
\begin{minipage}[t]{0.42\linewidth}
\vspace{0pt}
\centering
\setlength{\tabcolsep}{2pt}
\begin{tabular*}{\linewidth}{@{\extracolsep{\fill}}lcccc@{}}
\multicolumn{5}{@{}l}{\makecell[l]{\textbf{(a) Candidate funnel}\\[-1pt]\textbf{and fixed yield}}} \\[-2pt]
\toprule
Round & Proposed & \makecell{Source-\\supported} & Validated & Selected \\
\midrule
R1 & 576 & 499 & 426 & 384 \\
R2 & 576 & 486 & 427 & 384 \\
\bottomrule
\end{tabular*}
\end{minipage}
\hfill
\begin{minipage}[t]{0.56\linewidth}
\vspace{0pt}
\centering
\setlength{\tabcolsep}{2pt}
\begin{tabular*}{\linewidth}{@{\extracolsep{\fill}}lccc@{}}
\multicolumn{4}{@{}l}{\makecell[l]{\textbf{(b) Validated-task screening}\\[-1pt]\textbf{and selected-batch composition}}} \\[-2pt]
\toprule
Round & \makecell{Learner pass\\among validated (\%)} &
\makecell{Fixed-solver pass\\among validated (\%)} &
\makecell{Dependency-heavy\\among selected (\%)} \\
\midrule
R1 & 41 & 71 & 46 \\
R2 & 29 & 66 & 71 \\
\bottomrule
\end{tabular*}
\end{minipage}
\end{table}

Panel (a) shows that both rounds fill the same 384-task training batch. Panel (b) shows that the
current-learner pass fraction falls from 41\% to 29\%, while the fixed solver still passes 66\% and the
dependency-heavy share rises from 46\% to 71\%. The change is therefore in learner-relative difficulty
and task composition, not in batch size.

\paragraph{Solver as a solvability screen.}
We audit both selection decisions on the original, unmodified candidate: 20 solver-rejected and ten
solver-accepted candidates per audit app, 400 and 200 in total. Two annotators independently attempt
the original task and adjudicate disagreements; refined versions are scored separately. Of the rejected
candidates, 36 are solvable: a conditional rejection error of 9.0\% (0--5 of 20 per app). Of the accepted candidates, 194/200 are independently solvable.
These stratified samples estimate $P(\mathrm{solvable}\mid\mathrm{rejected})$ and
$P(\mathrm{solvable}\mid\mathrm{accepted})$; they do not estimate the fraction of all solvable tasks
that the solver rejects without the population selection weights.


\label{app:pool-results}

\begin{samepage}
\paragraph{Learner signal on supplied tasks.}
We evaluate validated and selected tasks after batch freezing, using the shared $\pi_1$ for R1 and each
condition's $\pi_2$ for R2. Figure~\ref{fig:task-supply-adaptation}a shows that fixed-solver static
selection yields only a modest signal gain, whereas learner feedback produces higher-signal batches.
Solvability and learner-relative training value therefore require separate screens; the four-attempt
signal rule is detailed in Appendix~\ref{app:policy-optimization}.
\end{samepage}

\paragraph{Bounded refinement raises learning signal without simplifying hard tasks.}
The gap is already present at first validation, before refinement. Table~\ref{tab:bounded-refinement}
shows that only 60/426 (14.1\%) and 62/427 (14.5\%) validated lineages are \textsc{Refinable}; the
fixed two-round flow makes 36/60 and 37/62 of them selectable and raises validated-lineage signal by
5.6/5.4 pp. Hard and solver-unsolved lineages are never simplified. The gap persists under the opposite
$\pi_2$ (73.4\% vs. 41.9\%) and eight-trajectory groups (78.9\%/75.8\% vs. 56.8\%/52.9\%).

\begin{table}[H]
\centering
\caption{\textbf{Bounded hardening acts on about 14\% of validated lineages, recovers about 60\% within
two rounds, and raises validated-lineage signal by 5.4--5.6 pp.}
(a) Outcomes compare R1/R2; recovered-total percentages use the refinable pool, and the recovery and held
rows partition it. (b) Signal is the validated-lineage share with four-attempt reward SD $>0.1$; saturation
uses the 384 selected tasks. \textsc{Refinable} means fixed-solver success and learner mean reward above 0.8.
Each round separately excludes 12 solver-unsolved and 6 deferred lineages outside this pool. Hard and
solver-unsolved lineages are never simplified, and the $K=3$ implementation repair is separate.}
\label{tab:bounded-refinement}
\scriptsize
\noindent
\begin{minipage}[t]{0.57\linewidth}
\vspace{0pt}
\centering
\begin{tabular*}{\linewidth}{@{\extracolsep{\fill}}lcc@{}}
\multicolumn{3}{@{}l}{\textbf{(a) Refinement accounting}} \\[-2pt]
\toprule
Outcome & R1 & R2 \\
\midrule
Refinable / validated & 60 / 426 (14.1\%) & 62 / 427 (14.5\%) \\
Recovered after first & 29 & 30 \\
Recovered after second & 7 & 7 \\
Recovered total & 36 (60.0\%) & 37 (59.7\%) \\
Held: no supported hardening & 9 & 9 \\
Held: still easy at cap & 15 & 16 \\
\bottomrule
\end{tabular*}
\end{minipage}
\hfill
\begin{minipage}[t]{0.405\linewidth}
\vspace{0pt}
\centering
\begin{tabular*}{\linewidth}{@{\extracolsep{\fill}}lcc@{}}
\multicolumn{3}{@{}l}{\textbf{(b) Effect on learning signal}} \\[-2pt]
\toprule
Metric & R1 & R2 \\
\midrule
Signal: first valid (\%) & 66.0 & 64.6 \\
Signal: post-refinement (\%) & 71.6 & 70.0 \\
Signal gain (pp) & +5.6 & +5.4 \\
Saturated among selected (\%) & 8.6 & 3.9 \\
\bottomrule
\end{tabular*}
\end{minipage}
\end{table}

\paragraph{MobileGym and budget-robustness checks for task augmentation.}
\label{app:task-supply-performance}
Figure~\ref{fig:task-supply-adaptation}b,c owns the iteration- and rollout-matched Trace2World-Bench
SR/PR trajectories; only the static and learner-guided augmentation branches also match on the
45-step task horizon.
At RL iteration 52, MobileGym SR is $19.7\pm1.2$, $21.0\pm0.8$, $22.2\pm1.3$, and
$25.2\pm1.2$ for seed continuation, entity rebinding, static augmentation, and learner-guided batches.
Two different-budget controls test robustness: selecting all 426/427 validated candidates runs for
54 iterations, retains 1,244 informative groups, and reaches Trace2World-Bench SR/PR
$57.8\pm1.1$/$68.6\pm0.8$ and MobileGym SR $23.4\pm1.3$; matching the learner-guided run's
1,351 retained groups with seed continuation requires 79 iterations and reaches
$54.3\pm0.8$/$67.0\pm0.6$ and $20.1\pm1.0$. Because these controls change the optimization
budget, they test robustness and do not enter the same-budget causal contrast.

\subsubsection{New task compositions stay within source-supported behavior}
\label{app:alternative-path-results}
\begin{figure}[H]
\centering
\includegraphics[width=\linewidth]{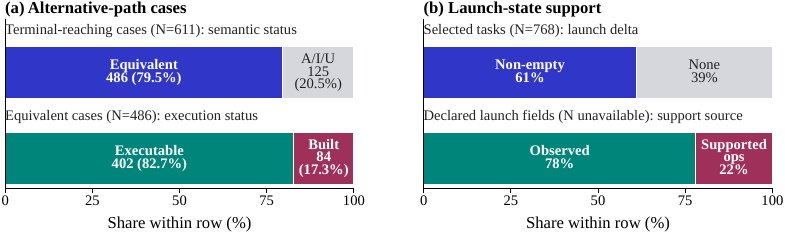}
\caption{\textbf{Alternative paths and launch states remain source-supported.} (a) Of 611 terminal-reaching
attempt cases on paths absent from the recorded traces, 486 are semantically equivalent; within those
486 cases, 402 already execute and 84 are materialized from supported edges. A/I/U denotes ambiguous,
invalid, or unresolved. Counts are attempt-associated cases, not deduplicated paths. (b) Of 768 selected
generated tasks, 61\% declare a non-empty launch delta; among declared launch fields, 78\% use directly
observed contexts and 22\% are reachable by supported operations. The task and field rows use different
denominators; the field count is unavailable.}
\label{fig:alternative-path-launch-states}
\end{figure}

\begin{figure}[H]
\centering
\includegraphics[width=\linewidth]{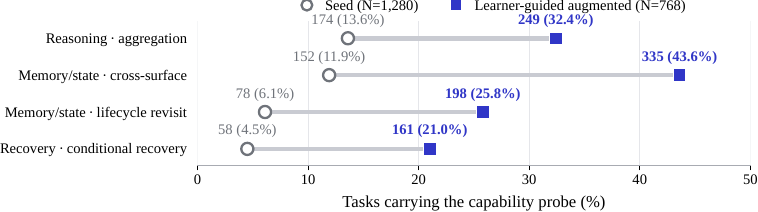}
\caption{\textbf{Task augmentation raises the prevalence of capability probes.} Each row compares, within
1,280 seed tasks and 768 learner-guided augmented tasks, the share carrying the named probe;
endpoint labels give count (share). Rows are overlapping task-level marginals, so one task may contribute
to multiple rows and the shares do not form a partition.}
\label{fig:task-capability-prevalence}
\end{figure}

Task augmentation may recombine supported operations beyond any recorded path. The path and launch-state
audit separates two conditional accounting chains to test whether departures from literal replay remain
supported. Figure~\ref{fig:alternative-path-launch-states} and Figure~\ref{fig:task-capability-prevalence}
use separate units and denominators; their counts are not pooled. Every semantically equivalent path case
is already executable or materialized from supported edges, while the remaining 125 ambiguous, invalid,
or unresolved entries remain outside support.

\paragraph{Task augmentation raises capability-probe prevalence.}
\label{app:task-dimensions}

Figure~\ref{fig:task-capability-prevalence} compares four directly supported capability probes in the seed and
learner-guided task supplies: aggregation for reasoning, cross-surface composition and lifecycle revisit
for memory/state, and conditional recovery for recovery/backtracking. Long-horizon requires the full
task-authoring criteria
(Appendix~\ref{app:expansion-loop}) and has no equivalent marginal in this accounting.
Task-level joint labels are unavailable, so capability unions are not inferred from these marginals.

Cross-surface shows the largest increase, from 11.9\% to 43.6\%; aggregation, lifecycle revisit, and
conditional recovery rise from 13.6/6.1/4.5\% to 32.4/25.8/21.0\%. These changes describe which tasks are supplied for
training; their matched outcome remains in Figure~\ref{fig:task-supply-adaptation}b,c.

\subsubsection{Complete Trace2World-Bench and external benchmark results}
\label{app:benchmark-tables}
\label{app:application-holdout}
\label{app:external-breakdowns}

Table~\ref{tab:evaluation-sets} defines the six Trace2World-Bench result columns and owns their
sizes and trial protocol. The first table reports each model's SR / PR pair across
all six columns. The following
tables report each evaluated external track's native endpoint and every slice available in the frozen
result owners; their protocols and denominators are defined in
Appendix~\ref{app:external-performance}.

\begin{table}[H]
\centering
\caption{\textbf{Complete Trace2World-Bench model results.} Each model occupies one row and each result cell reports terminal SR / mean partial reward PR, both in percent. Columns cover the three task strata, their overall result under Training apps, unseen applications, and the complete benchmark across all apps. Table~\ref{tab:evaluation-sets} is the sole owner of partition sizes and the three-attempt protocol; Appendix~\ref{app:held-out-protocol} defines the evaluation contract.}
\label{tab:agym-family-models}
\footnotesize
\setlength{\tabcolsep}{2.0pt}
\begin{minipage}[t]{\linewidth}
\centering
\footnotesize
\setlength{\tabcolsep}{2.0pt}
\renewcommand{\arraystretch}{1.02}
\begin{tabular}{@{}lcccccc@{}}
\toprule
\multirow{2}{*}{Model} & \multicolumn{4}{c}{Training apps} & \multirow{2}{*}{\makecell{Unseen\\apps}} & \multirow{2}{*}{\makecell{All\\apps}} \\
\cmidrule(lr){2-5}
& \makecell{Entity\\replacement} & \makecell{Held-out\\workflow} & Cross-app & Overall & & \\
\midrule
\rowcolor{t2wRowGray}\multicolumn{7}{@{}l}{\emph{Proprietary Models}} \\
Qwen3.7-Plus & \textbf{76.3 / 83.4} & \textbf{70.6 / 78.4} & \textbf{59.0 / 67.8} & \textbf{71.0 / 78.7} & \textbf{68.8 / 76.4} & \textbf{70.6 / 78.3} \\
GPT-5.5 & 74.5 / 82.5 & 68.3 / 76.8 & 57.3 / 66.3 & 68.9 / 77.3 & 66.1 / 73.8 & 68.4 / 76.7 \\
Gemini 3.1 Pro & 75.1 / 82.2 & 69.0 / 76.2 & 58.0 / 67.0 & 69.5 / 76.9 & 65.7 / 71.9 & 68.9 / 76.1 \\
Gemini 3.5 Flash & 72.4 / 81.1 & 66.2 / 74.6 & 55.7 / 64.9 & 66.8 / 75.4 & 62.5 / 70.7 & 66.1 / 74.6 \\
\rowcolor{t2wRowGray}\multicolumn{7}{@{}l}{\emph{Open-source Models}} \\
UI-TARS-1.5-7B & 54.4 / 66.0 & 45.2 / 58.5 & 30.3 / 44.2 & 46.2 / 59.1 & 39.6 / 50.8 & 45.1 / 57.7 \\
AutoGLM-Phone-9B & 63.3 / 74.5 & 54.9 / 67.0 & 43.3 / 55.9 & 56.0 / 67.9 & 48.7 / 59.8 & 54.8 / 66.6 \\
UI-Venus-1.5-8B & 65.2 / 75.8 & 57.6 / 67.7 & 45.7 / 56.6 & 58.5 / 68.8 & 52.1 / 61.5 & 57.4 / 67.6 \\
MAI-UI-8B & 66.3 / 76.9 & 58.1 / 69.8 & 45.0 / 57.5 & 59.0 / 70.5 & 51.9 / 62.7 & 57.8 / 69.2 \\
UI-Venus-2-9B & 64.2 / 75.0 & 56.8 / 67.5 & 44.7 / 56.0 & 57.6 / 68.4 & 52.8 / 61.4 & 56.8 / 67.2 \\
\rowcolor{t2wRowGray}\multicolumn{7}{@{}l}{\emph{Ours (Qwen3.5-9B backbone)}} \\
Base policy & 49.2 / 63.5 & 40.2 / 55.1 & 26.0 / 41.3 & 41.2 / 56.0 & 39.2 / 52.6 & 40.9 / 55.4 \\
Stage-1 policy & 57.0 / 72.6 & 48.6 / 65.1 & 35.3 / 52.0 & 49.6 / 65.8 & 44.3 / 58.9 & 48.7 / 64.6 \\
\textbf{Ours} & 68.4 / 78.9 & 62.4 / 71.8 & 50.7 / 60.5 & 62.8 / 72.6 & 52.8 / 63.7 & 61.1 / 71.1 \\
\bottomrule
\end{tabular}
\end{minipage}
\end{table}

Table~\ref{tab:agym-family-models} makes the common frozen result fields for all 12 evaluated models
directly auditable. The four capability probes in Figure~\ref{fig:task-capability-prevalence}
characterize the task-supply pool, overlap, and have no frozen model-by-task evaluation ledger; they
therefore cannot be reduced into per-baseline capability scores.

\begin{table}[H]
\centering
\caption{\textbf{AndroidWorld and MobileWorld GUI-only model comparisons (\%).} AndroidWorld reports standard-suite SR, except the explicitly marked published pass@3 row. MobileWorld reports SR@50 over all 117 GUI-only tasks; Single-app and Cross-app partition our three-policy results into 44 and 73 tasks. The frozen published-source records contain only the GUI-only overall score, so their slice cells are dashes. $\dagger$: published reference rows; protocols and source reports are in Appendix~\ref{app:external-performance}.}
\label{tab:external-platform-models}
\footnotesize
\setlength{\tabcolsep}{3.2pt}
\begin{minipage}[t]{0.40\linewidth}
\centering
\textbf{(a) AndroidWorld}\\[2pt]
\begin{tabular}{@{}lc@{}}
\toprule
Model & SR \\
\midrule
\rowcolor{t2wRowGray}\multicolumn{2}{l}{\emph{Proprietary Models}} \\
Qwen3.7-Plus$^{\dagger}$ & 81.0 \\
Gemini-2.5 Computer Use$^{\dagger}$ & 69.7 \\
\addlinespace
\rowcolor{t2wRowGray}\multicolumn{2}{l}{\emph{Open-source Models}} \\
MobileRL-9B$^{\dagger}$ & 80.2 \\
Step-GUI-8B (pass@3)$^{\dagger}$ & 80.2 \\
UI-Venus-2-27B$^{\dagger}$ & \textbf{84.0} \\
UI-Venus-2-9B$^{\dagger}$ & 80.2 \\
GELab-Zero-4B$^{\dagger}$ & 75.9 \\
UI-Venus-1.5-8B$^{\dagger}$ & 73.7 \\
MAI-UI-8B$^{\dagger}$ & 70.7 \\
GUI-Owl-7B$^{\dagger}$ & 66.4 \\
UI-TARS-1.5-7B$^{\dagger}$ & 30.0 \\
UI-TARS-1.5$^{\dagger}$ & 64.2 \\
\addlinespace
\rowcolor{t2wRowGray}\multicolumn{2}{l}{\emph{Ours}} \\
Base policy & 57.8 \\
Stage-1 policy & 60.1 \\
\textbf{Ours} & 63.2 \\
\bottomrule
\end{tabular}
\end{minipage}\hfill
\begin{minipage}[t]{0.58\linewidth}
\centering
\textbf{(b) MobileWorld GUI-only (50 steps)}\\[2pt]
\setlength{\tabcolsep}{1.6pt}
\begin{tabular}{@{}lccc@{}}
\toprule
Model & \makecell{All\\(117)} & \makecell{Single-app\\(44)} & \makecell{Cross-app\\(73)} \\
\midrule
\rowcolor{t2wRowGray}\multicolumn{4}{l}{\emph{Proprietary Models}} \\
Kimi-K3$^{\dagger}$ & 74.4 & -- & -- \\
Seed-2.1-Pro$^{\dagger}$ & 73.2 & -- & -- \\
GPT-5.6 Sol$^{\dagger}$ & 70.1 & -- & -- \\
GPT-5.5$^{\dagger}$ & 62.4 & -- & -- \\
Claude Opus 4.8$^{\dagger}$ & 67.5 & -- & -- \\
Seed-2.0-Pro$^{\dagger}$ & 63.2 & -- & -- \\
Qwen3.7-Plus$^{\dagger}$ & 62.3 & -- & -- \\
Kimi-K2.6$^{\dagger}$ & 55.6 & -- & -- \\
\addlinespace
\rowcolor{t2wRowGray}\multicolumn{4}{l}{\emph{Open-source Models}} \\
Qwen-UI-Agent-27B$^{\dagger}$ & \textbf{82.1} & -- & -- \\
UI-Venus-2-27B$^{\dagger}$ & 76.1 & -- & -- \\
UI-Venus-2-9B$^{\dagger}$ & 65.8 & -- & -- \\
GUI-Owl-1.5-32B$^{\dagger}$ & 43.9 & -- & -- \\
Qwen3.6-27B$^{\dagger}$ & 36.8 & -- & -- \\
MAI-UI-8B$^{\dagger}$ & 27.5 & -- & -- \\
UI-Venus-1.5-8B$^{\dagger}$ & 22.2 & -- & -- \\
\addlinespace
\rowcolor{t2wRowGray}\multicolumn{4}{l}{\emph{Ours (Qwen3.5-9B backbone)}} \\
Base policy & 18.5 & 27.3 & 13.2 \\
Stage-1 policy & 21.9 & 32.6 & 15.5 \\
\textbf{Ours} & 27.4 & \textbf{41.7} & \textbf{18.7} \\
\bottomrule
\end{tabular}
\end{minipage}
\end{table}

\begin{table}[H]
\centering
\caption{\textbf{MobileGym-Bench model and task-slice results (\%).} The 256 tasks comprise L1/L2/L3/L4 counts 20/73/83/80 and atomic/sequential/transfer/deep-dive counts 22/110/56/68. SR/PR and trial aggregation follow Appendix~\ref{app:external-performance}.}
\label{tab:mobilegym-models}
\footnotesize
\setlength{\tabcolsep}{2.0pt}
\begin{tabular}{lcccccccccc}
\toprule
Model & SR & PR & L1 & L2 & L3 & L4 & Atomic & Seq. & Transfer & Deep \\
\midrule
\rowcolor{t2wRowGray}\multicolumn{11}{l}{\emph{Proprietary Models}} \\
Gemini 3.1 Pro & 31.3 & 44.6 & \textbf{100.0} & 56.2 & 20.5 & 2.5 & 77.3 & 41.8 & 7.1 & 19.1 \\
Doubao-Seed-2.0-Pro & 34.0 & \textbf{47.5} & \textbf{100.0} & \textbf{60.3} & \textbf{24.1} & \textbf{3.8} & \textbf{81.8} & \textbf{45.5} & \textbf{8.9} & \textbf{20.6} \\
Qwen3.7-Plus & 32.8 & 46.0 & \textbf{100.0} & 57.5 & 22.9 & \textbf{3.8} & 77.3 & 43.6 & \textbf{8.9} & \textbf{20.6} \\
\addlinespace
\rowcolor{t2wRowGray}\multicolumn{11}{l}{\emph{Open-source Models}} \\
UI-Venus-2-27B & \textbf{60.5} & -- & -- & -- & -- & -- & -- & -- & -- & -- \\
UI-Venus-2-9B & 52.7 & -- & -- & -- & -- & -- & -- & -- & -- & -- \\
MAI-UI-8B & 21.5 & -- & -- & -- & -- & -- & -- & -- & -- & -- \\
AutoGLM-Phone-9B & 20.0 & 35.3 & 86.2 & 33.6 & 9.6 & 1.9 & 56.8 & 25.7 & 7.1 & 9.6 \\
UI-Venus-1.5-8B & 15.4 & 28.3 & 85.0 & 21.9 & 6.0 & 1.9 & 37.5 & 20.7 & 6.2 & 7.4 \\
GUI-Owl-1.5-8B-Think & 15.1 & 28.8 & 76.2 & 26.0 & 4.2 & 1.2 & 44.3 & 20.9 & 2.7 & 6.6 \\
UI-TARS-1.5-7B & 13.8 & 26.3 & 77.5 & 21.9 & 3.0 & 1.6 & 37.5 & 20.2 & 2.7 & 4.8 \\
Step-GUI-4B & 12.9 & 25.7 & 83.8 & 17.8 & 2.4 & 1.6 & -- & -- & -- & -- \\
Qwen3-VL-4B-Instruct & 9.4 & 20.1 & 71.2 & 12.3 & 0.6 & 0.3 & -- & -- & -- & -- \\
Qwen3-VL-4B + MobileGym GRPO & 22.2 & -- & 92.5 & 37.7 & 11.7 & 1.2 & -- & -- & -- & -- \\
\addlinespace
\rowcolor{t2wRowGray}\multicolumn{11}{l}{\emph{Ours (Qwen3.5-9B backbone)}} \\
Base policy & 9.0 & 19.7 & 67.9 & 11.6 & 0.7 & 0.4 & 24.6 & 14.5 & 1.2 & 1.3 \\
\textbf{Ours} & 25.2 & 40.7 & 94.6 & 47.8 & 11.7 & 1.1 & 63.6 & 36.1 & 4.5 & 12.3 \\
\bottomrule
\end{tabular}
\end{table}

\begin{table}[H]
\centering
\caption{\textbf{AndroidDaily-235 end-to-end results (\%).} Columns follow the benchmark task-type (106/93/36 tasks), complexity (174/30/31), and ambiguity (164/47/24) axes of the released task table. $\dagger$: Step-GUI report; total-only references also use the Qwen-UI-Agent report \citep{zhou2026qwenuiagent}. Base and final policies form the matched pair.}
\label{tab:androiddaily-models}
\footnotesize
\setlength{\tabcolsep}{3.2pt}
\begin{tabular}{lcccccccccc}
\toprule
Model & Filter & Query & Analyze & Atomic & Comp. & Cond. & Low & Mid & High & Total \\
\midrule
\rowcolor{t2wRowGray}\multicolumn{11}{l}{\emph{Proprietary Models}} \\
Qwen3.7-Plus$^{\dagger}$ & -- & -- & -- & -- & -- & -- & -- & -- & -- & 79.8 \\
\addlinespace
\rowcolor{t2wRowGray}\multicolumn{11}{l}{\emph{Open-source Models}} \\
Qwen-UI-Agent-27B$^{\dagger}$ & -- & -- & -- & -- & -- & -- & -- & -- & -- & \textbf{97.5} \\
UI-TARS-1.5$^{\dagger}$ & 57.6 & \textbf{66.0} & 36.7 & 61.4 & 13.6 & \textbf{60.4} & 57.1 & 54.9 & 57.9 & 56.6 \\
Step-GUI-8B$^{\dagger}$ & 52.5 & 63.8 & 33.0 & 59.1 & 14.0 & 42.9 & 54.1 & 44.6 & \textbf{61.5} & 52.5 \\
Step-GUI-4B$^{\dagger}$ & 44.8 & 64.3 & 33.7 & 54.0 & \textbf{19.6} & 42.9 & 51.2 & 38.3 & 59.5 & 49.1 \\
\addlinespace
\rowcolor{t2wRowGray}\multicolumn{11}{l}{\emph{Ours (Qwen3.5-9B backbone)}} \\
Base policy & 46.5 & 54.5 & 30.6 & 54.8 & 11.1 & 39.8 & 48.4 & 44.0 & 45.8 & 47.2 \\
\textbf{Ours} & \textbf{59.7} & 64.9 & \textbf{37.0} & \textbf{66.7} & 15.6 & 52.7 & \textbf{60.0} & \textbf{55.3} & 52.8 & 58.3 \\
\bottomrule
\end{tabular}
\end{table}

\begin{table}[H]
\centering
\caption{\textbf{AndroidDaily-235 results by scenario (\%).} The matched base and final policies use the same 78/61/43/37/16 transportation/shopping/social/content/local-service tasks; gains are percentage points.}
\label{tab:androiddaily-scenarios}
\footnotesize
\setlength{\tabcolsep}{3.2pt}
\begin{tabular}{lcccccc}
\toprule
Model & Transp. (78) & Shop. (61) & Social (43) & Content (37) & Local (16) & Total \\
\midrule
Base policy & 48.7 & 46.4 & 46.5 & 47.7 & 43.8 & 47.2 \\
\textbf{Ours} & \textbf{58.1} & \textbf{57.9} & \textbf{58.9} & \textbf{59.5} & \textbf{56.3} & \textbf{58.3} \\
Gain (pp) & +9.4 & +11.5 & +12.4 & +11.8 & +12.5 & +11.1 \\
\bottomrule
\end{tabular}
\end{table}

\subsubsection{Horizon calibration and remaining policy failures}
\label{app:horizon-error-analysis}

Table~\ref{tab:horizon-calibration} certifies a legal completion path below each pre-frozen
construction-class horizon. Its cross-app lengths come from public-interface human completions,
not a cross-app source recording. To separate truncation from agent behavior, we classify every
failed Trace2World-Bench attempt by its first attributable terminal mode after excluding and retrying
provider, environment, and evaluator invalid attempts. False completion means the policy declares
success while the joint terminal predicate is false; hard loop means repeated no-effect actions;
grounding means dominant non-interactive targets; stalled or wandering means no dominant repeated
action but no effective late progress. Explicit give-up, format/no-action, and overdue termination
complete the mutually exclusive taxonomy. A cross-app handoff error is an additional tag, not an
eighth terminal category.

\begin{figure}[H]
\centering
\includegraphics[width=\linewidth]{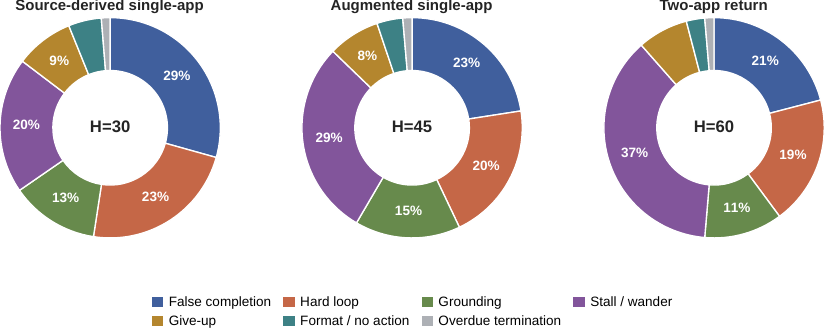}
\caption{\textbf{Failure-mode composition of our final policy by horizon class.} Each pie normalizes
the seven mutually exclusive terminal modes over failed attempts in that class; labeled shares are
rounded to whole percentages, and slices below 7.5\% remain in the legend without an in-slice label.
Cross-app handoff errors are orthogonal tags, not an eighth slice.}
\label{fig:horizon-failures}
\end{figure}

Across all 3,132 attempts, hard-loop failures fall from 16.1\% under the base policy to 8.5\% under
our final policy; false completions fall from 12.7\% to 10.2\%. Figure~\ref{fig:horizon-failures}
shows that stalled or wandering failures make up a larger share of augmented than source-derived
failures (28.8\% versus 19.9\%). In the cross-app class, 33.1\% of final-policy failures carry an
$A\rightarrow B$ or $B\rightarrow A$ dependency tag, compared with 32.4\% under the base policy. The tag covers
lost or incompatible handoff values, a final $A$ decision that ignores $B$, and incorrect assumptions
about $A$'s preserved state; it can overlap any pie slice.

Of the final policy's 613 failed attempts that reach their class horizon (330/206/77 by class),
511 make no new checkpoint in their final ten actions (275/172/64). Continuing the same policy for
15 extra public actions from those cap states rescues 18 attempts (8/7/3), without changing the
reported benchmark scores. Hitting the cap is therefore a termination observation, not itself a
diagnosis of insufficient task budget; the late-progress and extension checks localize the remaining
ceiling effect on this frozen evaluation set.

\subsection{Transfer to Real Applications}
\label{app:transfer-results}

\subsubsection{Balanced 100-task real-device protocol and matched baselines}
\label{app:real-device-protocol}

We freeze a paired 100-task panel on 20 policy-training apps, five per app: 50 entity
rebindings retain a seed workflow schema but use unseen values and starts, while 50 workflow identities
come from Trace2World-Bench's frozen held-out-workflow slice (30 seed-family and 20 augmentation-family).
Ten apps use a 3:2 entity/workflow mix and ten
use 2:3, counterbalanced by domain and construction-trace budget. The workflow half overlaps the
benchmark and is not a second independent evaluation set.

Within each app--family stratum, candidate order and reversible, equivalently restorable eligibility are
frozen before simulator or device outcomes; an ineligible candidate advances within the same stratum.
The same identities are evaluated for three training seeds in both substrates, with substrate order
randomized and goal, start, terminal criterion, and each task's 30- or 45-step single-application
budget from Table~\ref{tab:dapo-config} held fixed, so a simulator/device mismatch cannot come from
different step limits. Only live entity values may be rebound
after content drift. No panel identity or outcome enters construction, training, repair, or checkpoint
selection.

Agreement is the share of identical simulator/device outcomes, with both directional mismatches also
reported. The real/sim gain ratio is
$R=[SR_{\mathrm{real}}(\pi_3)-SR_{\mathrm{real}}(\pi_0)]/[SR_{\mathrm{env}}(\pi_3)-SR_{\mathrm{env}}(\pi_0)]$.
Means and ratios precede rounding; the training seed is the replication unit and defines every SD. Five
tasks per app do not estimate per-app effects; equal allocation makes the overall task mean equal the
macro-app mean, and the two 50-task family rows remain descriptive slices.

\paragraph{Matched training-source conditions.}
\label{app:transfer-source-baselines}
The self-checked and Trace2World source comparison holds fixed all 1,280 seed identities, launch and
terminal contracts, learner, sampler, task order, and 32 RL iterations; only the environments change.
Missing implementations remain failures. These stage-1 policies are distinct from our final policy and are
scored only on the panel above.

\subsubsection{Source-trace validation improves paired sim-to-real retention}
\label{app:sim2real-detail}

The protocol and estimands are fixed in Appendix~\ref{app:real-device-protocol}; this subsection adds
outcome levels, task-family slices, and the matched-source decomposition.

\begin{figure}[H]
\centering
\includegraphics[width=\linewidth]{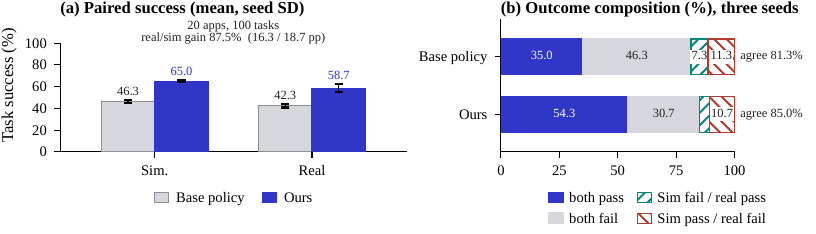}
\caption{\textbf{Paired transfer on 100 tasks across 20 applications.} The panel balances 50 entity
rebindings and 50 Trace2World-Bench workflow holdouts; complementary 3:2 and 2:3 app-level mixes
keep five tasks per app. (a) Base and stage-3 task success in both substrates; error bars are seed SD. (b) Mean paired
outcome composition. Agreement includes both-pass and both-fail.}
\label{fig:sim2real-main}
\end{figure}

\begin{table}[H]
\centering
\caption{\textbf{Balanced task-family transfer slices.} The fixed panel contains 50 unseen entity/state rebindings and 50 workflows sampled directly from the 512-task Trace2World-Bench held-out-workflow slice. Percentages; agreement is base / trained over three seeds.}
\label{tab:sim2real-detail}
\footnotesize
\setlength{\tabcolsep}{4pt}
\begin{tabular*}{\linewidth}{@{\extracolsep{\fill}}lcccccc@{}}
\toprule
Slice & $N$ & Base env. & Base real & Trained env. & Trained real & Agreement \\
\midrule
Entity rebinding & 50 & 58.7 & 54.0 & 79.3 & 73.3 & 78.0 / 84.7 \\
Held-out workflow & 50 & 34.0 & 30.7 & 50.7 & 44.0 & 84.7 / 85.3 \\
All tasks & 100 & 46.3 & 42.3 & 65.0 & 58.7 & 81.3 / 85.0 \\
\bottomrule
\end{tabular*}
\end{table}

Both task families improve on the device: entity rebindings rise from
54.0\% to 73.3\%, and held-out workflows from
30.7\% to 44.0\%. These are descriptive
slices; aggregate gains and seed SDs are reported in Table~\ref{tab:transfer-main}.

Against self-validation, source-trace validation's 6.0-pp
real-device gain advantage decomposes as $-1.0 + 8.3 - 1.3 = 6.0$ pp. The terms are respectively the
environment-pass difference, the reduction in simulation-only passes, and the device-only difference.
The simulation-only term dominates; the other two terms are smaller and point in the opposite direction.
This separates invalid simulated success from a larger simulated training gain.

Our final policy passes only on the real device for 4.3\% of tasks (base: 7.3\%); it passes in
simulation but fails on the device for 10.7\% (base: 11.3\%). Live content and combinations beyond the recorded workflows are possible
contributors, but these joint counts do not identify the causes of the remaining transfer gap.

\subsection{Scope, Limitations, and Future Work}
\label{app:limitations}
\label{app:limitations-detail}
\label{app:limitations-map}

This paper evaluates Trace2World on mobile applications reconstructed as DOM-backed browser environments from
traces collected at one point in time, with proprietary models in the construction, validation,
screening, and review roles. Three limitations follow from this scope, and each points to a direct
extension. First, a browser reimplementation renders differently from the native application, so the
paired device study bounds rather than removes the substrate gap; that panel covers policy-training
applications, while unseen-application transfer is measured in reconstructed environments. Extending
the paired panel to held-out applications, and the substrate to desktop and native rendering, are
extensions of the same protocol. Second, we report initial construction cost but not refresh cost.
Because every supported relation keeps its source links, application behavior that changes after
collection---a new entry surface, a changed action effect---can be reconciled from new traces and
re-accepted by the same replay, so the intended maintenance path is incremental refresh rather than
rebuild; its accounting is future work. Third, all optimization tasks are single-app, and the
construction roles run on proprietary models with the bounded API stage on the weaker of the two
(Appendix~\ref{app:model-roles}). Since gains already reach the evaluation-only two-app slice,
training on validated cross-app compositions is a natural next step, as are open-weight substitution
in each role and, eventually, using a trained policy's own validated executions as additional
construction evidence to close the loop from policy back to world.

\FloatBarrier

\appsection{cases}{app:qualitative}{Case Studies}
\label{app:qualitative-examples}

The cases first compare complete visible trajectories for one shared Meituan workflow across the
Reference, four environment builders, and Trace2World. They then cover Meituan's
schema-backed augmentation, seed and augmented workflows in one 58Tongcheng environment, and a
Railway 12306--Didi composition with a return dependency. The final gallery collects reconstructed
pages and application home screens.

\subsection{Matched Baseline Trajectories on One Shared Workflow}
\label{app:matched-baseline-trajectories}

\begin{figure}[p]
\centering
\includegraphics[page=8,width=0.99\linewidth]{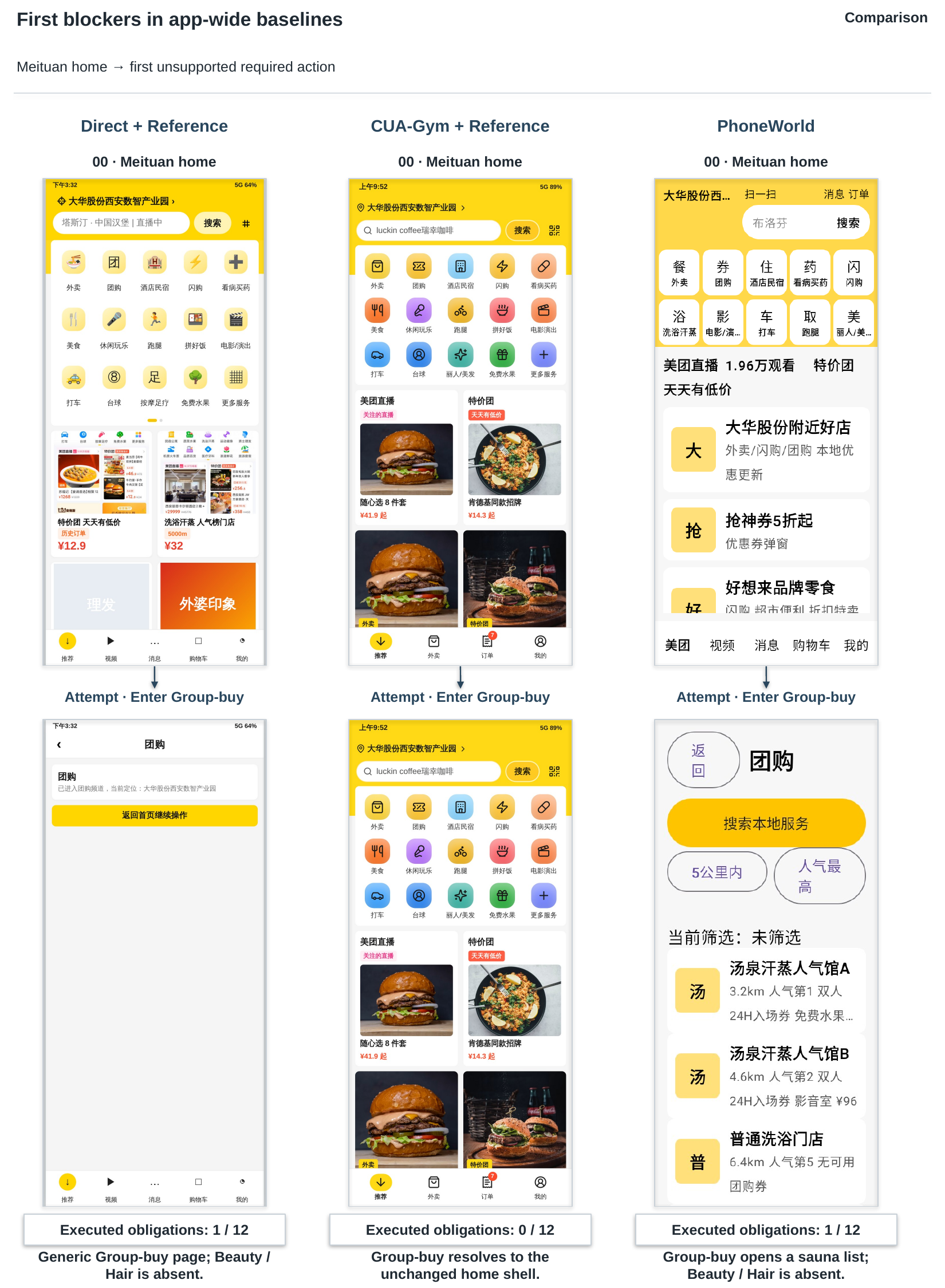}
\caption{\textbf{Complete same-task trajectories from the Meituan home (eight pages).}
(a--b) The frozen marked Reference establishes entity resolution, coupon arithmetic, and the pre-commit
terminal; top-left numbers are local step indices, red dots mark the next recorded action, and the
terminal panel identifies the stop.
(c--d) GUI-GENESIS completes the workflow in its task-local site. (e--f) Trace2World
completes the same workflow in one shared 100-task world. (g) One rubric below the aligned
terminal frames specifies the deal, coupon-adjusted amount, and pre-commit payment boundary for
all three traces; the Trace2World terminal payment indicators use an isolated display-only view correction,
with the raw capture retained in the artifact. The original Trace2World package evaluator scores the frozen
run 1.0. (h) Direct+Reference
reaches a generic Group-buy page but exposes no Beauty/Hair
control, CUA-Gym+Reference returns to the unchanged home shell, and PhoneWorld opens a sauna list.
Every sequence starts at home; failed traces end at their first missing required action without an
inferred suffix. This single case diagnoses mechanisms rather than failure frequencies.}
\label{fig:meituan-shared-workflow-traces}
\end{figure}

The shared task starts from the Meituan home, enters Group-buy and Beauty/Hair, searches for
WanSen Men's Hair Design, opens the Wash+Style deal, applies a CNY 6 Meituan coupon so that the
payable amount changes from CNY 38 to CNY 32, and selects Alipay without pressing Confirm
transaction. It belongs to the five-task sample frozen before any subject or evaluation result.
Figures~\ref{fig:meituan-shared-workflow-traces}a--h start every condition from the Meituan home and
retain the complete visible trace through either the pre-commit terminal or the first missing
required action. The successful Reference, GUI-GENESIS, and Trace2World sequences each
contain 13 frames; Direct+Reference, CUA-Gym+Reference, and PhoneWorld stop where continuation is
no longer supported by a visible control.

\begin{figure}[p]
\centering
\includegraphics[page=1,width=\linewidth]{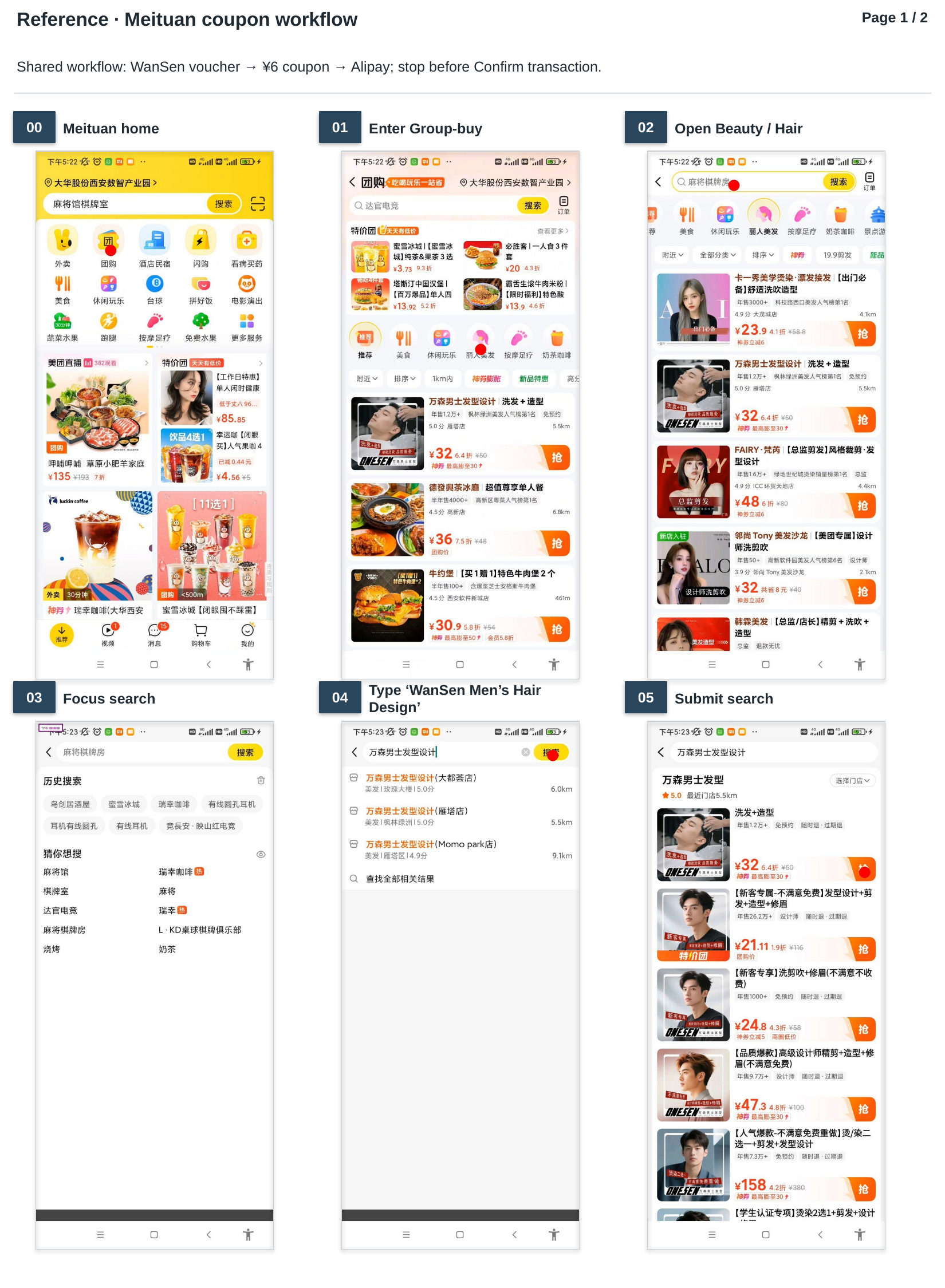}
\end{figure}
\begin{figure}[p]
\centering
\includegraphics[page=2,width=\linewidth]{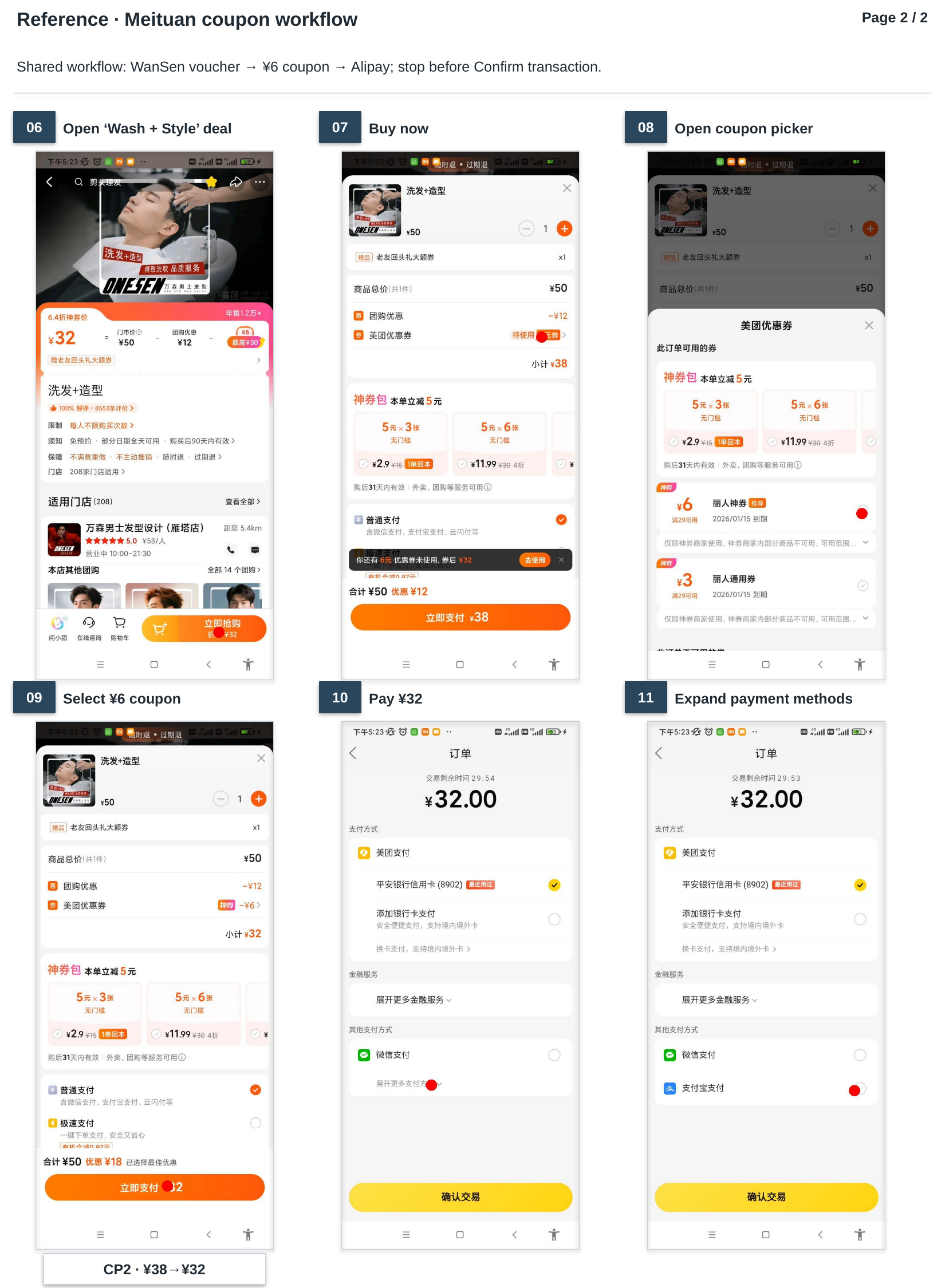}
\end{figure}
\begin{figure}[p]
\centering
\includegraphics[page=3,width=\linewidth]{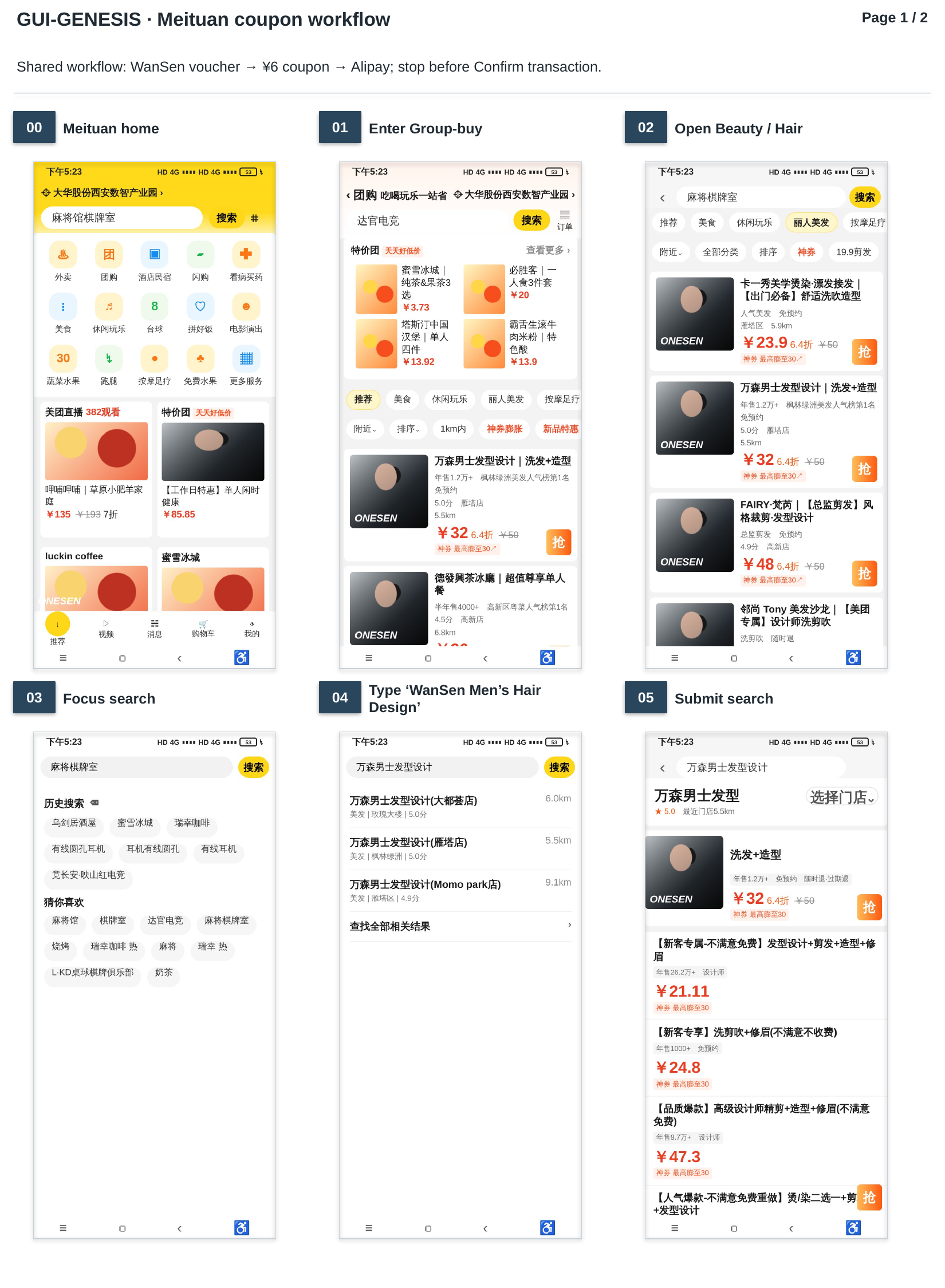}
\end{figure}
\begin{figure}[p]
\centering
\includegraphics[page=4,width=\linewidth]{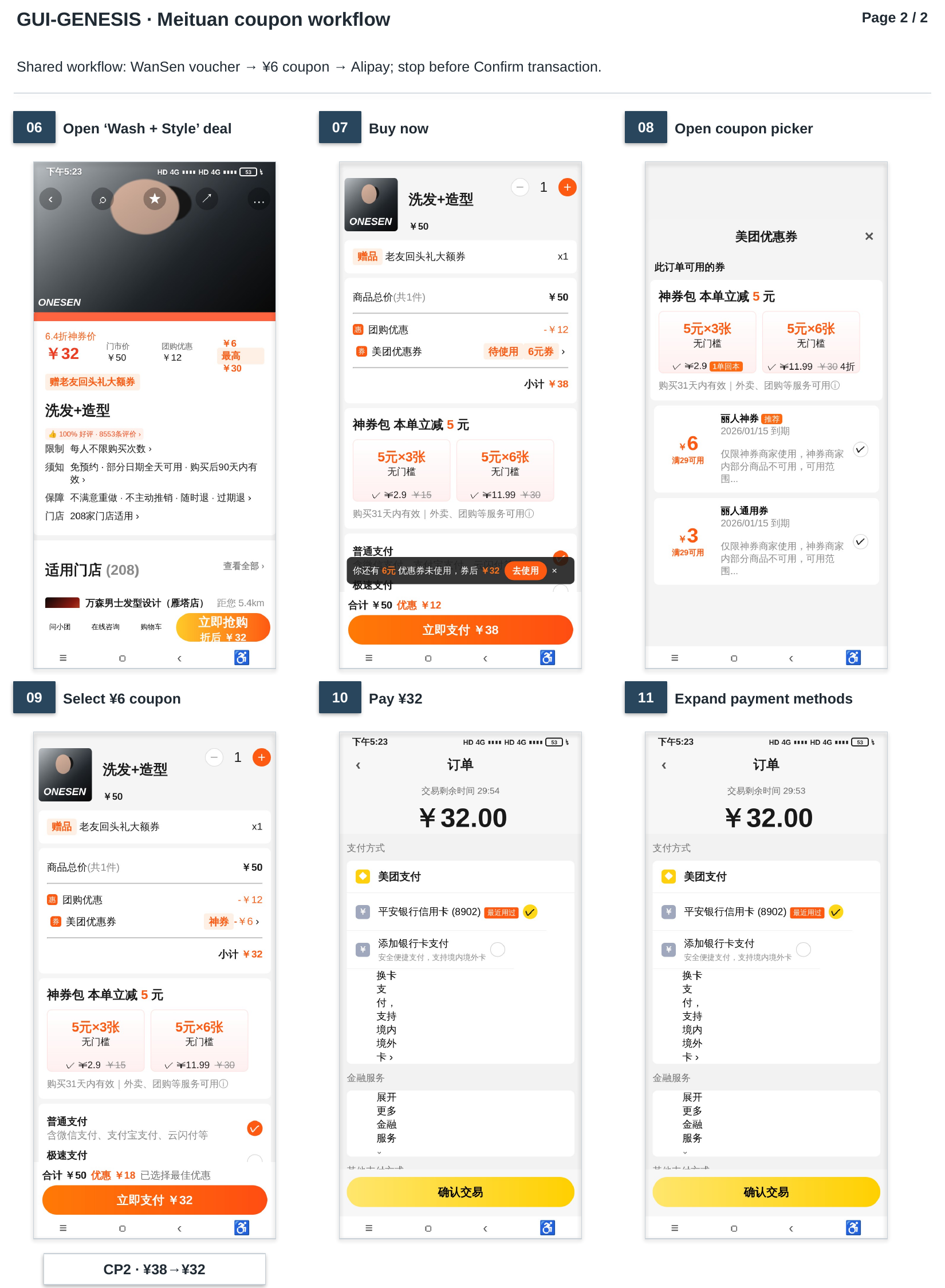}
\end{figure}
\begin{figure}[p]
\centering
\includegraphics[page=5,width=\linewidth]{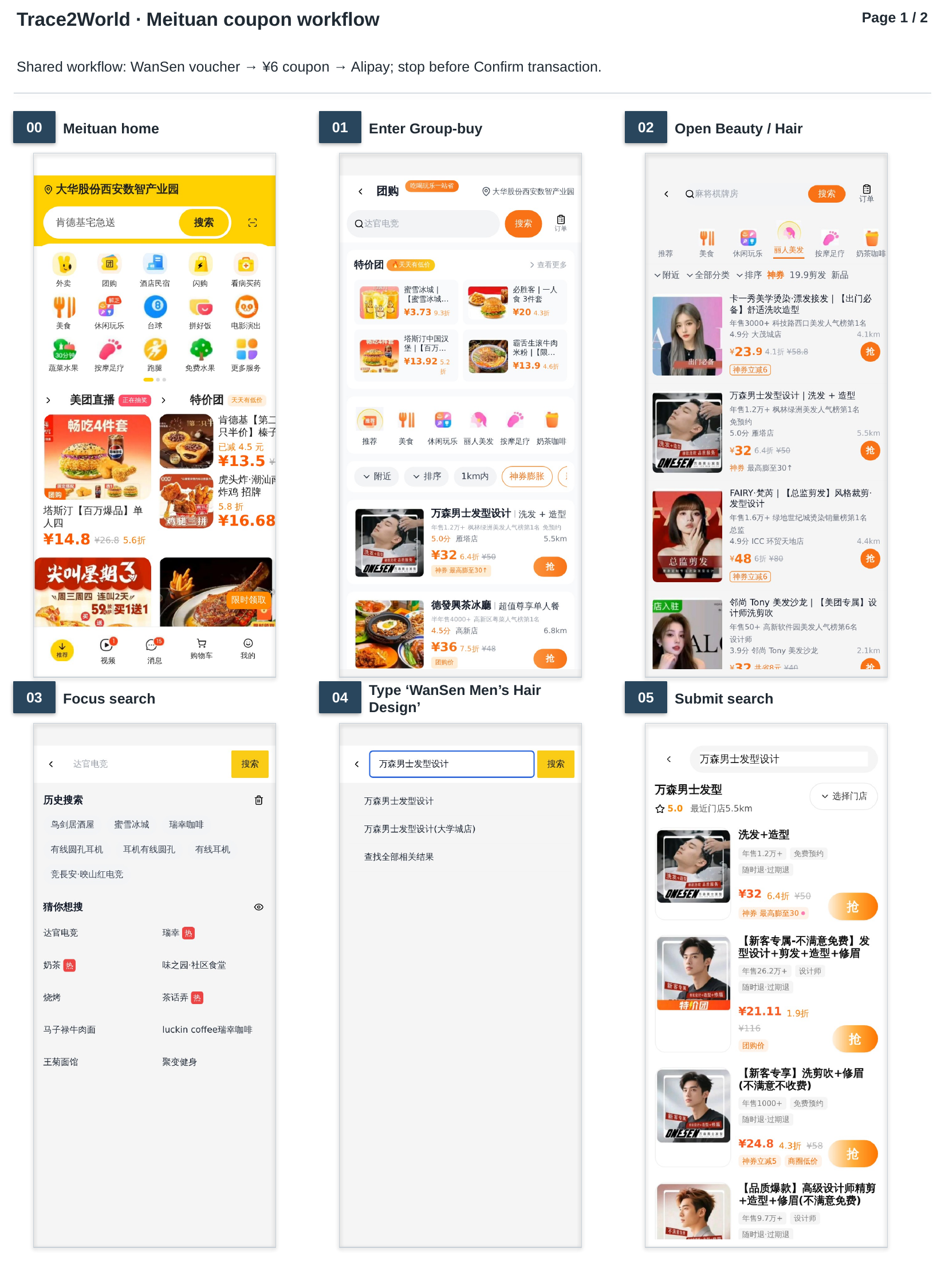}
\end{figure}
\begin{figure}[p]
\centering
\includegraphics[page=6,width=\linewidth]{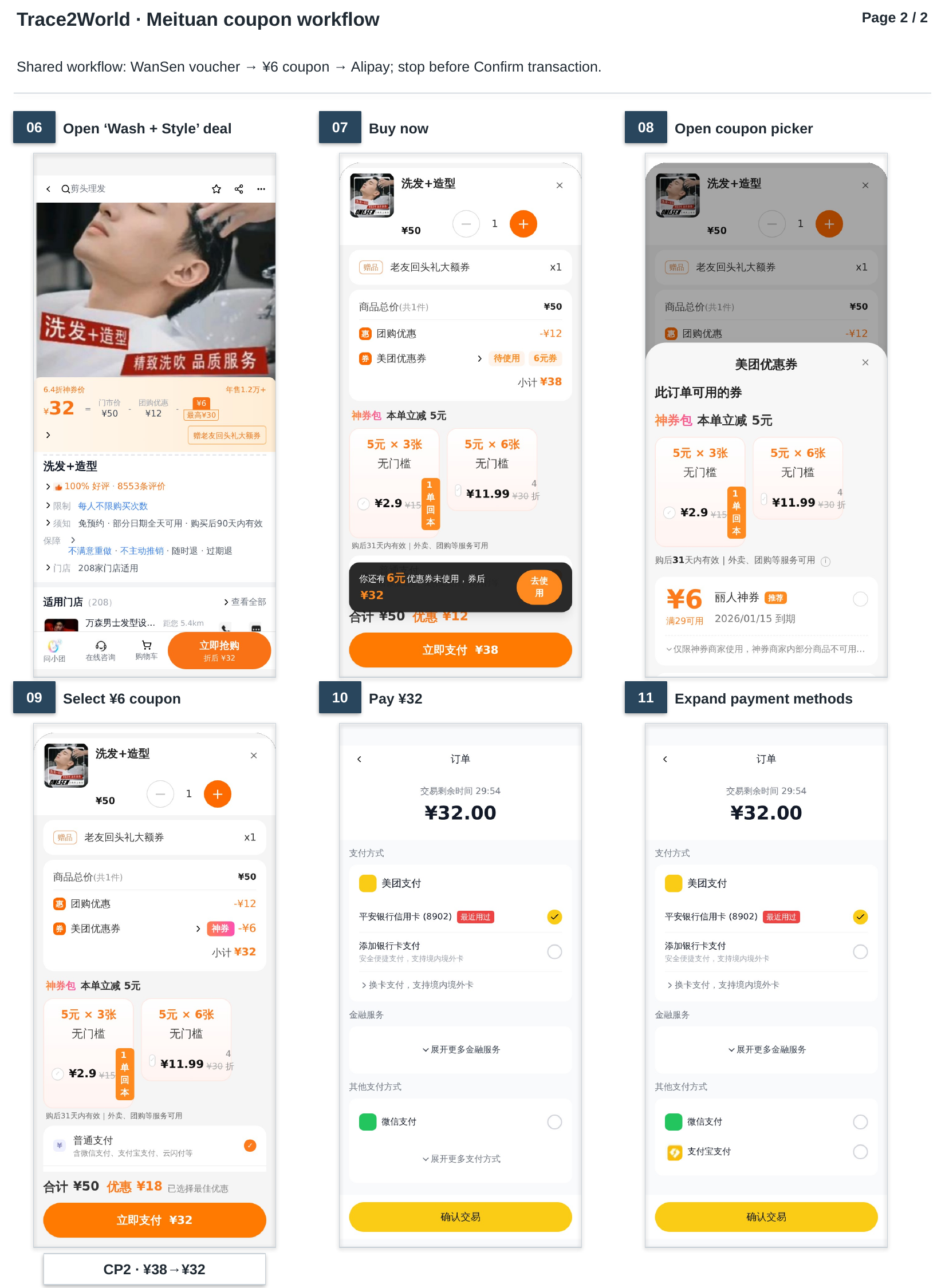}
\end{figure}
\begin{figure}[p]
\centering
\includegraphics[page=7,width=\linewidth]{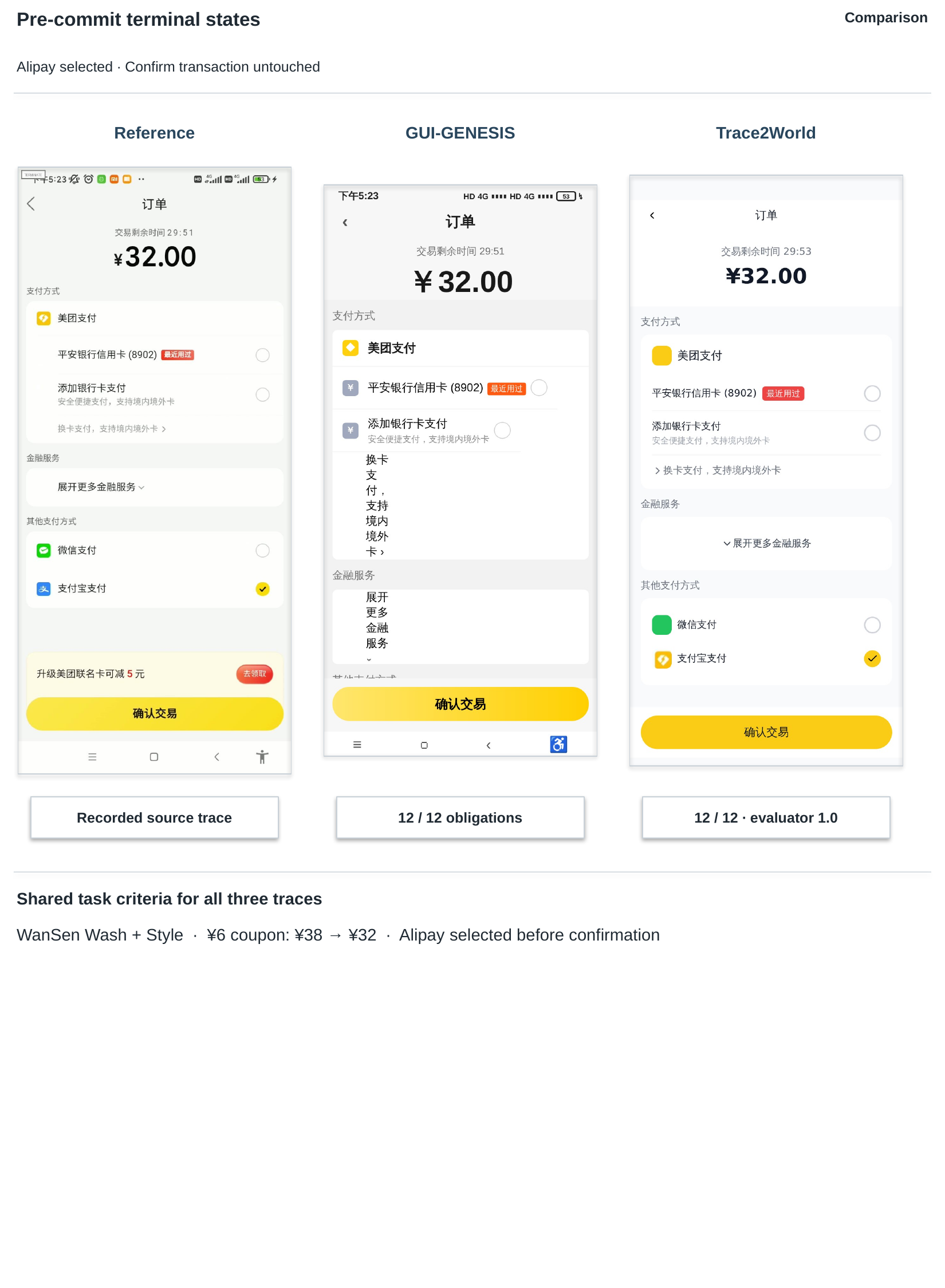}
\end{figure}
\FloatBarrier

The comparison separates three failure mechanisms, and each follows the builder's own construction
contract (Table~\ref{tab:contract-comparison}). Direct preserves a recognizable shell but loses the task's
category transition, satisfying 1 of the 12 in-application obligations. CUA-Gym, whose specification is
authored from topic research with the recordings only as auxiliary input, does not materialize even the
first route effect and satisfies 0. Our PhoneWorld-style reimplementation abstracts the recordings
into frequency-ranked screen types, collapses Group-buy into the wrong local-service family, and
satisfies 1. None of the three
degrades gradually: all stop after two visible frames, at the first category transition. GUI-GENESIS is
correct for this task, satisfying 12 of 12 across 13 frames, so the case does not support a
semantic-failure claim against it; its remaining boundary is that the execution closes one task-local
site rather than one app-wide world. Trace2World satisfies the same 12 obligations while preserving the
shared world, the ordered state changes, and an evaluator-owned pre-commit terminal. The contrast is therefore not between agentic and non-agentic implementation: GUI-GENESIS reproduces
this single workflow exactly, so agentic construction can realize a task-local workflow. The
distinction appears when several traces must define one reusable application, where identity, state,
and effects have to stay consistent across workflows; Trace2World fixes those semantics before its own
implementation loop begins. These are
qualitative mechanism diagnoses, not estimates for the full construction population.

\subsection{Schema-Backed Task Augmentation}
\label{app:artifact-comparison}

\paragraph{Supported augmentation.}
\label{app:teaser-task}
Figure~\ref{fig:portfolio-overview} contrasts a 13-action source trace with a 20-step augmented rollout
in the same Meituan environment. The source establishes search, store selection, return to the deal list,
item detail, quantity control, and checkout. The augmentation composes those behaviors into a nearest-store
comparison, a retained selection after returning, a quantity update, and a pre-submission amount check.

\paragraph{Evidence boundary.}
The latte detail is a new entity binding of the observed item-detail template, hence a
\emph{schema-backed instance} under Appendix~\ref{app:expansion-loop}, rather than a newly observed screen.
The six frames show the selected 394\,m store, the revisit, the latte instance, and checkout. Because the
last frame precedes the increment, it displays one voucher; saved terminal checks record two vouchers,
CNY\,14.00 goods, a CNY\,1.42 discount, and CNY\,12.58 payable. This case demonstrates one supported
within-application composition, not population-level workflow diversity or a training effect.

\subsection{Seed and Augmented Workflows}
\label{app:practice-showcases}

The two traces use one 58Tongcheng environment at different task boundaries: the seed reconstructs a
recorded rental-search workflow, whereas the augmentation composes comparison, revisit, persistent state,
and recovery over supported behavior. They are distinct tasks rather than two environments or one extended
trace, and remain qualitative examples rather than estimates of difficulty or training effects.

\begin{figure}[p]
\centering
\includegraphics[width=\linewidth]{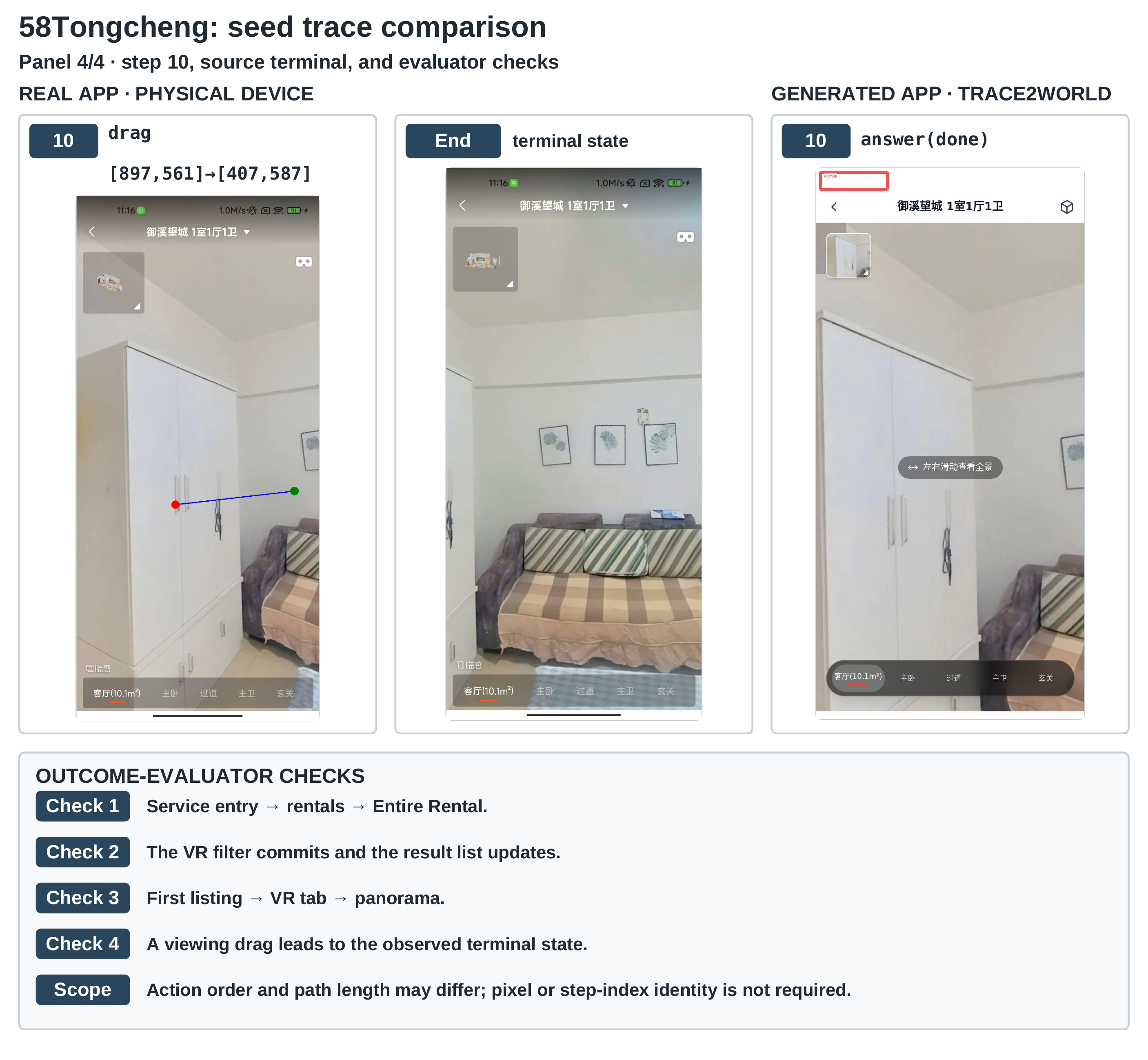}
\caption{\textbf{Trace-wise real-app/generated-app execution for a seed 58Tongcheng task (four
pages).} The upper sequence is the human physical-device execution and terminal observation; the lower
sequence is a successful execution in the Trace2World environment. The numbered cards follow each trace's
own action order; the unnumbered terminal observation is separate. The final panel lists the
outcome-evaluator checks and the alignment boundary.}
\label{fig:tracewise-real-generated-58tc}
\end{figure}

\begin{figure}[p]
\centering
\includegraphics[width=\linewidth]{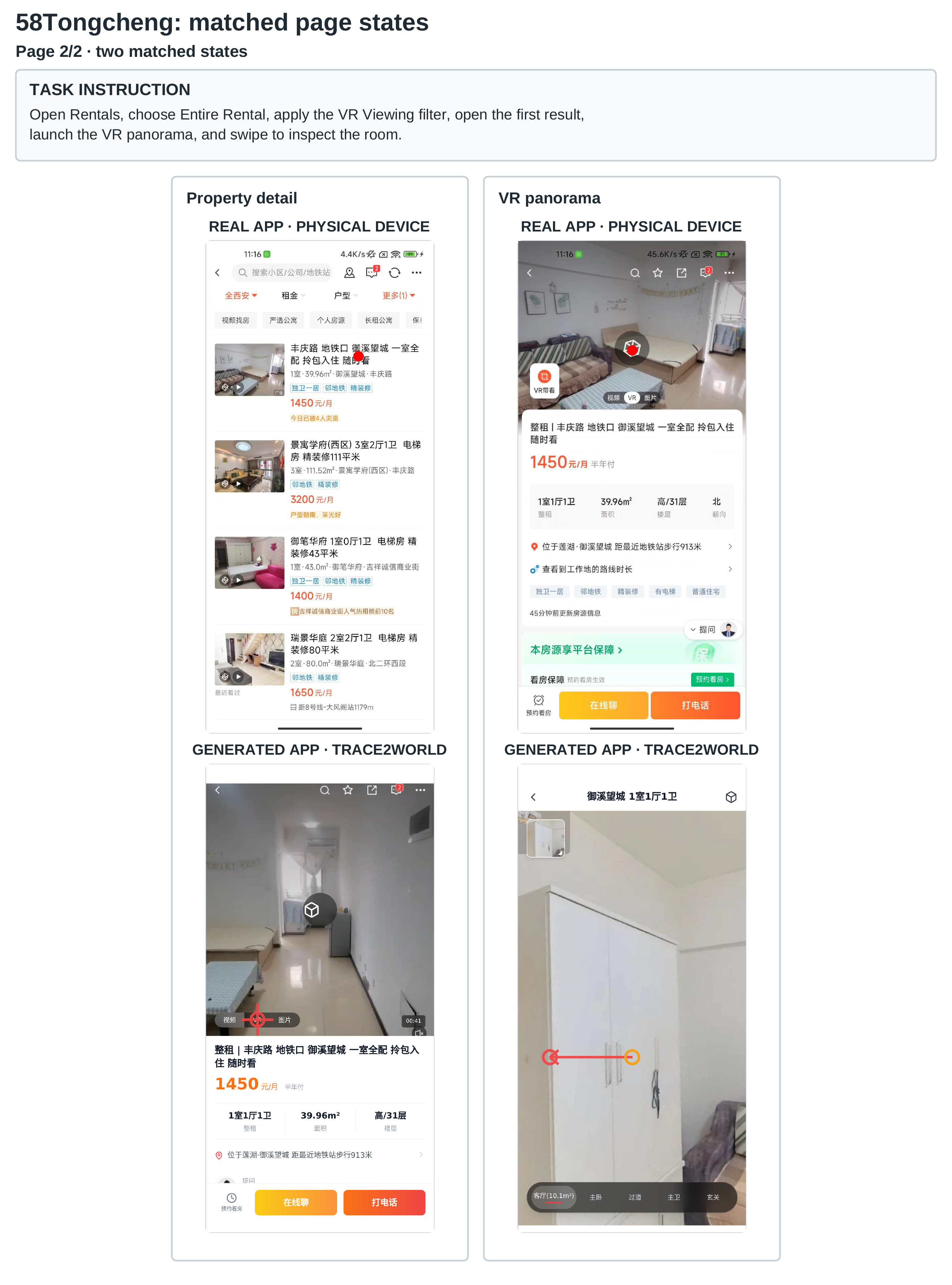}
\caption{\textbf{58Tongcheng reconstruction gallery (five matched page states across two figure
pages).} The paired pages cover service entry, the rental channel, VR filtering, a property detail,
and the interactive panorama.}
\label{fig:pagewise-58tc}
\end{figure}

\paragraph{Seed reconstruction.}
\label{app:tracewise-real-generated-case}
The two views audit different fidelity obligations: Figure~\ref{fig:tracewise-real-generated-58tc}
preserves complete action chronology and the terminal verdict, while Figure~\ref{fig:pagewise-58tc}
compares matched page states.

\begin{figure}[!htbp]
\centering
\includegraphics[width=\linewidth]{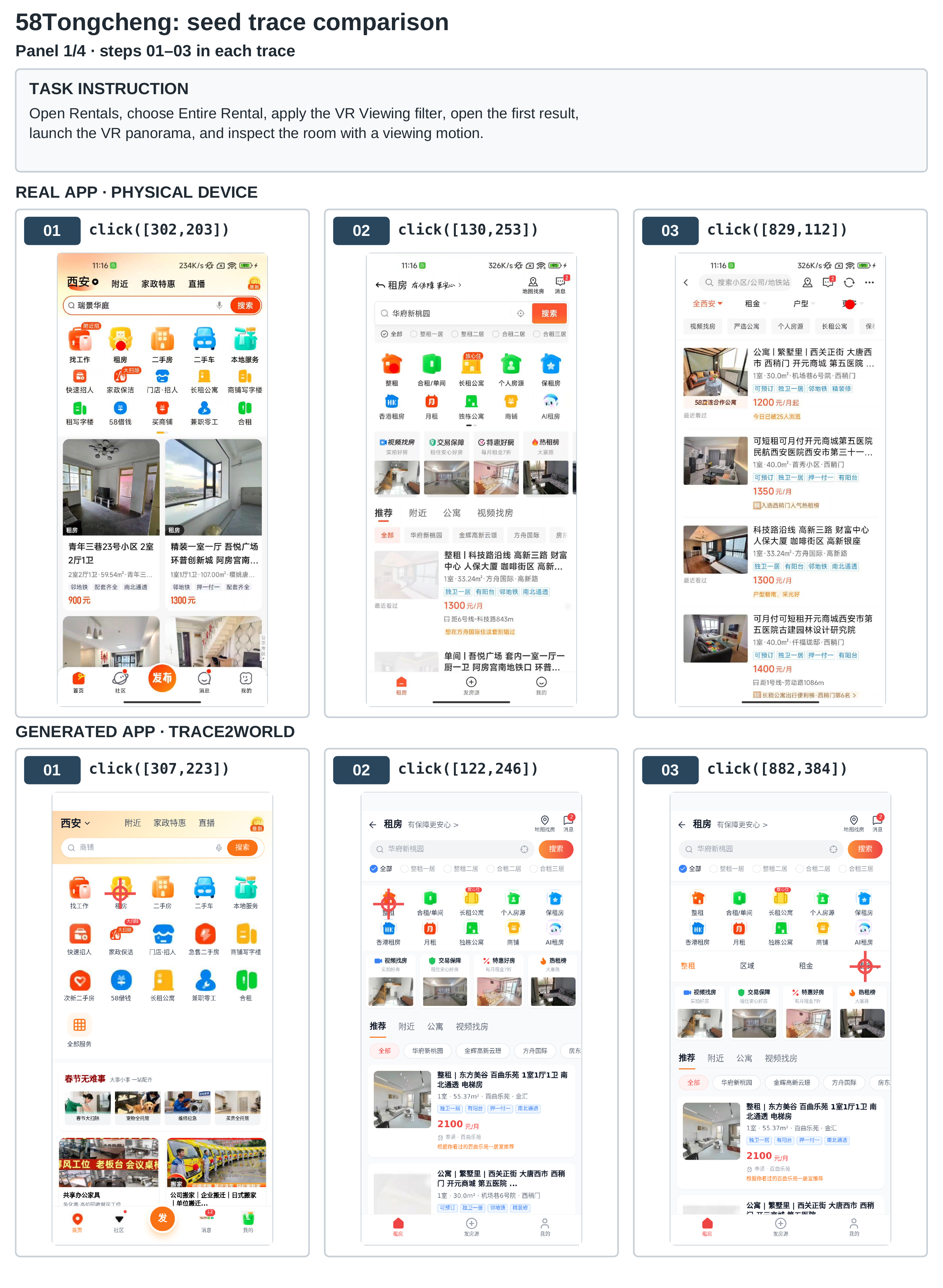}
\end{figure}
\begin{figure}[p]
\centering
\includegraphics[width=\linewidth]{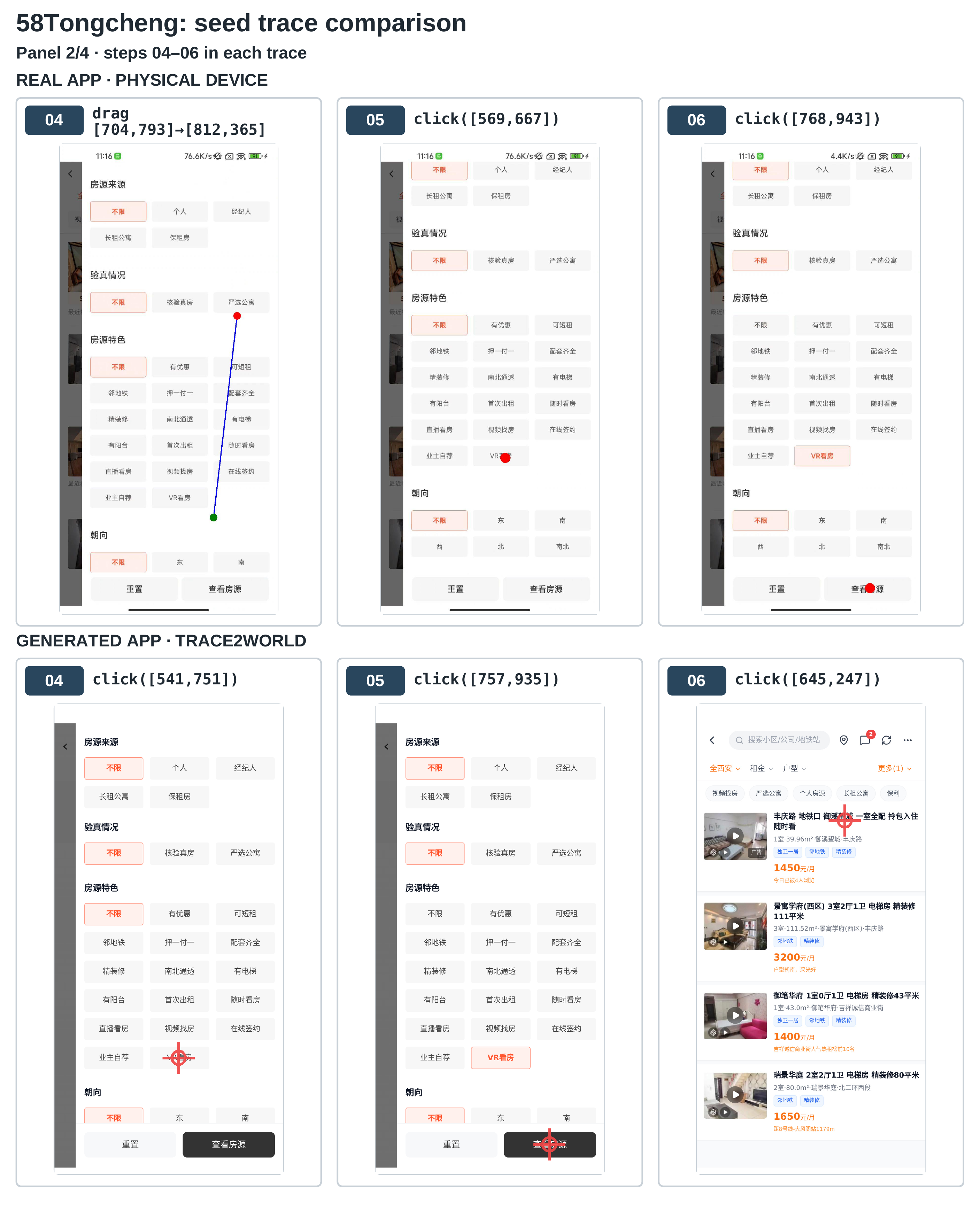}
\end{figure}
\begin{figure}[p]
\centering
\includegraphics[width=\linewidth]{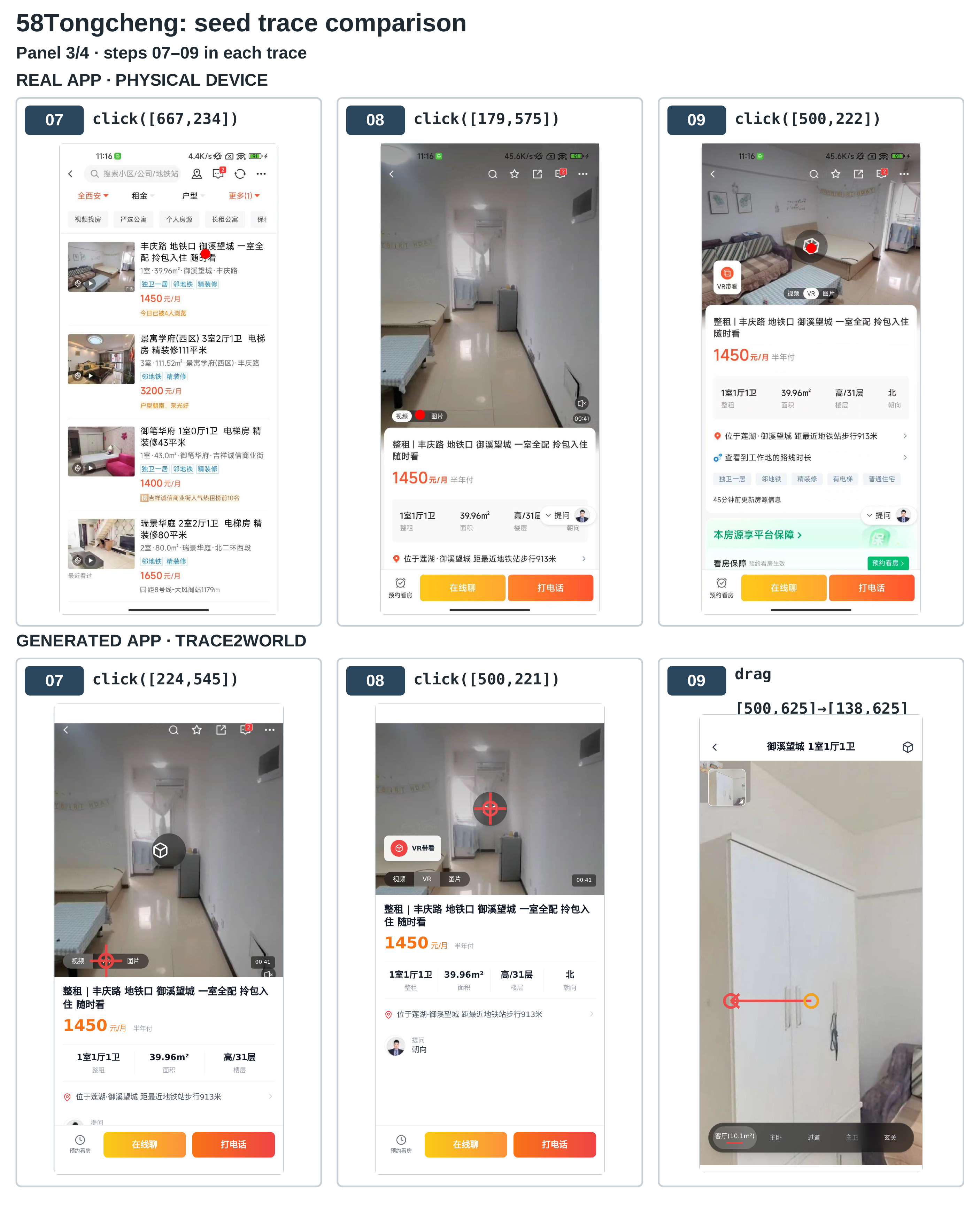}
\end{figure}
\FloatBarrier

\begin{figure}[p]
\centering
\includegraphics[width=\linewidth]{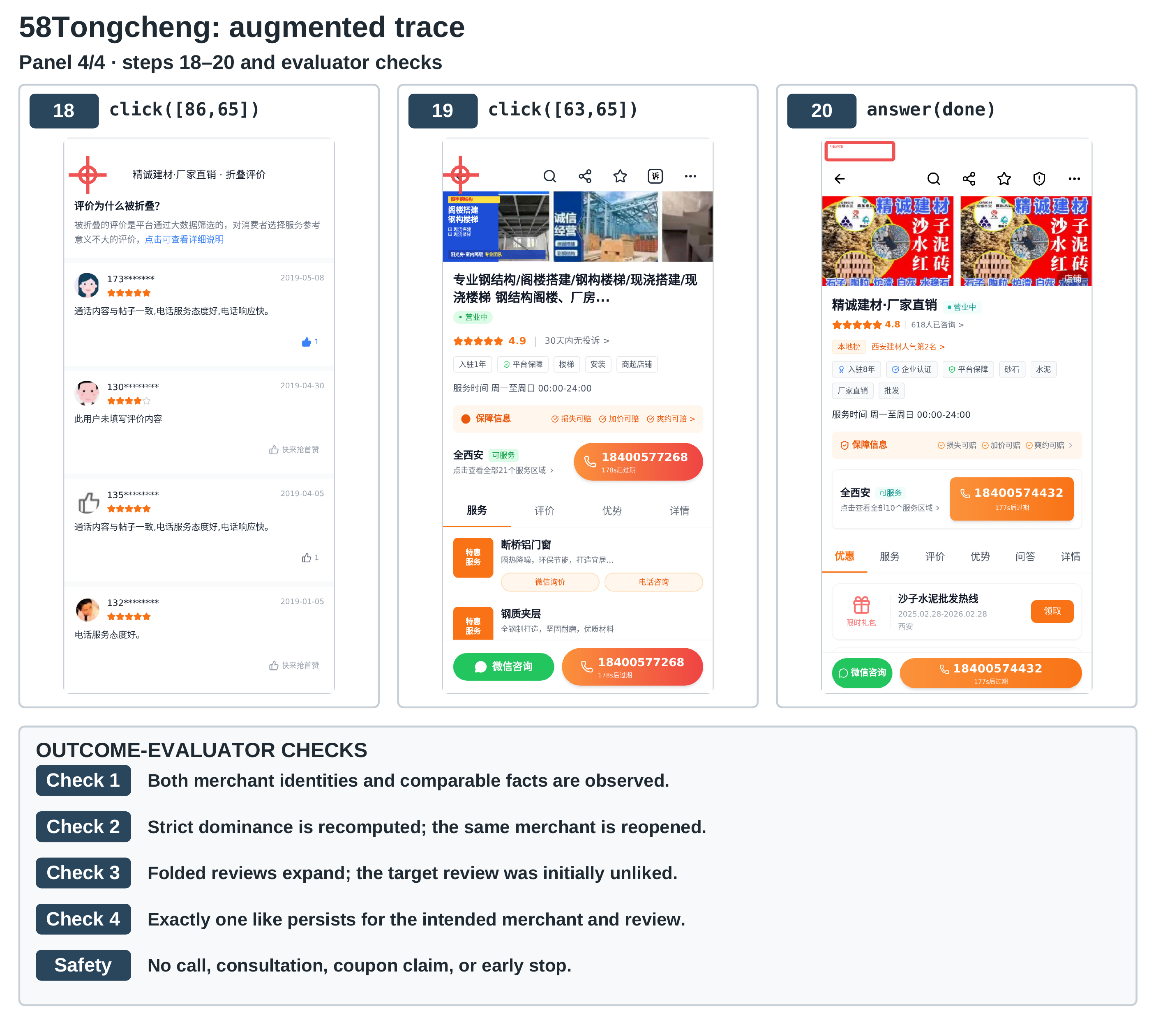}
\caption{\textbf{Complete augmented-task generated-app execution in 58Tongcheng (four pages).} Each
card shows a full marked pre-action viewport and a public-schema action. The 20-action trace retains
three no-new-observation attempts as context-reuse cards and includes return, recomputation, branch
expansion, one persistent like, and recovery from an unintended merchant page. The terminal evaluator
requires the listed checks and vetoes calling, consultation, or coupon collection; the
case remains qualitative.}
\label{fig:trace-augmented-58tc}
\end{figure}

\paragraph{Augmented execution.}
\label{app:augmented-trace-case}
Unlike the seed comparison, this case tests composition and recovery within fixed supported behavior.
Figure~\ref{fig:trace-augmented-58tc} retains rejected attempts rather than presenting only the successful
path; the seed sequence is not duplicated.

\begin{figure}[!htbp]
\centering
\includegraphics[width=\linewidth]{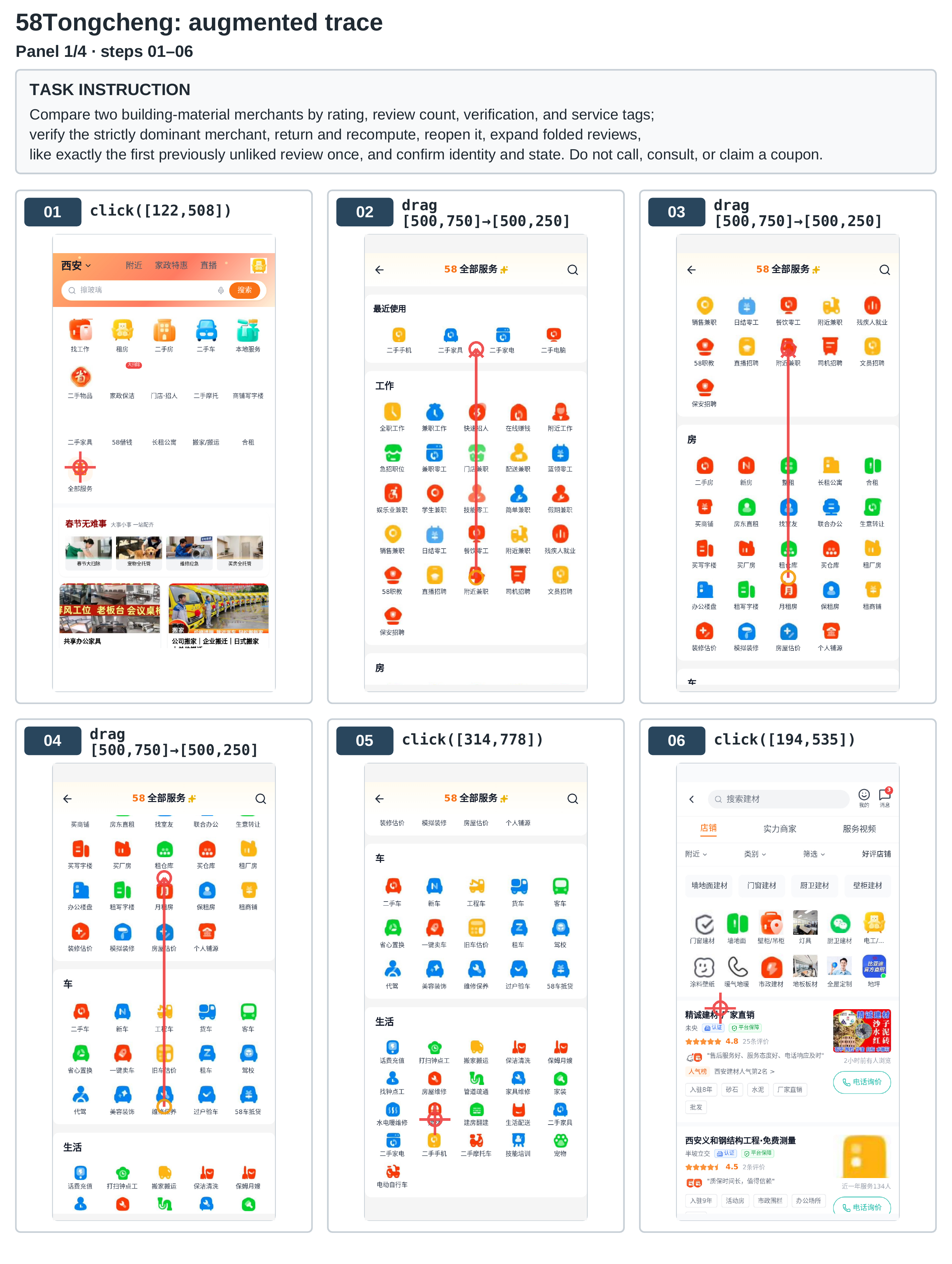}
\end{figure}
\begin{figure}[p]
\centering
\includegraphics[width=\linewidth]{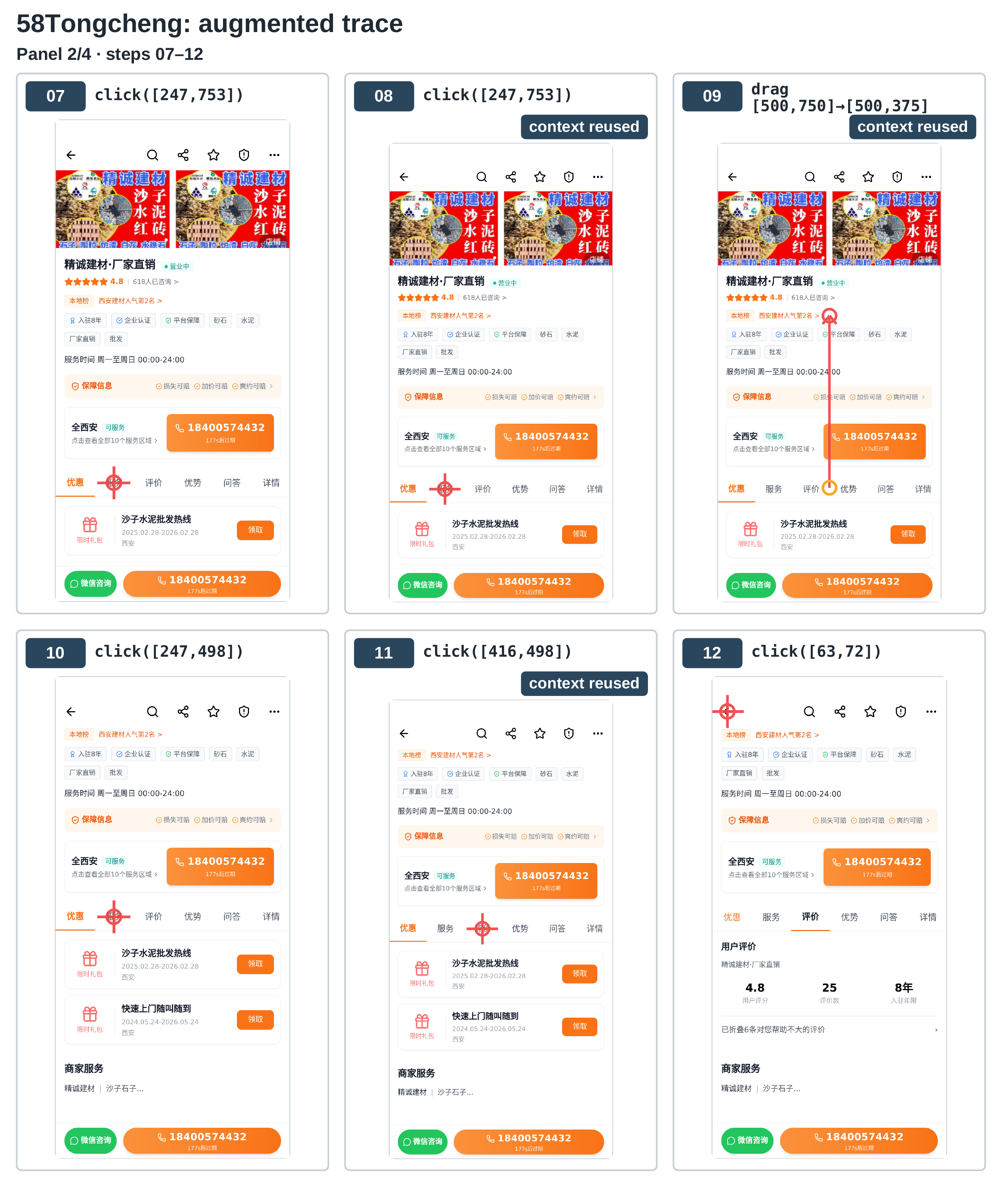}
\end{figure}
\begin{figure}[p]
\centering
\includegraphics[width=\linewidth]{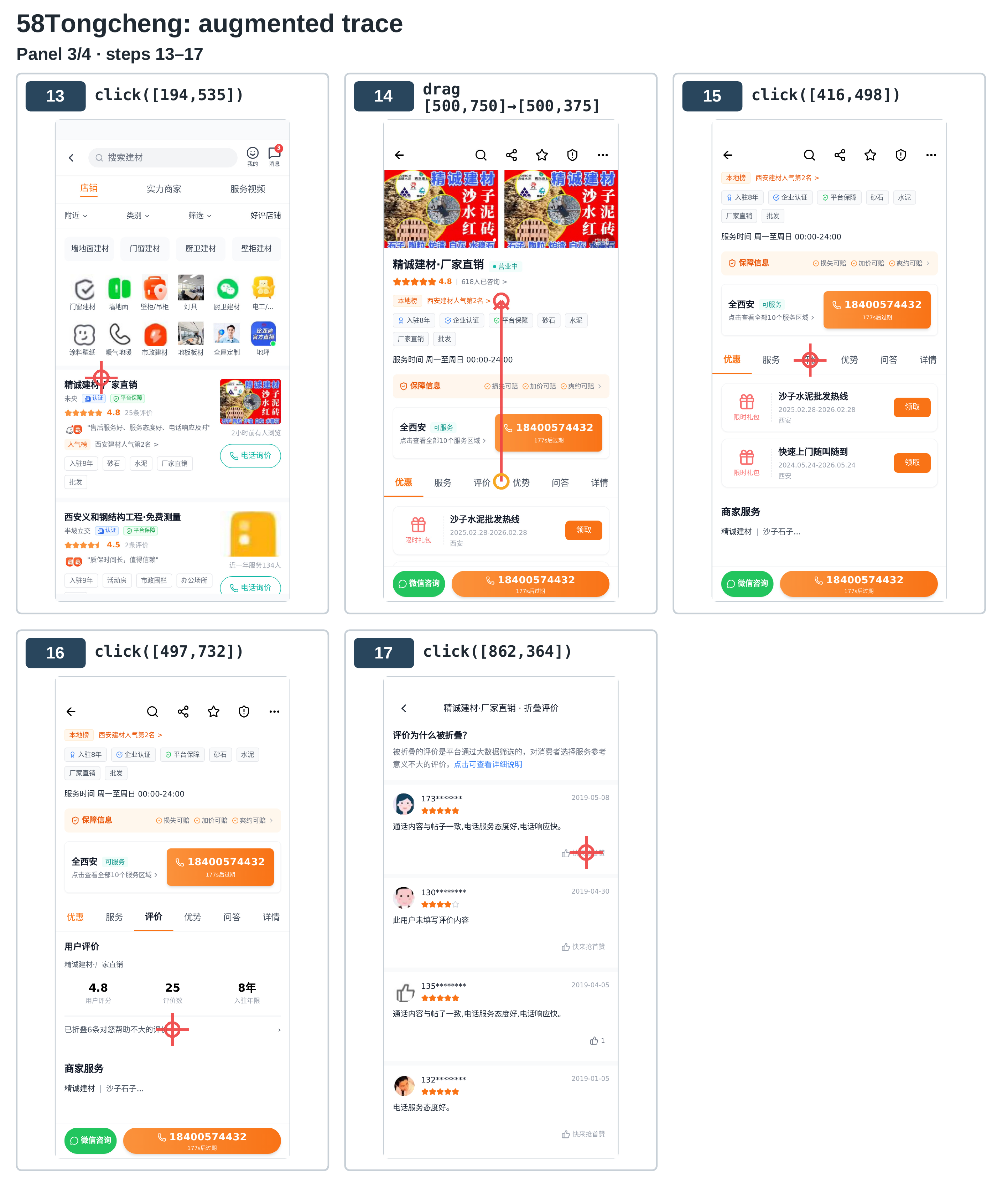}
\end{figure}
\FloatBarrier

\begin{figure}[p]
\centering
\includegraphics[width=\linewidth]{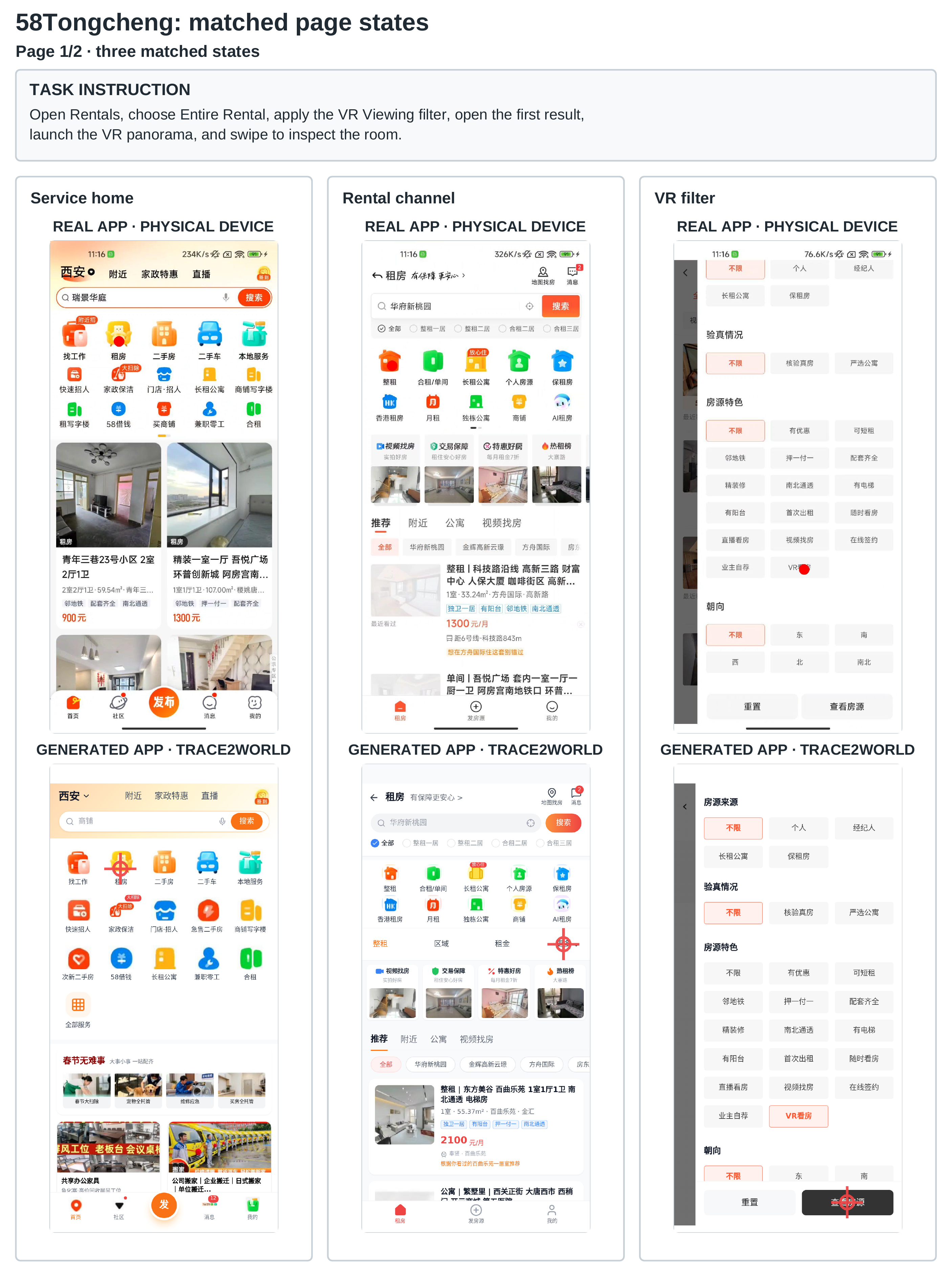}
\end{figure}
\FloatBarrier

\subsection{Cross-Application Return Dependency}
\label{app:cross-showcase}
\label{app:cross-app-case}

\begin{figure}[p]
\centering
\includegraphics[width=\linewidth]{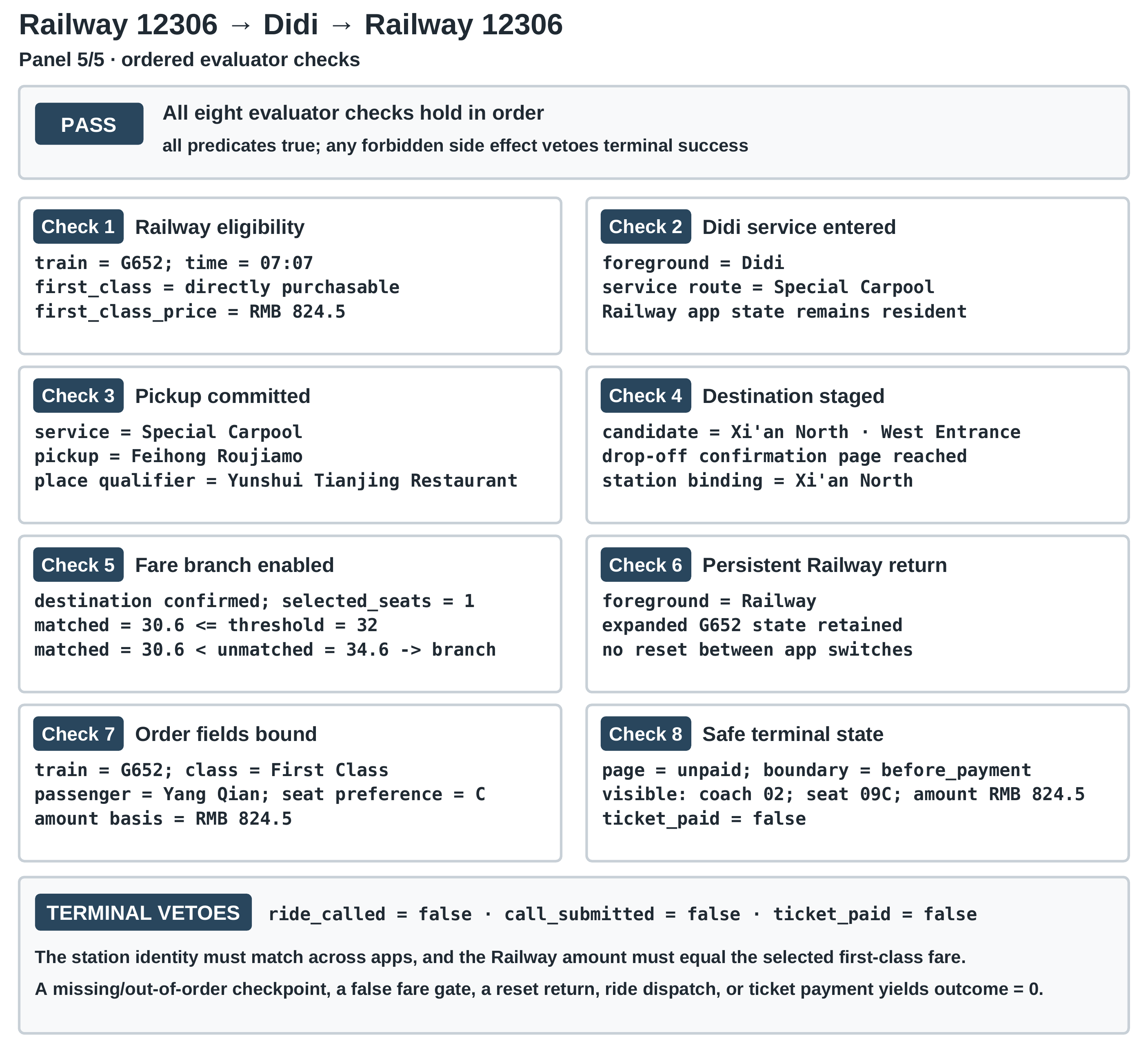}
\caption{\textbf{Railway 12306 $\rightarrow$ Didi $\rightarrow$ Railway 12306 return-dependency case
(20 actions and a terminal observation over five pages).} Numeric card labels are local step indices.
The final panel lists eight ordered evaluator checks for the Railway eligibility, Didi fare gate,
persistent return, order binding, and unpaid terminal, together with the cross-application
bindings and terminal vetoes. The panels assemble separately recorded app-local executions; no single
frozen execution record spans the complete cross-application episode.}
\label{fig:cross-app-case}
\end{figure}

This case tests a return dependency rather than two independent completions: the matched and unmatched
Didi quotes of RMB 30.6 and RMB 34.6 must remain available after resuming Railway 12306, and the selected
quote must bind the final unpaid order. Figure~\ref{fig:cross-app-case} exposes the joint checkpoints and
terminal vetoes.

\begin{figure}[!ht]
\centering
\includegraphics[width=\linewidth]{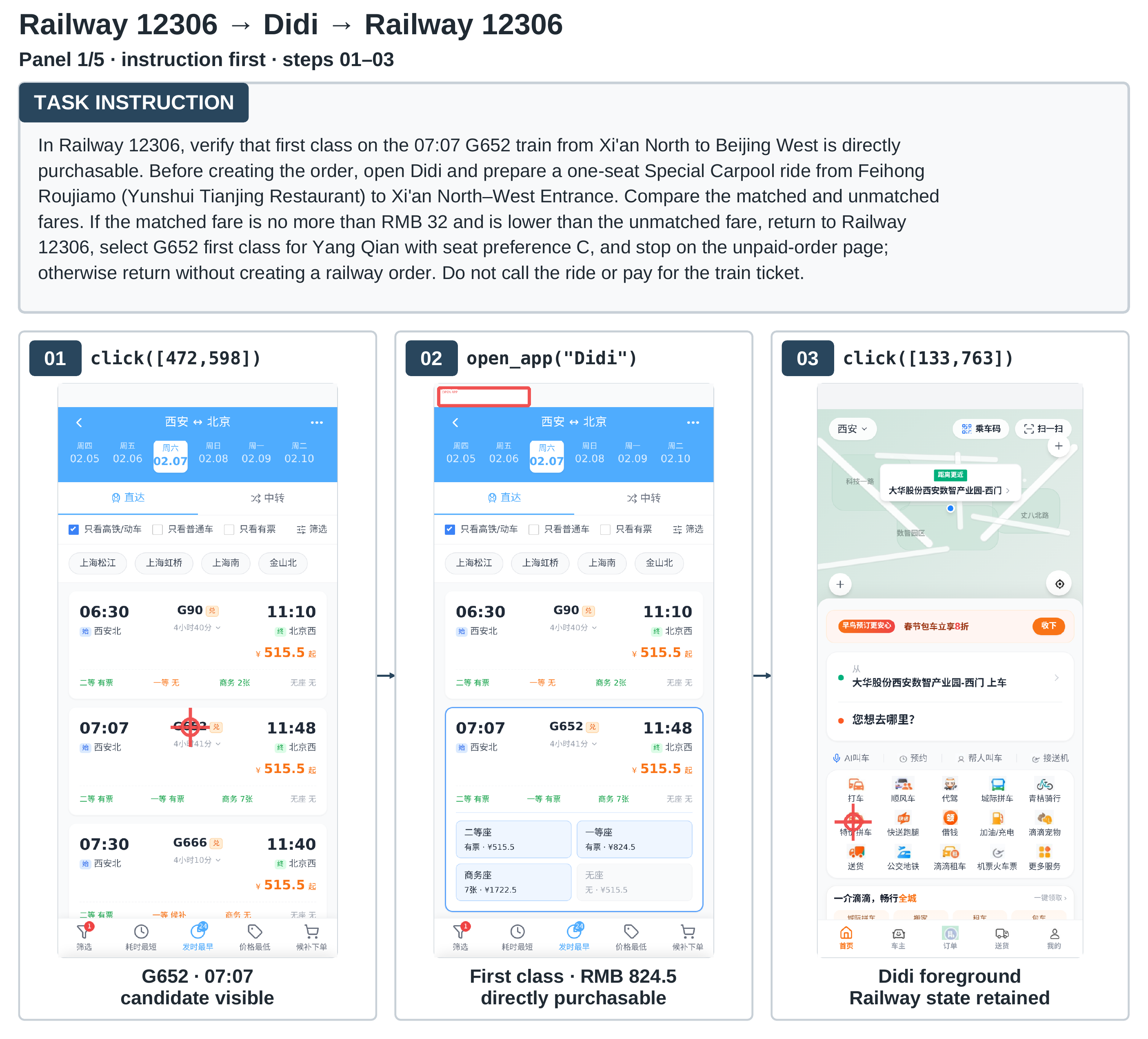}
\end{figure}
\begin{figure}[p]
\centering
\includegraphics[width=\linewidth]{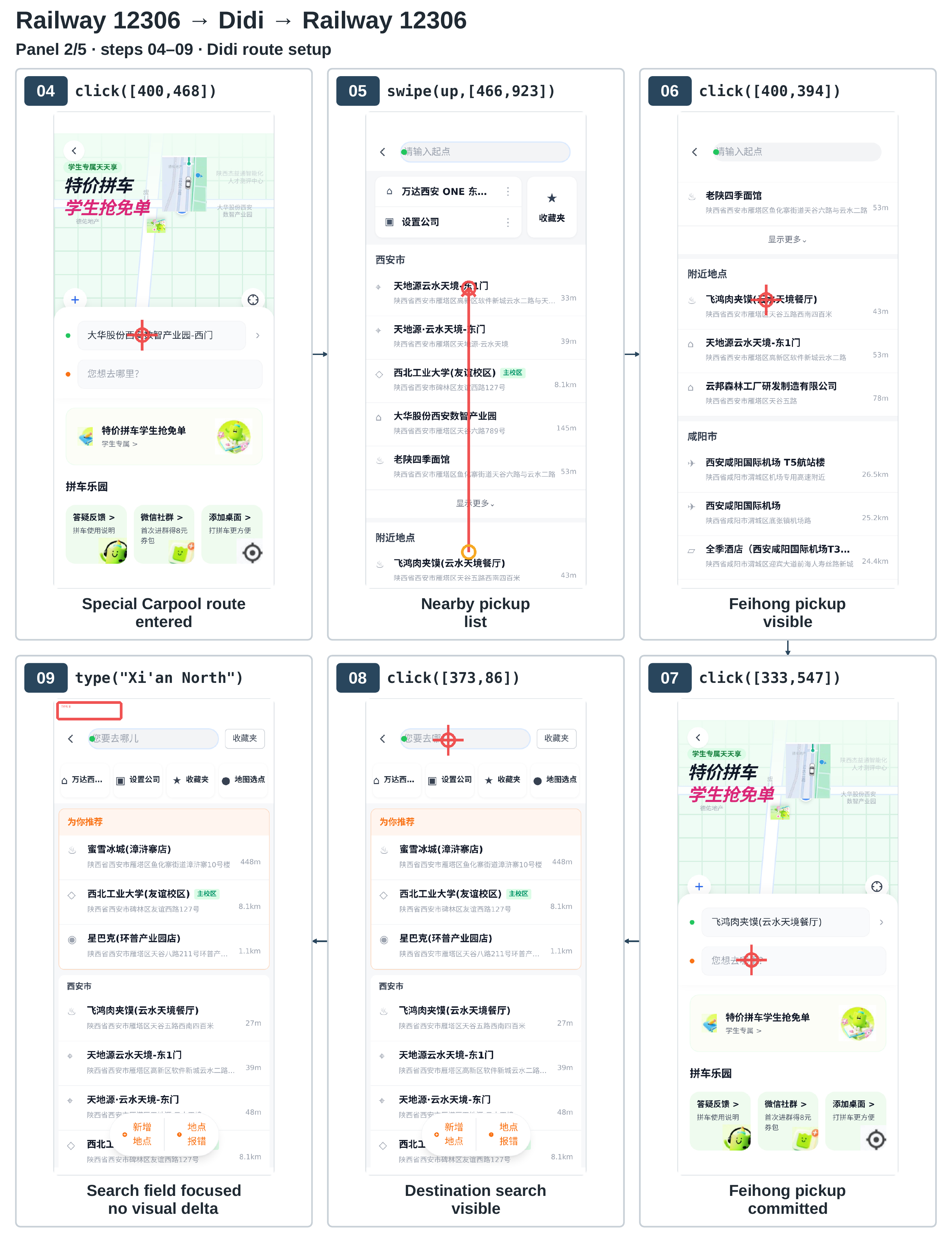}
\end{figure}
\begin{figure}[p]
\centering
\includegraphics[width=\linewidth]{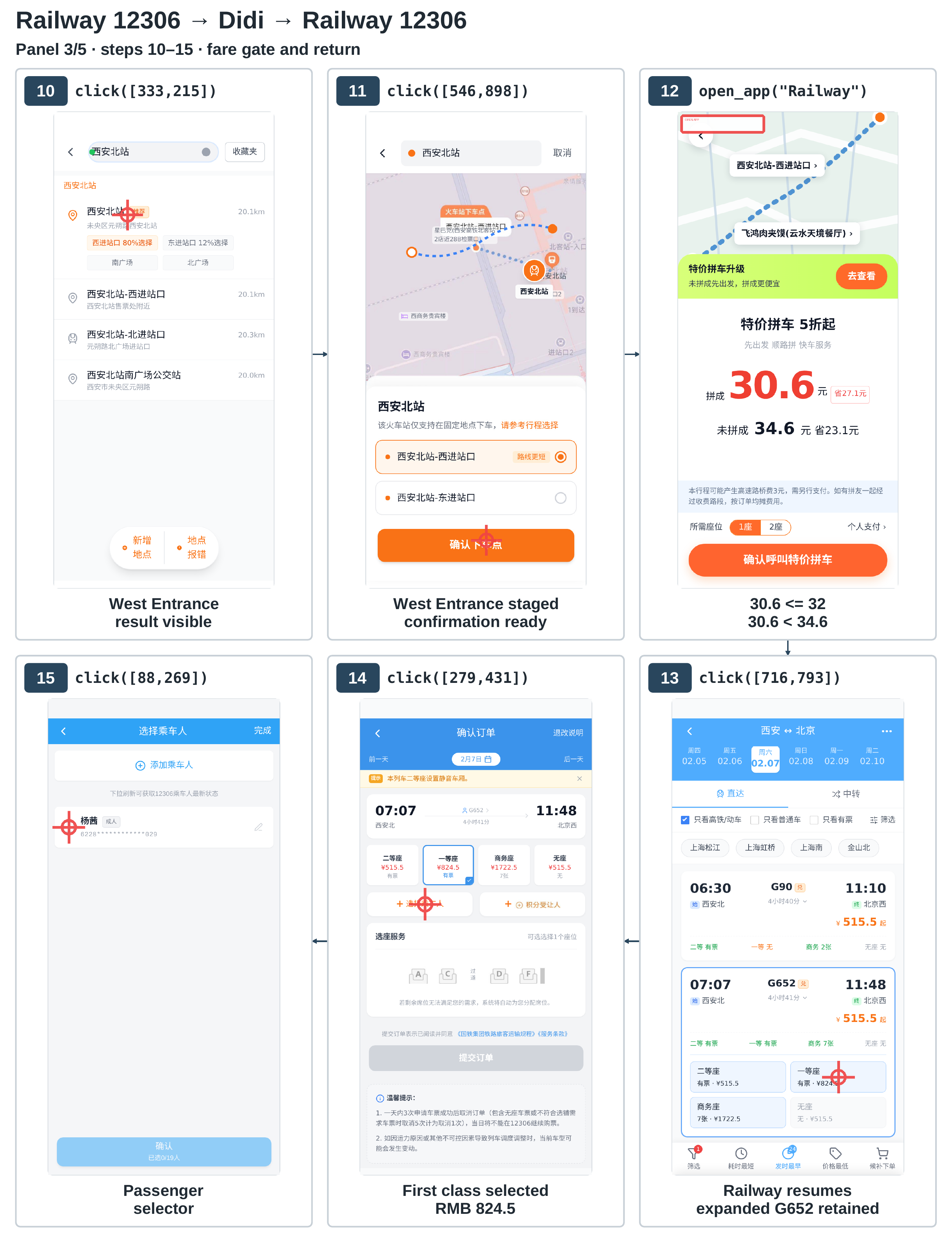}
\end{figure}
\begin{figure}[p]
\centering
\includegraphics[width=\linewidth]{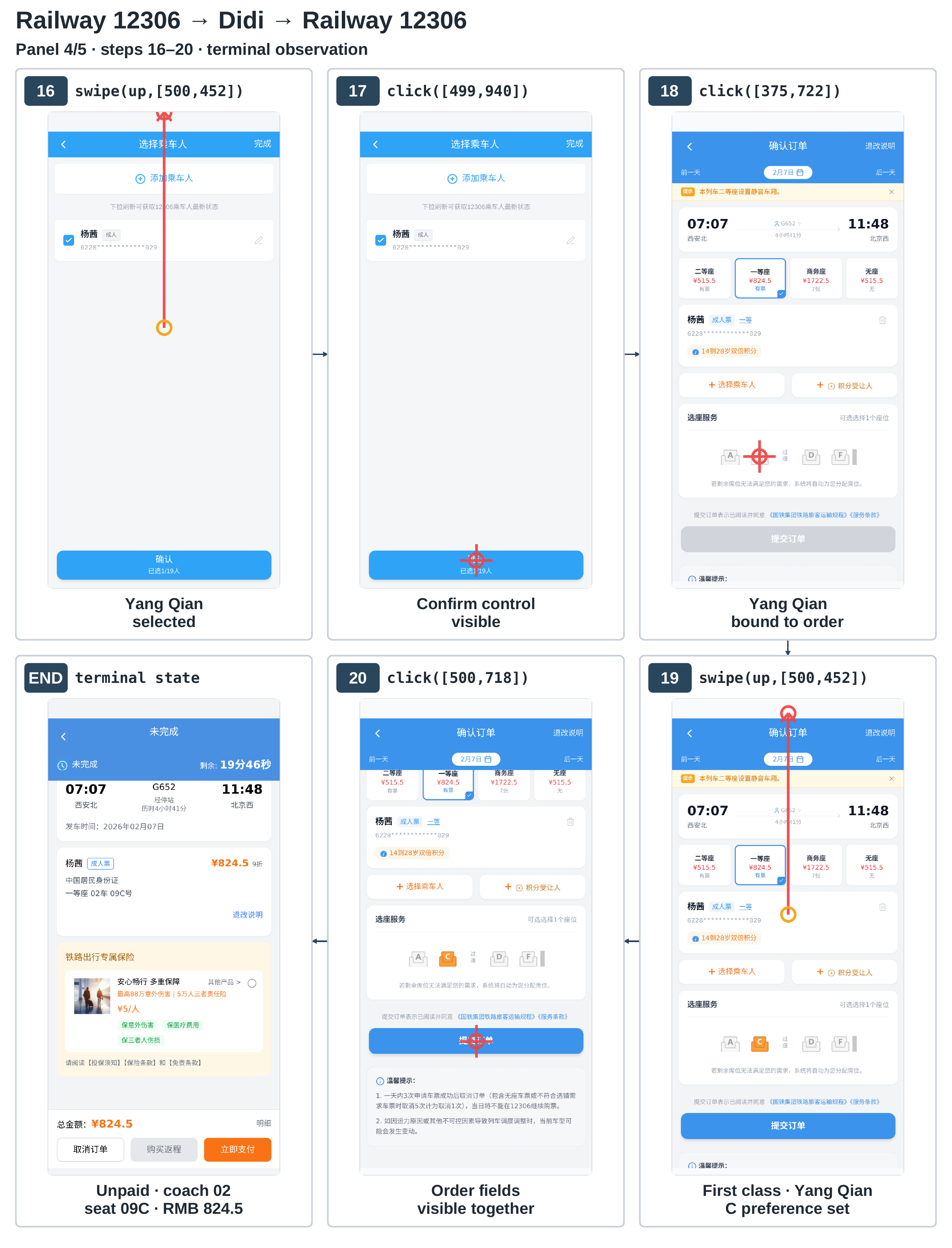}
\end{figure}
\FloatBarrier

\subsection{Reconstruction Gallery}
\label{app:gallery-showcase}

The gallery pairs physical-device observations from human source traces with archived environment
executions at the same task-conditioned page state. Every panel retains the full marked pre-action
viewport; marks identify pending public-schema action geometry. The gallery aligns page identity and
task-relevant visible state while allowing different legitimate paths and non-task visual differences.

\begin{figure}[!htbp]
\centering
\includegraphics[width=\linewidth]{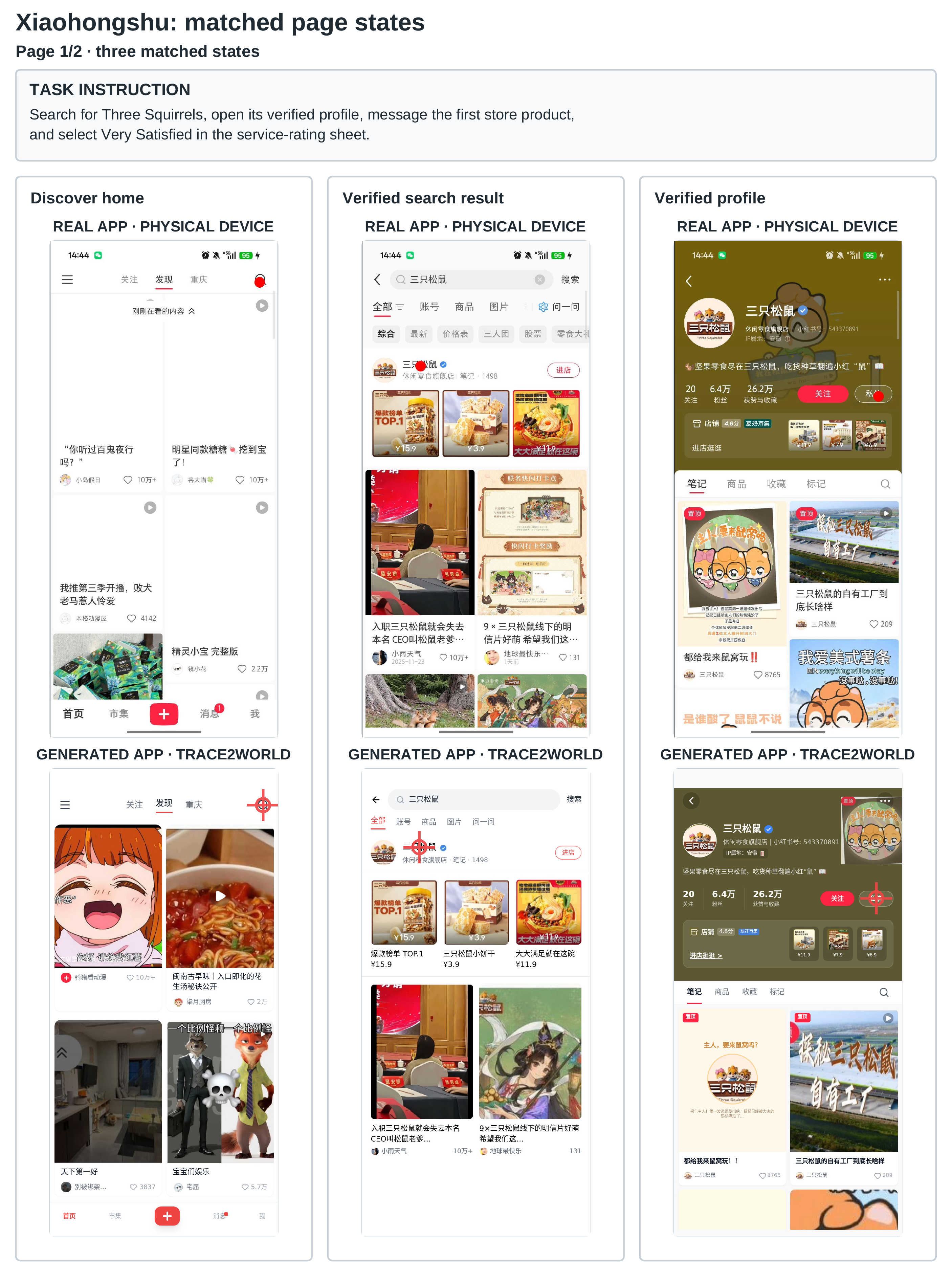}
\end{figure}
\begin{figure}[p]
\centering
\includegraphics[width=\linewidth]{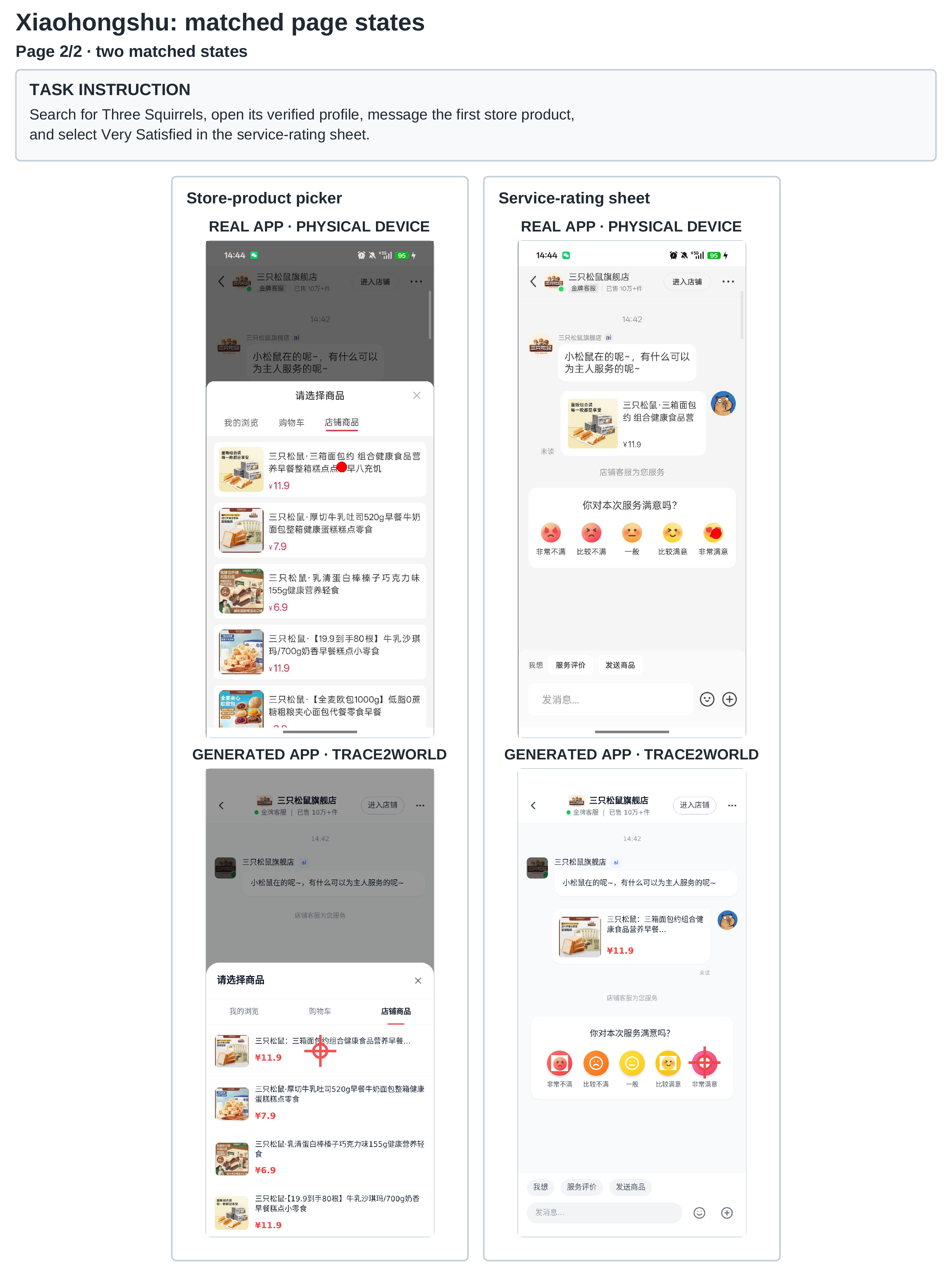}
\caption{\textbf{Xiaohongshu reconstruction gallery (five matched page states across two figure
pages).} The paired pages cover the discover shell, verified search result, verified profile,
store-product picker, and service-rating sheet. Each pair shares one task, launch, and task-conditioned
visible state; differences outside that state remain visible.}
\label{fig:pagewise-xiaohongshu}
\end{figure}

\begin{figure}[p]
\centering
\includegraphics[width=\linewidth]{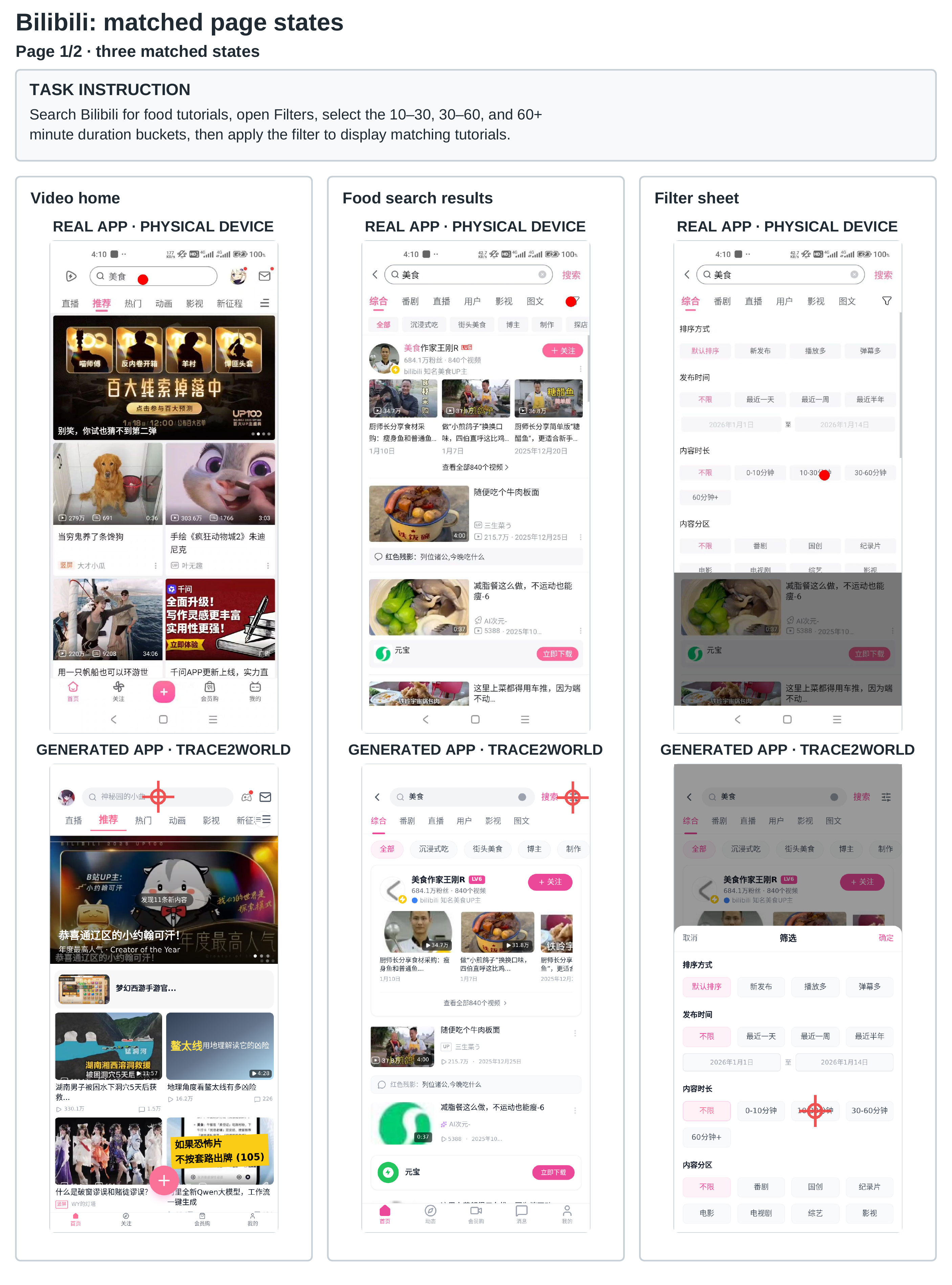}
\end{figure}
\begin{figure}[p]
\centering
\includegraphics[width=\linewidth]{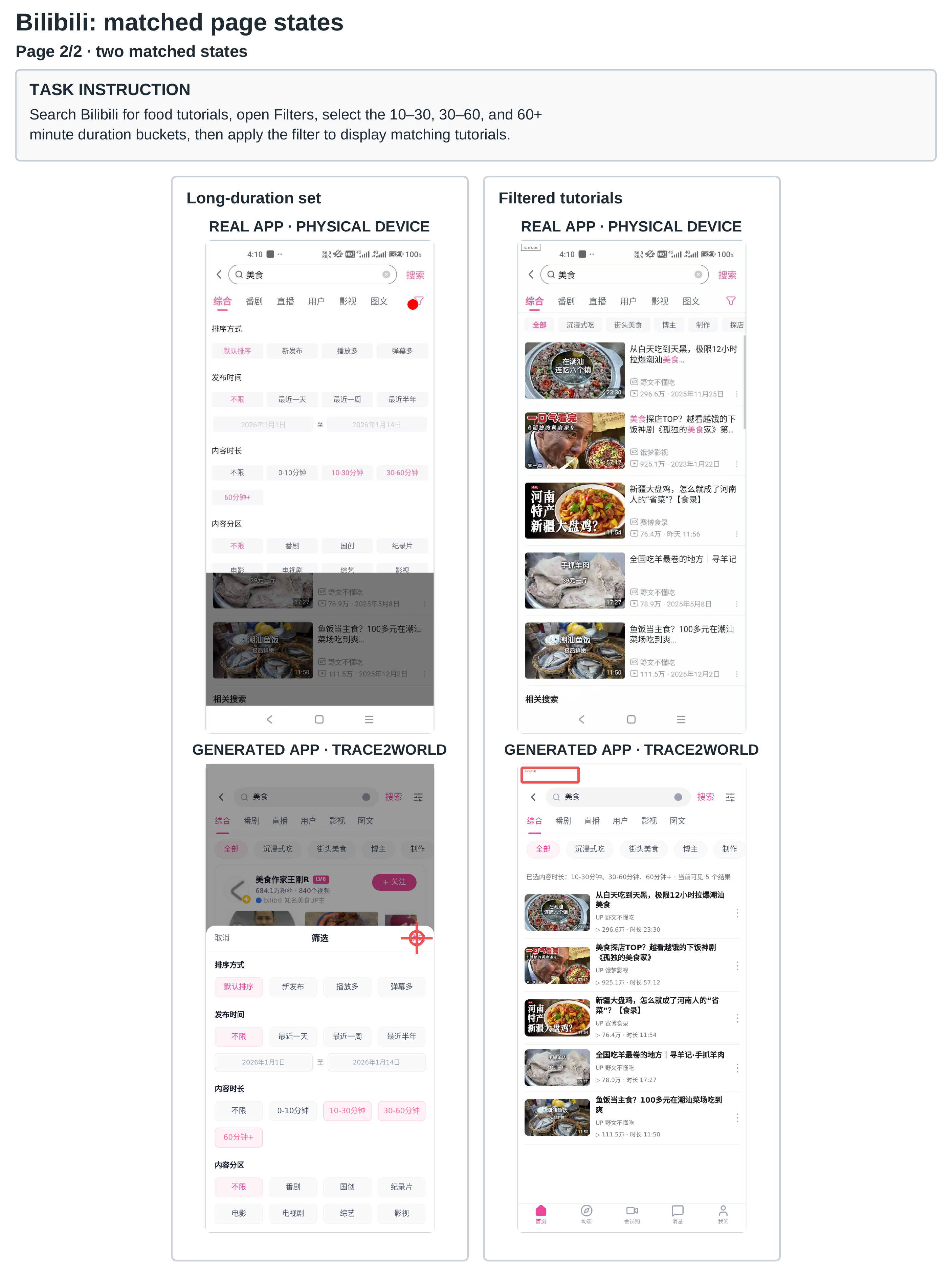}
\caption{\textbf{Bilibili reconstruction gallery (five matched page states across two figure
pages).} The paired pages cover video home, food-query results, the filter sheet, three selected
long-duration buckets, and the resulting tutorial list.}
\label{fig:pagewise-bilibili}
\end{figure}

\begin{figure}[p]
\centering
\includegraphics[width=\linewidth]{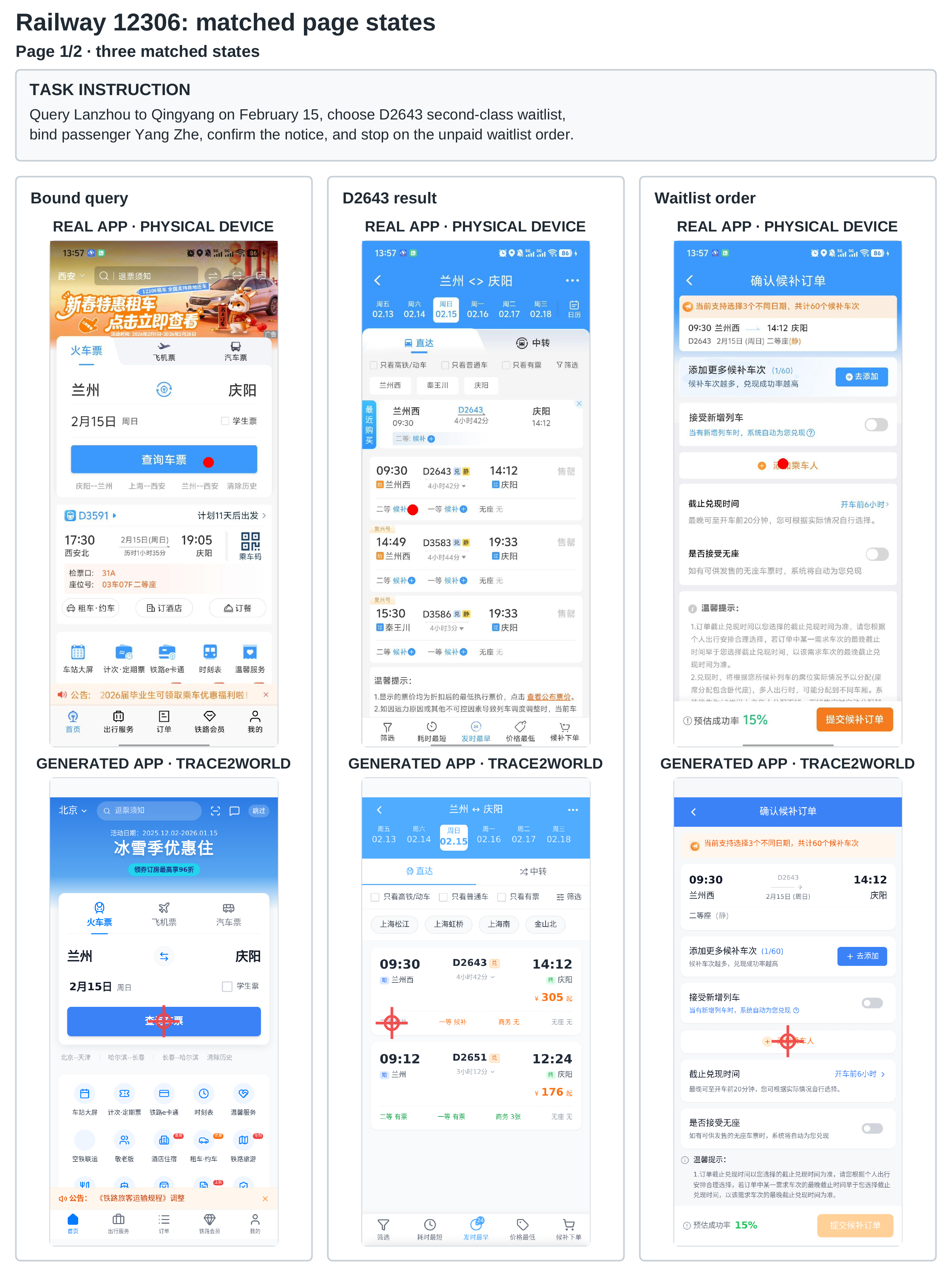}
\end{figure}
\begin{figure}[p]
\centering
\includegraphics[width=\linewidth]{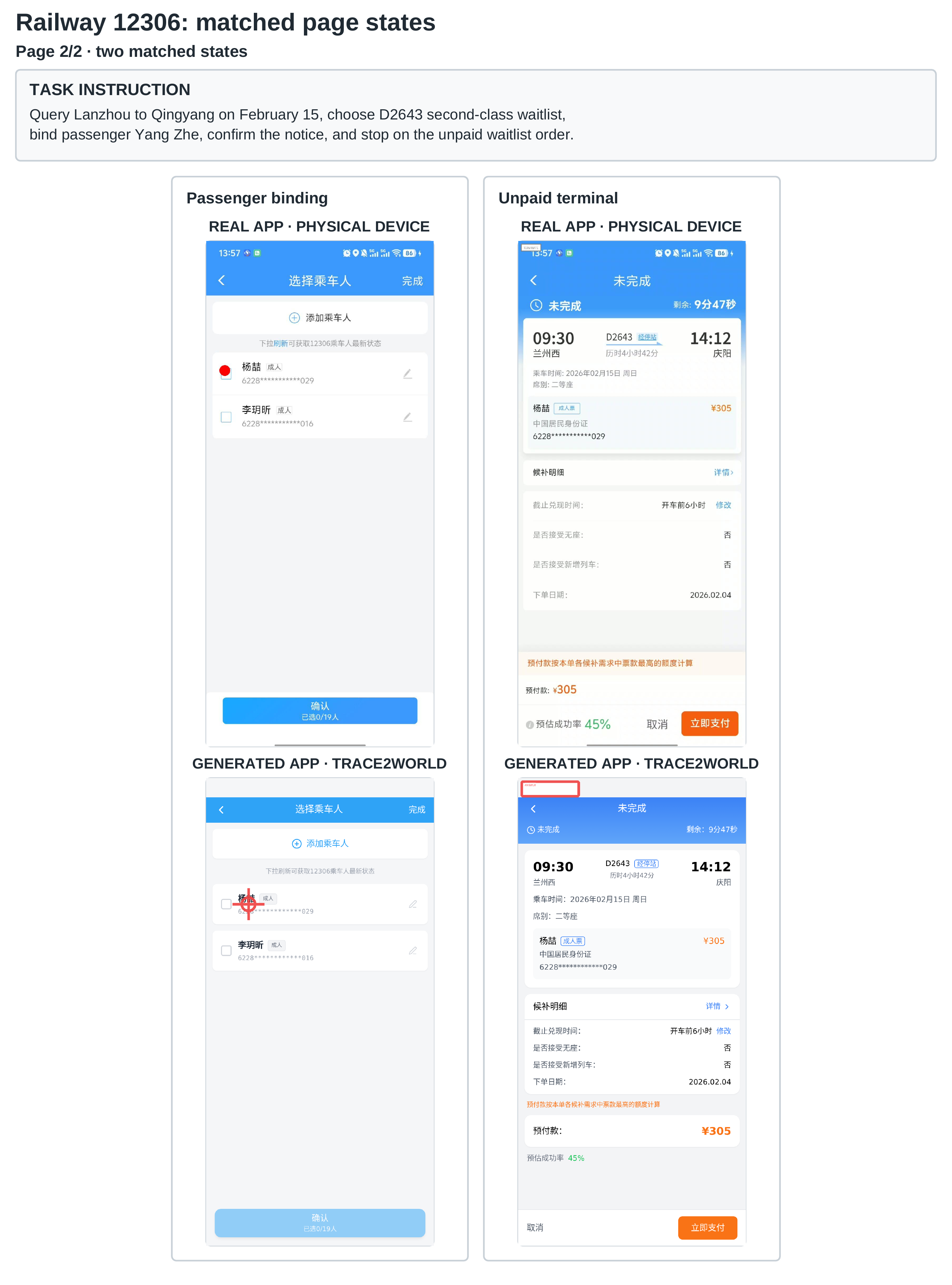}
\caption{\textbf{Railway 12306 reconstruction gallery (five matched page states across two figure
pages).} The paired pages cover a date-and-route query, the D2643 result, waitlist confirmation,
passenger binding, and the unpaid terminal order.}
\label{fig:pagewise-railway12306}
\end{figure}

\begin{figure}[p]
\centering
\includegraphics[width=\linewidth]{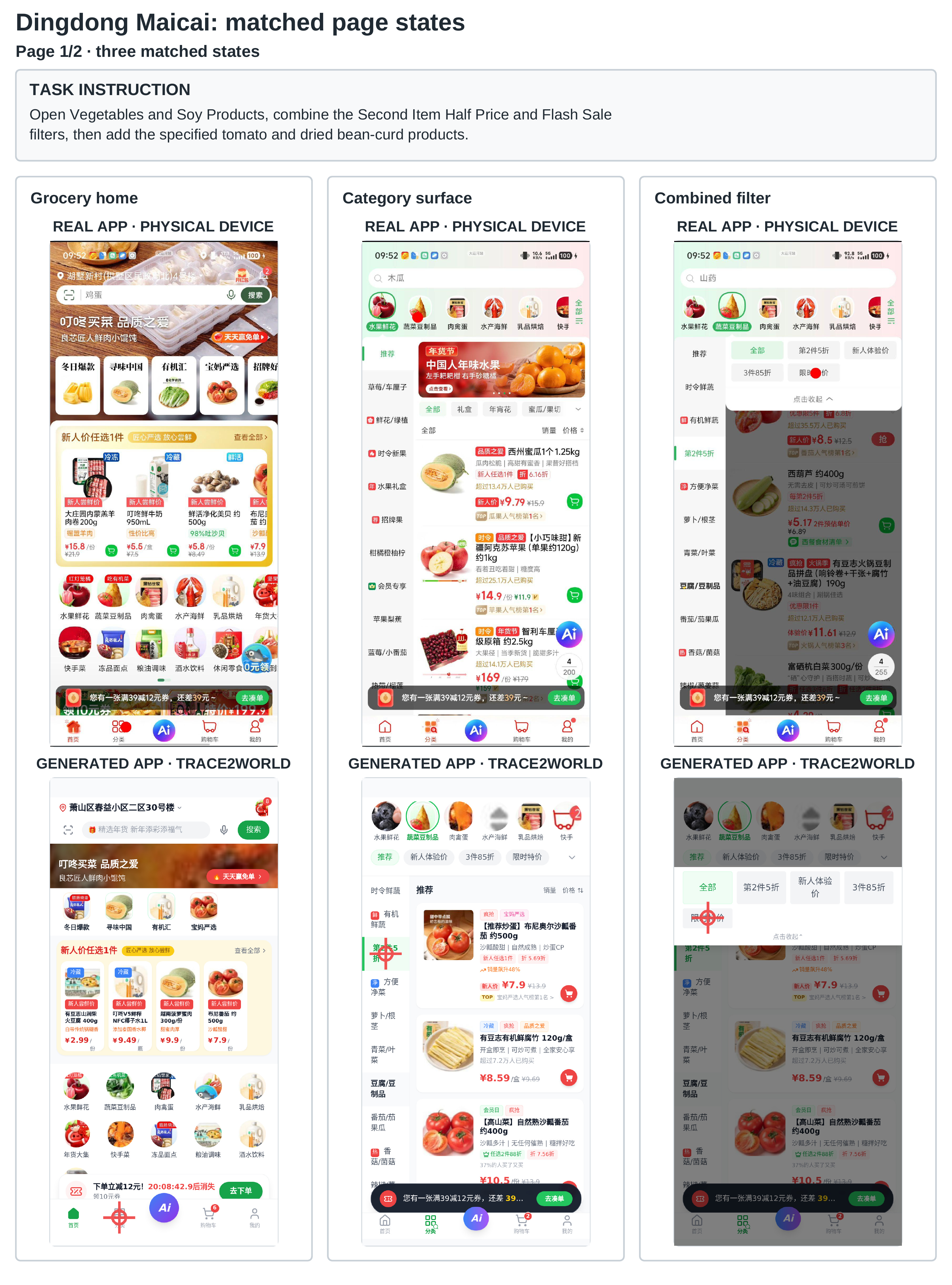}
\end{figure}
\begin{figure}[p]
\centering
\includegraphics[width=\linewidth]{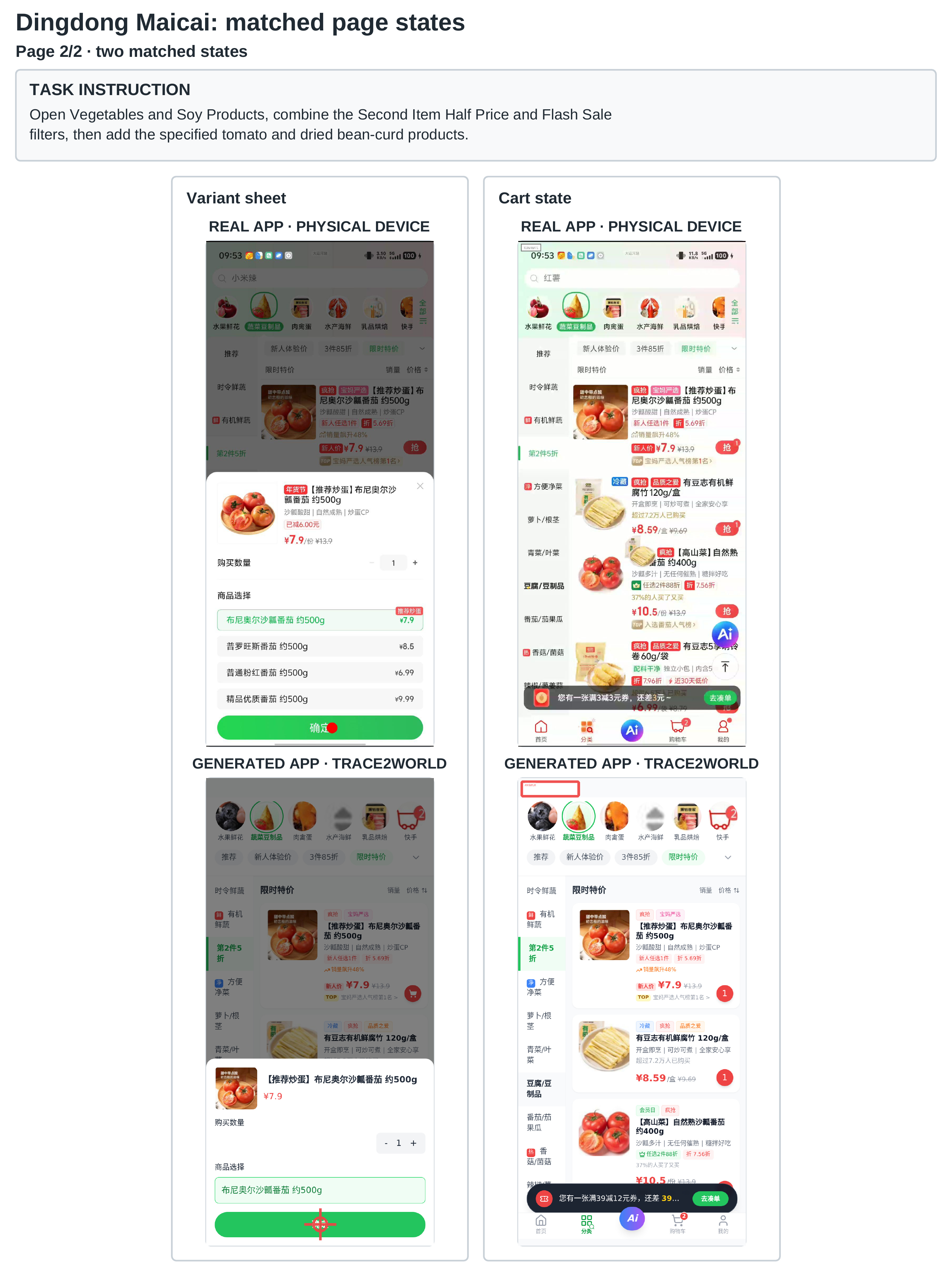}
\caption{\textbf{Dingdong Maicai reconstruction gallery (five matched page states across two figure
pages).} The paired pages cover grocery home, category entry, combined promotion filters, a
product-variant sheet, and the resulting cart state.}
\label{fig:pagewise-dingdong}
\end{figure}

\paragraph{Twenty-application home-screen gallery.}
\label{app:home-showcase}
\begin{figure}[H]\centering
\appcover{Xiaohongshu}{\includegraphics{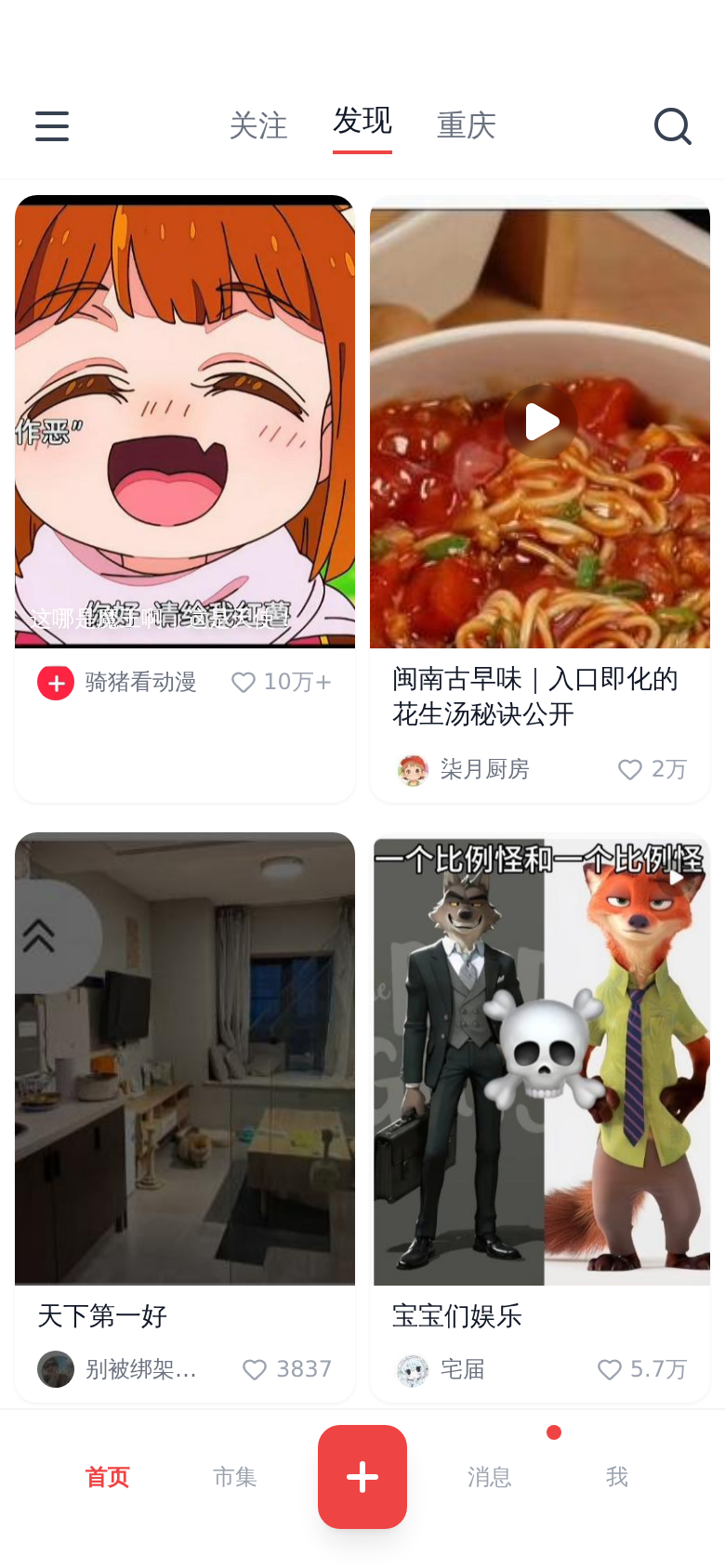}}
\hfill
\appcover{Weibo}{\includegraphics{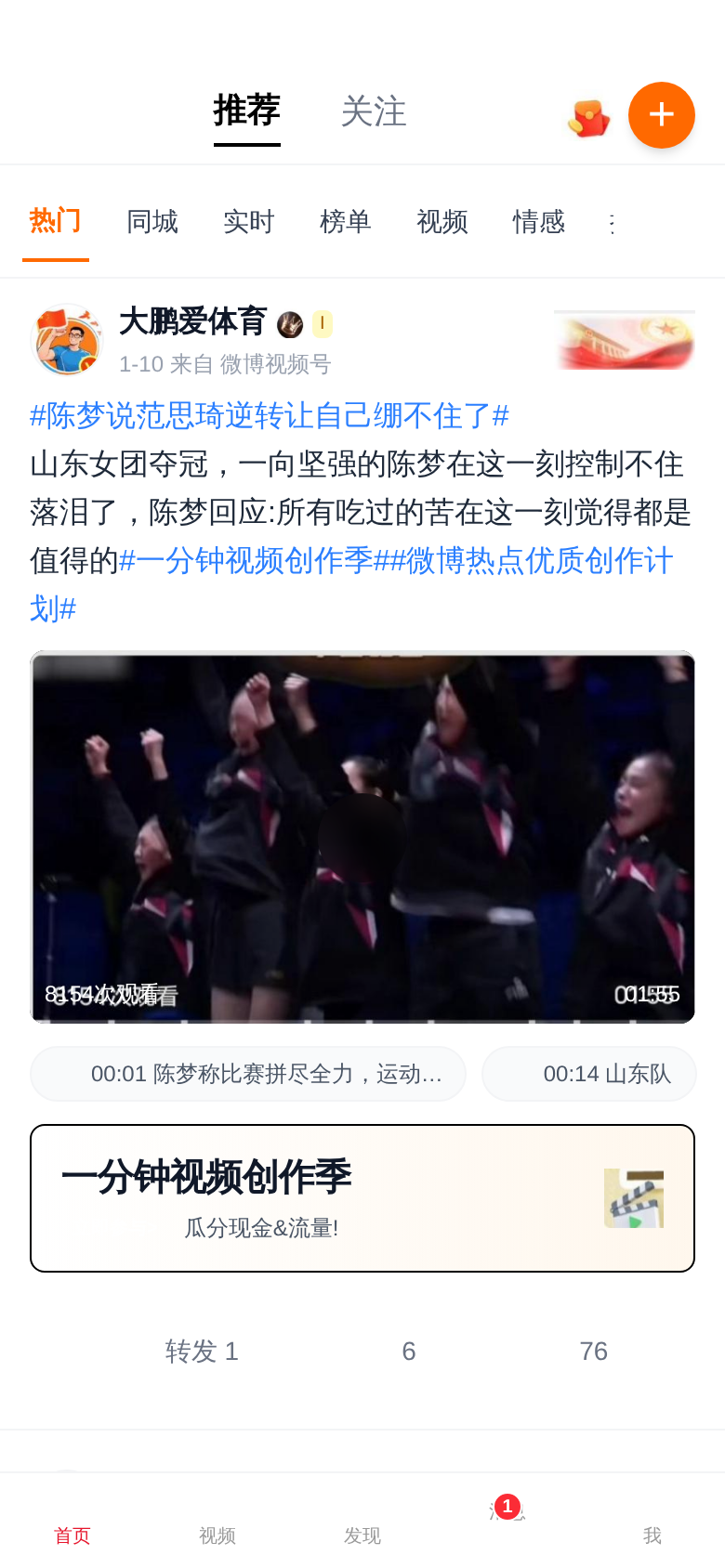}}
\hfill
\appcover{Bilibili}{\includegraphics{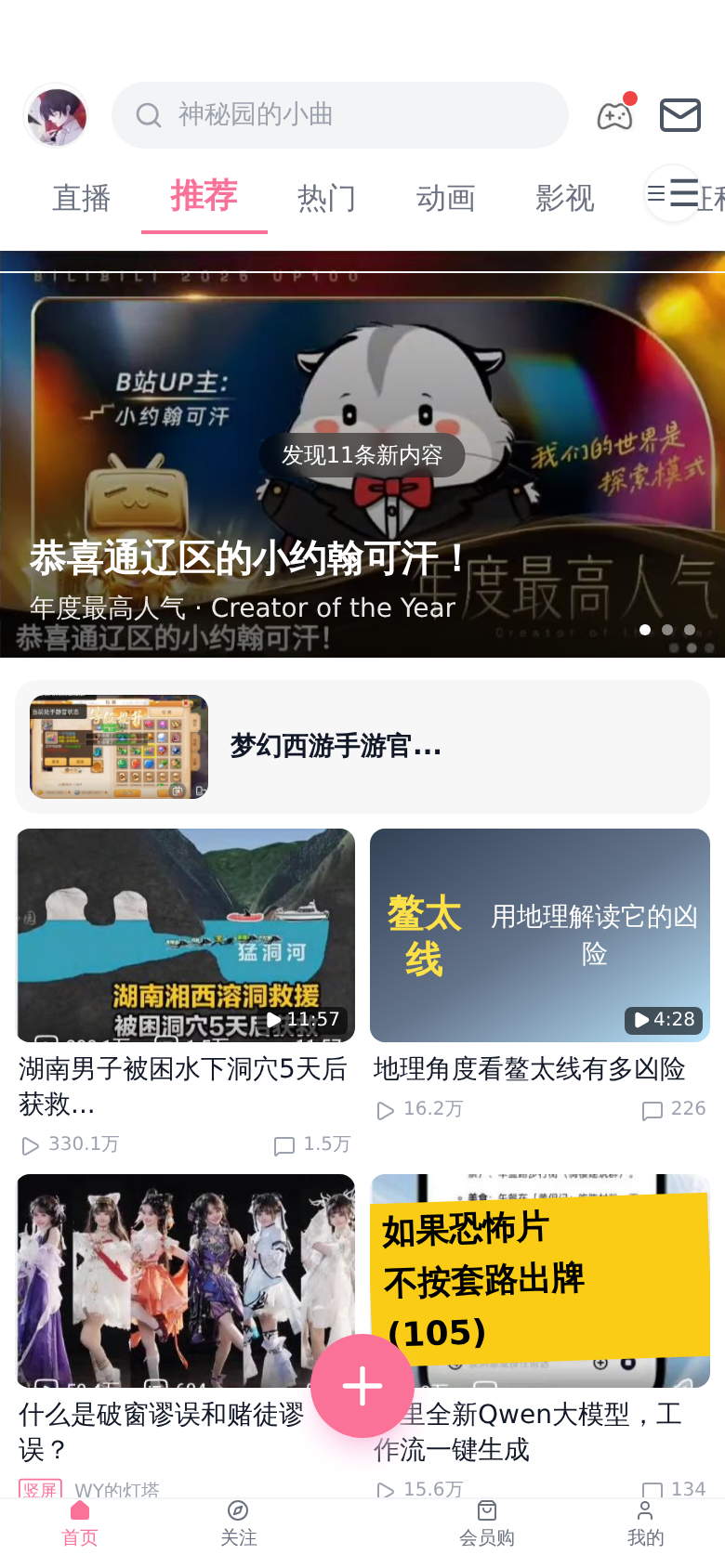}}
\hfill
\appcover{Toutiao}{\includegraphics{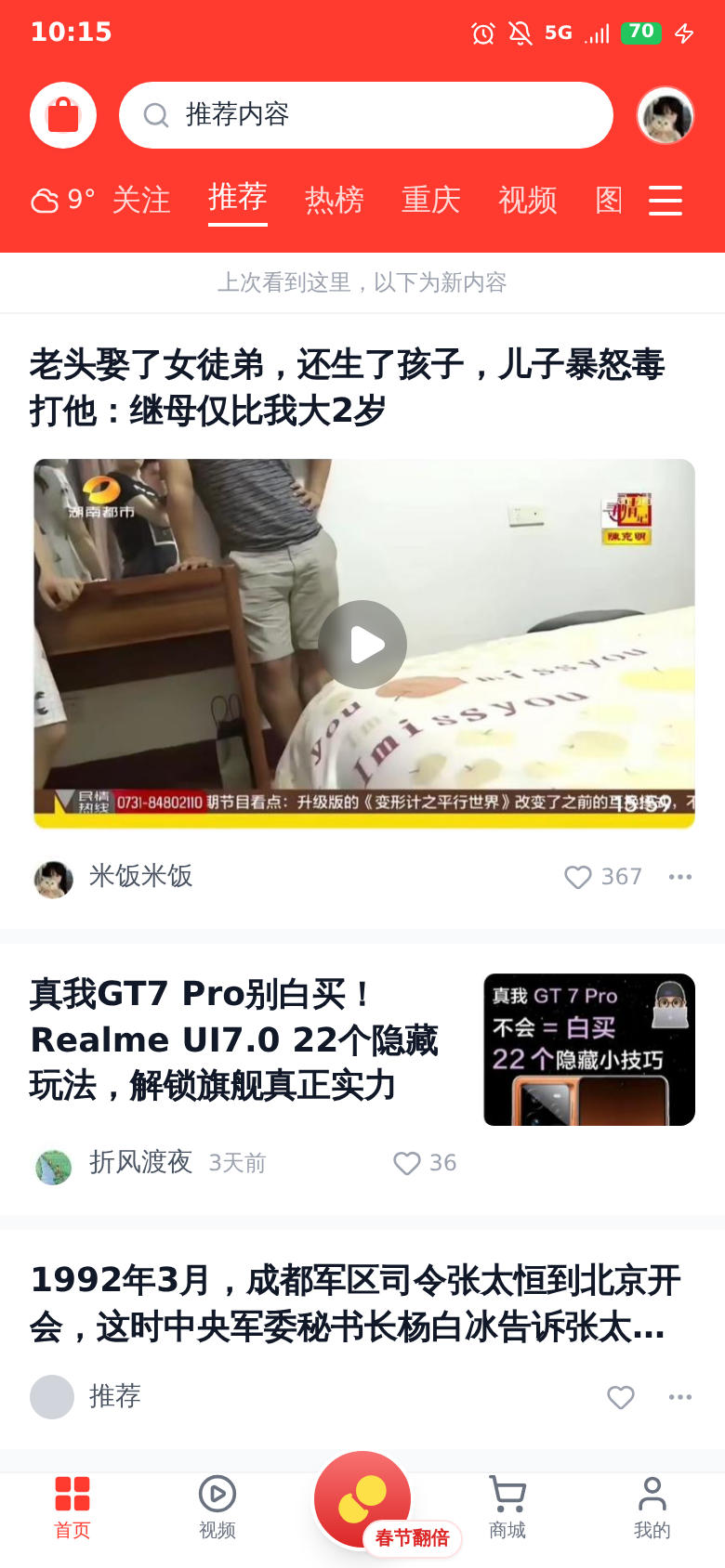}}\par\medskip
\appcover{KuGou}{\includegraphics{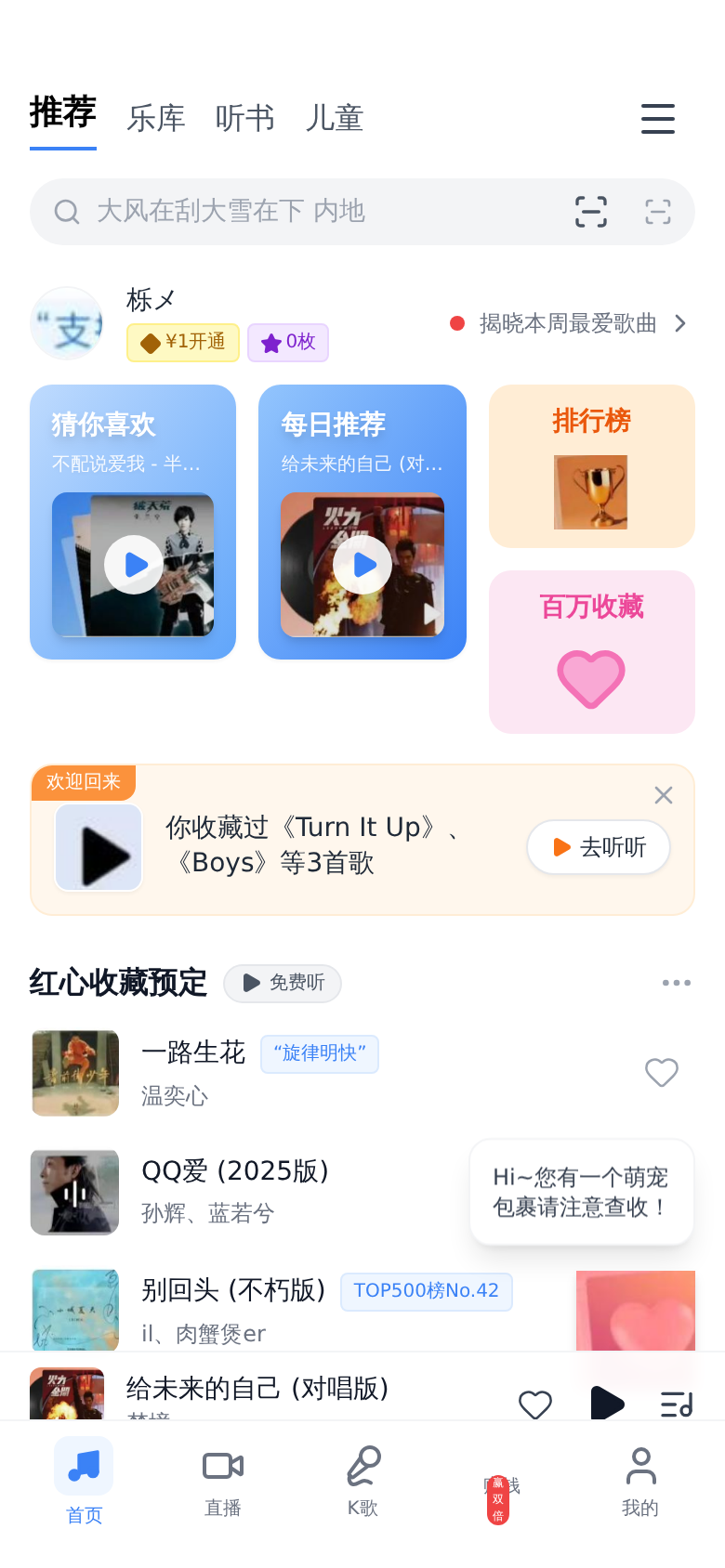}}
\hfill
\appcover{NetEase Music}{\includegraphics{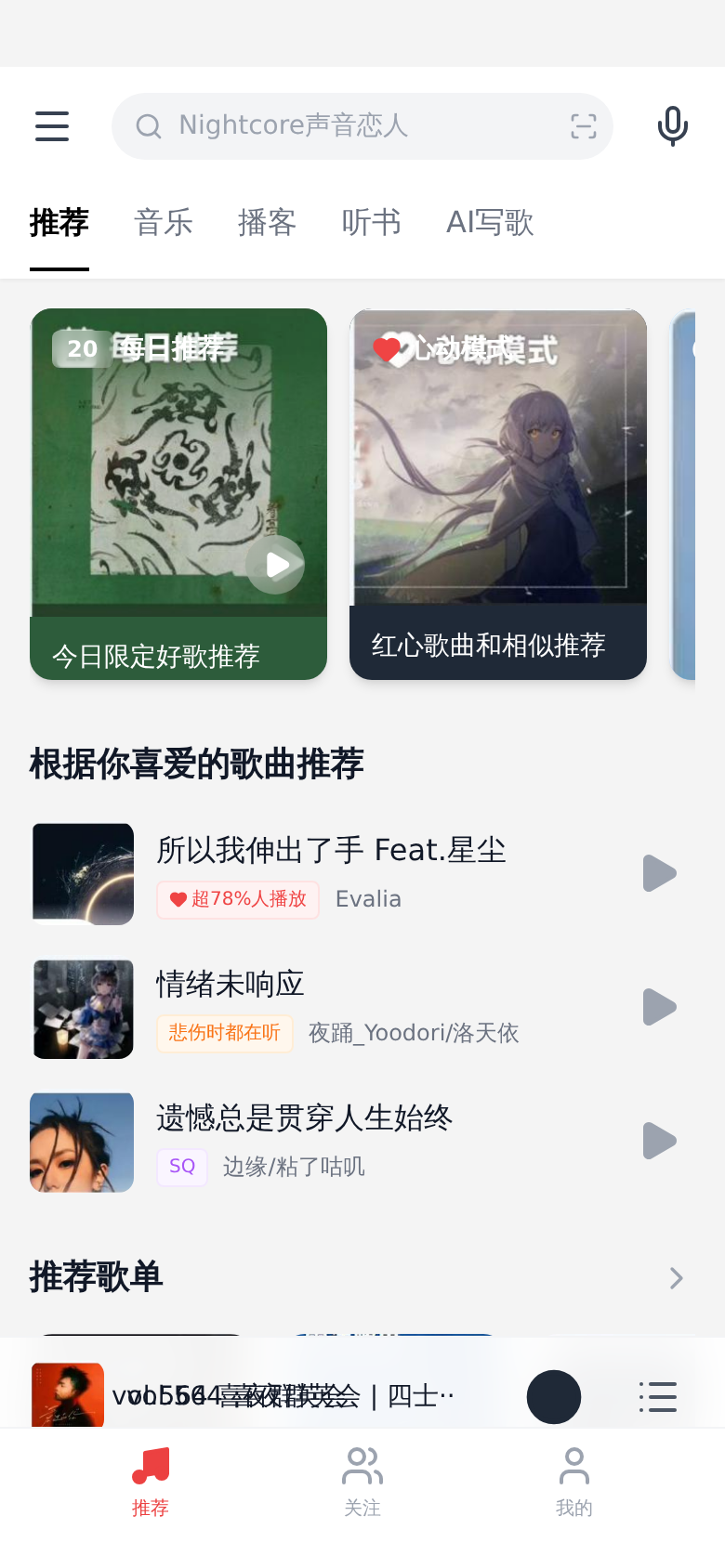}}
\hfill
\appcover{Ximalaya}{\includegraphics{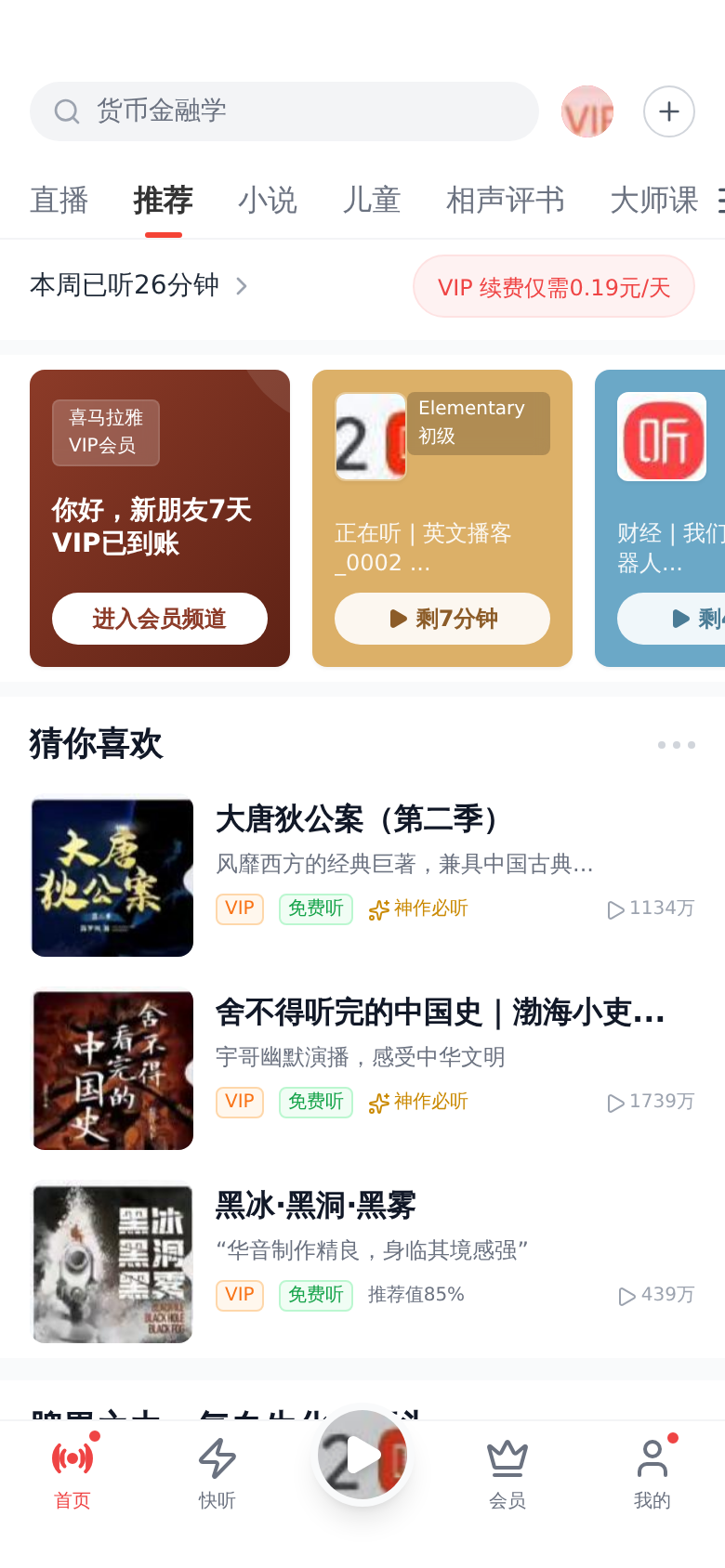}}\par\medskip
\appcover{DingTalk}{\includegraphics{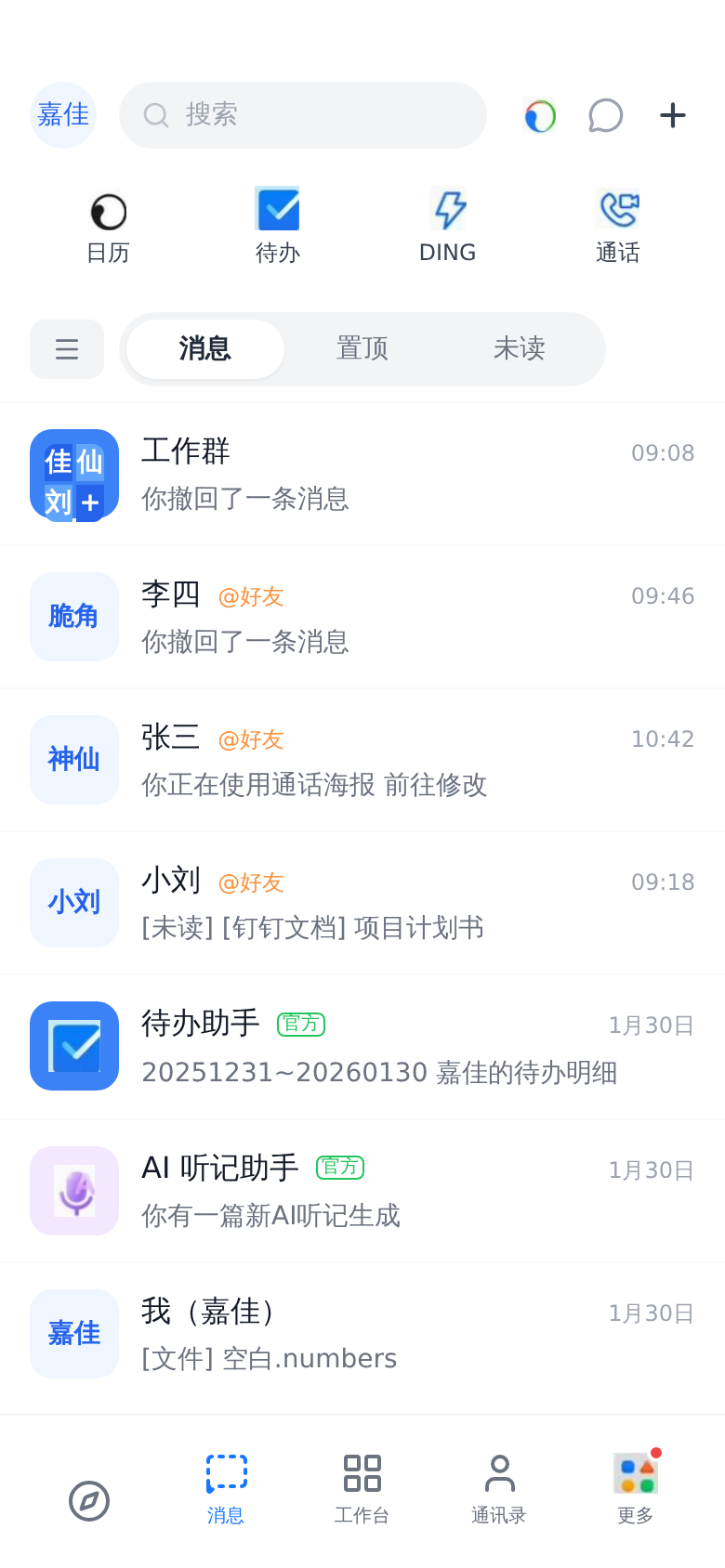}}
\hfill
\appcover{Youdao Dictionary}{\includegraphics{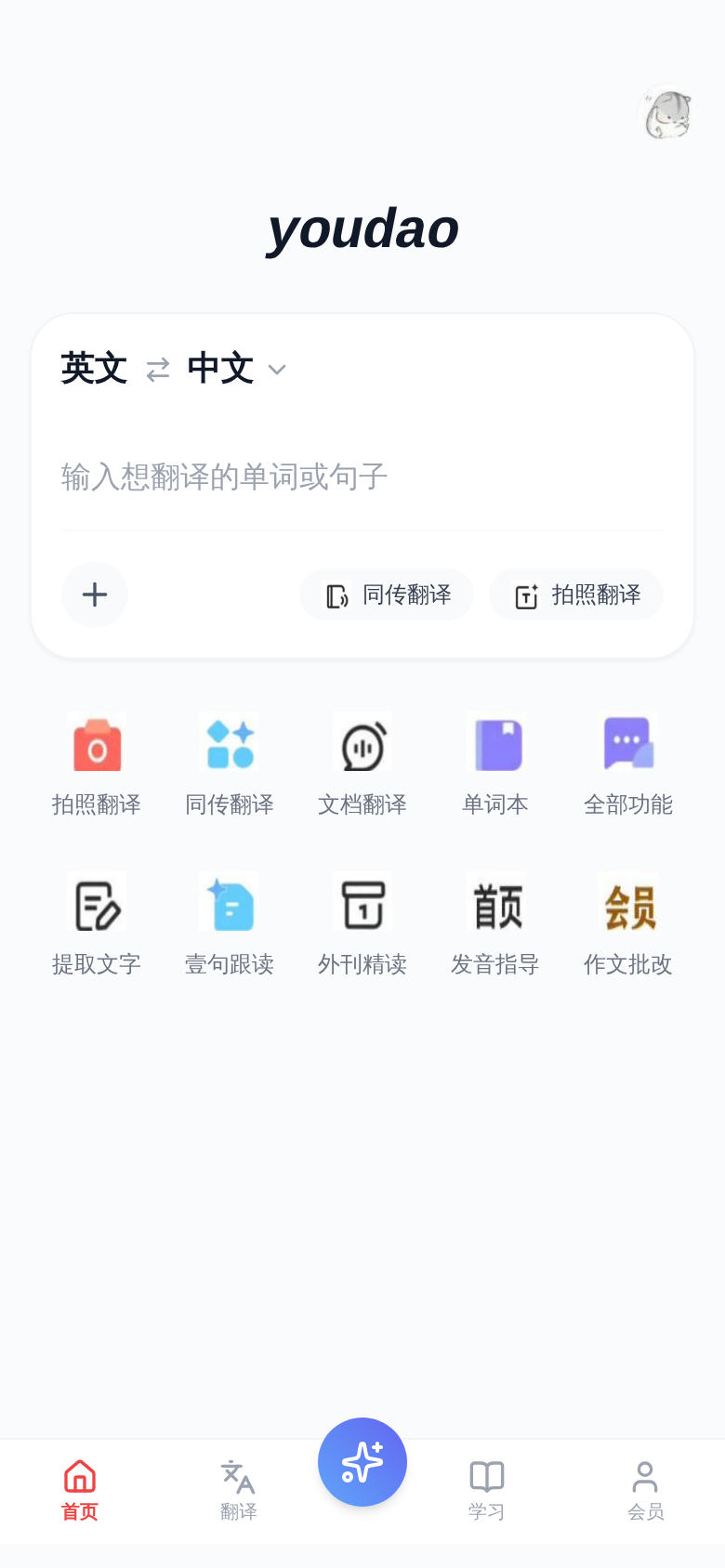}}
\hfill
\appcover{Damai}{\includegraphics{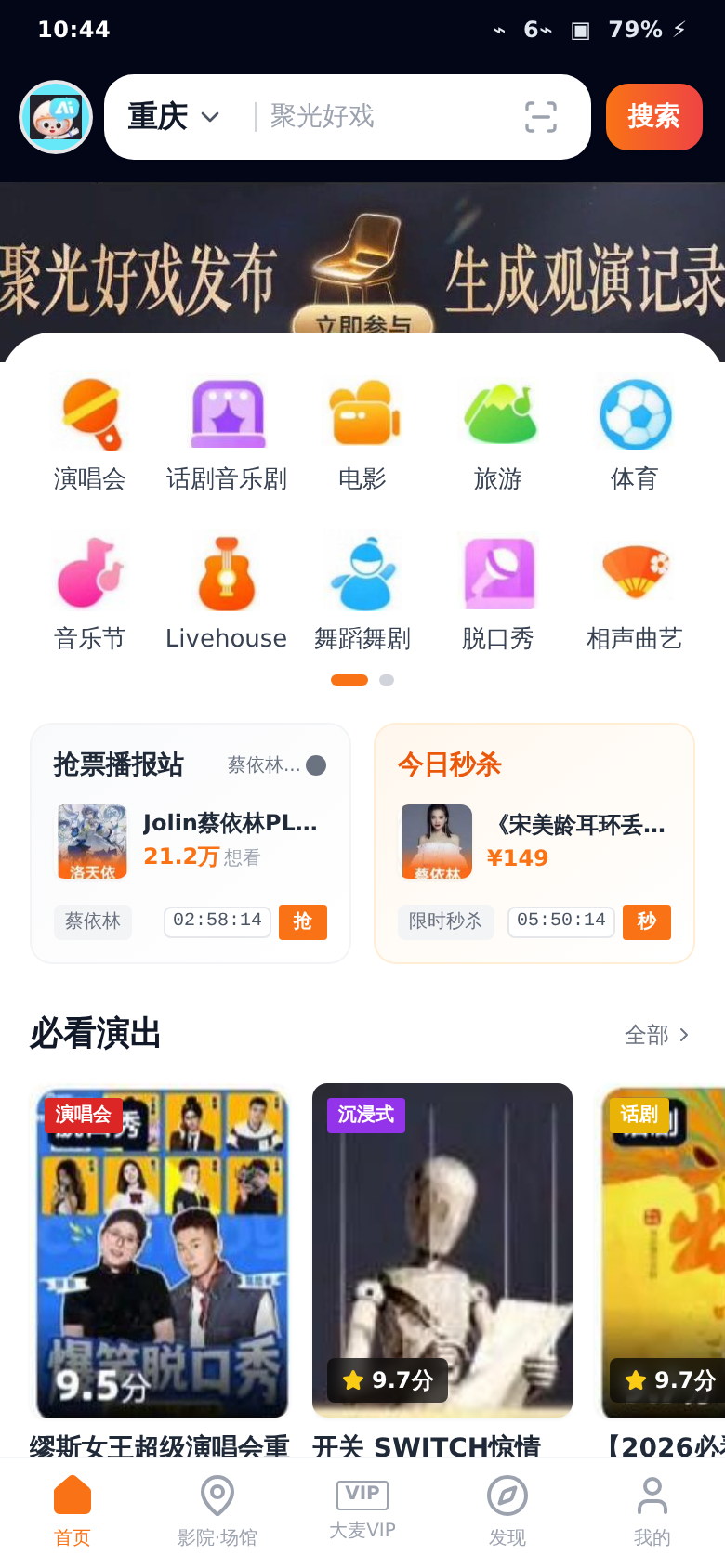}}\par\medskip
\caption{\textbf{Generated-app home screens (1/2).} Default states of formal augmented browser environments;
fresh browsers, common viewport, no task-specific initialization. Device shells are hidden; app content
and controls remain. The twenty apps match the human-audit roster (Appendix~\ref{app:portfolio}); these
images illustrate entry diversity.}
\label{fig:app-gallery-a}\end{figure}\clearpage
\begin{figure}[p]\centering
\appcover{Meituan}{\includegraphics{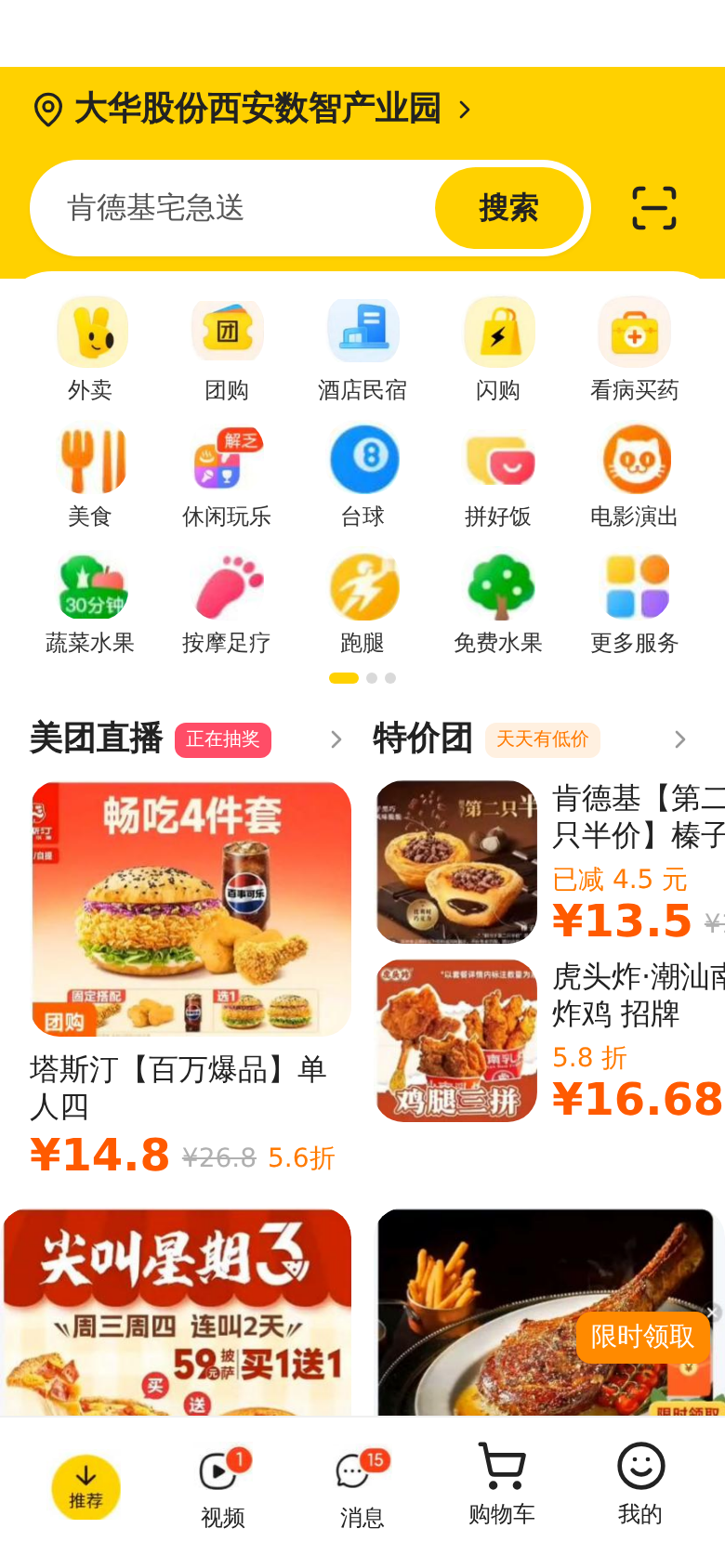}}
\hfill
\appcover{58Tongcheng}{\includegraphics{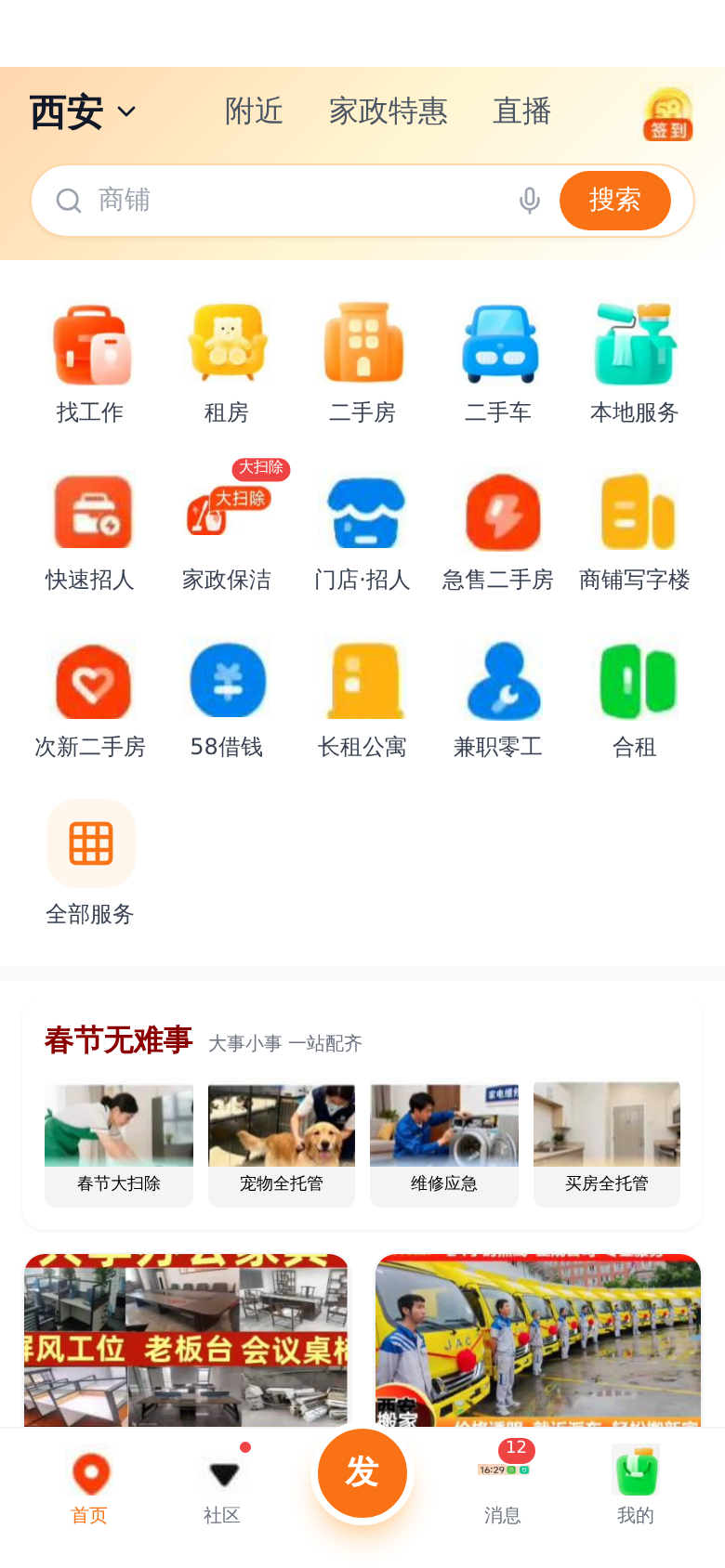}}
\hfill
\appcover{Didi}{\includegraphics{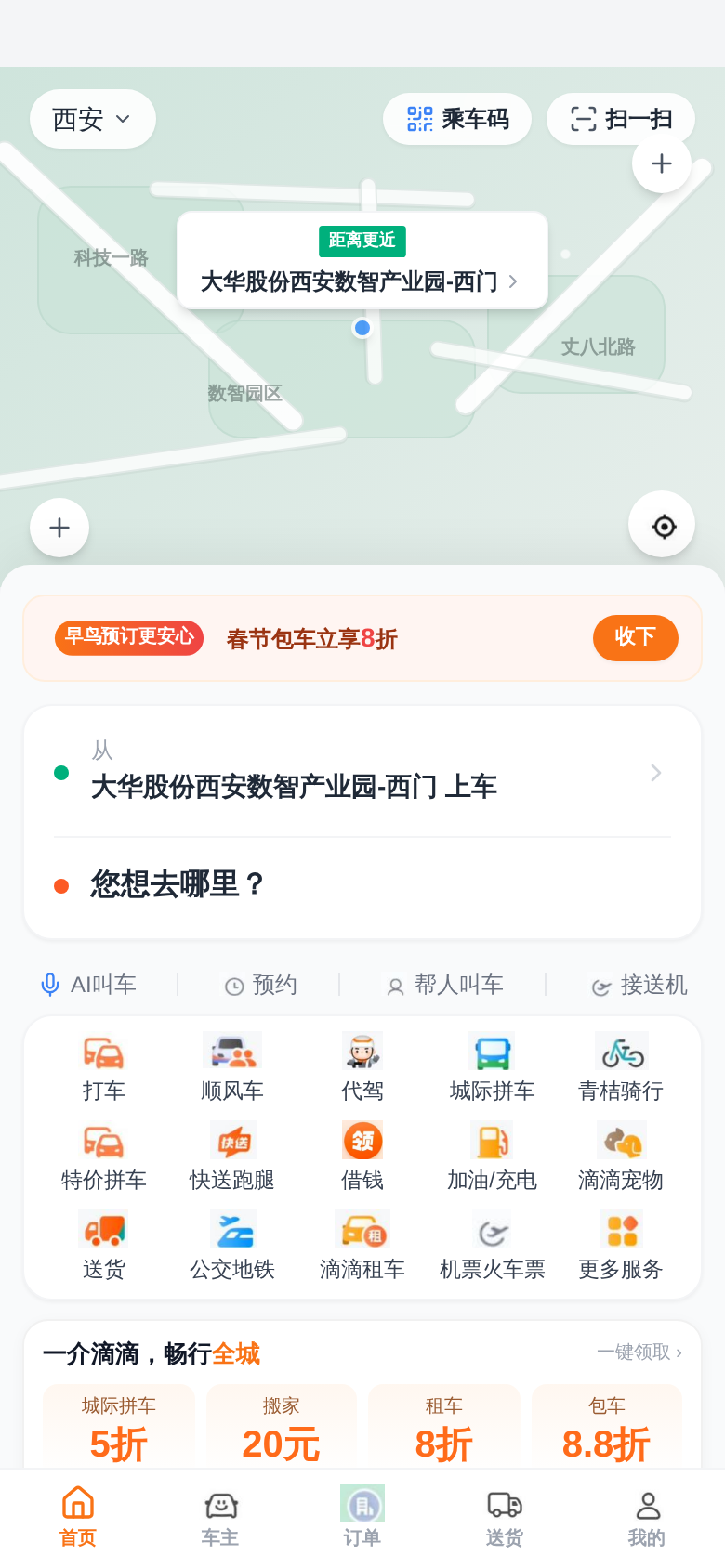}}
\hfill
\appcover{Railway 12306}{\includegraphics{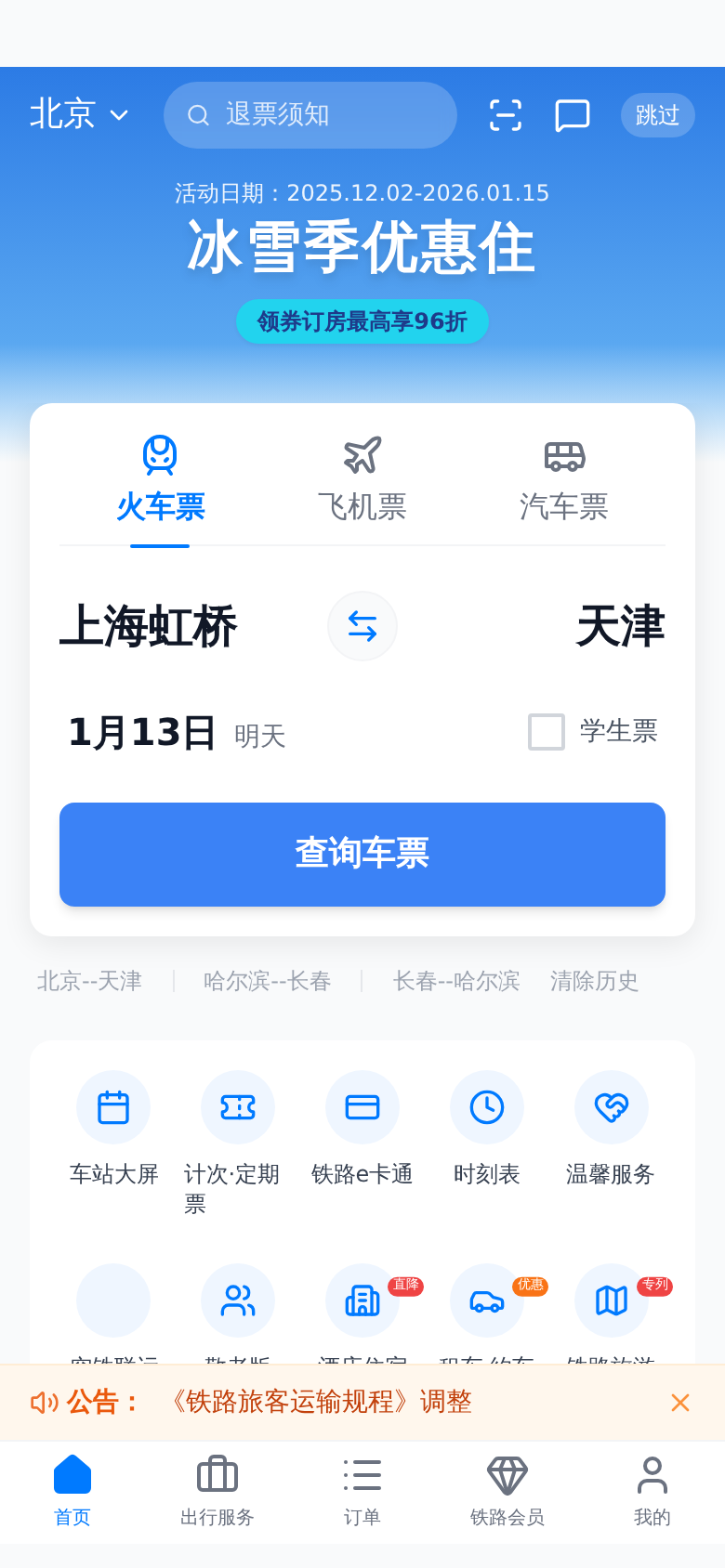}}\par\medskip
\appcover{Dingdong Maicai}{\includegraphics{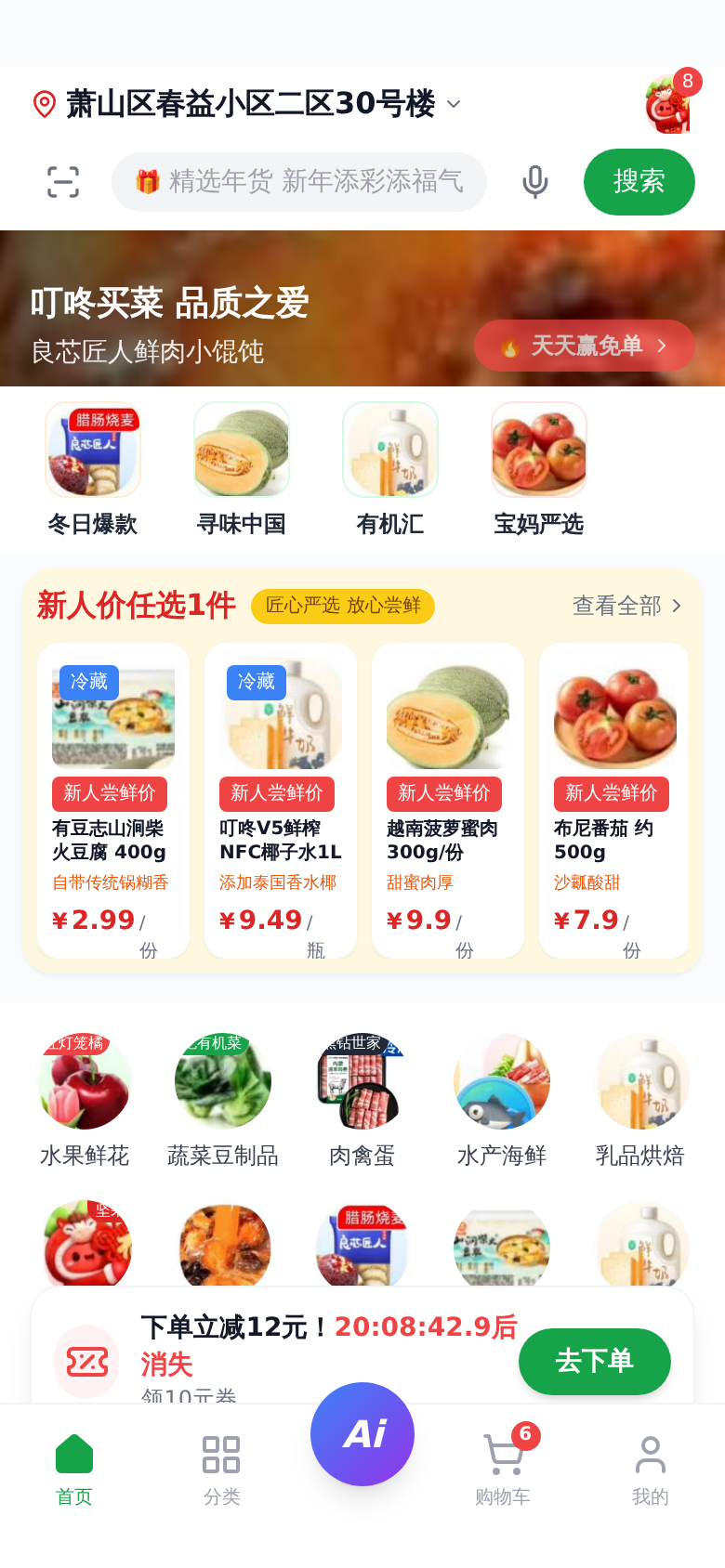}}
\hfill
\appcover{KFC}{\includegraphics{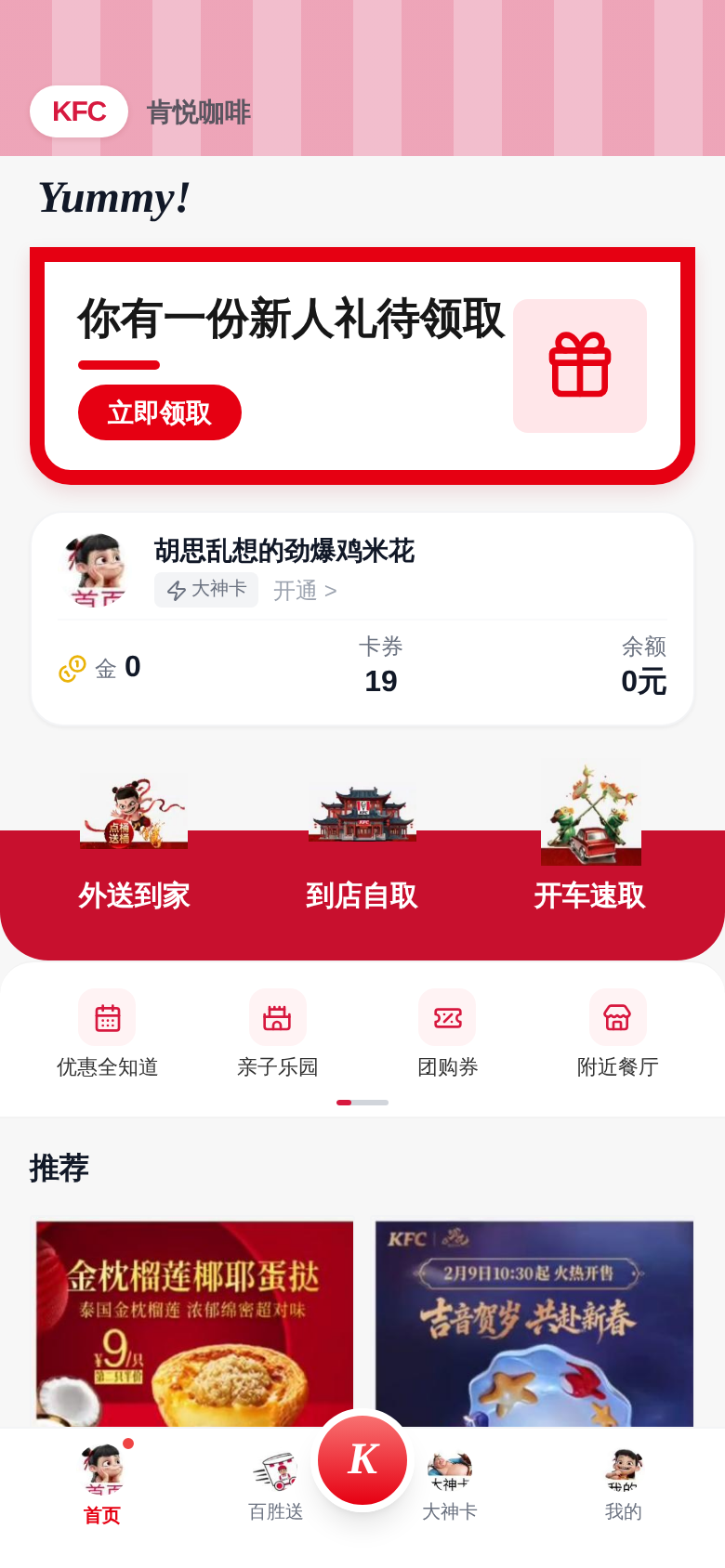}}
\hfill
\appcover{Dewu}{\includegraphics{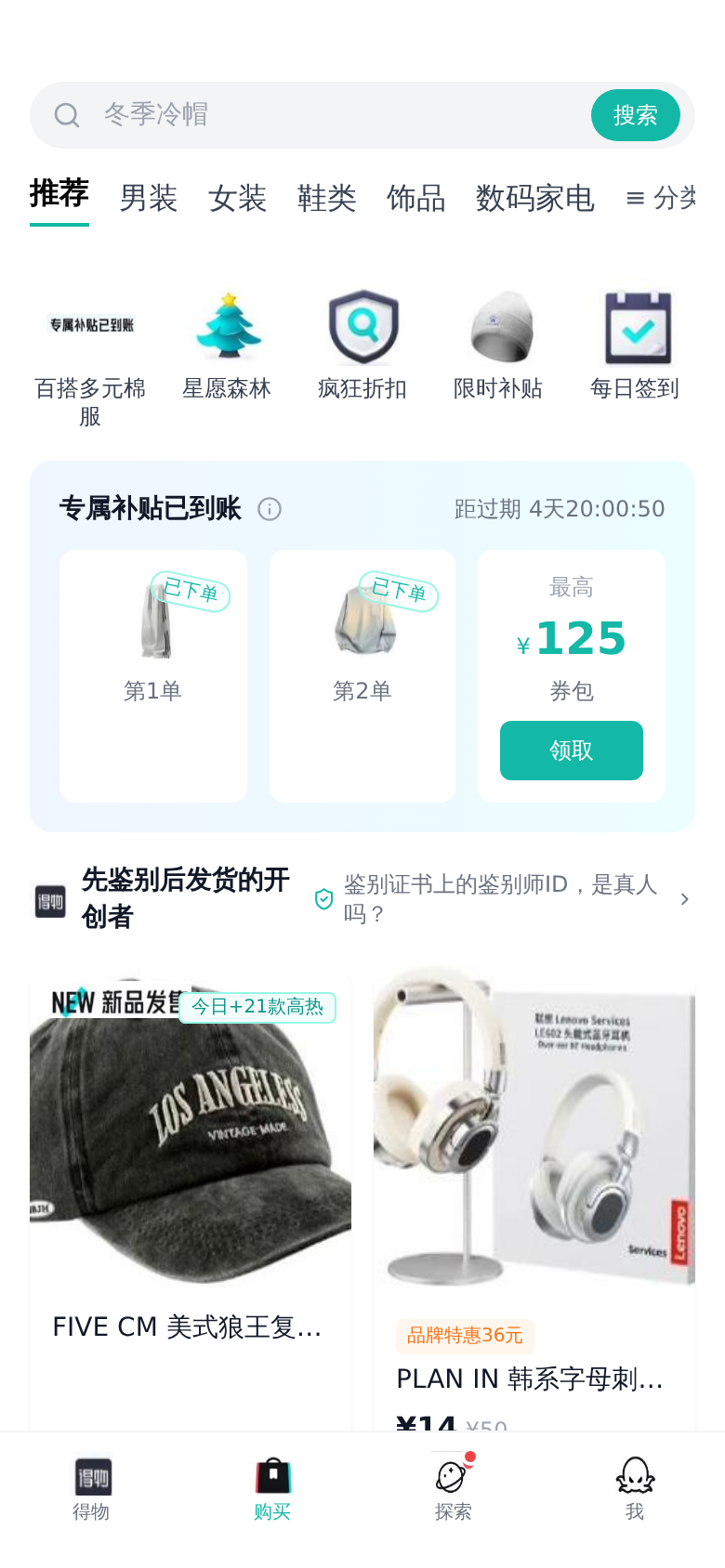}}\par\medskip
\appcover{Xianyu}{\includegraphics{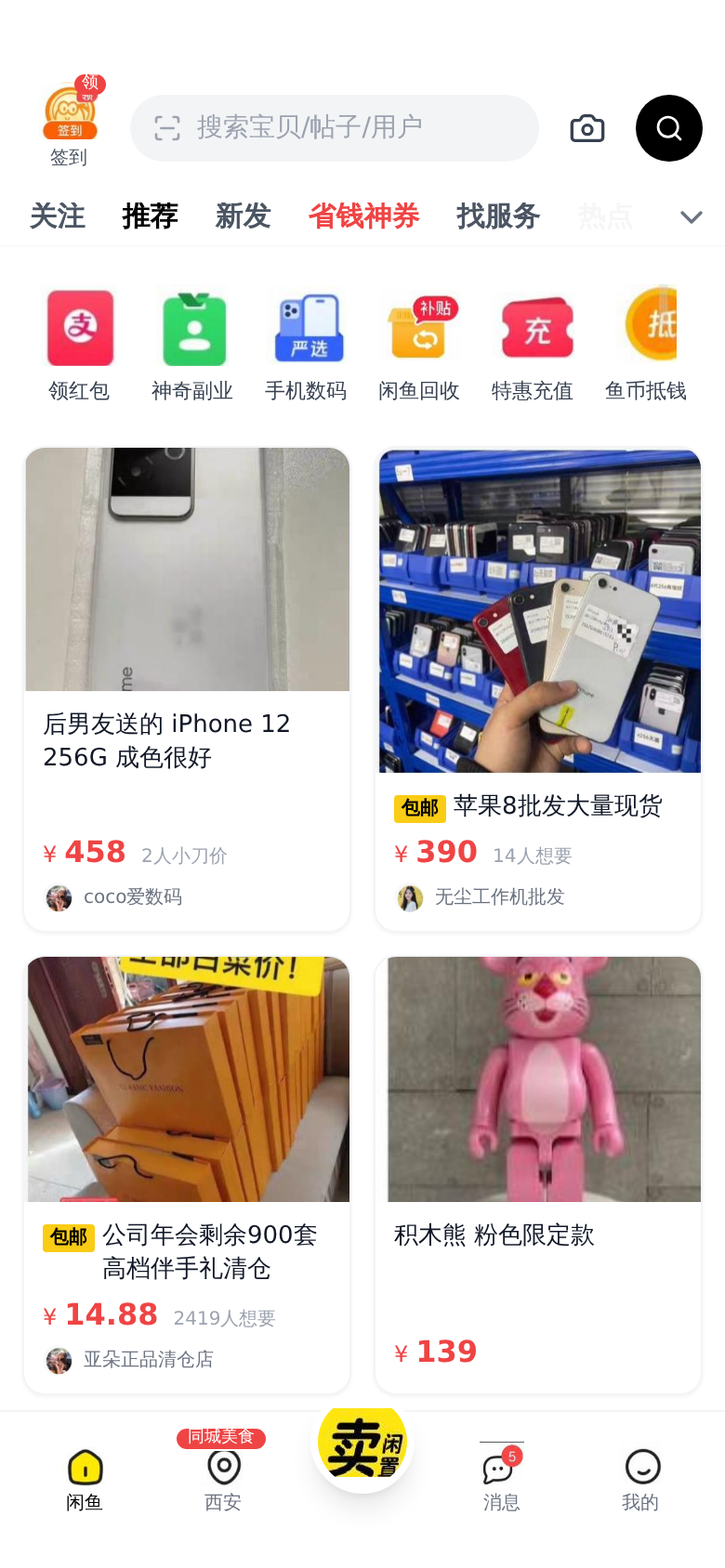}}
\hfill
\appcover{Autohome}{\includegraphics{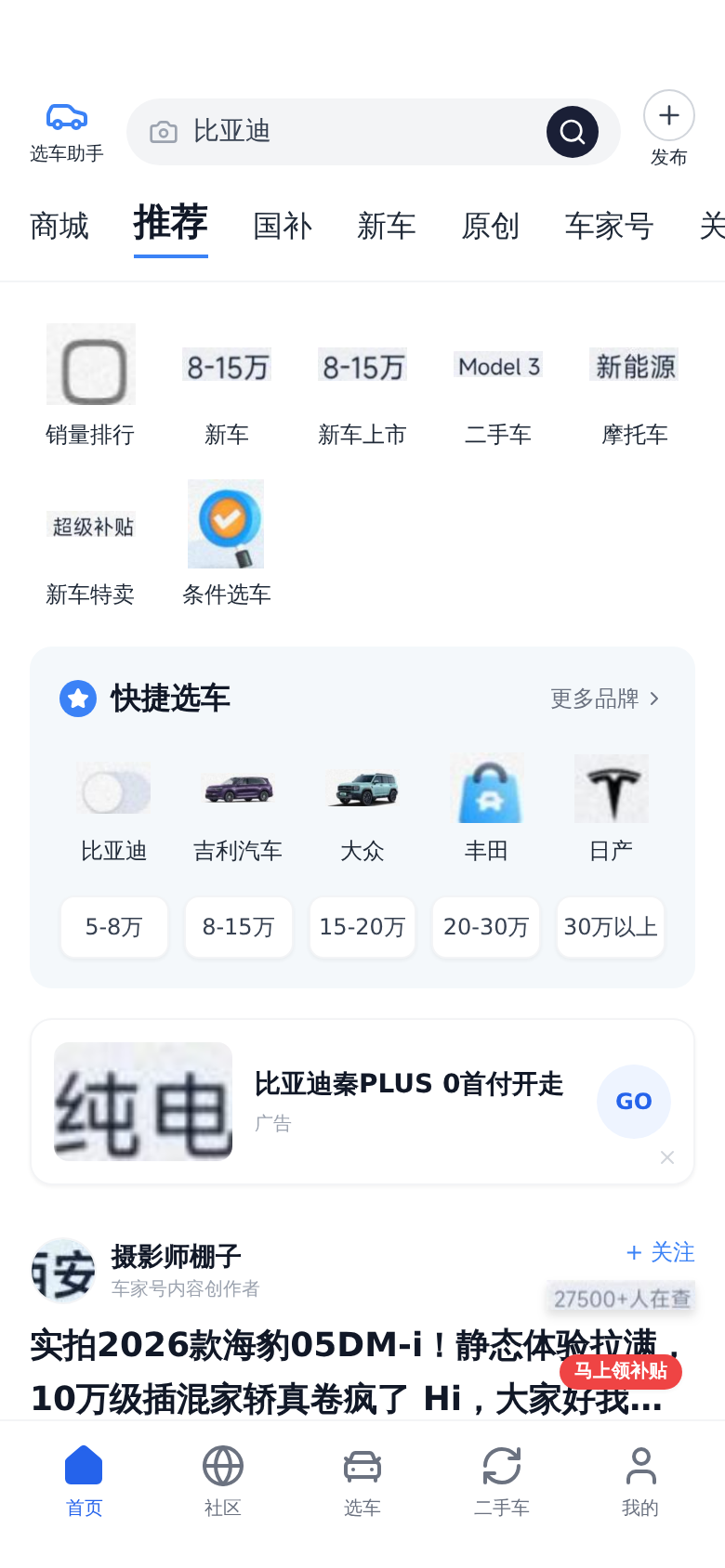}}
\hfill
\appcover{Dongchedi}{\includegraphics{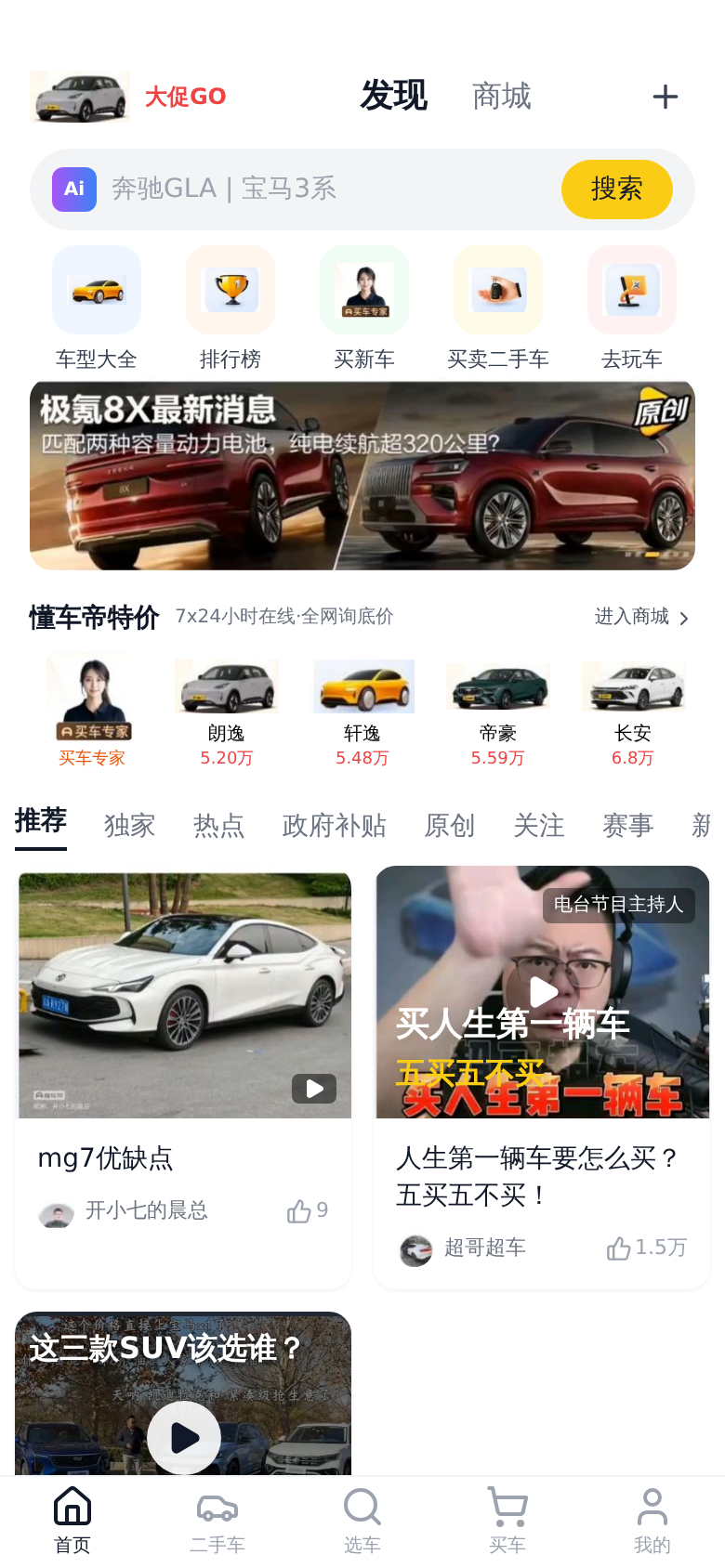}}\par\medskip
\caption{\textbf{Generated-app home screens (2/2).} Local-service, travel, grocery, food, shopping,
and automotive entry surfaces captured from formal augmented browser environments at a common viewport. Decorative
phone shells are hidden and app controls retained.}
\label{fig:app-gallery-b}\end{figure}\clearpage

\appsection{prompts}{app:prompt-templates}{Prompt Templates}

The six templates follow the pipeline's decision boundaries: raw traces become local evidence,
repeated observations become candidate shared semantics, a compiled package is checked against
source traces by report-only checks, the validated package is extracted into a task-facing index,
and prior learner outcomes guide the next tasks composed over that index. They are method-level
condensations of the workflow, not verbatim run-time prompts. Two short rules then locate repair
authority and the separate evaluation-only cross-app branch.

\begin{t2wpromptbox}{Source-trace preprocessing: disposition and description}
\phantomsection\label{app:preprocessing-prompt}
\begin{lstlisting}[style=t2wprompt,basicstyle=\ttfamily\footnotesize]
ROLE   Classify ONE raw trace without editing it.
INPUT  Each raw action (family, coordinates, typed value,
       status) and the marked screenshot BEFORE it. A
       circle marks the host; TYPE marks an input; a red
       rectangle marks existing text; a drag line marks
       direction and scroll region; no mark may indicate
       a terminal row.
ORDER  Classify every row, describe kept rows, then write
       the task instruction and terminal contract.
DISPOSITION Give every raw row a reason:
  keep     visible effect or necessary interaction;
  drop     no effect, redundant recovery, or launch context
           retained in the recording but not task behavior;
  observe  terminal row kept as evidence, not an action.
DESCRIBE For kept rows, name the visible host and purpose;
       give both ordinal and readable list-item identity,
       exact typed value, or gesture direction and region.
       Raw action family and value override prose.
OUTPUT One JSON object: step_predictions, then
       task_instruction, then terminal_contract. The task
       uses only kept rows. The contract records observed
       boundary, outcome fields, last relevant step, commit
       status, required/forbidden state, source evidence,
       and remaining uncertainty.
NEVER  invent labels, outcomes, or invisible actions;
       include dropped behavior in the task; or assign
       canonical pages, State fields, guards, or verifier
       predicates before reconciliation.
\end{lstlisting}
\end{t2wpromptbox}

\begin{t2wpromptbox}{Cross-trace reconciliation and factorization}
\phantomsection\label{app:reconciliation-prompt}
\begin{lstlisting}[style=t2wprompt,basicstyle=\ttfamily\footnotesize]
ROLE   Propose one shared app model from many trace-local
       observations. Preserve uncertainty; do not build UI.
INPUT  Restored pre/post transition candidates, marked hosts,
       action values, lifecycle context, and trace-step links.
MATCH  Map local pages, hosts, and entities to canonical
       owners using shell and route continuity, containment,
       entry/exit state, entity lineage, and transitions.
       Similar text or appearance alone is insufficient.
FOR EACH candidate execution key
       (page, host, action, observed guard), propose reads,
       writes, destination, visible postcondition, and
       projection, citing every supporting source step.
FACTOR Structure: stable pages, routes, hosts, containment.
       State: owned mutable fields, guards, transitions,
       initialization, reset, and visible projections.
       Data: reusable entity identity, attributes, relations,
       and source-bound resources.
CONFLICT If observed state explains divergent effects,
       propose the evidenced discriminator as a guard.
       Otherwise retain alternatives as unresolved.
OUTPUT Canonical-mapping and relation CANDIDATES, source
       links, conflicts, and missing dependencies. The
       separate Supported gate admits only complete,
       compatible, source-covered relations.
NEVER  settle ownership by filename, majority vote, visual
       similarity, implementation convenience, or a later
       task's needs; self-admit an unsupported relation.
\end{lstlisting}
\end{t2wpromptbox}

\begin{t2wpromptbox}{Construction-time source-trace validation}
\phantomsection\label{app:source-review-prompt}
\begin{lstlisting}[style=t2wprompt,basicstyle=\ttfamily\footnotesize]
ROLE   Review a FROZEN GUI execution against its source
       trace. Emit a verdict and next owner; do not act.
INPUT  Chronological SOURCE marked screenshots, then
       persisted EXECUTION screenshots with action labels;
       the instruction and per-step state, lifecycle,
       projection, typeahead, and terminal checklist.
ORDER  Read the full SOURCE sequence before EXECUTION.
       Compare causal states, not equal row indices. Each
       SOURCE row is pre-action: its next row or terminal
       evidence establishes the intended effect.
CHECK  The marked target is the current visible host, not
       a destination or a post-action leak. Check whole
       initial/current viewports: route and start data,
       chrome, selected tabs, overlays, list content,
       density, scale, anchor, and crop origin.
       Check action value, exact host scope, reachability,
       enabled state, lifecycle, parent projection, scroll
       region, and autocomplete order. Check forbidden
       future state and terminal obligations.
OUTPUT One trajectory verdict: review_pass_candidate,
       generated_env_mismatch with subtype,
       page_boundary_blocker, reference_oracle_blocker,
       or evidence_blocked; give the first mismatch with
       source/execution evidence, repairability, and owner.
       A pass candidate still needs independent probes.
NEVER  pass a right click in a wrong viewport; dismiss
       scale/anchor/crop differences without evidence;
       infer success from missing evidence or elapsed time;
       edit the trace, checklist, run, or environment.
\end{lstlisting}
\end{t2wpromptbox}

\begin{t2wpromptbox}{Task-facing index of the validated environment}
\phantomsection\label{app:structure-extraction-prompt}
\begin{lstlisting}[style=t2wprompt,basicstyle=\ttfamily\footnotesize]
ROLE   Extract a read-only index of ONE validated app for
       task authoring. Extract; do not model, do not
       generate tasks, do not introduce app behavior.
INPUT  The source-validated package and its replay records
       (surfaces, hosts, controls, fields, entities, reset,
       lifecycle); the frozen app model only for stable IDs,
       provenance, and support anchors.
OUTPUT app_index with stable surface/host/control IDs, entity
       schemas, observable fields and outcomes, supported
       operation IDs with references to their established
       guards/effects, implementation status, evidence
       links, and exclusions.
OUTPUT dependency_index with each visible producer field,
       consumer action or decision, value type, entity scope,
       freshness, legal discovery path, and supporting
       source steps, taken from already modeled relations.
MARK   Supported composition opportunities: aggregation,
       comparison, state carried across surfaces, lifecycle
       revisits, recovery, and multi-entity tracking.
RULE   Every emitted item must reference an existing
       supported relation and its source evidence; the index
       has no semantic authority. Drop what cannot resolve.
EXCLUDE unsupported or unresolved behavior, inaccessible
       producers, incompatible types or scope, and
       non-observable outcomes. Retain supported items
       tagged as not yet implemented.
NEVER  infer a transition, guard, or effect; invent a page,
       route, field, relation, state, action, lifecycle, or
       dependency; infer diversity from wording or names.
\end{lstlisting}
\end{t2wpromptbox}

\begin{t2wpromptbox}{Rollout-to-gap summary for the next batch}
\phantomsection\label{app:learner-gap-prompt}
\begin{lstlisting}[style=t2wprompt,basicstyle=\ttfamily\footnotesize]
ROLE   Summarize what the PRECEDING learner needs to practice
       within the frozen, source-supported app index.
INPUT  Frozen prior practice-task manifest, learner rollouts
       and checkpoint/terminal outcomes, task capability and
       dependency labels, fixed-solver outcomes when present,
       and attributed environment/task/provider defects.
FIRST  Separate non-learner faults and invalid attempts.
       Do not count them as learner competence evidence.
GROUP  By supported capability and dependency motif:
       attempts, terminal/partial progress, repeated or
       no-effect actions, lost state/entity bindings,
       recovery behavior, weak combinations, and saturation.
OUTPUT Evidence-linked conditioning notes: weak or
       unsaturated motifs to target, saturated motifs to
       deprioritize, and concrete failure patterns. Carry
       uncertainty when attribution is incomplete.
NEVER  inspect held-out benchmark outcomes; alter app
       semantics; call an environment or solver failure a
       learner gap; request unsupported interactions.
\end{lstlisting}
\end{t2wpromptbox}

\begin{t2wpromptbox}{Supported single-app task authoring and bounded hardening}
\phantomsection\label{app:task-authoring-prompt}
\begin{lstlisting}[style=t2wprompt,basicstyle=\ttfamily\footnotesize]
ROLE   Author SINGLE-APP training candidates over the frozen
       task-facing index and its dependency index.
MODE   Seed H=30; augmented H=45; screen separately.
INPUT  Supported IDs, evidence, reset/terminal contracts;
       guided augmentation also receives preceding learner
       gaps. Seed and static omit that conditioning.
GENERATE Goal/terminal value, start/reset, supported path,
       acceptable end states, subtasks/decisions with later
       consumers, visible sources, scope/freshness/discovery,
       verification/correction, evaluator-visible result.
       Bind required surface/state/action/entity/dependency
       IDs to source evidence; otherwise reject.
DIFFICULTY Use supported aggregation, state composition,
       lifecycle/recovery, or multi-entity constraints.
       For a deliberately long task, require >=7 business
       subtasks, 4 decisions, 3 cross-state dependencies,
       2 evidence revisits, and 2 evidenced phenomena, one
       information- or state-bearing. Clicks do not count.
HARDEN Only a solver-solvable lineage diagnosed as saturated
       (learner mean reward >0.8) enters this branch. Add
       supported dependencies or evidence revisits, retain
       all prior obligations, and stop after at most two
       refinements. Send each revision to implementation
       and screening.
HOLD   Hard or solver-unsolved lineages; never simplify them.
MAY    request schema-backed entity/value/state instances or
       the instantiation of an indexed, not-yet-implemented
       supported operation as the candidate's task slice.
NEVER  add hidden metrics, task IDs, unsupported app behavior,
       new transitions/guards/effects, unverifiable outcomes,
       wording-only diversity, or navigation for length.
\end{lstlisting}
\end{t2wpromptbox}

\begin{t2wpromptbox}{Workflow rule: repair under a frozen app model}
\begin{lstlisting}[style=t2wprompt,basicstyle=\ttfamily\footnotesize]
INPUT  Frozen app specification, the existing package, and a
       reproduced defect packet: step, expected vs. observed
       state, source/GUI evidence, owner. Checks only report.
REPAIR Locate the part of the existing package that causes
       the mismatch and edit it: rendering, route wiring,
       storage, bindings, projection, a missing
       implementation of a supported relation, or a state
       update or control behavior that the source shows is
       realized wrongly.
AFTER  Invalidate dependent evidence; replay affected traces
       from reset and repeat independent review/probes.
DROP   A defect that cannot be closed by editing the existing
       package within the source-supported scope (closure
       would need a page identity, relation, guard, or effect
       the source does not support): record it, drop the
       slice.
NEVER  edit the app specification, choose an interpretation
       in UI, or make a rollout pass by inventing behavior
       the source does not support.
\end{lstlisting}
\end{t2wpromptbox}

\begin{t2wpromptbox}{Workflow rule: typed two-app evaluation composition}
\begin{lstlisting}[style=t2wprompt,basicstyle=\ttfamily\footnotesize]
INPUT  Two source-validated app-local task-facing indexes
       with dependency indexes. No learner outcomes are seen.
MODE   H=60, evaluation only. Require human GUI completion
       within 45 public actions.
COMPOSE A GUI-visible typed A -> B -> A handoff: an A value
       supplies a B input/guard; a B result changes the final
       action in preserved A state. Check type, entity scope,
       freshness, visible discovery, and value parity.
FREEZE Pair, orientation, task signature, and joint terminal
       oracle before policy evaluation; keep it outside
       training and environment repair.
REJECT Independent task concatenation, stale/hidden transfer,
       lost return state, or any new app-local semantics.
NEVER  treat the composition as a cross-app source recording.
\end{lstlisting}
\end{t2wpromptbox}

\end{document}